\documentclass[lettersize,journal]{IEEEtran}
\usepackage{amsmath,amsfonts}
\usepackage{algorithmic}
\usepackage{algorithm}
\usepackage{array}
\usepackage[caption=false,font=normalsize,labelfont=sf,textfont=sf]{subfig}
\usepackage{textcomp}
\usepackage{stfloats}
\usepackage{url}
\usepackage{verbatim}
\usepackage{graphicx}
\usepackage{cite}

\usepackage[colorlinks=True]{hyperref}
\usepackage{upgreek}
\usepackage{wasysym}
\usepackage{xcolor}
\usepackage{multirow}
\usepackage{multicol}
\usepackage{makecell}
\usepackage{booktabs}
\usepackage{amssymb}
\usepackage{enumitem}
\begin{document}

\title{FusionMamba: Efficient Remote Sensing Image Fusion with State Space Model}

\author{Siran Peng, 
	    Xiangyu Zhu,~\IEEEmembership{Senior Member,~IEEE,}
	    Haoyu Deng,
	    Liang-Jian Deng,~\IEEEmembership{Senior Member,~IEEE,}
	    and Zhen Lei,~\IEEEmembership{Fellow,~IEEE}
\thanks{Siran Peng and Xiangyu Zhu are with the State Key Laboratory of Multimodel Artificial Intelligence Systems, Institute of Automation, Chinese Academy of Sciences (CASIA), Beijing 100190, China; the School of Artificial Intelligence, University of Chinese Academy of Sciences (UCAS), Beijing 100049, China. (Emails: pengsiran2023@ia.ac.cn, xiangyu.zhu@nlpr.ia.ac.cn)}
\thanks{Haoyu Deng is with the School of Information and Communication Engineering, University of Electronic Science and Technology of China (UESTC), Chengdu, Sichuan 611731, China. (Email:  haoyu\_deng@std.uestc.edu.cn)}
\thanks{Liang-Jian Deng is with the School of Mathematical Sciences/Multi-Hazard Early Warning Key Laboratory of Sichuan Province, University of Electronic Science and Technology of China (UESTC), Chengdu, Sichuan 611731, China. (Email: liangjian.deng@uestc.edu.cn)}
\thanks{Zhen Lei is with the State Key Laboratory of Multimodel Artificial Intelligence Systems, Institute of Automation, Chinese Academy of Sciences (CASIA), Beijing 100190, China; the School of Artificial Intelligence, University of Chinese Academy of Sciences (UCAS), Beijing 100049, China; the Centre for Artificial Intelligence and Robotics, Hong Kong Institute of Science and Innovation, Chinese Academy of Sciences, Hong Kong, China. (Email: zhen.lei@ia.ac.cn)}
\thanks{Corresponding author: Liang-Jian Deng.}
}



\maketitle

\begin{abstract}
Remote sensing image fusion aims to generate a high-resolution multi/hyper-spectral image by combining a high-resolution image with limited spectral data and a low-resolution image rich in spectral information. Current deep learning (DL) methods typically employ convolutional neural networks (CNNs) or Transformers for feature extraction and information integration. While CNNs are efficient, their limited receptive fields restrict their ability to capture global context. Transformers excel at learning global information but are computationally expensive. Recent advancements in the state space model (SSM), particularly Mamba, present a promising alternative by enabling global perception with low complexity. However, the potential of SSM for information integration remains largely unexplored. Therefore, we propose FusionMamba, an innovative method for efficient remote sensing image fusion. Our contributions are twofold. First, to effectively merge spatial and spectral features, we expand the single-input Mamba block to accommodate dual inputs, creating the FusionMamba block, which serves as a plug-and-play solution for information integration. Second, we incorporate Mamba and FusionMamba blocks into an interpretable network architecture tailored for remote sensing image fusion. Our designs utilize two U-shaped network branches, each primarily composed of four-directional Mamba blocks, to extract spatial and spectral features separately and hierarchically. The resulting feature maps are sufficiently merged in an auxiliary network branch constructed with FusionMamba blocks. Furthermore, we improve the representation of spectral information through an enhanced channel attention module. Quantitative and qualitative valuation results across six datasets demonstrate that our method achieves state-of-the-art (SOTA) performance, underscoring the effectiveness of FusionMamba. The code is available at {\url{https://github.com/PSRben/FusionMamba}}.

\end{abstract}


\begin{figure}[t]
	\begin{center}
		\begin{minipage}{1\linewidth}
			{\includegraphics[width=0.9\linewidth]{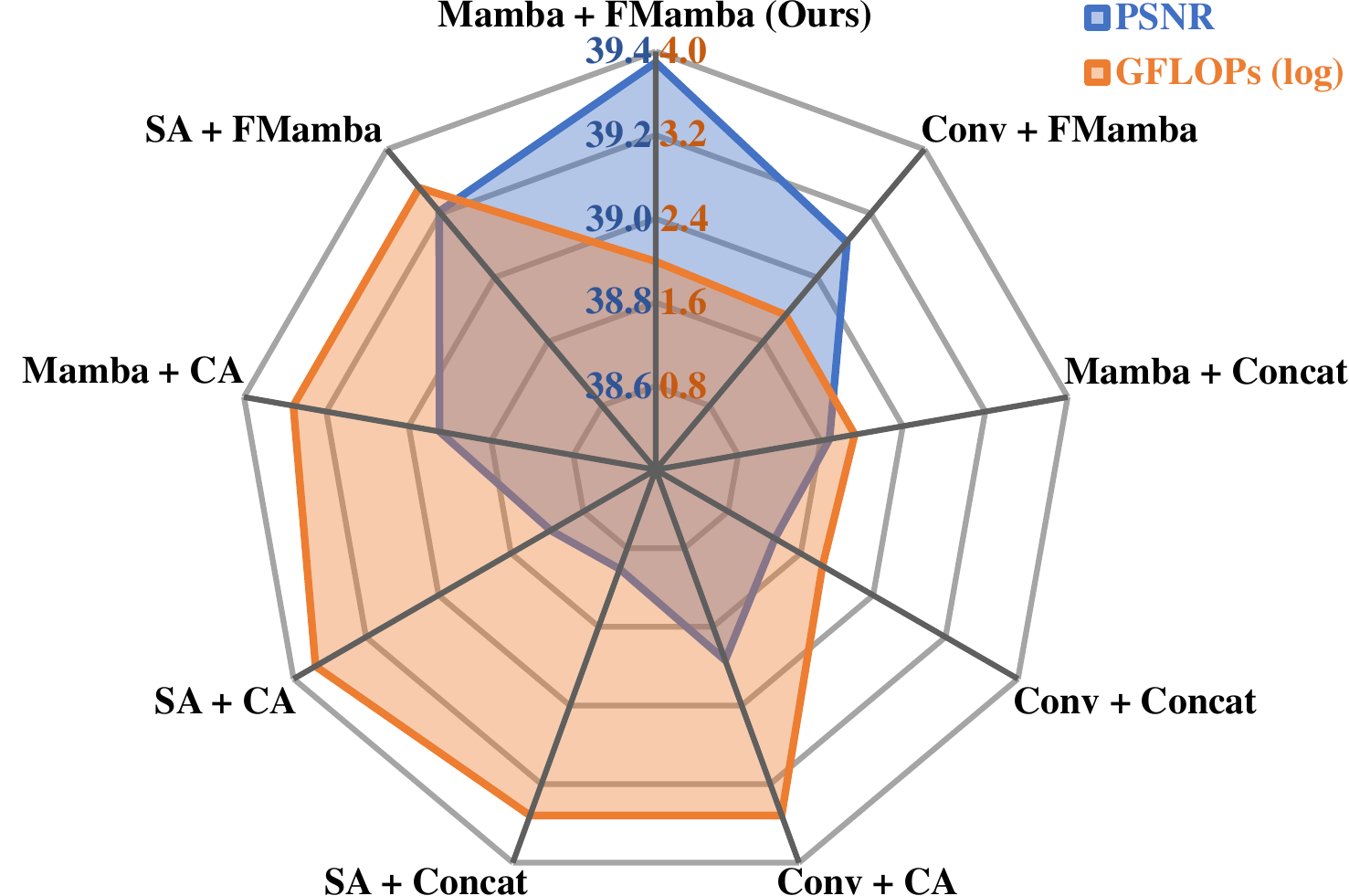}}
			\centering
		\end{minipage}
	\end{center}
	\caption{
	Different combinations of feature extraction methods and information integration approaches for remote sensing image fusion. The candidate feature extraction methods include the convolution (Conv) layer, self-attention (SA) module \cite{vaswani2017attention}, and four-directional Mamba (Mamba) block \cite{liu2024vmamba}. For information integration, the options comprise the concatenation (Concat) operation, cross-attention (CA) module, and the proposed FusionMamba (FMamba) block. For fairness, all combinations are designed with the same number of parameters.	Quantitative evaluation results on 20 reduced-resolution samples from the WorldView-3 (WV3) dataset \cite{9844267} demonstrate the superior efficacy and efficiency of our method. For precise values, please refer to Table \ref{abl5}.  \label{titlefigure} 
}
\end{figure}

\section{Introduction}
Due to hardware limitations, satellite sensors often struggle to capture high-resolution multi/hyper-spectral images. As an alternative approach, they can simultaneously acquire a high-resolution image with limited spectral data and a low-resolution image with extensive spectral information. Remote sensing image fusion aims to merge these two types of images, generating a high-resolution result enriched with spectral characteristics. This study primarily investigates two remote sensing image fusion tasks: pansharpening \cite{9245579} and hyper-spectral pansharpening \cite{7284770}. Pansharpening involves creating a high-resolution multi-spectral (HRMS) image by combining a high-resolution panchromatic (PAN) image with a low-resolution multi-spectral (LRMS) image. Hyper-spectral pansharpening extends this process to hyper-spectral images, producing a high-resolution hyper-spectral (HRHS) image from a PAN image and a low-resolution hyper-spectral (LRHS) image.

Traditional (hyper-spectral) pansharpening studies can be broadly classified into three categories: component substitution (CS) methods, multi-resolution analysis (MRA) approaches, and variational optimization (VO) techniques. The CS-based methods \cite{aiazzi2006mtf, 2010A, 2019Robust} project the LRMS/LRHS image into a transformed domain, where the spatial information is treated as an independent component. By substituting this component with the PAN image, a desired HRMS/HRHS result is produced. These methods are known for their simplicity, low computational requirements, and high spatial fidelity. However, they often lead to significant spectral distortions. The MRA-based approaches \cite{6616569, vivone2018full} utilize an MRA framework to inject spatial details from the PAN image into the LRMS/LRHS image, thus generating an HRMS/HRHS output. These techniques are effective at preserving spectral characteristics but may experience issues with spatial distortions. The VO-based techniques \cite{palsson2013new, 9384242, 9722977, NEURIPS2022_89ef9ce3, 10410873} aim to uncover the intrinsic relationships between two types of images. They typically rely on various forms of prior information to construct optimization models that integrate spatial and spectral data. Despite their meticulous designs, VO-based techniques often fail to deliver satisfactory fusion results and are hindered by slow inference speeds.

Over the past few years, deep learning (DL) has become the leading solution for addressing image fusion problems within the remote sensing domain. By leveraging the powerful feature learning and non-linear fitting capabilities of neural networks, DL-based methods have consistently yielded impressive outcomes \cite{2016Pansharpening, 8237455, 8667448, 8731649, rs12040676, rs12172804, 9106801, 2020Detail, Xu_2021_CVPR, zhou2022pan, 9785810, 9851415, 10034991, 10137388, 10.1145/3581783.3612084, 10243544, 10415854, 10410890, Duan_2024_CVPR}. 
{Analyzing these studies reveals two key insights. \emph{First, networks with global perception abilities are particularly effective, as they leverage holistic information rather than solely relying on localized features. Second, since low-level tasks necessitate processing at relatively high resolutions, it is essential to keep computational complexity within manageable limits.}}
Most DL-based methods utilize convolutional neural networks (CNNs) \cite{726791} or Transformers \cite{vaswani2017attention} for feature extraction and information integration. Although CNNs are computationally efficient, they are hindered by limited receptive fields, restricting their ability to capture global context. On the contrary, Transformers excel at extracting global features but are burdened by quadratic complexity with respect to the length of input tokens. Recent breakthroughs in the state space model (SSM) \cite{gu2021combining, gu2021efficiently, fu2022hungry, smith2022simplified}, particularly Mamba \cite{gu2023mamba}, offer a promising solution to this issue by achieving global perception with linear complexity. The SSM has demonstrated remarkable success across a range of computer vision tasks \cite{zhu2024vision, liu2024vmamba, 10537177, 10542538, he2024pan}, delivering outstanding performance while requiring significantly fewer computational resources compared to Transformers. \emph{However, there has been limited exploration into the potential of the SSM for integrating different types of information, which is a crucial aspect of image fusion.}

Given the aforementioned situation, we propose FusionMamba, a novel method for efficient remote sensing image fusion. Our innovations focus on two aspects. First, we expand the single-input Mamba block to support dual inputs, creating the FusionMamba block. This new module effectively merges spatial and spectral features, demonstrating superiority over existing fusion techniques like concatenation and cross-attention \cite{vaswani2017attention}, as illustrated in Fig.~\ref{titlefigure}. Moreover, experimental results presented in Table~\ref{abl3} indicate that the FusionMamba block can function as a plug-and-play module for information integration. Second, based on the intrinsic properties of image fusion, we meticulously design an interpretable network architecture that incorporates Mamba and FusionMamba blocks. For feature extraction, we embed four-directional Mamba blocks \cite{liu2024vmamba} into two U-shaped network branches: the spatial branch and the spectral branch. The former emphasizes capturing spatial details from the PAN image, while the latter focuses on extracting spectral features from the LRMS/LRHS image. This design allows for the separate and hierarchical learning of spatial and spectral information. The resulting feature maps from both branches are sufficiently merged in an auxiliary combination branch, which is constructed using several FusionMamba blocks. To further improve the representation of spectral information, we introduce a Mamba-driven channel attention (MCA) module, where the traditional multi-layer perceptron (MLP) is replaced with a bidirectional Mamba block \cite{zhu2024vision}. The contributions of this study are as follows:
\begin{enumerate}
	\item To effectively merge spatial and spectral information, we expand the Mamba block to accommodate dual inputs, resulting in the innovative FusionMamba block. This module demonstrates superior effectiveness over existing fusion techniques, representing a significant advancement in the application of the SSM for information integration.
	\item According to the properties of image fusion, we develop an interpretable network architecture that incorporates Mamba and FusionMamba blocks. Our designs employ two U-shaped network branches to extract spatial and spectral features separately and hierarchically. The resulting feature maps are sufficiently merged in a combination branch. Additionally, an enhanced channel attention module is utilized to improve spectral representation.
	\item To the best of our knowledge, this study represents the first application of the SSM in hyper-spectral pansharpening and hyper-spectral image super-resolution (HISR) tasks. The proposed method achieves state-of-the-art (SOTA) performance across six datasets, thereby convincingly demonstrating the superiority of FusionMamba.
\end{enumerate}

The rest of this paper is structured as follows. Section~\ref{s2} reviews the related works and outlines our motivations. Section~\ref{s3} provides a detailed explanation of our method. In Section~\ref{s4}, we present the experimental results for both pansharpening and hyper-spectral pansharpening tasks, accompanied by comprehensive ablation studies. Finally, Sections~\ref{s5} and \ref{s6} cover the discussion and conclusion, respectively.

\section{Related Works and Motivations}
\label{s2}
\subsection{DL Methods for Remote Sensing Image Fusion}
In recent years, DL-based methods have dominated the remote sensing image fusion community. These techniques leverage the powerful feature learning and non-linear fitting capabilities inherent in neural networks, significantly outperforming traditional approaches. Broadly speaking, DL-based methods for remote sensing image fusion can be classified into two categories: CNN-based approaches and Transformer-based techniques. Notable examples in the first category include PNN \cite{2016Pansharpening}, PanNet \cite{8237455}, and FusionNet \cite{2020Detail}. PNN represents a pioneering advancement by integrating DL into the pansharpening field. It employs three stacked convolutional layers to achieve SOTA performance at the time of its publication. PanNet creatively uses high-pass filters to capture edge information and incorporates residual network (ResNet) blocks \cite{he2016deep} to extract spatial and spectral features. FusionNet embeds CNNs into the architecture of traditional algorithms, yielding remarkable outcomes. However, due to the limited receptive field size of convolutional kernels, these CNN-based approaches often struggle to capture global information, resulting in significant spatial distortions. Transformer-based techniques address this problem by calculating the correlation between any two pixels. Noteworthy works in this category include INNformer \cite{zhou2022pan} and U2Net \cite{10.1145/3581783.3612084}. The former pioneers the application of Transformers for pansharpening, utilizing modified cross-attention blocks to sufficiently merge spatial and spectral feature maps. The latter incorporates enhanced cross-attention modules into a U-shaped network, attaining SOTA results in pansharpening. Despite their exceptional performance, these techniques are burdened by a high volume of floating-point operations (FLOPs) due to the quadratic complexity. 
{Some methods in the remote sensing domain, such as STT \cite{rs13214441} and LeMeViT \cite{jiang_lemevit_2024}, attempt to alleviate this computational burden by discarding redundant tokens. However, these approaches are not suitable for image fusion tasks and still exhibit quadratic complexity.} Over the past two years, the development of DL-based methods for remote sensing image fusion has encountered a bottleneck, with none successfully achieving both global perception and low computational cost.

\begin{figure}[t]
	\begin{center}
		\begin{minipage}{1\linewidth}
			{\includegraphics[width=0.92\linewidth]{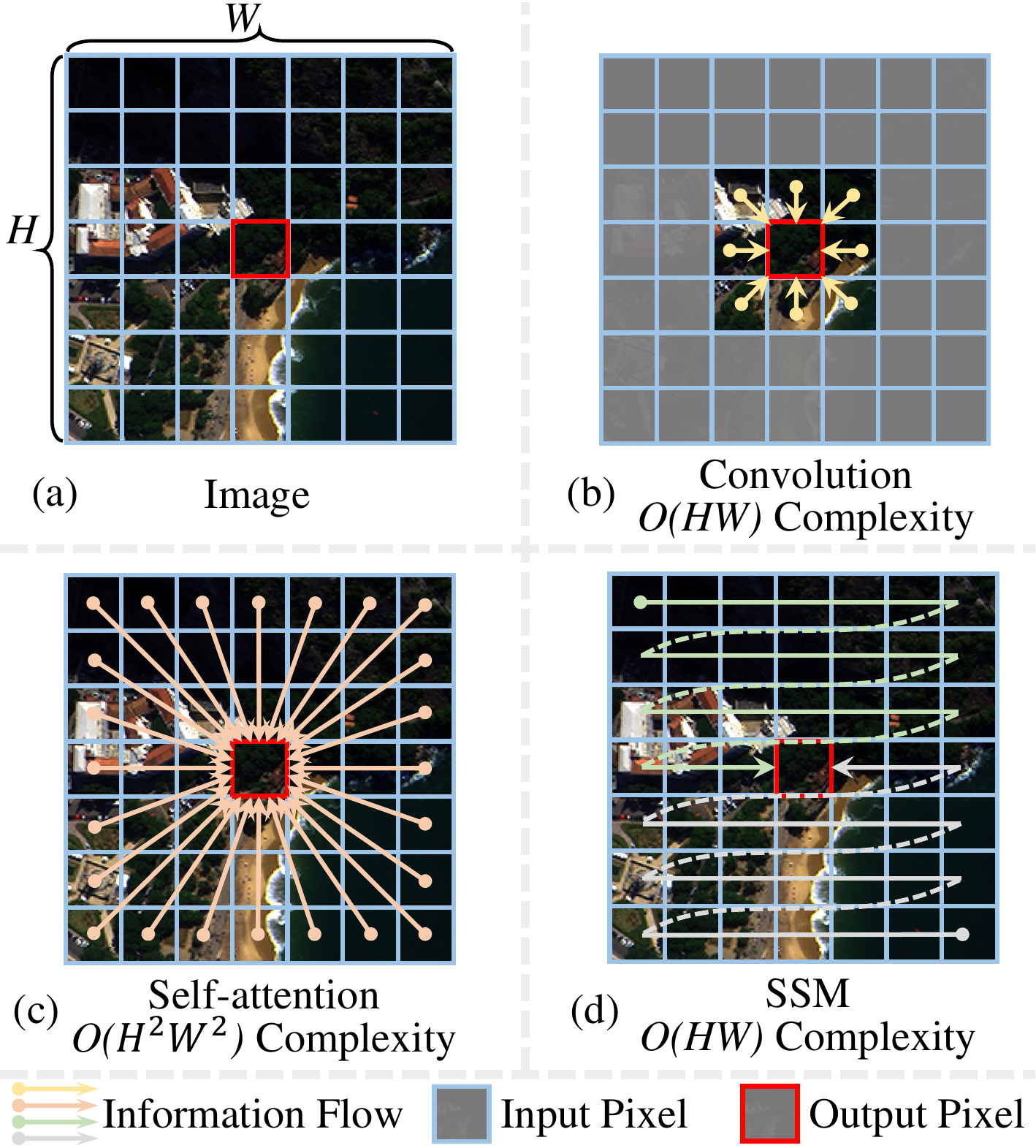}}
			\centering
		\end{minipage}
	\end{center}
	\caption{Comparison among the convolution layer in CNNs, the self-attention module in Transformers, and the SSM in bidirectional Mamba \cite{zhu2024vision}. (a) Suppose we have an image with a resolution of $H\times W$. (b) The convolution operation integrates pixels within a limited receptive field, resulting in a computational complexity of $O(HW)$. (c) The self-attention mechanism uniformly integrates all pixels, which leads to a significantly higher computational complexity of $O(H^2W^2)$. {(d) The SSM integrates all pixels along specific directions, with those closer to the output pixel contributing more significantly to the final result. Additionally, its computational complexity is $O(HW)$.} \label{rcf}}
\end{figure}

\subsection{State Space Model}
The SSM is a foundational scientific model primarily utilized in control theory and econometrics. Recently, its application has extended into the field of DL, thanks to the pioneering research of LSSL \cite{gu2021combining} and S4 \cite{gu2021efficiently}. Based on a series of mathematical derivations, LSSL approaches the SSM as a foundational DL framework. Building on this, S4 introduces the concept of \emph{normal plus low-rank}, substantially reducing the computational complexity during the training phase of the SSM. Subsequent studies, such as S5 \cite{smith2022simplified} and H3 \cite{fu2022hungry}, further explore the potential of the SSM in DL, effectively narrowing the performance gap between the SSM and Transformers. This line of research has culminated in the development of Mamba \cite{gu2023mamba}, which synthesizes pivotal findings from earlier works and proposes a selection mechanism for dynamic feature extraction. Mamba not only outperforms Transformers across various 1D tasks but also demands significantly fewer computational resources. The success of Mamba has captivated the computer vision community, leading to its widespread adoption in a variety of 2D vision tasks. 
Since Mamba was originally designed for 1D tasks with inherent directionality, applying it directly to 2D vision tasks, which typically lack such directionality, will result in incomplete global perception. To address this limitation, Vision Mamba \cite{zhu2024vision} flattens spatial feature maps from both positive and negative directions, introducing a bidirectional Mamba approach that ensures complete global perception. VMamba \cite{liu2024vmamba} further improves upon this by proposing a four-directional Mamba technique, which enables the discovery of more spatial connections. {In the remote sensing domain, notable contributions include RSCaMa \cite{10537177}, RSMamba \cite{10542538}, and Pan-Mamba \cite{he2024pan}. RSCaMam introduces the SSM into remote sensing change captioning, achieving commendable performance by employing Mamba for joint spatial-temporal modeling. RSMamba proposes a shuffle flattening method to explore unconventional spatial connections, yielding SOTA results in remote sensing image classification tasks.} Additionally, Pan-Mamba represents a pioneering effort in utilizing Mamba for pansharpening, demonstrating impressive performance even with the original Mamba blocks. However, the methods discussed above primarily concentrate on the application and directionality of the SSM, leaving its potential for information integration largely unexplored.

\subsection{Motivations}
Existing DL-based methods for remote sensing image fusion primarily utilize CNNs or Transformers for feature extraction and information integration. While CNNs are efficient, they often struggle to capture global information. Conversely, Transformers exhibit outstanding global perception but come with significant computational costs. Fortunately, recent advancements in the SSM, particularly Mamba, offer a promising solution to this dilemma, achieving both global perception and high efficiency. For a clearer understanding, Fig.~\ref{rcf} provides a visual comparison of the convolution layer in CNNs, the self-attention module in Transformers, and the SSM in bidirectional Mamba. It is evident that the SSM integrates the strengths of both CNNs and Transformers. Although the SSM has demonstrated notable success in a range of computer vision tasks, its potential for information integration, an essential aspect of image fusion, remains largely untapped. Therefore, we expand the single-input Mamba block to support dual inputs, resulting in the FusionMamba block, which effectively merges spatial and spectral information. Next, we need to determine an interpretable network architecture that leverages Mamba and FusionMamba blocks for feature extraction and information integration, respectively. Given that images from different sources exhibit distinct characteristics, we employ two U-shaped network branches, each primarily composed of Mamba blocks, to separately and hierarchically learn spatial and spectral features. Additionally, we utilize an auxiliary network branch built with FusionMamba blocks to achieve sufficient information integration. Moreover, to mitigate the distortion caused by encoding spectral data into the channels of feature maps, we develop an enhanced channel attention module to improve the representation of spectral information.

\section{Methodology}
\label{s3}
{In this section, we first introduce the notations (Section \ref{s31}) and explore the mathematical foundations of the SSM (Section \ref{s32}). Next, we provide a detailed explanation of the network architecture (Section \ref{s33}), followed by an in-depth discussion of the Mamba and FusionMamba blocks (Section \ref{s34}). Finally, we describe the loss function (Section \ref{s35}).}

\subsection{Notations}
\label{s31}
The PAN image is denoted as $\mathbf{P} \in \mathbb{R}^{{H \times W}}$, where ${H}$ and ${W}$ represent its height and width. In addition, $\mathbf{M} \in \mathbb{R}^{{h \times w\times S}}$ denotes the LRMS/LRHS image, with ${S}$ representing the number of spectral bands, ${h=\frac{H}{4}}$, and ${w=\frac{W}{4}}$. Furthermore, the up-sampled LRMS/LRHS image, the generated HRMS/HRHS image, and the ground-truth (GT) image are defined as ${\mathbf{M}}^{\rm{U}} \in \mathbb{R}^{{H \times W\times S}}$, $\mathbf{\tilde{O}} \in \mathbb{R}^{{H \times W\times S}}$, and $\mathbf{O} \in \mathbb{R}^{{H \times W\times S}}$, respectively. 
{Our network takes $\mathbf{P}$ and $\mathbf{M}$ as inputs to produce an output $\mathbf{\tilde{O}}$, which is supervised by the GT image $\mathbf{O}$.} The network performs feature extraction and information integration through five cascading stages. At the $i$-th stage, the spatial, spectral, and fusion feature maps are denoted as ${\mathbf{F}}^{\rm{a}}_{{i}}$, ${\mathbf{F}}^{\rm{b}}_{{i}}$, and ${\mathbf{F}}^{\rm{c}}_{{i}}$, respectively. The dimensions of ${\mathbf{F}}^{\{\rm{a, b, c}\}}_{\{1, 5\}}$, ${\mathbf{F}}^{\{\rm{a, b, c}\}}_{\{2, 4\}}$, and ${\mathbf{F}}^{\{\rm{a, b, c}\}}_3$ are ${H\times W\times C}$, ${\frac{H}{2}\times \frac{W}{2}\times 2C}$, and ${\frac{H}{4}\times \frac{H}{4}\times 4C}$, where $C$ represents the number of channels. Additionally, ${N}$ denotes the size of hidden states in the SSM.

\begin{figure*}[t]
	\begin{center}
		\begin{minipage}{1\linewidth}
			{\includegraphics[width=0.99\linewidth]{Fig/u2net2.pdf}}
			\centering
		\end{minipage}
	\end{center}
	\caption{The proposed network architecture. Our designs comprise two U-shaped network branches dedicated to feature extraction, a combination branch for information integration, and an MCA module for spectral enhancement. Detailed structures of the Mamba and FusionMamba blocks are depicted in Fig.~\ref{mamba}.
		\label{pipeline}}
\end{figure*}

\subsection{Preliminaries}
\label{s32}
\subsubsection{State Space Model} 
The SSM is a continuous system that maps a 1D input $x(t)\in \mathbb{R}$ into an output $y(t)\in \mathbb{R}$ via intermediate hidden states $h(t)\in \mathbb{R}^{{N}}$. This process is frequently described using ordinary differential equations (ODEs), as illustrated below:
\begin{equation}\label{E1}
	\begin{aligned}
		h'(t)&=\mathbf{A}h(t)+\mathbf{B}x(t), \\
		y(t)&=\mathbf{C}h(t).
	\end{aligned}
\end{equation} 
Here, $\mathbf{A}\in \mathbb{R}^{{N\times N}}$ denotes the state matrix governing the system's evolution. $\mathbf{B}\in \mathbb{R}^{{N\times 1}}$ and $\mathbf{C}\in \mathbb{R}^{{1\times N}}$ are projection parameters that regulate the system updates. Eq.~\ref{E1} indicates that the SSM possesses global perception, as its current output is influenced by all preceding inputs. When $\mathbf{A}$, $\mathbf{B}$, and $\mathbf{C}$ are constant, this equation characterizes a linear time-invariant (LTI) system, as exemplified in LSSL \cite{gu2021combining} and S4 \cite{gu2021efficiently}. Conversely, when these parameters change over time, the equation describes a linear time-varying (LTV) system, which is the case in Mamba \cite{gu2023mamba}. LTI systems inherently lack the ability to perceive input content, whereas input-aware LTV systems are designed to possess this capability. 

\subsubsection{Discretization} 
When employing the SSM in the field of DL, discretization is required. To facilitate this process, a timescale parameter, denoted as $\mathbf{\Delta}\in \mathbb{R}$, is introduced to convert the continuous parameters $\mathbf{A}$ and $\mathbf{B}$ into their discrete counterparts, represented as $\mathbf{\overline{A}}$ and $\mathbf{\overline{B}}$. Using the zero-order hold (ZOH) method as the transformation algorithm, the discrete parameters are calculated as follows:
\begin{equation}\label{E2}
	\begin{aligned}
		\mathbf{\overline{A}}&=\rm{exp}(\mathbf{\Delta}\mathbf{A}), \\
		\mathbf{\overline{B}}&={(\mathbf{\Delta}\mathbf{A})}^{-1}(\rm{exp}(\mathbf{\Delta}\mathbf{A})-{\mathbf{I}})\cdot \mathbf{\Delta}\mathbf{B}\approx\mathbf{\Delta}\mathbf{B}.
	\end{aligned}
\end{equation} 
Then, the discrete form of Eq.~\ref{E1} can be expressed as:
\begin{equation}\label{E3}
	\begin{aligned}
		h_t&=\mathbf{\overline{A}}h_{t-1}+\mathbf{\overline{B}}x_{t}, \\
		y_t&=\mathbf{C}h_t.
	\end{aligned}
\end{equation} 
In practice, $x_t$ is a feature vector with ${C}$ components, and Eq.~\ref{E3} processes each of these components independently.

\subsubsection{Selective Scan} 
In Mamba, the variability of parameters with the input prevents the reformulation of Eq.~\ref{E3} into a convolutional form, thereby impeding the parallelization of the SSM. To overcome this obstacle, Mamba introduces the selective scan mechanism, which incorporates three hardware-related techniques: kernel fusion, parallel scan, and recomputation. This selective scan enables Mamba to achieve impressive speed while maintaining a relatively low memory requirement.

\subsection{Network Architecture}
\label{s33}
To fully exploit the potential of the SSM in remote sensing image fusion, we design an interpretable network architecture, which consists of two U-shaped network branches (the spatial branch and the spectral branch) for feature extraction, a combination branch for information integration, and an MCA module for spectral enhancement, as shown in Fig.~\ref{pipeline}. Next, we will provide a detailed explanation of these network components.

\begin{figure*}[t]
	\begin{center}
		\begin{minipage}{1\linewidth}
			{\includegraphics[width=1\linewidth]{Fig/mamba3.pdf}}
			\centering
		\end{minipage}
	\end{center}
	\caption{The schematic diagram of the bidirectional Mamba block (first from the left), the four-directional Mamba block (second from the left), and the proposed FusionMamba block (second from the right), along with an illustration depicting the four flattening directions (first from the right). FSSM stands for the fusion state space model. Additionally, the specifics of the SSM and FSSM blocks are detailed in Algorithms \ref{ssmblock} and \ref{fssmblock}, respectively. \label{mamba}}
\end{figure*}

\begin{algorithm}[t]
	\small
	\caption{SSM Block}
	\label{ssmblock}
	\renewcommand{\algorithmicrequire}{\textbf{Input:}}
	\renewcommand{\algorithmicensure}{\textbf{Output:}}
	\begin{algorithmic}[1]
		\REQUIRE $\mathbf{x}:\textcolor[RGB]{0, 100, 0}{(HW, C)}$
		\ENSURE $\mathbf{y}:\textcolor[RGB]{0, 100, 0}{(HW, C)}$
		\STATE  $\mathbf{A}:\textcolor[RGB]{0, 100, 0}{(C, N)}\gets \mathbf{Parameter}_{\rm{A}}$ \\
		\textcolor{gray}{\text{/* $\mathbf{A}$ represents $\textcolor[RGB]{0, 100, 0}{C}$ sets of structured $\textcolor[RGB]{0, 100, 0}{N\times N}$ matrices \cite{gu2021efficiently} */}}
		\STATE  $\mathbf{B}:\textcolor[RGB]{0, 100, 0}{(HW, N)}\gets \mathbf{Linear}_{\rm{B}}(\mathbf{x})$ 
		\STATE  $\mathbf{C}:\textcolor[RGB]{0, 100, 0}{(HW, N)}\gets \mathbf{Linear}_{\rm{C}}(\mathbf{x})$ 
		\STATE  $\mathbf{\Delta}:\textcolor[RGB]{0, 100, 0}{(HW, C)}\gets {\rm{log}}(1+{\rm{exp}}(\mathbf{Linear}_{\rm{\Delta}}(\mathbf{x})+\mathbf{Parameter}_{\rm{\Delta}}))$ \\
		\textcolor{gray}{\text{/* $\mathbf{Parameter}_{\rm{\Delta}}$ is a bias vector with a size of $\textcolor[RGB]{0, 100, 0}{C}$ */}}
		\STATE $\mathbf{\overline{A}}:\textcolor[RGB]{0, 100, 0}{(HW, C, N)}\gets {\rm{exp}}(\mathbf{\Delta}\otimes \mathbf{A})$ 
		\STATE $\mathbf{\overline{B}}:\textcolor[RGB]{0, 100, 0}{(HW, C, N)}\gets \mathbf{\Delta}\otimes \mathbf{B}$ 
		\STATE  $\mathbf{y} \gets {\rm{SSM}}(\mathbf{\overline{A}}, \mathbf{\overline{B}}, \mathbf{C})(\mathbf{x})$ \\
		\textcolor{gray}{\text{/* $\rm{SSM}$ represents Eq.~\ref{E3} implemented using selective scan */}}
		\RETURN  $\mathbf{y}$
	\end{algorithmic}
	
\end{algorithm}

\begin{algorithm}[t]
	\small
	\caption{FSSM Block}
	\label{fssmblock}
	\renewcommand{\algorithmicrequire}{\textbf{Inputs:}}
	\renewcommand{\algorithmicensure}{\textbf{Output:}}
	\begin{algorithmic}[1]
		\REQUIRE $\mathbf{x}^{\rm{a}}, \mathbf{x}^{\rm{b}}:\textcolor[RGB]{0, 100, 0}{(HW, C)}$
		\ENSURE $\mathbf{y}^{\rm{a}}:\textcolor[RGB]{0, 100, 0}{(HW, C)}$
		\STATE  $\mathbf{A}:\textcolor[RGB]{0, 100, 0}{(C, N)}\gets \mathbf{Parameter}_{\rm{A}}$ \\
		\textcolor{gray}{\text{/* $\mathbf{A}$ represents $\textcolor[RGB]{0, 100, 0}{C}$ sets of structured $\textcolor[RGB]{0, 100, 0}{N\times N}$ matrices \cite{gu2021efficiently} */}}
		\STATE  $\mathbf{B}:\textcolor[RGB]{0, 100, 0}{(HW, N)}\gets \mathbf{Linear}_{\rm{B}}(\mathbf{x}^{\rm{b}})$ 
		\STATE  $\mathbf{C}:\textcolor[RGB]{0, 100, 0}{(HW, N)}\gets \mathbf{Linear}_{\rm{C}}(\mathbf{x}^{\rm{b}})$ 
		\STATE  $\mathbf{\Delta}:\textcolor[RGB]{0, 100, 0}{(HW, C)}\gets {\rm{log}}(1+{\rm{exp}}(\mathbf{Linear}_{\rm{\Delta}}(\mathbf{x}^{\rm{b}})+\mathbf{Parameter}_{\rm{\Delta}}))$ \\
		\textcolor{gray}{\text{/* $\mathbf{Parameter}_{\rm{\Delta}}$ is a bias vector with a size of $\textcolor[RGB]{0, 100, 0}{C}$ */}}
		\STATE $\mathbf{\overline{A}}:\textcolor[RGB]{0, 100, 0}{(HW, C, N)}\gets {\rm{exp}}(\mathbf{\Delta}\otimes \mathbf{A})$ 
		\STATE $\mathbf{\overline{B}}:\textcolor[RGB]{0, 100, 0}{(HW, C, N)}\gets \mathbf{\Delta}\otimes \mathbf{B}$ 
		\STATE  $\mathbf{y}^{\rm{a}} \gets {\rm{SSM}}(\mathbf{\overline{A}}, \mathbf{\overline{B}}, \mathbf{C})(\mathbf{x}^{\rm{a}})$ \\
		\textcolor{gray}{\text{/* $\rm{SSM}$ represents Eq.~\ref{E3} implemented using selective scan */}}
		\RETURN  $\mathbf{y}^{\rm{a}}$
	\end{algorithmic}
	
\end{algorithm}

\subsubsection{U-shaped Network Branches} 
This design facilitates the efficient learning of spatial and spectral information in a separate and hierarchical manner. Specifically, the spatial branch focuses on extracting spatial details from $\mathbf{P}$, while the spectral branch is dedicated to capturing spectral characteristics from $\mathbf{M}$. To acquire sufficient deep-level information without significantly increasing network parameters, we extract features at three different scales. This means that each U-shaped network branch comprises a total of five stages. At each stage, the spatial or spectral feature map is initially processed by a four-directional Mamba block. The resulting feature map then passes through a FusionMamba block to generate a fusion output, which is subsequently added back to the original input. Finally, a convolution layer of varying types is employed to adjust both the spatial resolution and the number of channels.

\subsubsection{Combination Branch} 
This design enables the comprehensive integration of spatial and spectral information. To align with the U-shaped network branches, it incorporates five FusionMamba blocks. Each block receives its corresponding spatial and spectral feature maps as inputs, generating a fusion output that is subsequently added back to the original inputs. From a holistic perspective, the combination branch effectively simulates the progressive merging of different features.

\subsubsection{Mamba-driven Channel Attention} 
The MCA module is designed to improve the representation of spectral information. Based on the widely used channel attention mechanism \cite{Hu_2018_CVPR}, the MCA module replaces the MLP with a bidirectional Mamba block. Additionally, several modifications are made to better accommodate the characteristics of the SSM in data processing. Specifically, we first utilize global max pooling to eliminate spatial information from $\mathbf{M}^{\rm{U}}$, resulting in a $1\times 1\times S$ feature map. This map is then reshaped into a 1D sequence of size $S\times 1$. Next, we employ a fully connected layer to increase the channel number of this sequence to $C$. The augmented 1D sequence is subsequently passed through a bidirectional Mamba block to extract spectral features. Finally, the output is projected and reshaped back into a $1\times 1\times S$ feature map, which is multiplied with $\mathbf{F}^{\rm{c}}_5$ to complete the spectral enhancement.

\subsection{Mamba and FusionMamba Blocks}
\label{s34}
In this section, we detail the bidirectional Mamba block, the four-directional Mamba block, and the proposed FusionMamba block, all of which are depicted in Fig.~\ref{mamba}. Additionally, we compare the FLOPs required by different DL models.

\subsubsection{Bidirectional Mamba Block} 
For an input 1D sequence $\mathbf{x}_{\rm{in}}\in \mathbb{R}^{S\times C}$, we first normalize it using layer normalization. Next, it is processed by two parallel fully connected layers, producing two distinct sequences, denoted as $\mathbf{x}\in\mathbb{R}^{S\times C}$ and $\mathbf{z}\in\mathbb{R}^{S\times C}$. This procedure can be expressed as follows:
\begin{equation}\label{E4}
	\begin{aligned}
		\mathbf{x}, \mathbf{z} = \mathbf{Linear}_{\rm{x}}(\mathbf{Norm}(\mathbf{x}_{\rm{in}})), \mathbf{Linear}_{\rm{z}}(\mathbf{Norm}(\mathbf{x}_{\rm{in}})).
	\end{aligned}
\end{equation} 
Here, $\mathbf{Norm}$ denotes layer normalization, while $\mathbf{Linear}_{\rm{x}}$ and $\mathbf{Linear}_{\rm{z}}$ represent two separate fully connected layers. After that, we flip $\mathbf{x}$ vertically, generating $\mathbf{\hat{x}}$ of size $S\times C$. Subsequently, $\mathbf{x}$ and $\mathbf{\hat{x}}$ are processed separately through two SSM blocks for feature extraction, resulting in two output sequences, denoted as $\mathbf{\overline{y}}$ and $\mathbf{\hat{y}}$. This process is expressed as:
\begin{equation}\label{E5}
	\begin{aligned}
		\mathbf{\hat{x}}&={\rm{VFlip}}(\mathbf{x}), \\
		\mathbf{\overline{y}}, \mathbf{\hat{y}}&={\mathbf{SSM}}_1(\mathbf{x}), {\mathbf{SSM}}_2(\mathbf{\hat{x}}).
	\end{aligned}
\end{equation}
Here, ${\rm{VFlip}}$ refers to the operation of flipping a matrix vertically. ${\mathbf{SSM}}_1$ and ${\mathbf{SSM}}_2$ represent two separate SSM blocks, which are thoroughly detailed in Algorithm~\ref{ssmblock}. Next, we vertically flip $\mathbf{\hat{y}}$ and add it to $\mathbf{\overline{y}}$, producing a new sequence denoted as $\mathbf{y}\in\mathbb{R}^{S\times C}$. After gating by $\mathbf{z}$, this sequence undergoes a fully connected layer and is added to $\mathbf{x}_{\rm{in}}$, resulting in the final output represented as $\mathbf{x}_{\rm{out}}\in \mathbb{R}^{S\times C}$:
\begin{equation}\label{E6}
	\begin{aligned}
		\mathbf{y}&=\mathbf{\overline{y}} + {\rm{VFlip}}(\mathbf{\hat{y}}), \\
		\mathbf{x}_{\rm{out}}&={\mathbf{Linear}}_{\rm{o}}(\mathbf{y}\cdot {\rm{SiLU}}(\mathbf{z}))+\mathbf{x}_{\rm{in}}.
	\end{aligned}
\end{equation} 
Here, ${\mathbf{Linear}}_{\rm{o}}$ denotes the fully connected layer, and ${\rm{SiLU}}$ stands for the ``SiLU" activation function.

\subsubsection{Four-directional Mamba Block} 
Given an input feature map $\mathbf{F}_{\rm{in}}\in \mathbb{R}^{H\times W\times C}$, we normalize it using layer normalization and process it via two parallel $1\times 1$ convolution layers, yielding two distinct feature maps, denoted as $\mathbf{X}\in\mathbb{R}^{H\times W\times C}$ and $\mathbf{Z}\in\mathbb{R}^{H\times W\times C}$. This process can be expressed as follows:
\begin{equation}\label{E7}
	\begin{aligned}
		\mathbf{X}, \mathbf{Z} = \mathbf{Conv}_{\rm{x}}(\mathbf{Norm}(\mathbf{F}_{\rm{in}})), \mathbf{Conv}_{\rm{z}}(\mathbf{Norm}(\mathbf{F}_{\rm{in}})).
	\end{aligned}
\end{equation} 
Here, $\mathbf{Conv}_{\rm{x}}$ and $\mathbf{Conv}_{\rm{z}}$ represents two separate convolution layers. Following this, $\mathbf{X}$ is flattened in four directions, producing $\mathbf{x}_1$, $\mathbf{x}_2$, $\mathbf{x}_3$, and $\mathbf{x}_4$, each with dimension of ${HW\times C}$. These sequences are then separately processed by SSM blocks, resulting in four outputs denoted as $\mathbf{y}_1$, $\mathbf{y}_2$, $\mathbf{y}_3$, and $\mathbf{y}_4$:
\begin{equation}\label{E8}
	\begin{cases}
		{\mathbf{x}_i}={\rm{Flatten}}_i(\mathbf{X}), \\
		{\mathbf{y}_i}={\mathbf{SSM}}_i(\mathbf{x}_i).
	\end{cases}
i = 1, 2, 3, 4.
\end{equation} 
Here, ${\rm{Flatten}}_i$ represents the flattening operation along the $i$-th direction. Subsequently, we unflatten the outputs of SSM blocks and combine them to obtain a new feature map, denoted as $\mathbf{Y}\in\mathbb{R}^{H\times W\times C}$. After gating by $\mathbf{Z}$, this feature map undergoes a $1\times 1$ convolution layer and is added to $\mathbf{F}_{\rm{in}}$, yielding the final output represented as $\mathbf{F}_{\rm{out}}\in\mathbb{R}^{H\times W\times C}$:
\begin{equation}\label{E9}
	\begin{aligned}
		\mathbf{Y}&=\sum_{i=1}^{4}{\rm{Unflatten}}_i(\mathbf{y}_i), \\
		\mathbf{F}_{\rm{out}}&={\mathbf{Conv}}_{\rm{o}}(\mathbf{Y}\cdot {\rm{SiLU}}(\mathbf{Z}))+\mathbf{F}_{\rm{in}}.
	\end{aligned}
\end{equation} 
Here, ${\rm{Unflatten}}_i$ denotes the operation of unflattening along the $i$-th direction and ${\mathbf{Conv}}_{\rm{o}}$ represents the convolution layer.

\begin{figure}[t]
	\begin{center}
		\begin{minipage}{1\linewidth}
			{\includegraphics[width=0.9\linewidth]{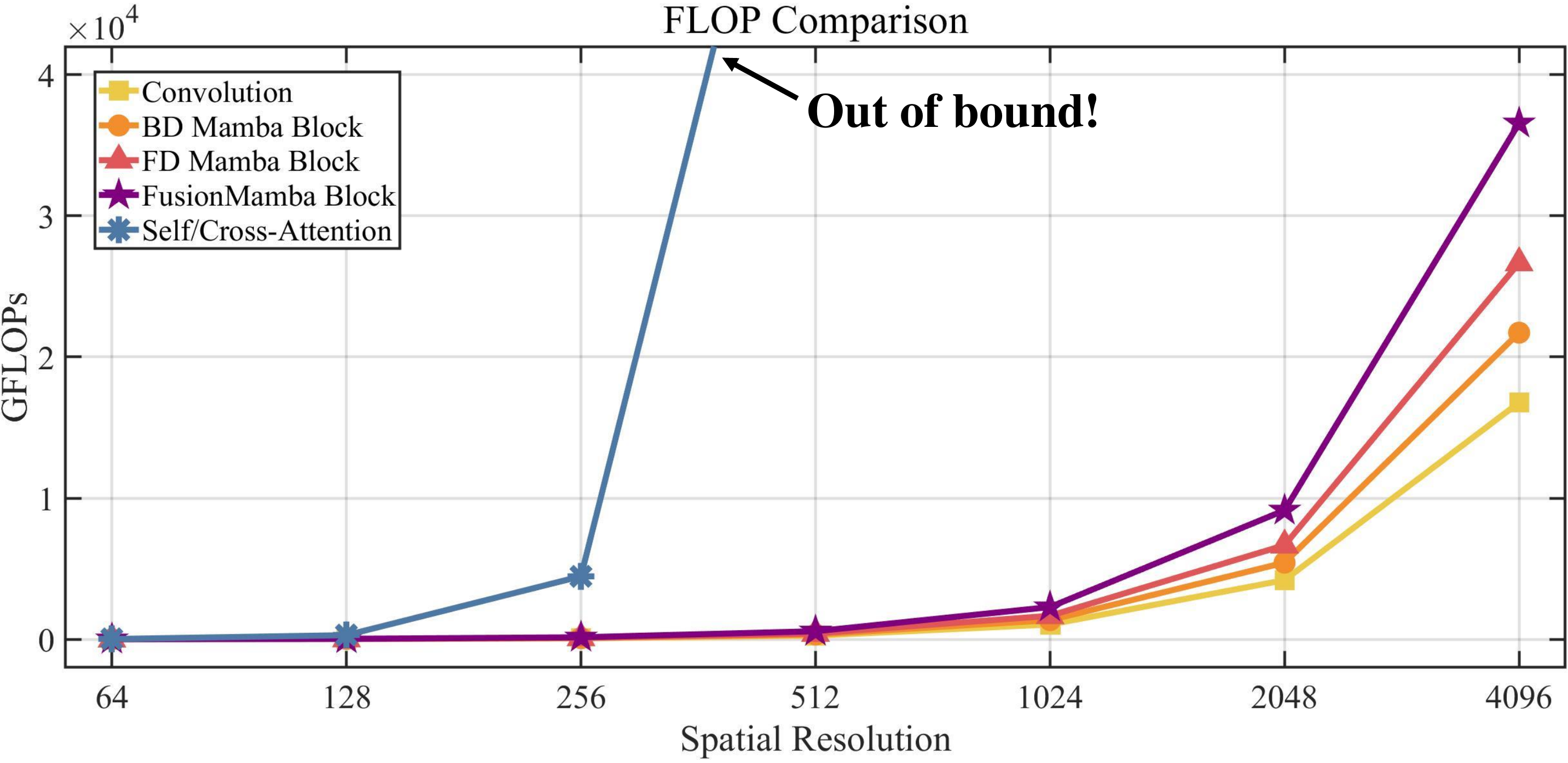}}
			\centering
		\end{minipage}
	\end{center}
	\caption{Comparison of FLOPs among the convolution layer, bidirectional (BD) Mamba block, four-directional (FD) Mamba block, FusionMamba block, and self/cross-attention module at various spatial resolutions. For optimal visual effects, we configure $D$, $C$, and $N$ to be 0.5M, 256, and 64, respectively.  \label{flopxs}}
\end{figure}

\subsubsection{FusionMamba Block} 
The original SSM can only handle a single input. To effectively integrate different types of information, we expand it to accommodate dual inputs, resulting in the fusion state space model (FSSM), as detailed in Algorithm~\ref{fssmblock}. Within the FSSM block, one input is responsible for generating the projection and timescale parameters, while the other input is the sequence to be processed. The FusionMamba block, consisting of eight FSSM blocks, is designed with a symmetrical structure. For the input spatial and spectral feature maps, denoted as $\mathbf{F}_{\rm{in}}^{\rm{a}}\in \mathbb{R}^{H\times W\times C}$ and $\mathbf{F}_{\rm{in}}^{\rm{b}}\in \mathbb{R}^{H\times W\times C}$, we employ a method similar to the four-directional Mamba block to generate two sets of feature maps as follows:
\begin{equation}\label{E10}
	\begin{aligned}
		\mathbf{X}^{\rm{a}}, \mathbf{Z}^{\rm{a}} = \mathbf{Conv}_{\rm{x}}^{\rm{a}}(\mathbf{Norm}(\mathbf{F}_{\rm{in}}^{\rm{a}})), \mathbf{Conv}_{\rm{z}}^{\rm{a}}(\mathbf{Norm}(\mathbf{F}_{\rm{in}}^{\rm{a}})); \\
		\mathbf{X}^{\rm{b}}, \mathbf{Z}^{\rm{b}} = \mathbf{Conv}_{\rm{x}}^{\rm{b}}(\mathbf{Norm}(\mathbf{F}_{\rm{in}}^{\rm{b}})), \mathbf{Conv}_{\rm{z}}^{\rm{b}}(\mathbf{Norm}(\mathbf{F}_{\rm{in}}^{\rm{b}})). \\
	\end{aligned}
\end{equation} 
Since this equation is a direct extension of Eq.~\ref{E7}, explanations for the symbols are omitted. Next, $\mathbf{X}^{\rm{a}}$ and $\mathbf{X}^{\rm{b}}$ are flattened separately in four directions. The resulting 1D sequences are then forwarded to FSSM blocks for information integration: 
\begin{equation}\label{E11}
	\begin{cases}
		{\mathbf{x}_i^{\rm{a}}},{\mathbf{x}_i^{\rm{b}}}={\rm{Flatten}}_i(\mathbf{X}^{\rm{a}}), {\rm{Flatten}}_i(\mathbf{X}^{\rm{b}}), \\
		{\mathbf{y}_i^{\rm{a}}},{\mathbf{y}_i^{\rm{b}}}={\mathbf{FSSM}}_i^{\rm{a}}(\mathbf{x}_i^{\rm{a}}, \mathbf{x}_i^{\rm{b}}), {\mathbf{FSSM}}_i^{\rm{b}}(\mathbf{x}_i^{\rm{b}}, \mathbf{x}_i^{\rm{a}}).
	\end{cases}
	i = 1, 2, 3, 4.
\end{equation} 
Here, $\mathbf{FSSM}^{\rm{a}}$ and $\mathbf{FSSM}^{\rm{b}}$ refer to the FSSM blocks on the left and right halves of the FusionMamba block in Fig.~\ref{mamba}. After that, we process the two sets of outputs separately, producing two new feature maps denoted as $\mathbf{Y}^{\rm{a}}\in\mathbb{R}^{H\times W\times C}$ and $\mathbf{Y}^{\rm{b}}\in\mathbb{R}^{H\times W\times C}$. These maps are finally combined to form $\mathbf{F}_{\rm{out}}$:
\begin{equation}\label{E12}
	\begin{aligned}
		\mathbf{Y}^{\rm{a}}, \mathbf{Y}^{\rm{b}}&=\sum_{i=1}^{4}{\rm{Unflatten}}_i(\mathbf{y}_i^{\rm{a}}),\sum_{i=1}^{4}{\rm{Unflatten}}_i(\mathbf{y}_i^{\rm{b}}), \\
		\mathbf{F}_{\rm{out}}^{\rm{a}}&={\mathbf{Conv}}_{\rm{o}}^{\rm{a}}(\mathbf{Y}^{\rm{a}}\cdot {\rm{SiLU}}(\mathbf{Z}^{\rm{a}}))+\mathbf{F}_{\rm{in}}^{\rm{a}}, \\
		\mathbf{F}_{\rm{out}}^{\rm{b}}&={\mathbf{Conv}}_{\rm{o}}^{\rm{b}}(\mathbf{Y}^{\rm{b}}\cdot {\rm{SiLU}}(\mathbf{Z}^{\rm{b}}))+\mathbf{F}_{\rm{in}}^{\rm{b}}, \\
		\mathbf{F}_{\rm{out}}&=\mathbf{Conv}_{\rm{o}}(\mathbf{F}_{\rm{out}}^{\rm{a}} + \mathbf{F}_{\rm{out}}^{\rm{b}}).
	\end{aligned}
\end{equation} 
Here, ${\mathbf{Conv}}_{\rm{o}}^{\rm{a}}$, ${\mathbf{Conv}}_{\rm{o}}^{\rm{b}}$, and ${\mathbf{Conv}}_{\rm{o}}$ represent $1\times 1$ convolution layers that generate $\mathbf{F}_{\rm{out}}^{\rm{a}}$, $\mathbf{F}_{\rm{out}}^{\rm{b}}$, and $\mathbf{F}_{\rm{out}}$, respectively.

\subsubsection{Analysis of FLOPs} 
\label{flopsexp}
In a convolution layer with $D$ parameters, the FLOP count is commonly calculated as $2HWD$. Given that a selective scan costs $9HWCN$ FLOPs \cite{gu2023mamba}, the overall FLOP counts for a bidirectional Mamba block, a four-directional block, and a FusionMamba block, each with $D$ parameters, are $2HWD+18HWCN$, $2HWD+36HWCN$, and $2HWD+72HWCN$, respectively. As for a self/cross-attention block in Transformers, the total FLOP count is estimated to be around $2HWD+4H^2W^2C$. The FLOP comparison among these modules, as depicted in Fig.~\ref{flopxs}, indicates that the Mamba and FusionMamba blocks possess FLOP costs comparable to that of the convolution layer and are significantly more efficient than the self/cross-attention block. 

\subsection{Loss Function}
\label{s35}
The main contributions of this study lie in the application and innovation of the SSM. Therefore, we employ the simplest $\ell_{1}$ loss function for network training, as shown below:
\begin{equation}\label{loss}
	\begin{aligned}
		\mathcal{L}oss =\frac{1}{T}\sum_{i=1}^{T}{\| f_{\mathbf{\Theta}}(\mathbf{P}_i,\mathbf{M}_i) - \mathbf{O}_i\|}_{1}.
	\end{aligned}
\end{equation} 
Here, ${T}$ denotes the total number of training examples, and $f_{\mathbf{\Theta}}$ represents our network with learnable parameters $\mathbf{\Theta}$. Additionally, $\mathbf{P}_i$, $\mathbf{M}_i$, and $\mathbf{O}_i$ refer to the $i$-th PAN image, LRMS/LRHS image, and GT image in the training dataset, respectively. Furthermore, $\|\cdot\|_1$ defines the $\ell_{1}$ normalization.

\section{Experiments}
\label{s4}
In this section, we present the quantitative and qualitative evaluation results for representative remote sensing image fusion approaches on the pansharpening and hyper-spectral pansharpening tasks. Additionally, we conduct comprehensive ablation studies to demonstrate the superiority of our method.

\subsection{Pansharpening}

\begin{table*}[t]	
	\centering\renewcommand\arraystretch{1.2}\setlength{\tabcolsep}{5.4pt}
	\belowrulesep=0pt\aboverulesep=0pt
	\caption{Quantitative evaluation results on 20 reduced-resolution and 20 full-resolution samples from the WV3 dataset, which belongs to the pansharpening task. The best results are highlighted in \textbf{bold}, and the second-best results are \underline{underlined}. Additionally, the methods above the dividing line represent traditional approaches, while the methods below it are DL-based techniques.}
	\label{rr}	
	\begin{tabular}{l|c|cccc|ccc}
		\toprule
		\multirow{2}{*}{\textbf{Methods}} & 
		\multirow{2}{*}{\textbf{Params}} & 
		\multicolumn{4}{c|}{\textbf{Reduced-Resolution}} & \multicolumn{3}{c}{\textbf{Full-Resolution (Real Data)}}\\
		\cmidrule(lr){3-6}\cmidrule(lr){7-9}
		&\multicolumn{1}{c|}{} 
		&\multicolumn{1}{c}{PSNR($\pm$std)} 
		&\multicolumn{1}{c}{Q2n($\pm$std)} 
		&\multicolumn{1}{c}{SAM($\pm$std)} 
		&\multicolumn{1}{c|}{ERGAS($\pm$std)} 
		&\multicolumn{1}{c}{${{\rm{D}}_{\rm{\lambda}}}$($\pm$std)} 
		&\multicolumn{1}{c}{${{\rm{D}}_{\rm{s}}}$($\pm$std)} 
		&\multicolumn{1}{c}{QNR($\pm$std)} \\
		\midrule
		\textbf{TV} \cite{palsson2013new} & $-$ & 32.381$\pm$2.328 & 0.795$\pm$0.120 & 5.692$\pm$1.808 & 4.855$\pm$1.434
		& 0.0234$\pm$0.0061  & 0.0393$\pm$0.0227 & 0.9383$\pm$0.0269\\ 
		\textbf{GLP-HPM} \cite{6616569} & $-$ & 33.095$\pm$2.800 & 0.835$\pm$0.092 & 5.333$\pm$1.761 & 4.616$\pm$1.503
		 & 0.0206$\pm$0.0082  & 0.0630$\pm$0.0284 & 0.9180$\pm$0.0346\\ 
		\textbf{GLP-FS} \cite{vivone2018full}& $-$  & 32.963$\pm$2.753 & 0.833$\pm$0.092 & 5.315$\pm$1.765 & 4.700$\pm$1.597
		& 0.0197$\pm$0.0078  & 0.0630$\pm$0.0289 & 0.9187$\pm$0.0347\\ 
		\textbf{BDSD-PC} \cite{2019Robust} & $-$ & 32.970$\pm$2.784 & 0.829$\pm$0.097 & 5.428$\pm$1.822 & 4.697$\pm$1.617
		& 0.0625$\pm$0.0235  & 0.0730$\pm$0.0356 & 0.8698$\pm$0.0531\\ 
		\midrule
		\textbf{PanNet} \cite{8237455} & 0.08M & 37.346$\pm$2.688 & 0.891$\pm$0.093 & 3.613$\pm$0.766 & 2.664$\pm$0.688 & \textbf{0.0165}$\pm$0.0074  & 0.0470$\pm$0.0210 & 0.9374$\pm$0.0271\\ 
		\textbf{MSDCNN} \cite{8127731} & 0.23M & 37.068$\pm$2.686 & 0.890$\pm$0.090 & 3.777$\pm$0.803 & 2.760$\pm$0.689
		& 0.0230$\pm$0.0091 & 0.0467$\pm$0.0199 & 0.9316$\pm$0.0271\\
		\textbf{BDPN} \cite{8667448} & 1.49M & 36.191$\pm$2.702 & 0.871$\pm$0.100 & 4.201$\pm$0.857 & 3.046$\pm$0.732
		& 0.0364$\pm$0.0142  & 0.0459$\pm$0.0192 & 0.9196$\pm$0.0308\\ 
		\textbf{FusionNet} \cite{2020Detail} & 0.08M & 38.047$\pm$2.589 & 0.904$\pm$0.090 & 3.324$\pm$0.698 & 2.465$\pm$0.644
		& 0.0239$\pm$0.0090  & 0.0364$\pm$0.0137 & 0.9406$\pm$0.0197\\   
		\textbf{MUCNN} \cite{10.1145/3474085.3475600} & 2.32M & 38.262$\pm$2.703 & 0.911$\pm$0.089 & 3.206$\pm$0.681 & 2.400$\pm$0.617 
		& 0.0258$\pm$0.0111  & {0.0327}$\pm$0.0140 & {0.9424}$\pm$0.0205\\   
		\textbf{LAGNet} \cite{jin2022aaai}& 0.15M & 38.592$\pm$2.778 & 0.910$\pm$0.091 & 3.103$\pm$0.558 & {2.292}$\pm$0.607
		& 0.0368$\pm$0.0148  & 0.0418$\pm$0.0152& 0.9230$\pm$0.0247\\  
		\textbf{PMACNet} \cite{9764690} & 0.94M & {38.595}$\pm$2.882 & {0.912}$\pm$0.092 & {3.073}$\pm$0.623 & 2.293$\pm$0.532
		& 0.0540$\pm$0.0232  & 0.0336$\pm$0.0115 & 0.9143$\pm$0.0281\\  	
		\textbf{U2Net} \cite{10.1145/3581783.3612084}& 0.66M & \underline{39.117}$\pm$3.009 & \underline{0.920}$\pm$0.085 & \underline{2.888}$\pm$0.581 & \underline{2.149}$\pm$0.525
		& \underline{0.0178}$\pm$0.0072  & {0.0313}$\pm$0.0075 & {0.9514}$\pm$0.0115\\   
		\textbf{Pan-Mamba} \cite{he2024pan} & 0.48M & {39.012}$\pm$2.986 & \underline{0.920}$\pm$0.085 & {2.914}$\pm$0.592 & 2.184$\pm$0.521
		& 0.0183$\pm$0.0071  & 0.0307$\pm$0.0108 & \underline{0.9516}$\pm$0.0146\\ 
		\textbf{CANNet} \cite{Duan_2024_CVPR} & 0.78M & {39.003}$\pm$2.900 & {0.919}$\pm$0.084 & {2.941}$\pm$0.590 & 2.175$\pm$0.530
		& 0.0196$\pm$0.0083  & \underline{0.0301}$\pm$0.0074 & 0.9510$\pm$0.0126\\ 
		\textbf{FusionMamba} & 0.73M & \textbf{39.374}$\pm$2.973 & \textbf{0.922}$\pm$0.084 & \textbf{2.843}$\pm$0.577 & \textbf{2.092}$\pm$0.510 
		& 0.0186$\pm$0.0078  & \textbf{0.0269}$\pm$0.0058 & \textbf{0.9550}$\pm$0.0110\\      
		\midrule
		\textbf{Ideal Values} 
		& $-$
		&\multicolumn{1}{c}{\textbf{+$\infty$}}
		&\multicolumn{1}{c}{\textbf{\textbf{1}}}
		&\multicolumn{1}{c}{\textbf{\textbf{0}}}
		&\multicolumn{1}{c|}{\textbf{\textbf{0}}}
		&\multicolumn{1}{c}{\textbf{\textbf{0}}}
		&\multicolumn{1}{c}{\textbf{\textbf{0}}}
		&\multicolumn{1}{c}{\textbf{\textbf{1}}}
		\\ 
		\bottomrule
	\end{tabular}
\end{table*}

\begin{figure*}[t]
	\begin{center}
		\begin{minipage}[t]{1\linewidth}
			\begin{minipage}[t]{0.12\linewidth}
				{\includegraphics[width=1\linewidth]{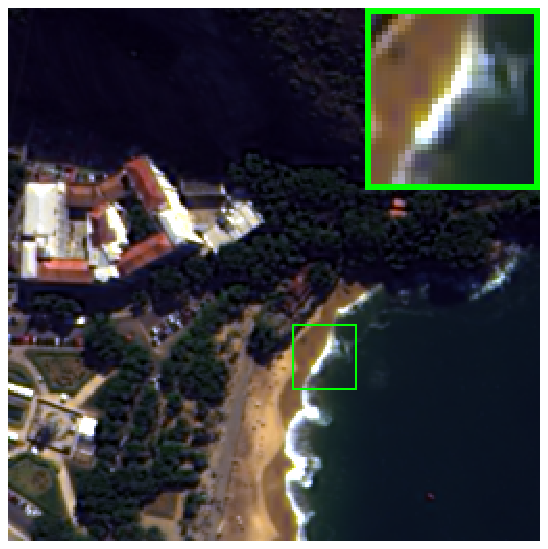}}
				{\includegraphics[width=1\linewidth]{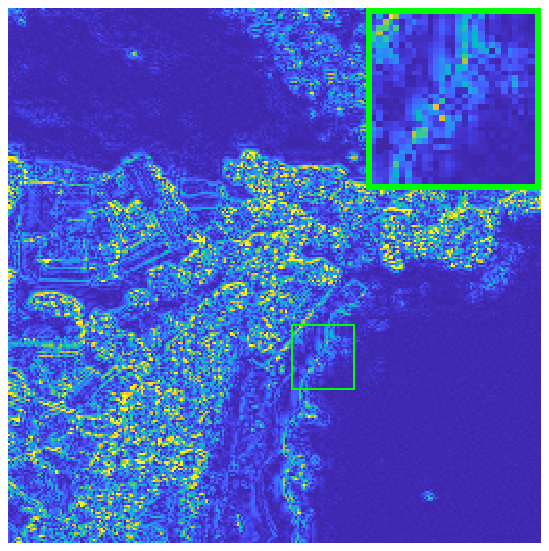}}
				\vspace{4pt}
				{TV}
				{\includegraphics[width=1\linewidth]{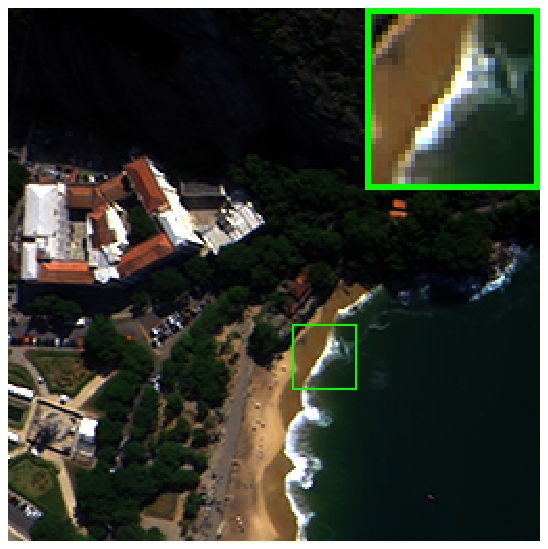}}
				{\includegraphics[width=1\linewidth]{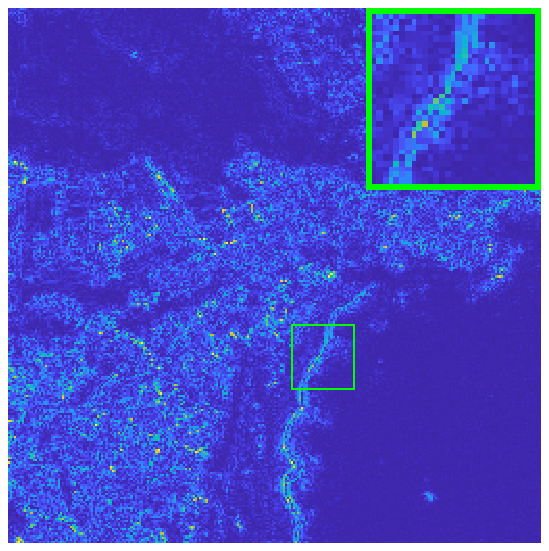}}
				\vspace{4pt}
				{MUCNN}
				\centering
				
			\end{minipage}
			\begin{minipage}[t]{0.12\linewidth}
				{\includegraphics[width=1\linewidth]{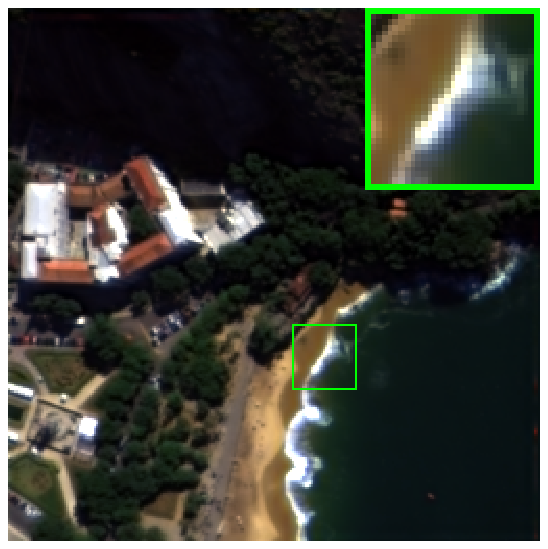}}
				{\includegraphics[width=1\linewidth]{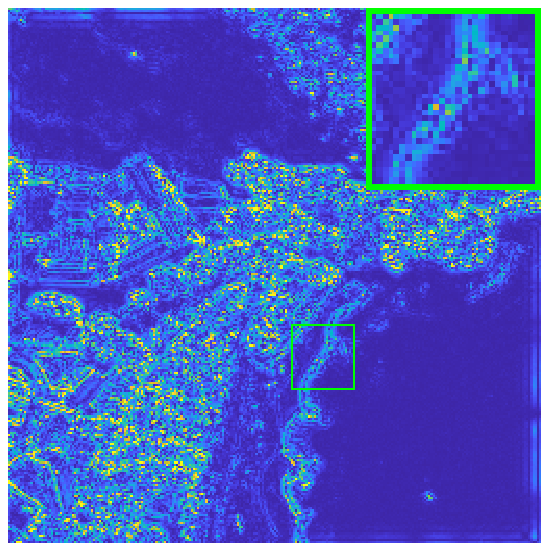}}
				\vspace{4pt}
				{GLP-HPM}
				{\includegraphics[width=1\linewidth]{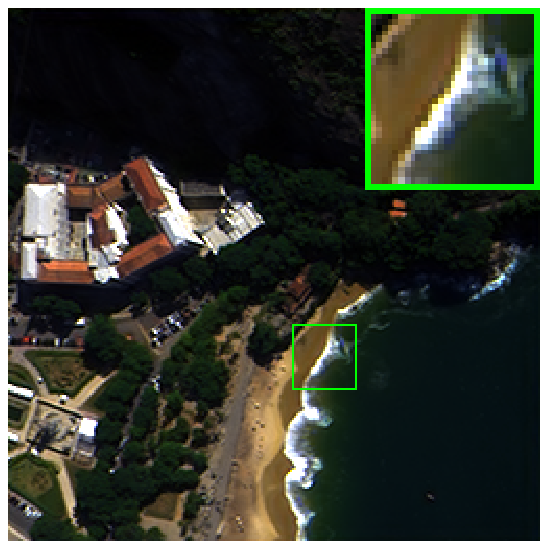}}
				{\includegraphics[width=1\linewidth]{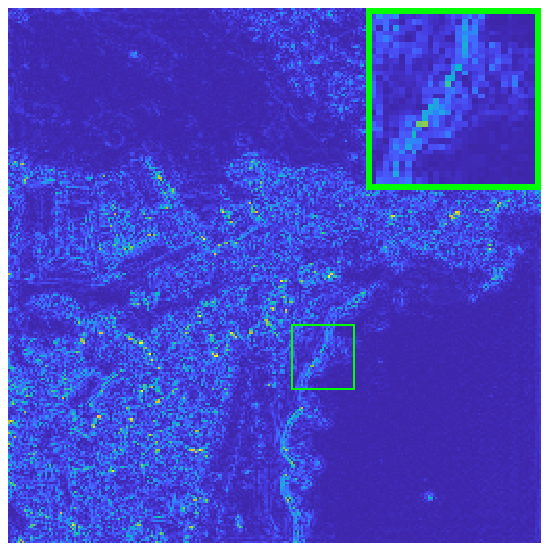}}
				\vspace{4pt}
				{LAGNet}
				\centering
				
			\end{minipage}
			\begin{minipage}[t]{0.12\linewidth}
				{\includegraphics[width=1\linewidth]{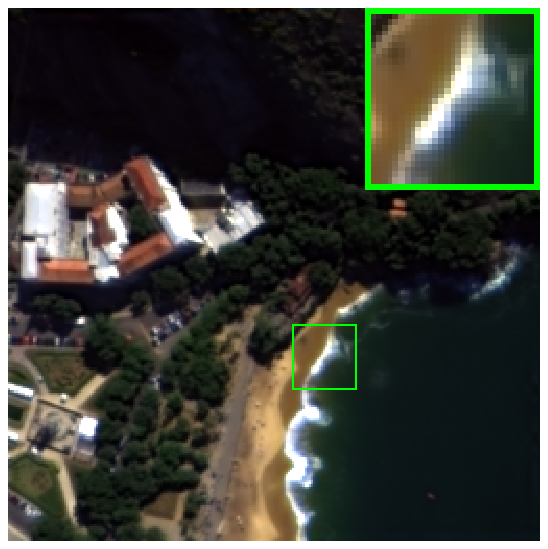}}
				{\includegraphics[width=1\linewidth]{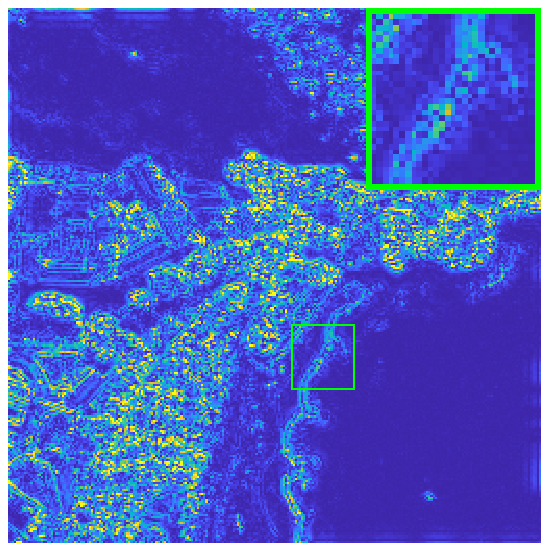}}
				\vspace{4pt}
				{GLP-FS}
				{\includegraphics[width=1\linewidth]{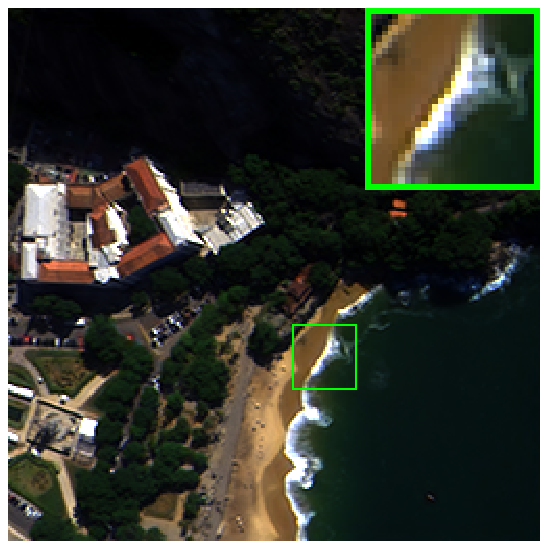}}
				{\includegraphics[width=1\linewidth]{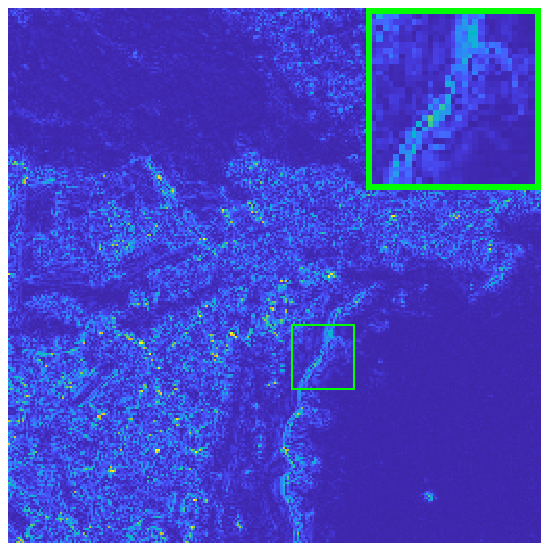}}
				\vspace{4pt}
				{PMACNet}
				\centering
				
			\end{minipage}
			\begin{minipage}[t]{0.12\linewidth}
				{\includegraphics[width=1\linewidth]{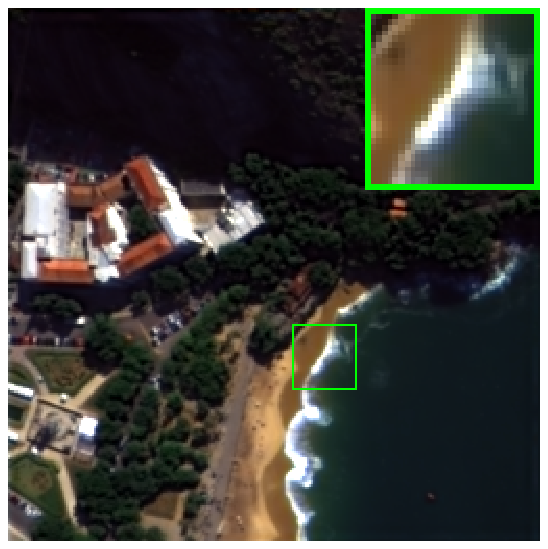}}
				{\includegraphics[width=1\linewidth]{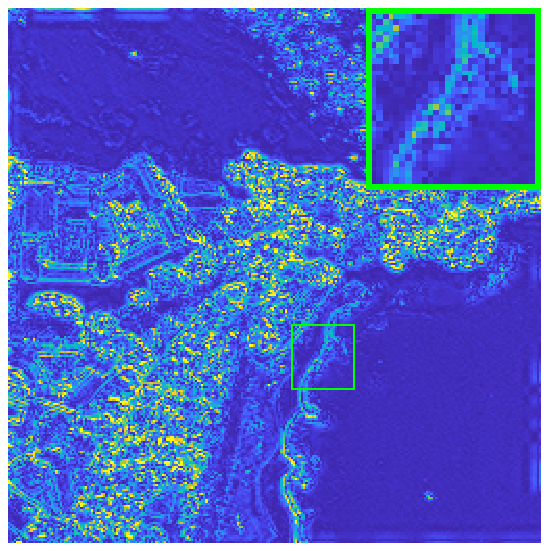}}
				\vspace{4pt}
				{BDSD-PC}
				{\includegraphics[width=1\linewidth]{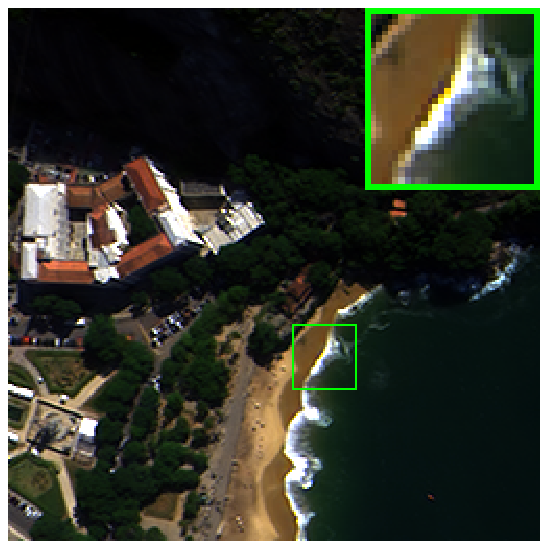}}
				{\includegraphics[width=1\linewidth]{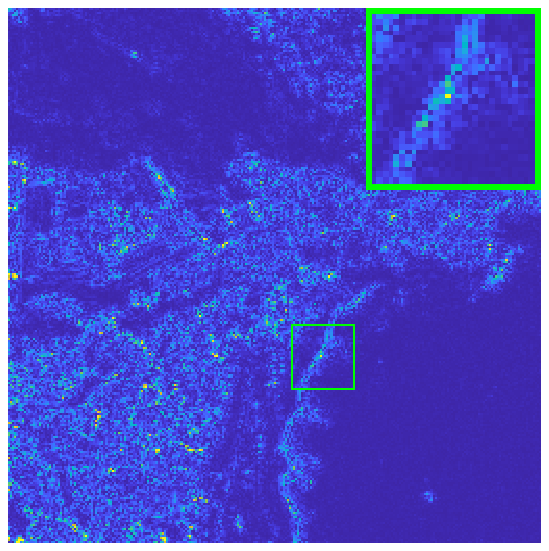}}
				\vspace{4pt}
				{U2Net}
				\centering
				
			\end{minipage}
			\begin{minipage}[t]{0.12\linewidth}
				{\includegraphics[width=1\linewidth]{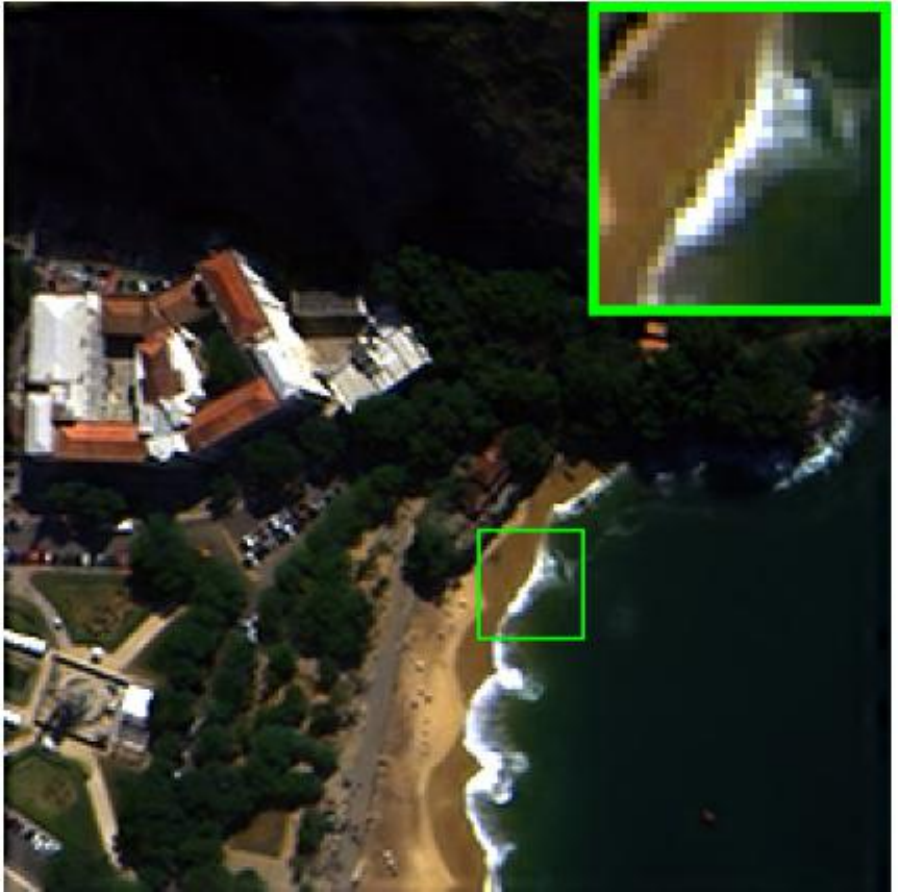}}
				{\includegraphics[width=1\linewidth]{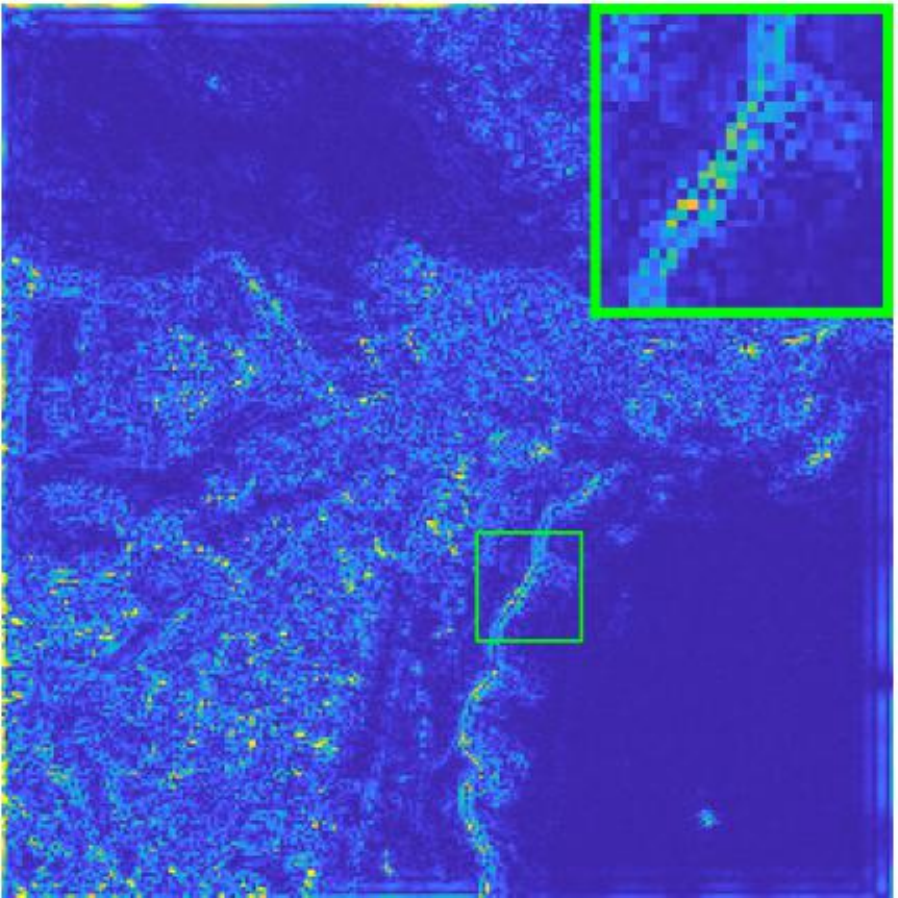}}
				\vspace{4pt}
				{PanNet}
				{\includegraphics[width=1\linewidth]{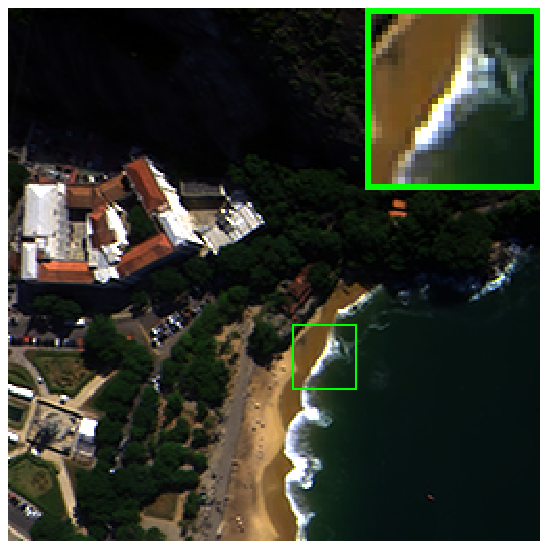}}
				{\includegraphics[width=1\linewidth]{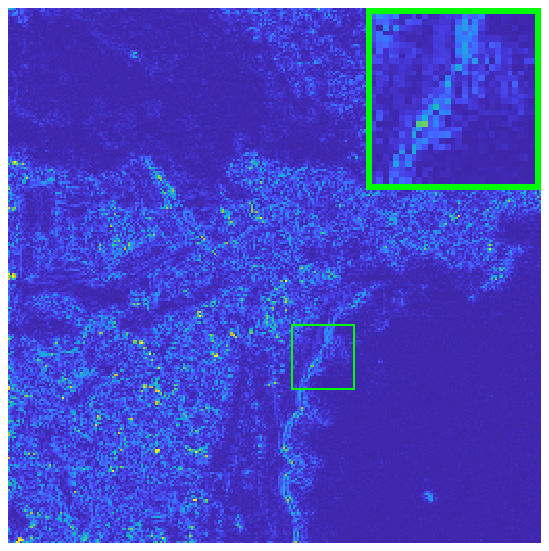}}
				\vspace{4pt}
				{Pan-Mamba}
				\centering
				
			\end{minipage}
			\begin{minipage}[t]{0.12\linewidth}
				{\includegraphics[width=1\linewidth]{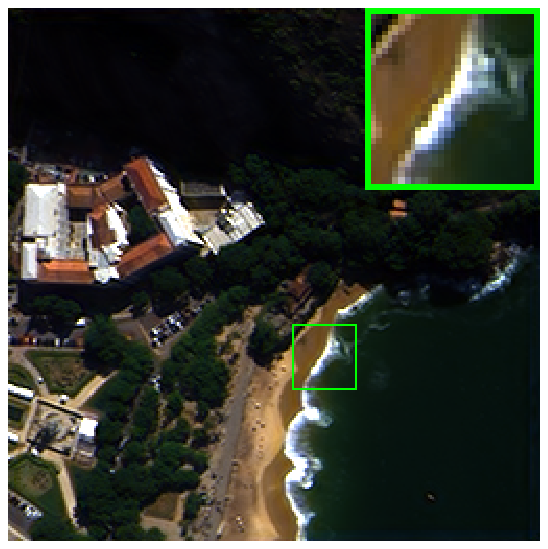}}
				{\includegraphics[width=1\linewidth]{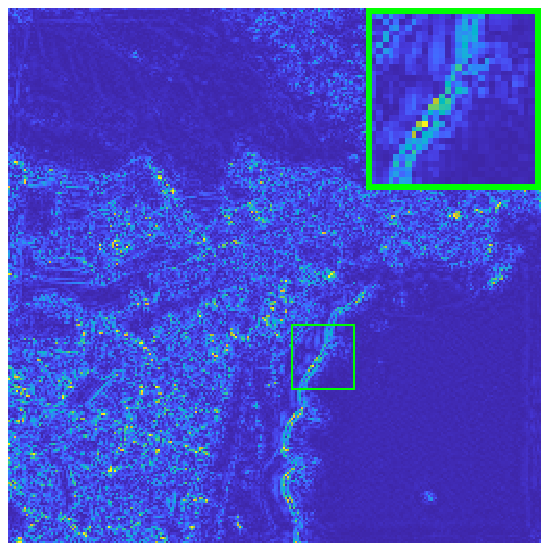}}
				\vspace{4pt}
				{MSDCNN}
				{\includegraphics[width=1\linewidth]{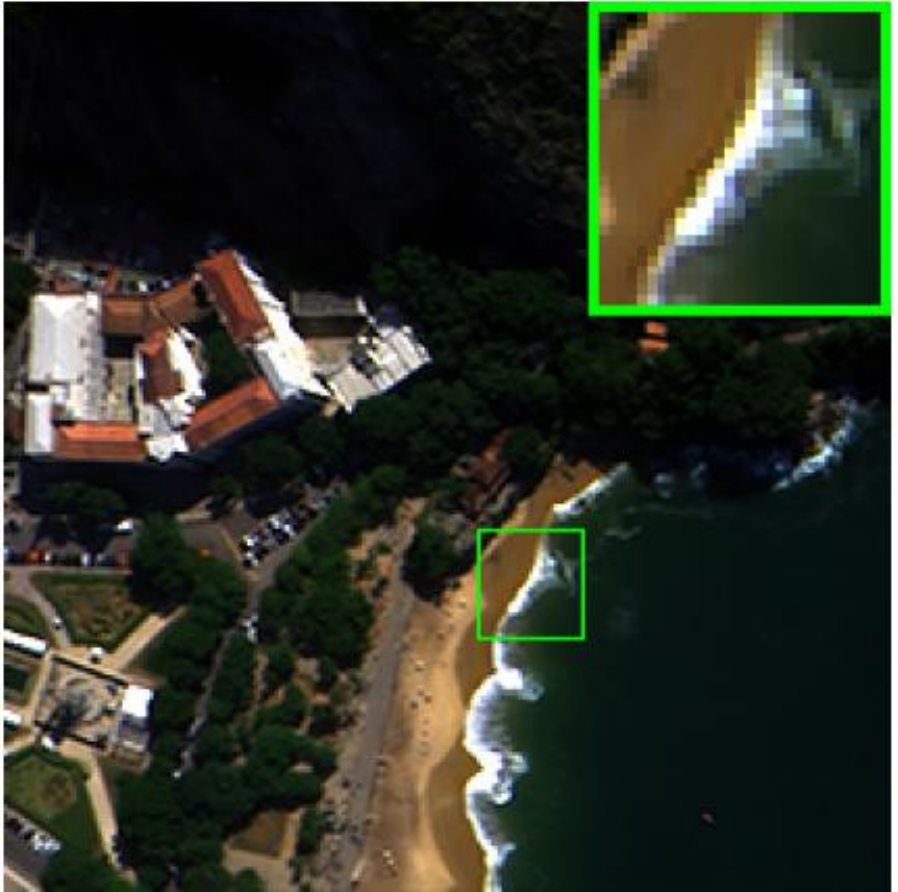}}
				{\includegraphics[width=1\linewidth]{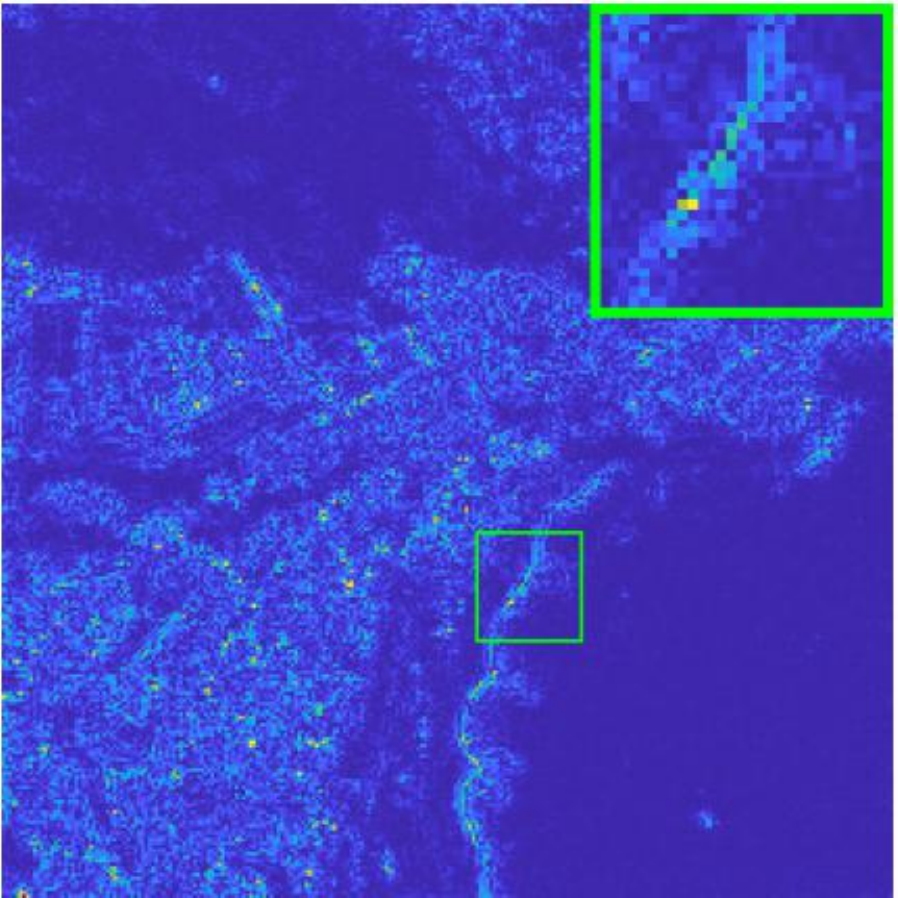}}
				\vspace{4pt}
				{CANNet}
				\centering
				
			\end{minipage}
			\begin{minipage}[t]{0.12\linewidth}
				{\includegraphics[width=1\linewidth]{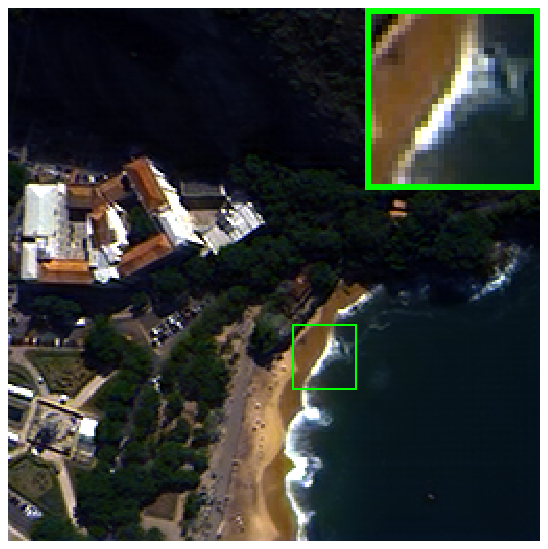}}
				{\includegraphics[width=1\linewidth]{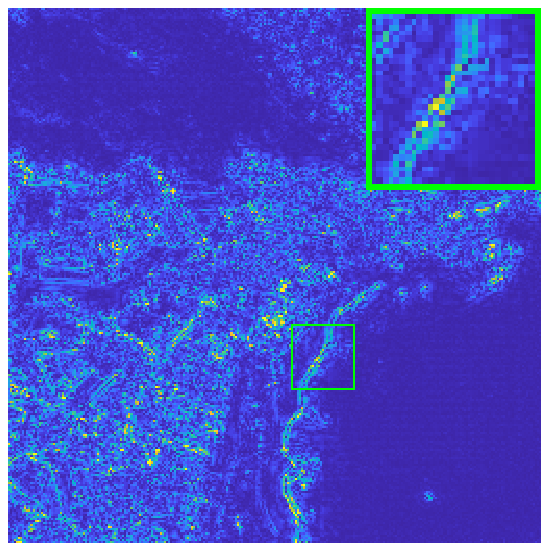}}
				\vspace{4pt}
				{BDPN}
				{\includegraphics[width=1\linewidth]{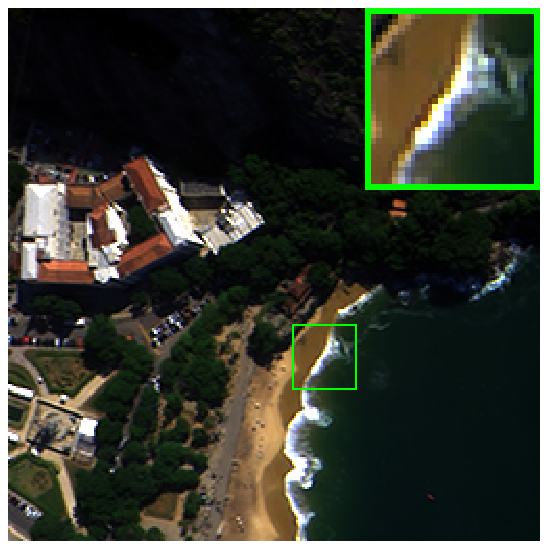}}
				{\includegraphics[width=1\linewidth]{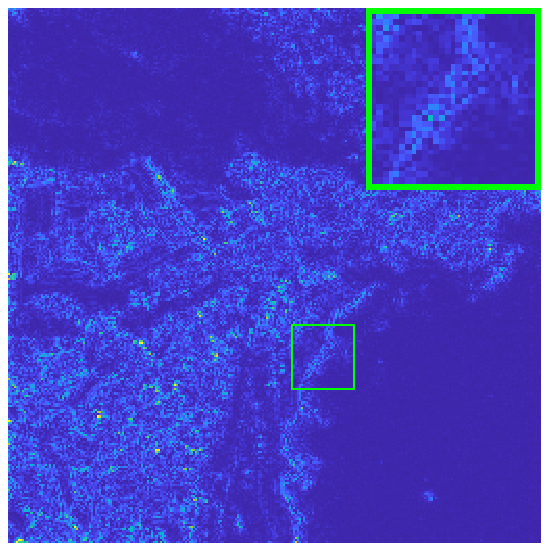}}
				\vspace{4pt}
				{FusionMamba}
				\centering
				
			\end{minipage}
			\begin{minipage}[t]{0.12\linewidth}
				{\includegraphics[width=1\linewidth]{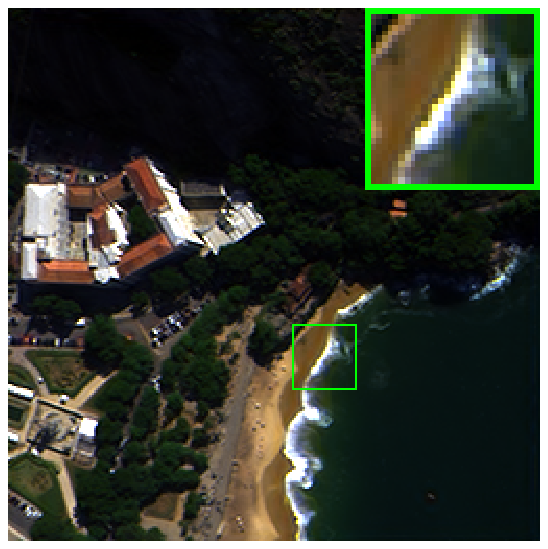}}
				{\includegraphics[width=1\linewidth]{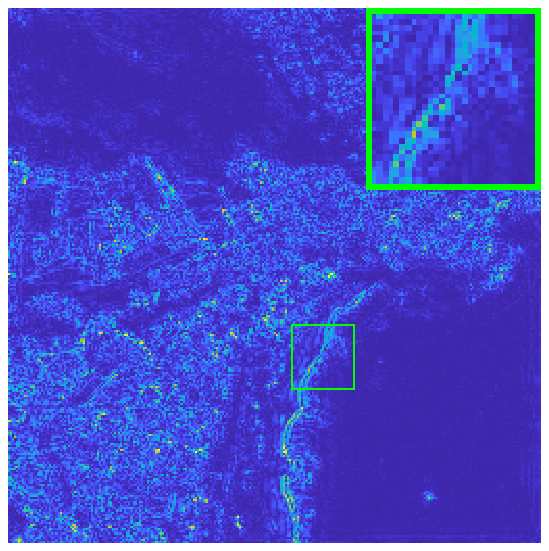}}
				\vspace{4pt}
				{FusionNet}
				{\includegraphics[width=1\linewidth]{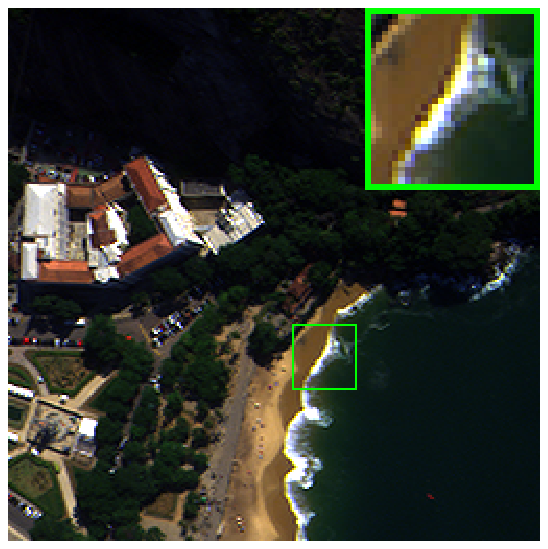}}
				{\includegraphics[width=1\linewidth]{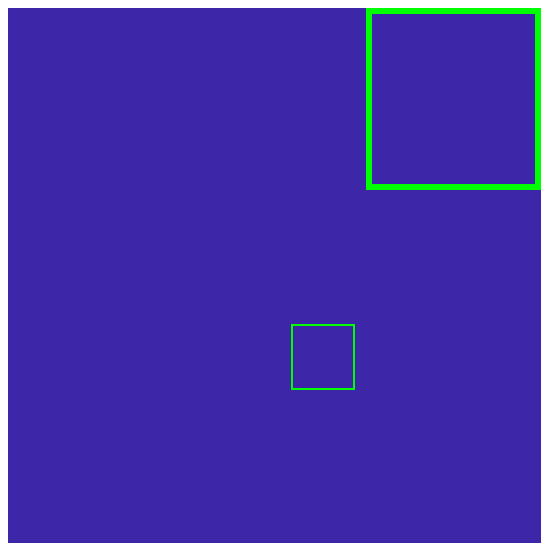}}
				\vspace{4pt}
				{GT}
				\centering
				
			\end{minipage}
		\end{minipage}
		\begin{minipage}[t]{1\linewidth}
			{\includegraphics[width=1\linewidth]{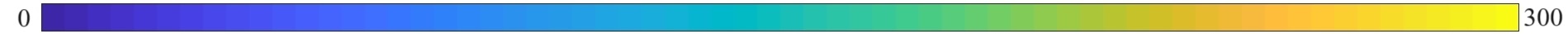}}
			\centering
		\end{minipage}
	\end{center}
    \vspace{-2pt}
	\caption{Qualitative results on a reduced-resolution example from the WV3 dataset. Rows 1 and 3: Pseudo-color images representing spectral bands 1, 3, and 5. Rows 2 and 4: The corresponding absolute error maps (AEMs) for spectral band 7. The values on both sides of the color bar indicate the degree of errors. \label{rr_v}}
\end{figure*}

\begin{table*}[t]	
	\centering\renewcommand\arraystretch{1.2}\setlength{\tabcolsep}{5.4pt}
	\belowrulesep=0pt\aboverulesep=0pt
	\caption{Results on 20 reduced-resolution and 20 full-resolution samples from the GF2 dataset, which belongs to pansharpening.}
	\label{gf2}	
	\begin{tabular}{l|c|cccc|ccc}
		\toprule
		\multirow{2}{*}{\textbf{Methods}} & 
		\multirow{2}{*}{\textbf{Params}} & 
		\multicolumn{4}{c|}{\textbf{Reduced-Resolution}} & \multicolumn{3}{c}{\textbf{Full-Resolution (Real Data)}}\\
		\cmidrule(lr){3-6}\cmidrule(lr){7-9}
		&\multicolumn{1}{c|}{} 
		&\multicolumn{1}{c}{PSNR($\pm$std)} 
		&\multicolumn{1}{c}{Q2n($\pm$std)} 
		&\multicolumn{1}{c}{SAM($\pm$std)} 
		&\multicolumn{1}{c|}{ERGAS($\pm$std)} 
		&\multicolumn{1}{c}{${{\rm{D}}_{\rm{\lambda}}}$($\pm$std)} 
		&\multicolumn{1}{c}{${{\rm{D}}_{\rm{s}}}$($\pm$std)} 
		&\multicolumn{1}{c}{QNR($\pm$std)} \\
		\midrule
		\textbf{TV} \cite{palsson2013new} & $-$ & 41.262$\pm$2.264 & 0.907$\pm$0.029 & 1.911$\pm$0.447 & 1.737$\pm$0.447
		& 0.0553$\pm$0.0430  & 0.1118$\pm$0.0226 & 0.8392$\pm$0.0441\\ 
		\textbf{GLP-HPM} \cite{6616569} & $-$ & 41.582$\pm$2.217 & 0.900$\pm$0.034 & 1.650$\pm$0.392 & 1.588$\pm$0.405
		& 0.0336$\pm$0.0129  & 0.1404$\pm$0.0277 & 0.8309$\pm$0.0334\\ 
		\textbf{GLP-FS} \cite{vivone2018full}& $-$  & 41.565$\pm$2.125 & 0.897$\pm$0.035 & 1.655$\pm$0.385 & 1.589$\pm$0.395
		& 0.0346$\pm$0.0137  & 0.1429$\pm$0.0282 & 0.8276$\pm$0.0348\\ 
		\textbf{BDSD-PC} \cite{2019Robust} & $-$ & 41.205$\pm$2.317 & 0.892$\pm$0.035 & 1.681$\pm$0.360 & 1.667$\pm$0.445
		& 0.0759$\pm$0.0301  & 0.1548$\pm$0.0280 & 0.7812$\pm$0.0409\\ 
		\midrule
		\textbf{PanNet} \cite{8237455} & 0.08M & 46.268$\pm$2.031 & 0.967$\pm$0.010 & 0.997$\pm$0.212 & 0.919$\pm$0.191 & \underline{0.0179}$\pm$0.0110  & 0.0799$\pm$0.0178 & 0.9036$\pm$0.0198\\ 
		\textbf{MSDCNN} \cite{8127731} & 0.23M & 45.247$\pm$2.228 & 0.961$\pm$0.011 & 1.047$\pm$0.221 & 1.041$\pm$0.231
		& 0.0243$\pm$0.0133 & 0.0730$\pm$0.0093 & 0.9044$\pm$0.0126\\
		\textbf{BDPN} \cite{8667448} & 1.49M & 42.080$\pm$2.625 & 0.923$\pm$0.024 & 1.481$\pm$0.326 & 1.546$\pm$0.432
		& 0.0330$\pm$0.0223  & 0.0765$\pm$0.0199 & 0.8929$\pm$0.0250\\ 
		\textbf{FusionNet} \cite{2020Detail} & 0.08M & 45.663$\pm$2.270 & 0.964$\pm$0.009 & 0.974$\pm$0.212 & 0.988$\pm$0.222
		& 0.0350$\pm$0.0124  & 0.1013$\pm$0.0134 & 0.8673$\pm$0.0179\\   
		\textbf{MUCNN} \cite{10.1145/3474085.3475600} & 2.32M & 48.256$\pm$1.930 & 0.979$\pm$0.008 & 0.808$\pm$0.171 & 0.731$\pm$0.146 
		& 0.0181$\pm$0.0093  & {0.0515}$\pm$0.0088 & \underline{0.9312}$\pm$0.0107\\   
		\textbf{LAGNet} \cite{jin2022aaai}& 0.15M & 48.760$\pm$1.447 & 0.980$\pm$0.009 & 0.786$\pm$0.148 & {0.687}$\pm$0.113
		& 0.0284$\pm$0.0130  & 0.0792$\pm$0.0136& 0.8947$\pm$0.0200\\  
		\textbf{PMACNet} \cite{9764690} & 0.94M & {45.041}$\pm$2.135 & {0.963}$\pm$0.011 & {1.359}$\pm$0.133 & 1.248$\pm$0.204
		& 0.0981$\pm$0.0215  &\underline{0.0474}$\pm$0.0115 & 0.8590$\pm$0.0171\\  	
		\textbf{U2Net} \cite{10.1145/3581783.3612084}& 0.66M & {49.404}$\pm$1.730 & {0.982}$\pm$0.009 & {0.714}$\pm$0.138 & {0.632}$\pm$0.117
		& {0.0236}$\pm$0.0172  & {0.0510}$\pm$0.0101 & {0.9265}$\pm$0.0172\\   
		\textbf{Pan-Mamba} \cite{he2024pan} & 0.48M & {48.931}$\pm$1.811 & {0.982}$\pm$0.008 & {0.743}$\pm$0.156 & 0.684$\pm$0.129
		& 0.0231$\pm$0.0110  & 0.0573$\pm$0.0116 & {0.9209}$\pm$0.0148\\ 
		\textbf{CANNet} \cite{Duan_2024_CVPR} & 0.78M & \underline{49.520}$\pm$1.932 & \underline{0.983}$\pm$0.006 & \underline{0.708}$\pm$0.156 & \underline{0.630}$\pm$0.128
		& 0.0194$\pm$0.0101  & {0.0630}$\pm$0.0094 & 0.9188$\pm$0.0110\\ 
		\textbf{FusionMamba} & 0.73M & \textbf{49.678}$\pm$1.708 & \textbf{0.984}$\pm$0.007 & \textbf{0.705}$\pm$0.137 & \textbf{0.615}$\pm$0.108
		& \textbf{0.0174}$\pm$0.0094  & \textbf{0.0295}$\pm$0.0073 & \textbf{0.9536}$\pm$0.0086\\      
		\midrule
		\textbf{Ideal Values} 
		& $-$
		&\multicolumn{1}{c}{\textbf{+$\infty$}}
		&\multicolumn{1}{c}{\textbf{\textbf{1}}}
		&\multicolumn{1}{c}{\textbf{\textbf{0}}}
		&\multicolumn{1}{c|}{\textbf{\textbf{0}}}
		&\multicolumn{1}{c}{\textbf{\textbf{0}}}
		&\multicolumn{1}{c}{\textbf{\textbf{0}}}
		&\multicolumn{1}{c}{\textbf{\textbf{1}}}
		\\ 
		\bottomrule
	\end{tabular}
\end{table*}

\begin{figure*}[t]
	\begin{center}
		\begin{minipage}[t]{1\linewidth}
			\begin{minipage}[t]{0.12\linewidth}
				{\includegraphics[width=1\linewidth]{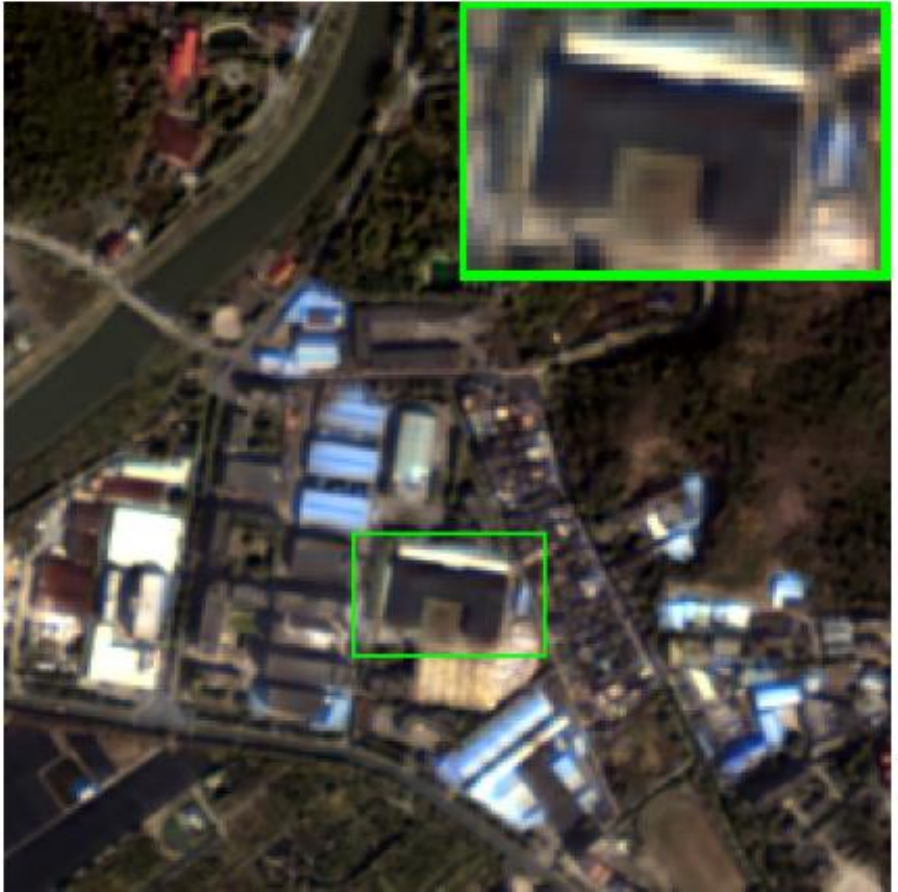}}
				{\includegraphics[width=1\linewidth]{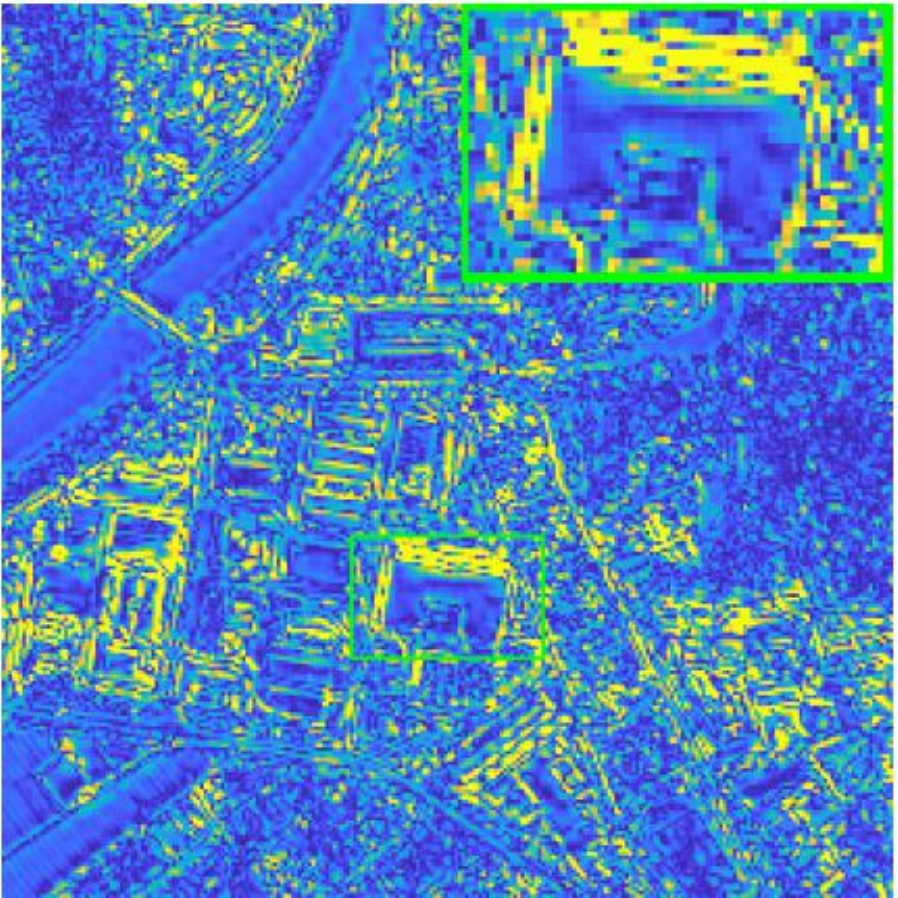}}
				\vspace{4pt}
				{TV}
				{\includegraphics[width=1\linewidth]{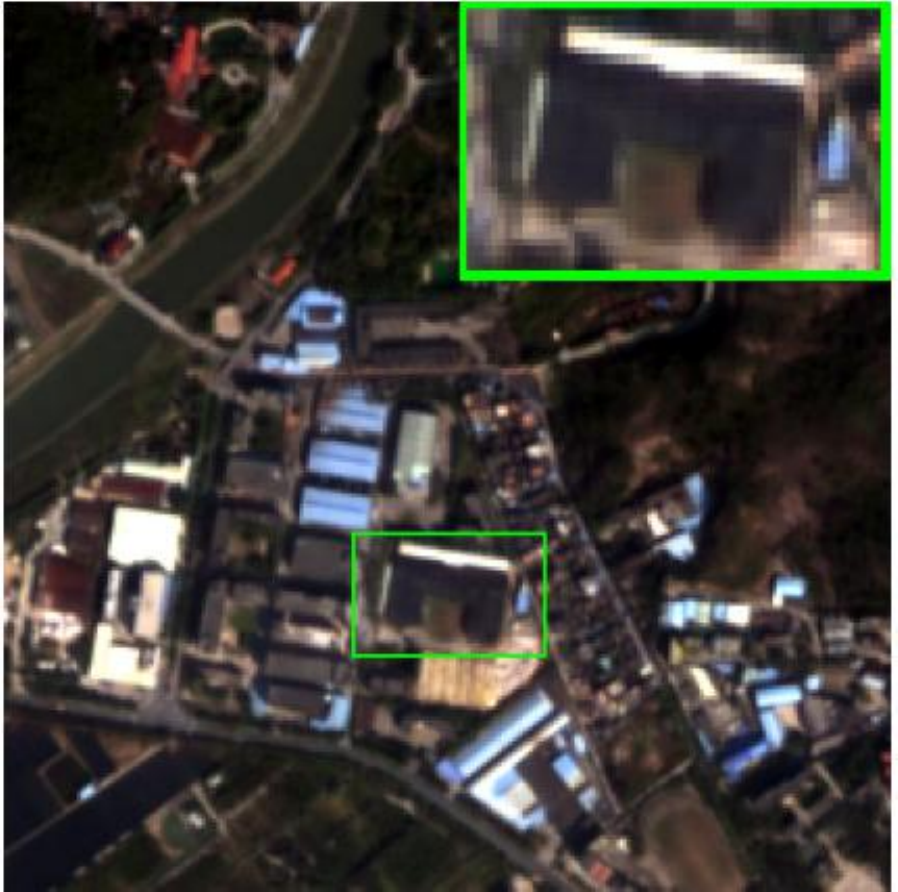}}
				{\includegraphics[width=1\linewidth]{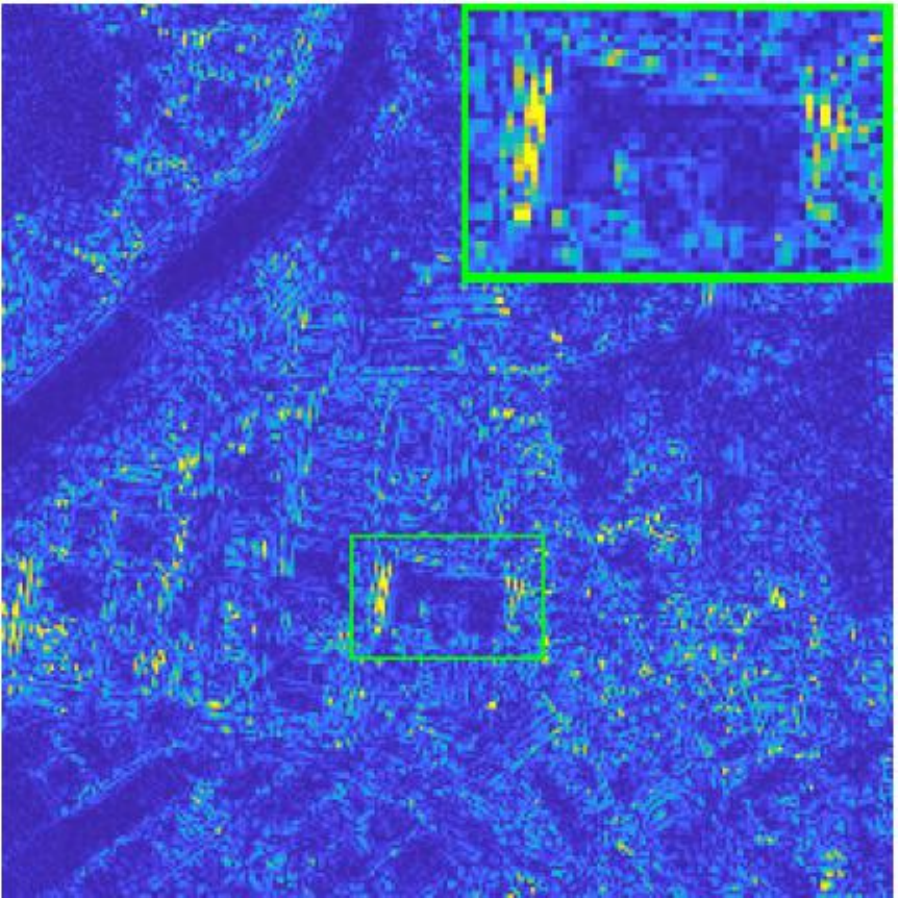}}
				\vspace{4pt}
				{MUCNN}
				\centering
				
			\end{minipage}
			\begin{minipage}[t]{0.12\linewidth}
				{\includegraphics[width=1\linewidth]{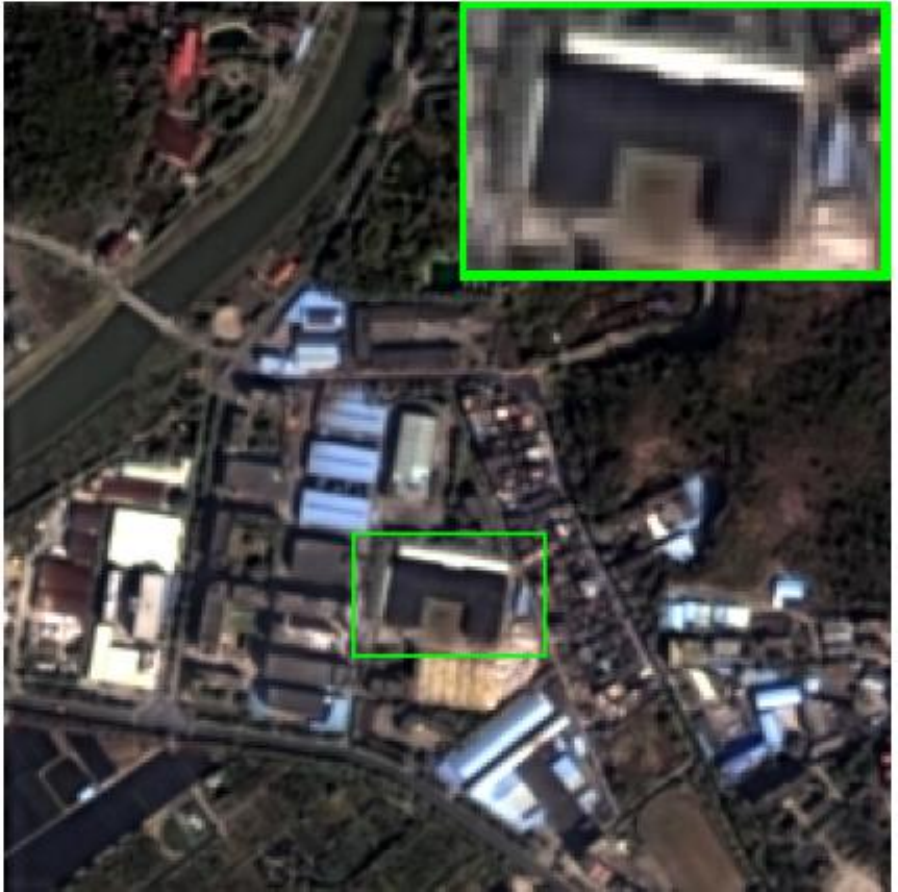}}
				{\includegraphics[width=1\linewidth]{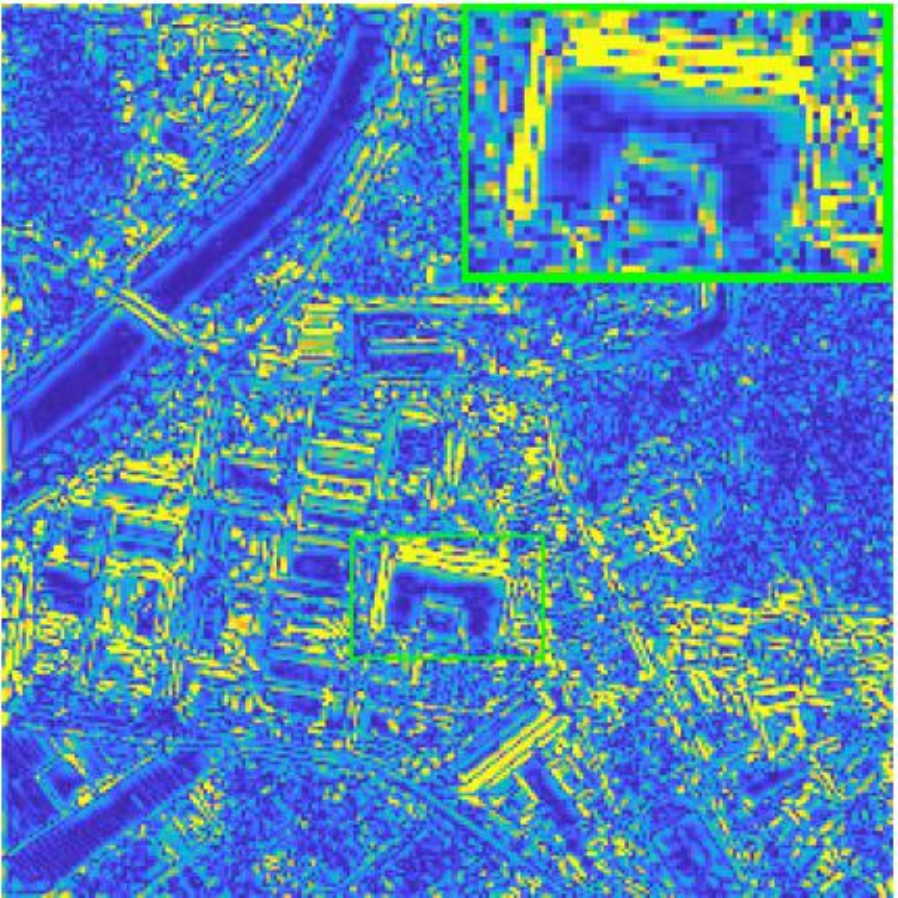}}
				\vspace{4pt}
				{GLP-HPM}
				{\includegraphics[width=1\linewidth]{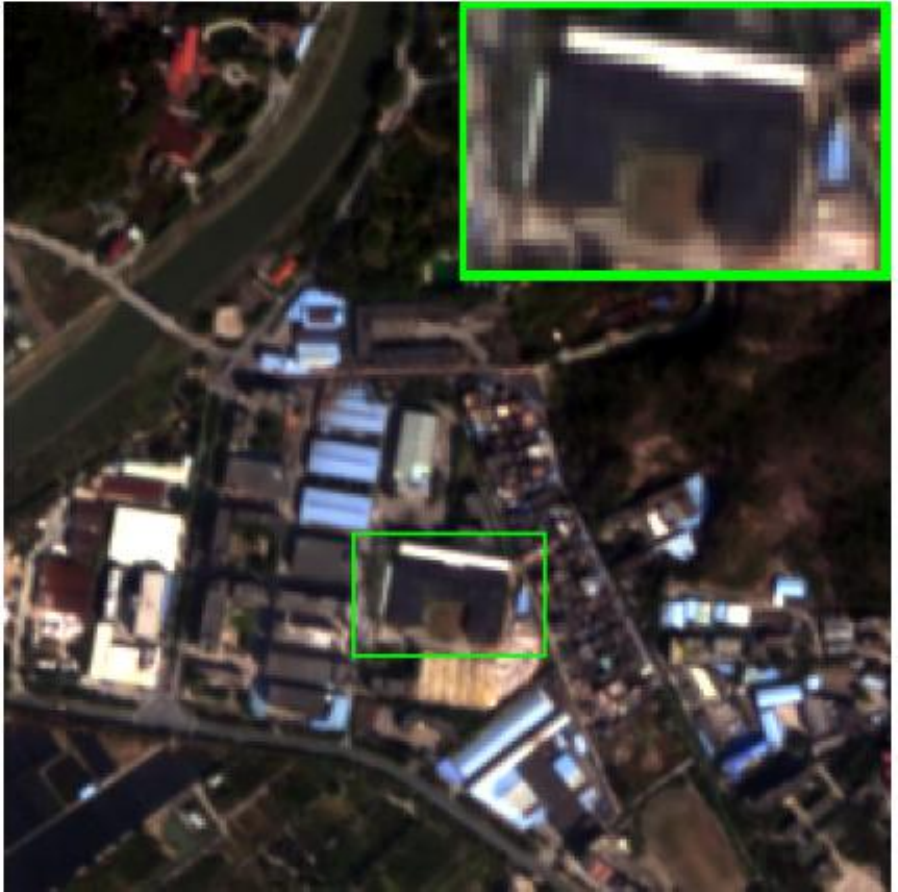}}
				{\includegraphics[width=1\linewidth]{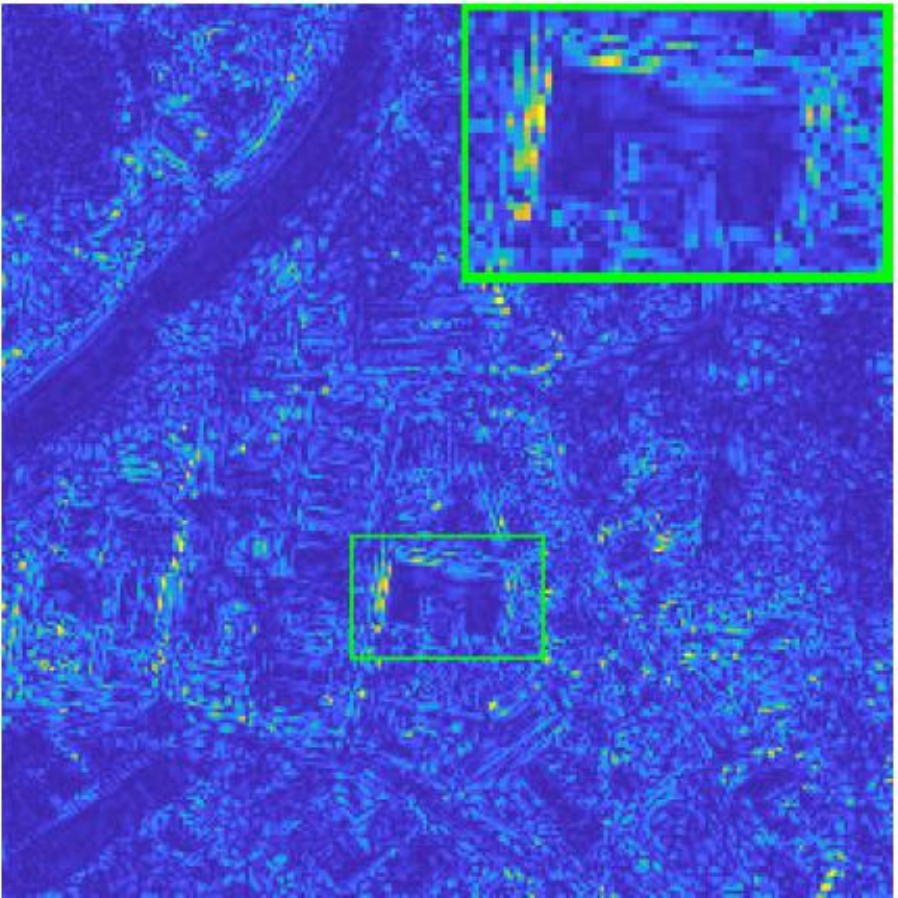}}
				\vspace{4pt}
				{LAGNet}
				\centering
				
			\end{minipage}
			\begin{minipage}[t]{0.12\linewidth}
				{\includegraphics[width=1\linewidth]{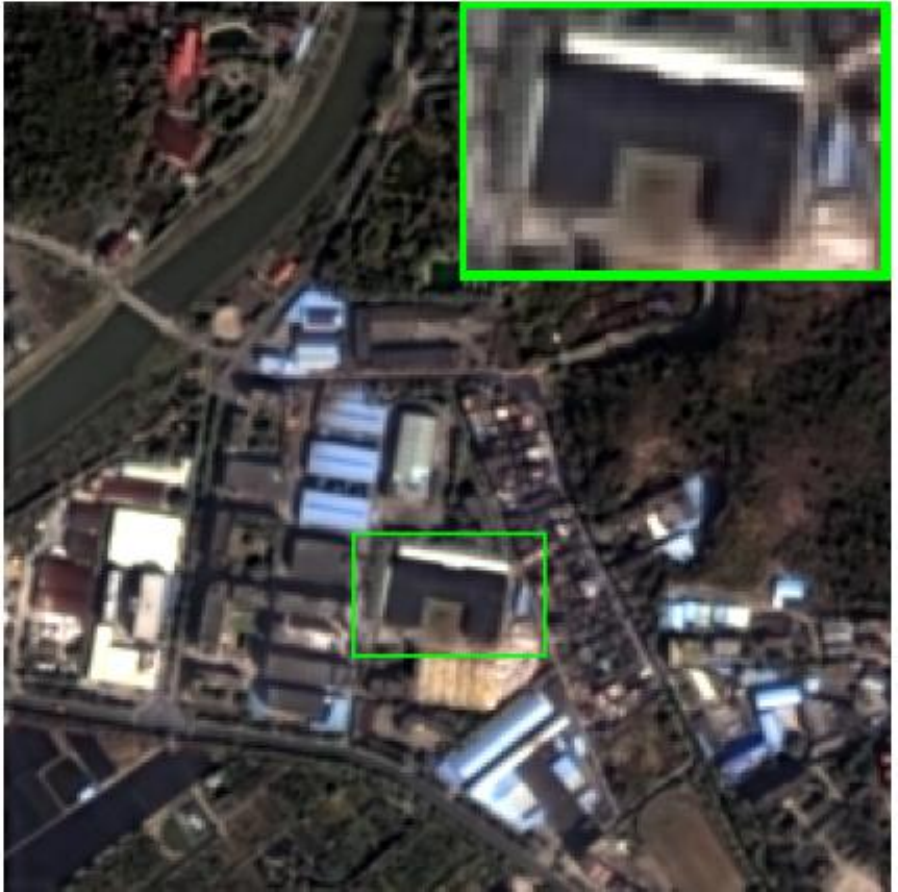}}
				{\includegraphics[width=1\linewidth]{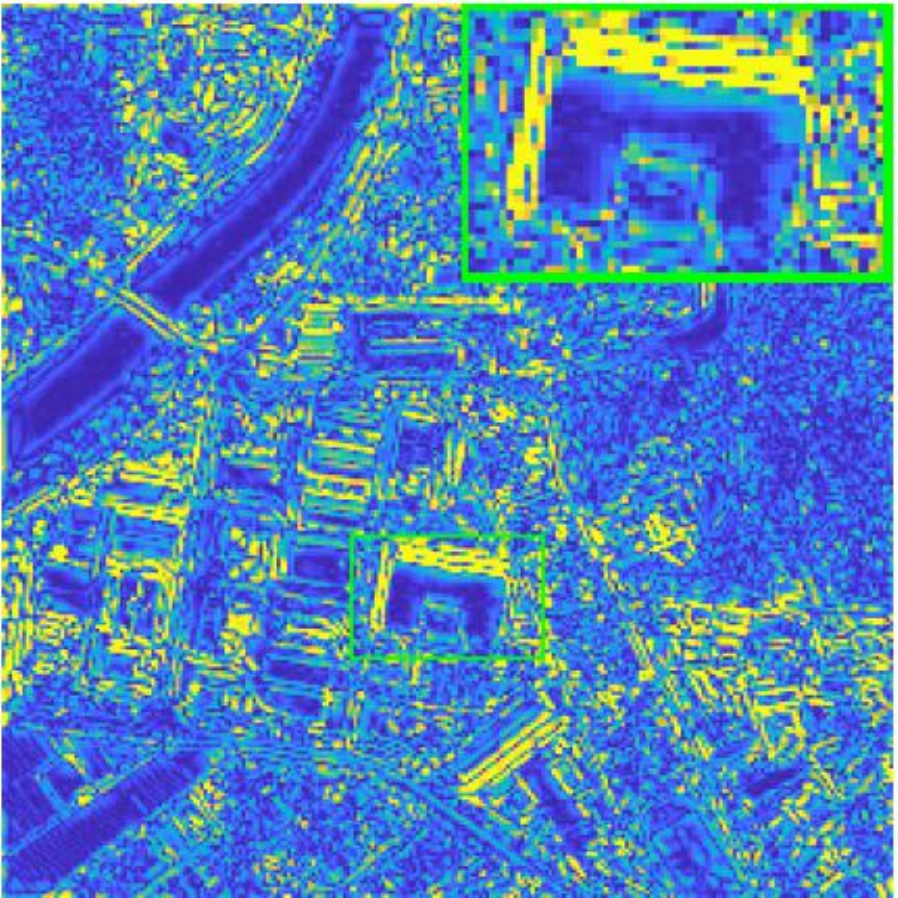}}
				\vspace{4pt}
				{GLP-FS}
				{\includegraphics[width=1\linewidth]{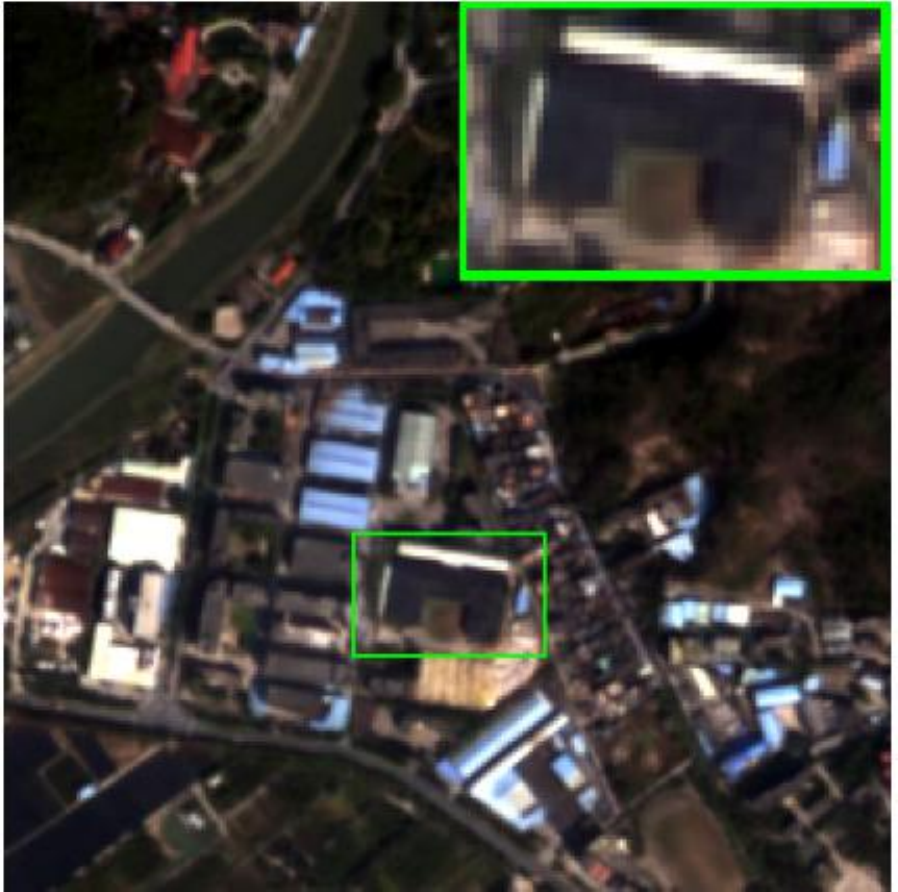}}
				{\includegraphics[width=1\linewidth]{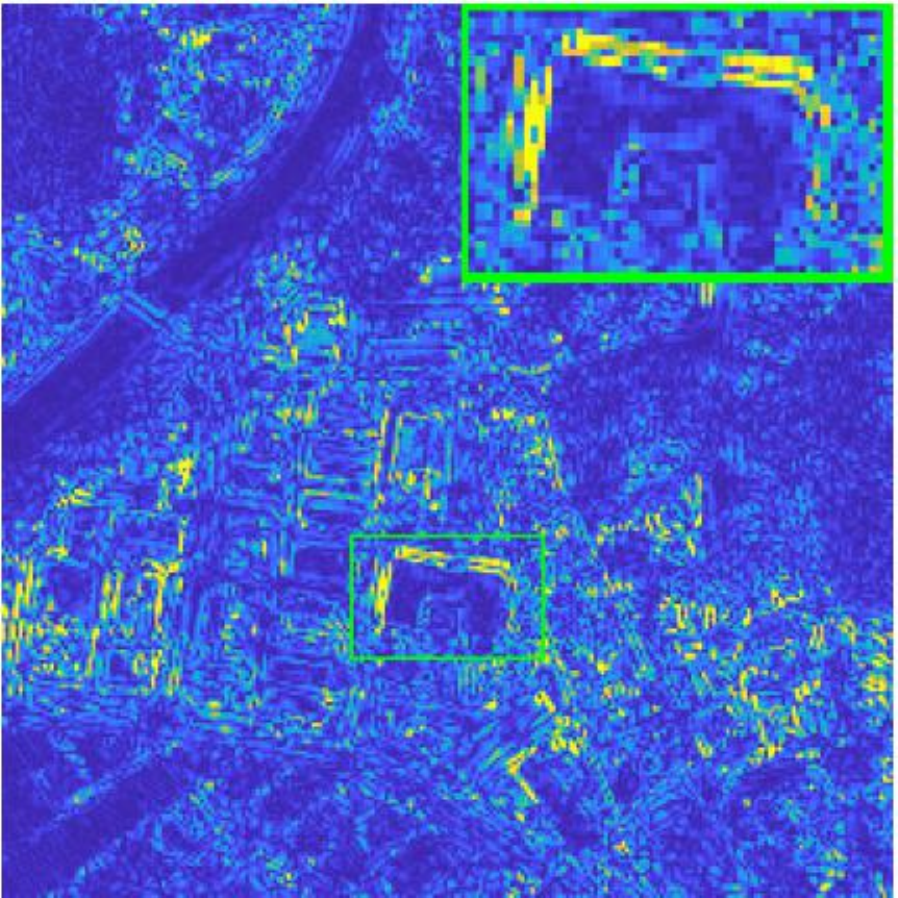}}
				\vspace{4pt}
				{PMACNet}
				\centering
				
			\end{minipage}
			\begin{minipage}[t]{0.12\linewidth}
				{\includegraphics[width=1\linewidth]{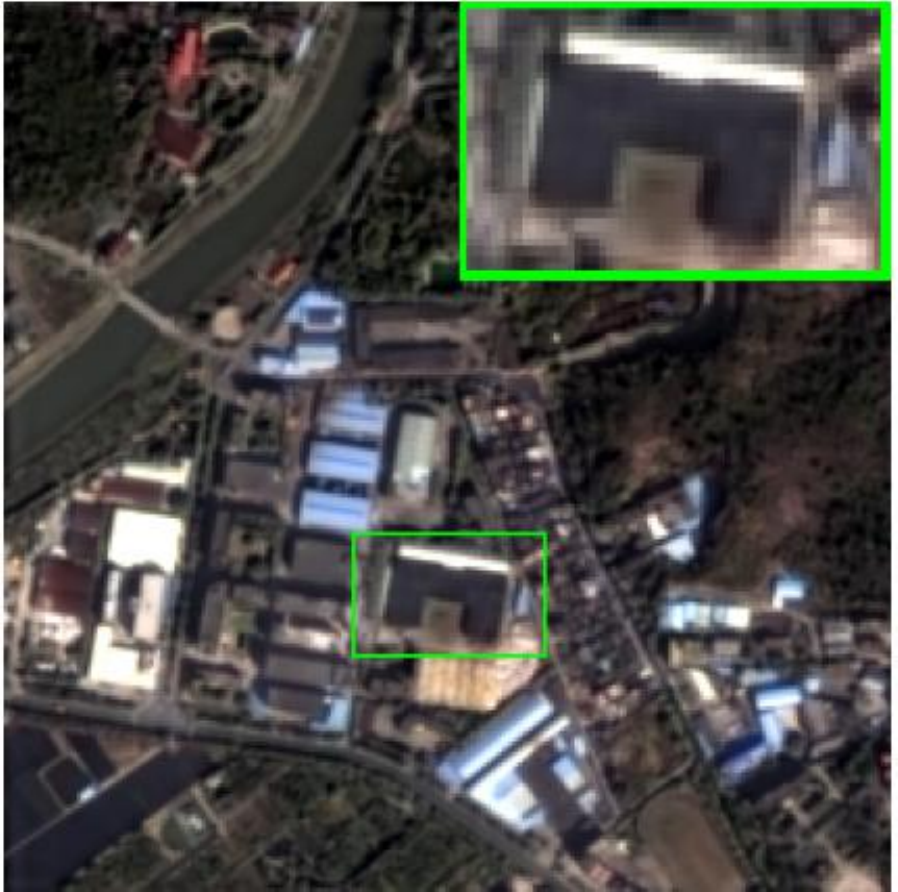}}
				{\includegraphics[width=1\linewidth]{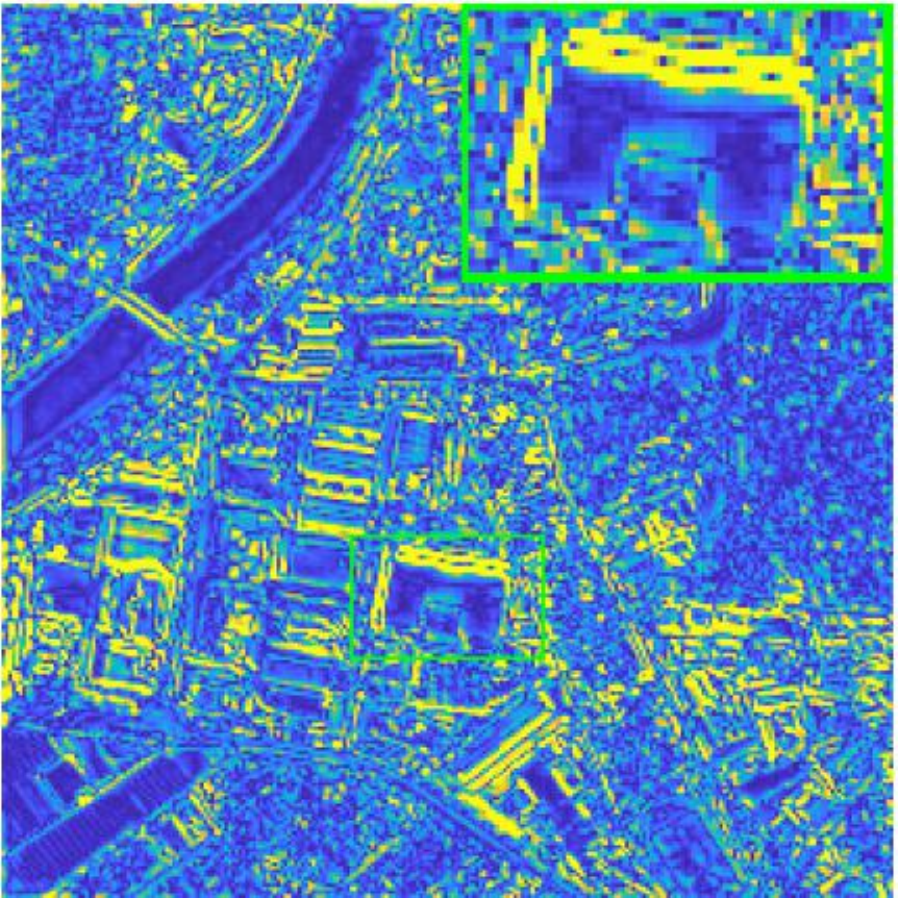}}
				\vspace{4pt}
				{BDSD-PC}
				{\includegraphics[width=1\linewidth]{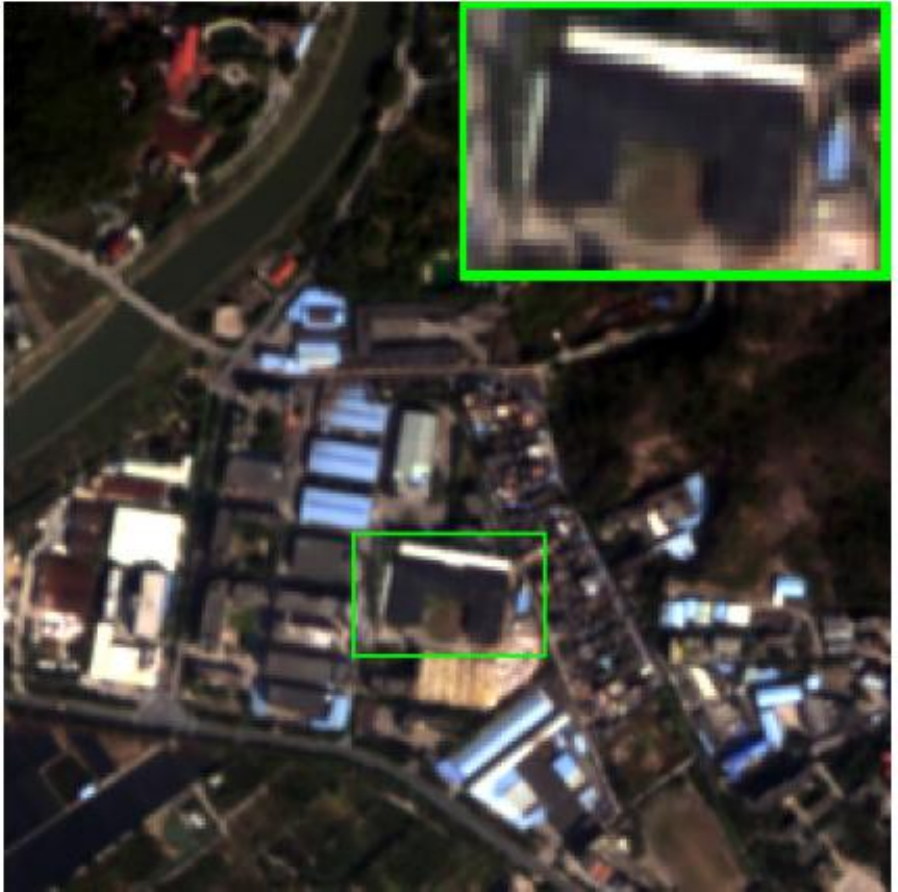}}
				{\includegraphics[width=1\linewidth]{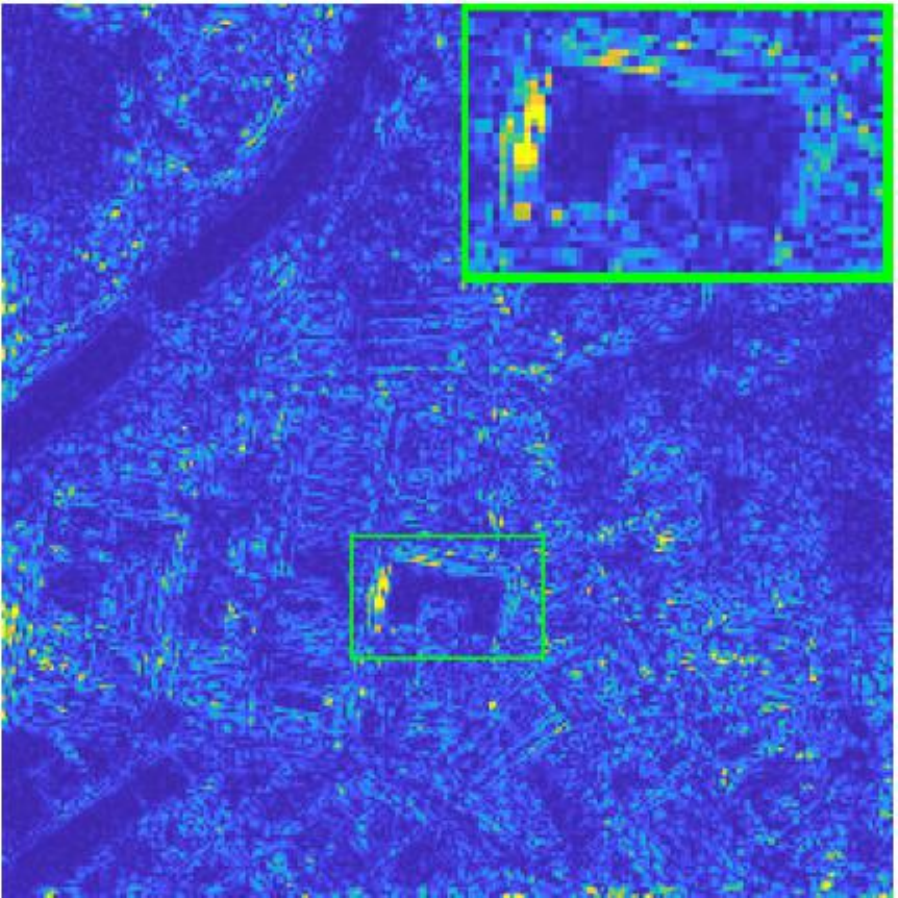}}
				\vspace{4pt}
				{U2Net}
				\centering
				
			\end{minipage}
			\begin{minipage}[t]{0.12\linewidth}
				{\includegraphics[width=1\linewidth]{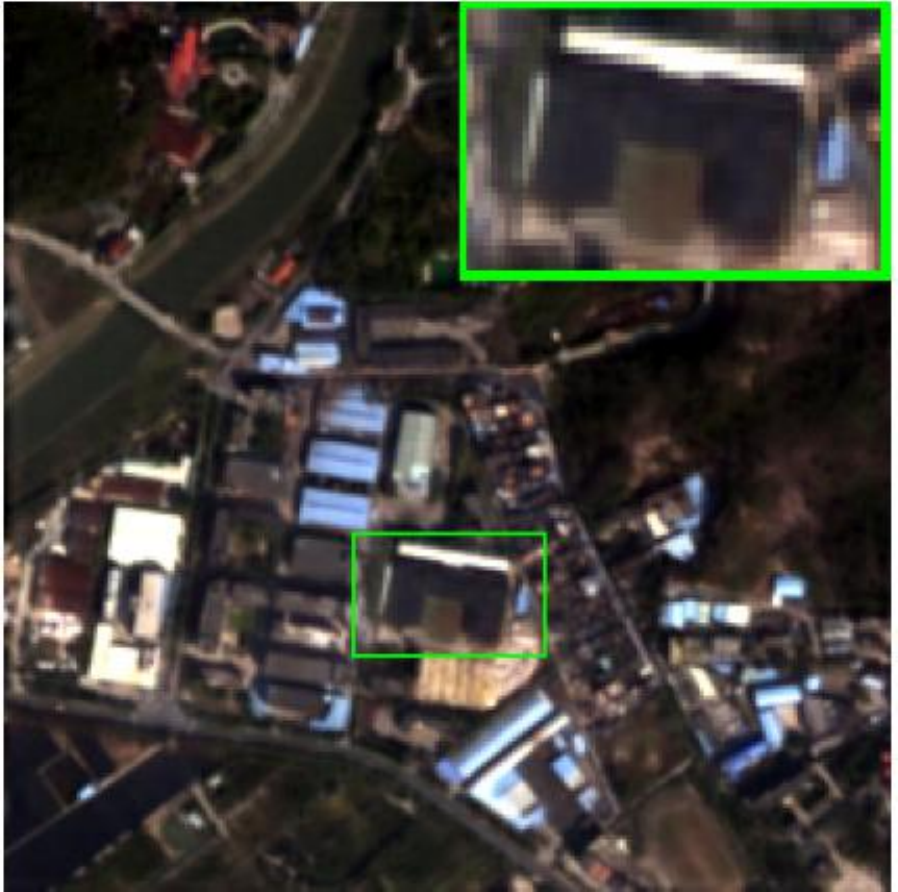}}
				{\includegraphics[width=1\linewidth]{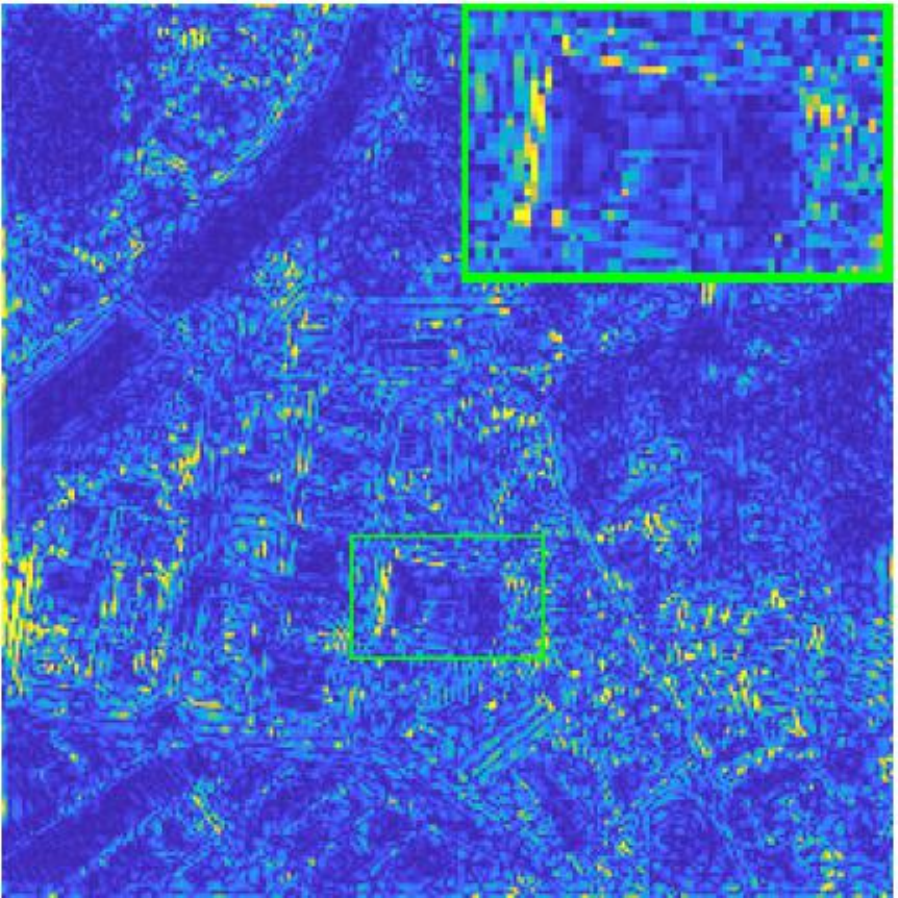}}
				\vspace{4pt}
				{PanNet}
				{\includegraphics[width=1\linewidth]{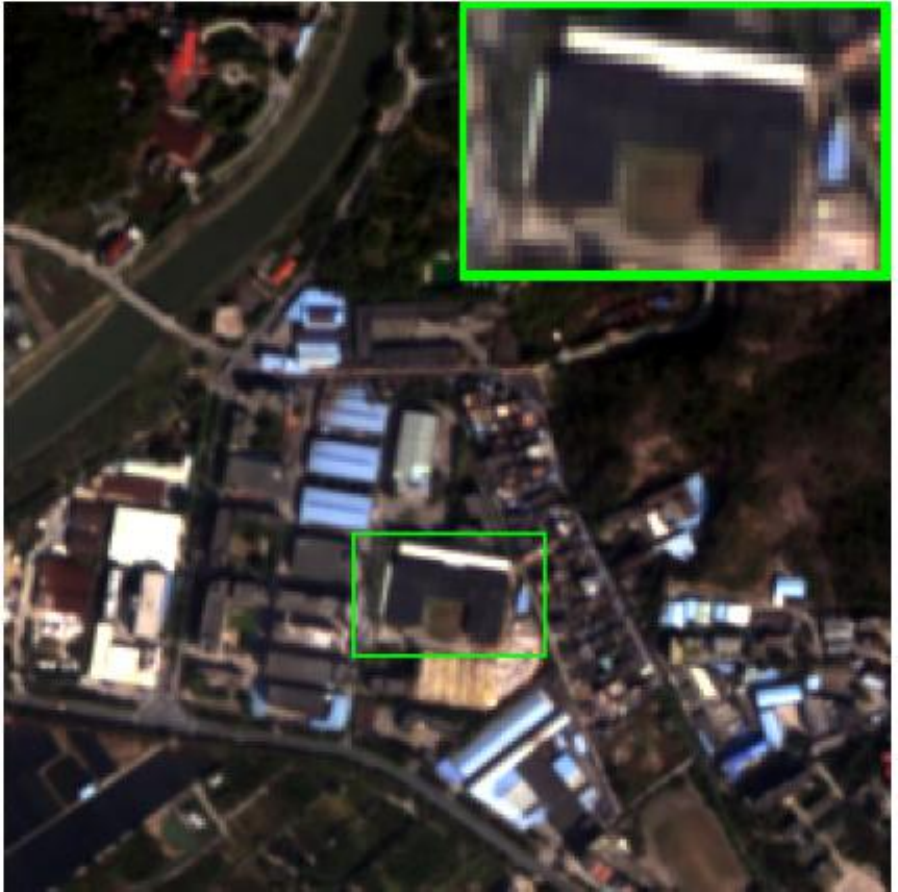}}
				{\includegraphics[width=1\linewidth]{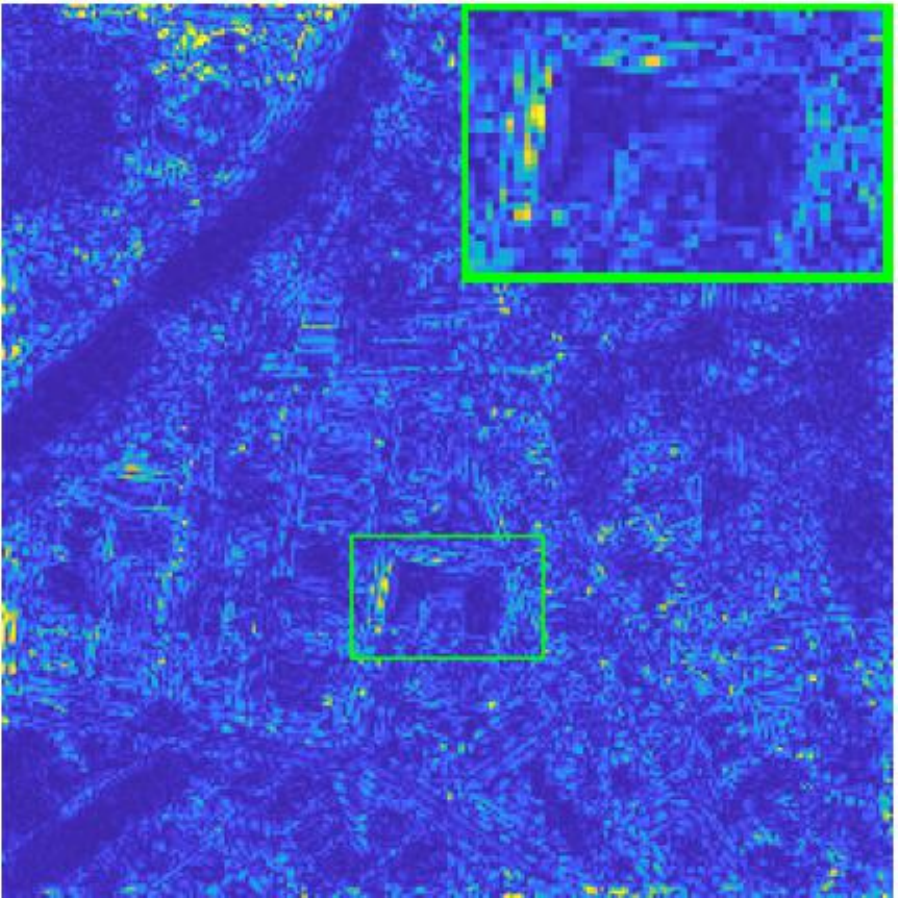}}
				\vspace{4pt}
				{Pan-Mamba}
				\centering
				
			\end{minipage}
			\begin{minipage}[t]{0.12\linewidth}
				{\includegraphics[width=1\linewidth]{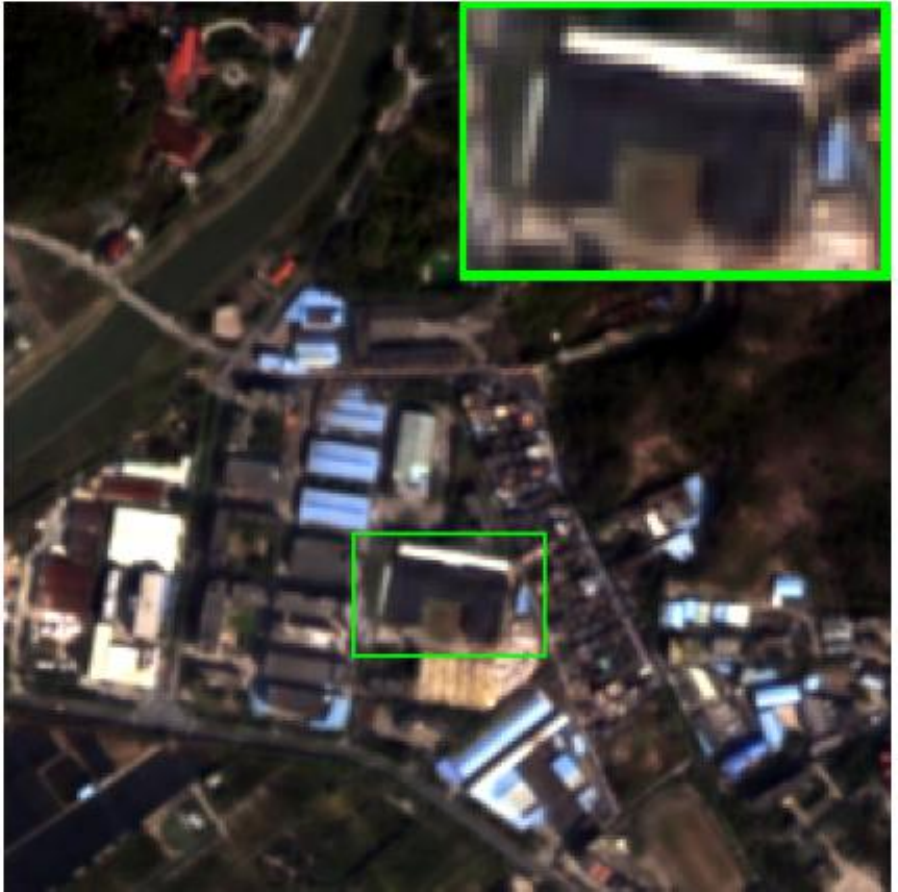}}
				{\includegraphics[width=1\linewidth]{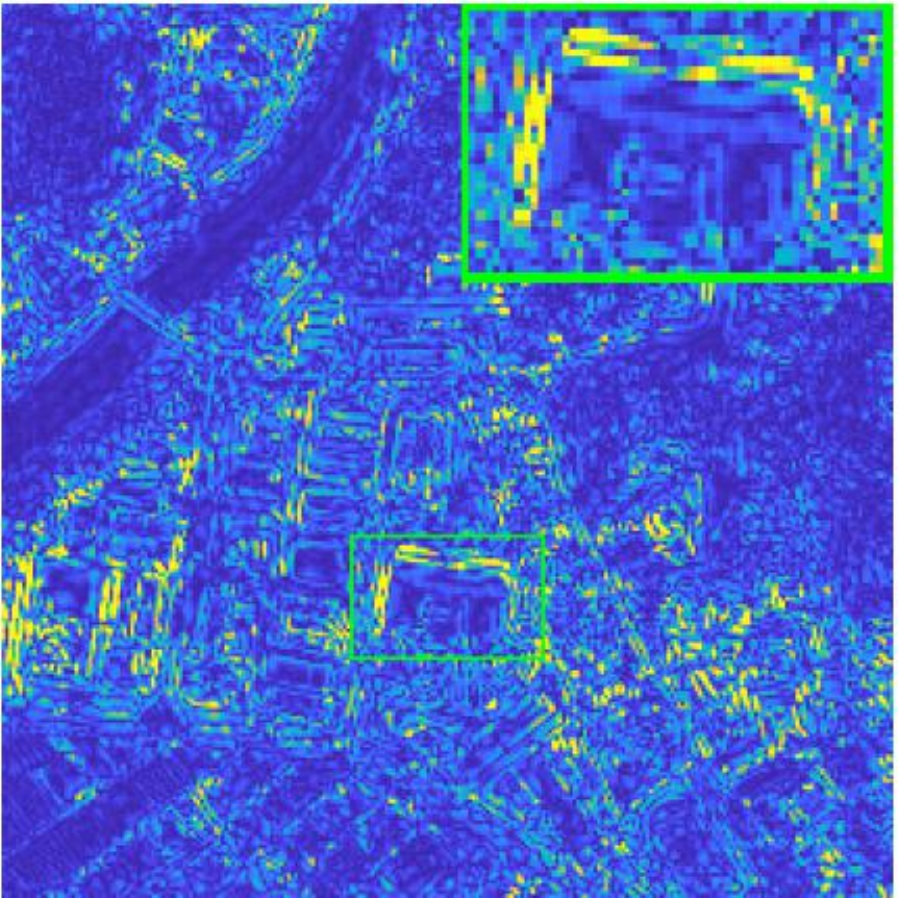}}
				\vspace{4pt}
				{MSDCNN}
				{\includegraphics[width=1\linewidth]{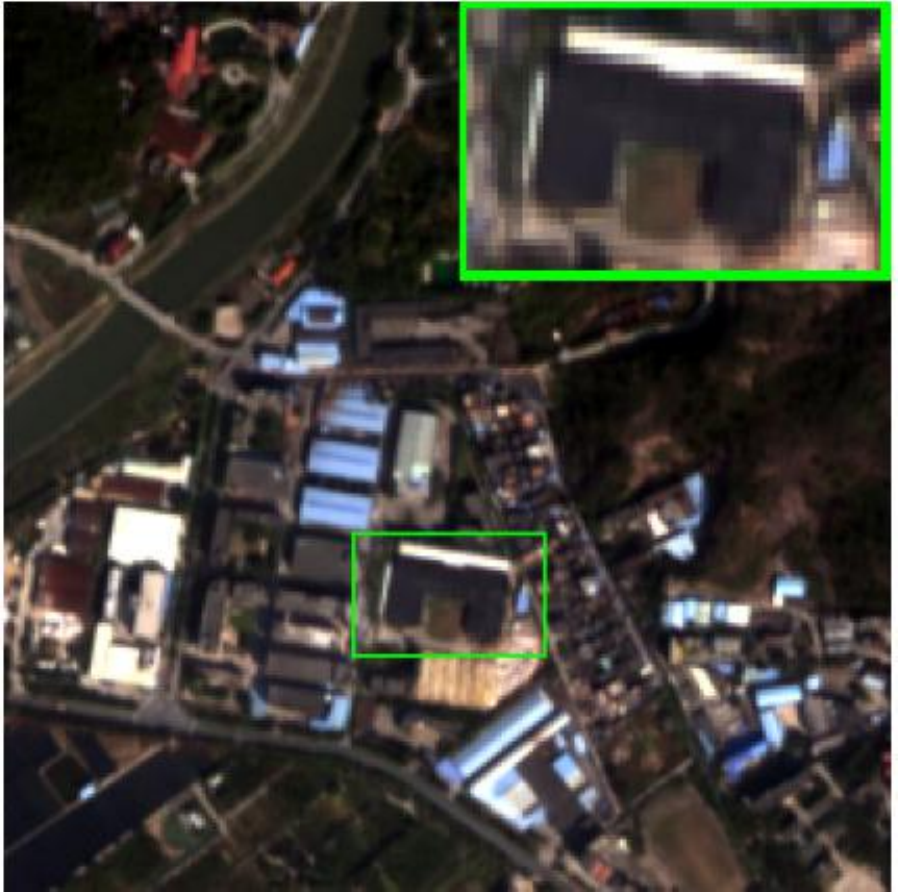}}
				{\includegraphics[width=1\linewidth]{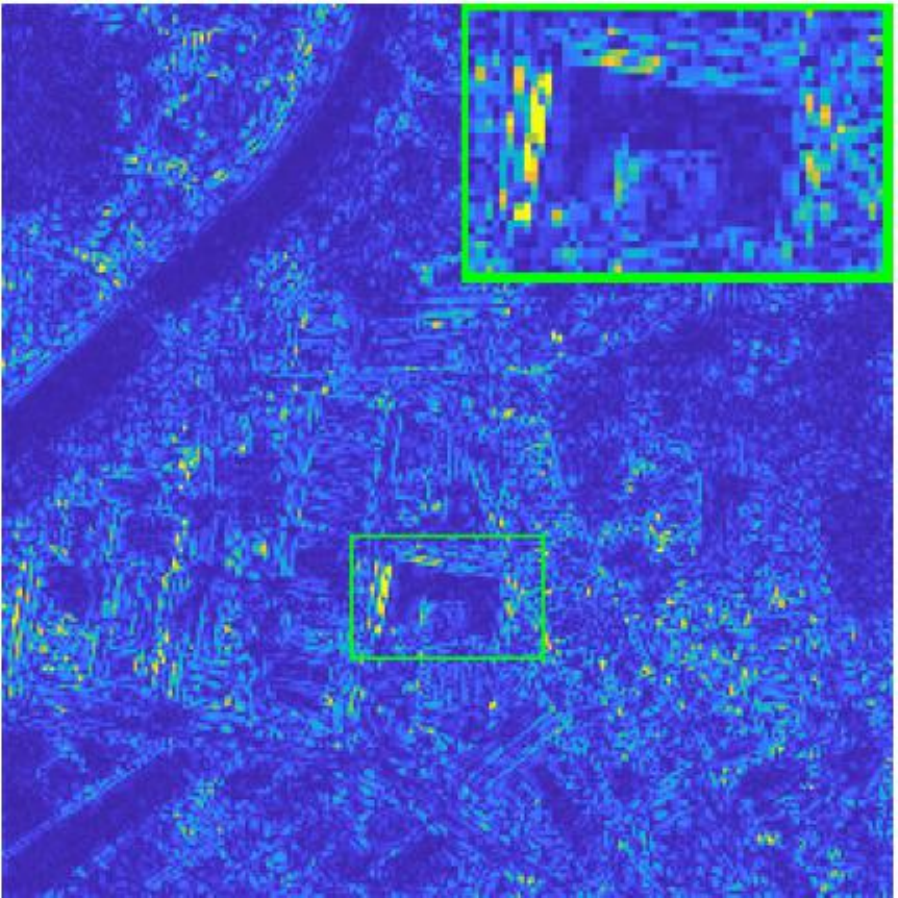}}
				\vspace{4pt}
				{CANNet}
				\centering
				
			\end{minipage}
			\begin{minipage}[t]{0.12\linewidth}
				{\includegraphics[width=1\linewidth]{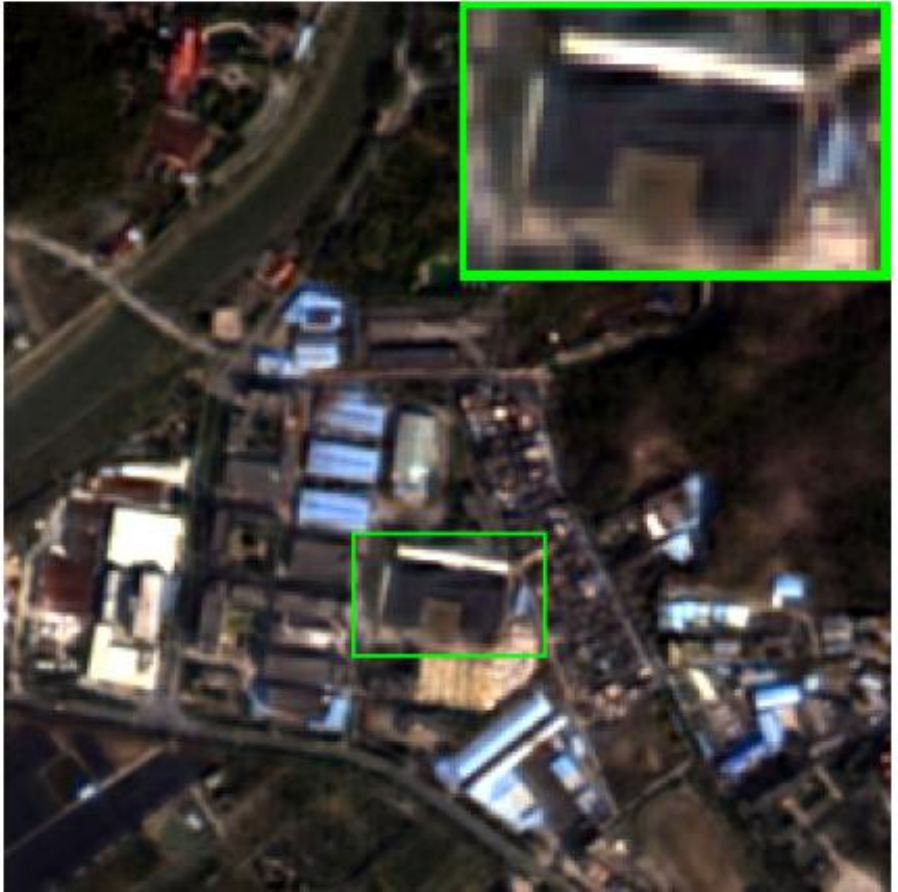}}
				{\includegraphics[width=1\linewidth]{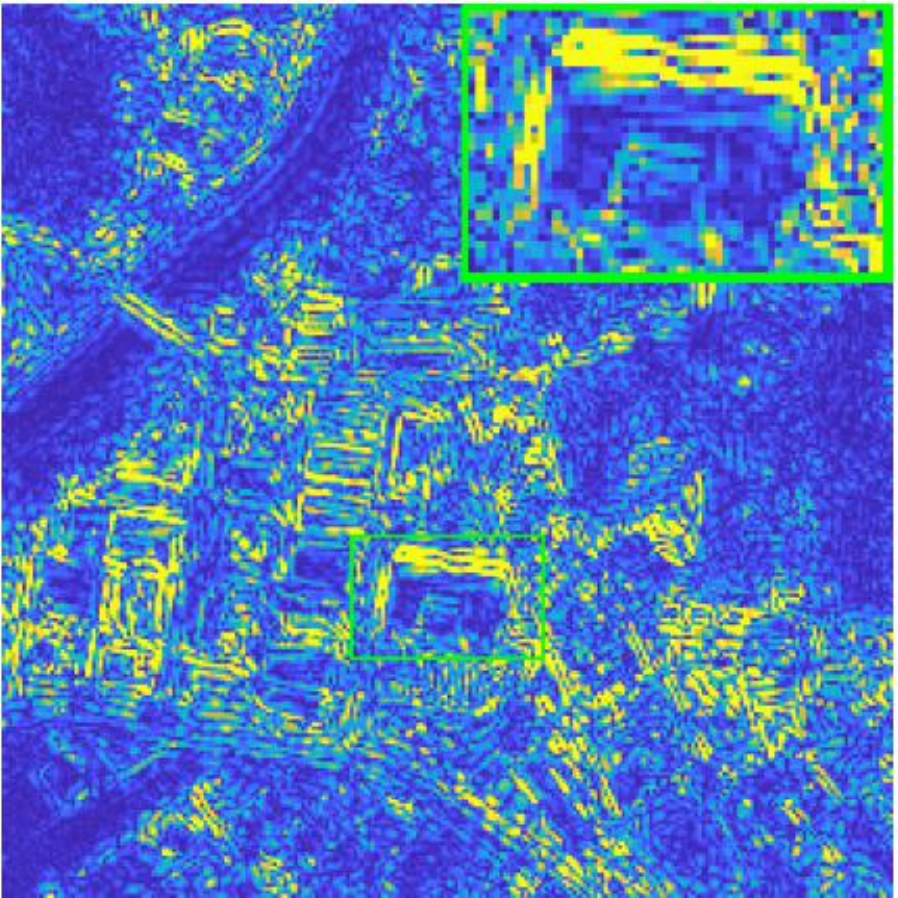}}
				\vspace{4pt}
				{BDPN}
				{\includegraphics[width=1\linewidth]{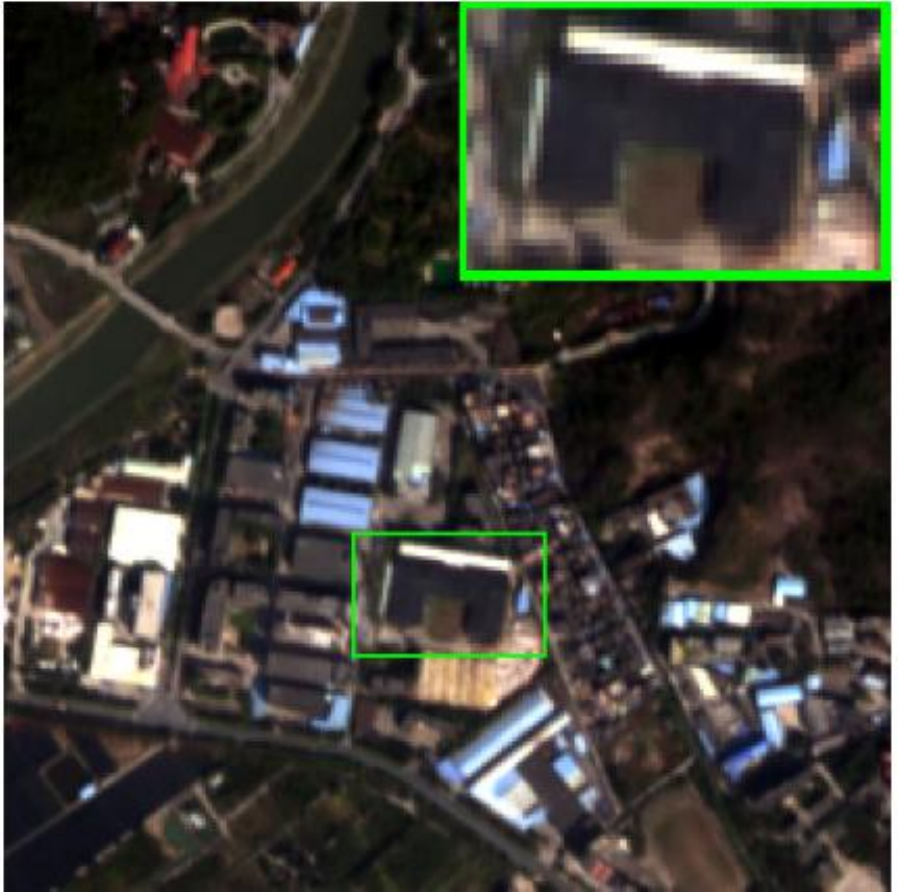}}
				{\includegraphics[width=1\linewidth]{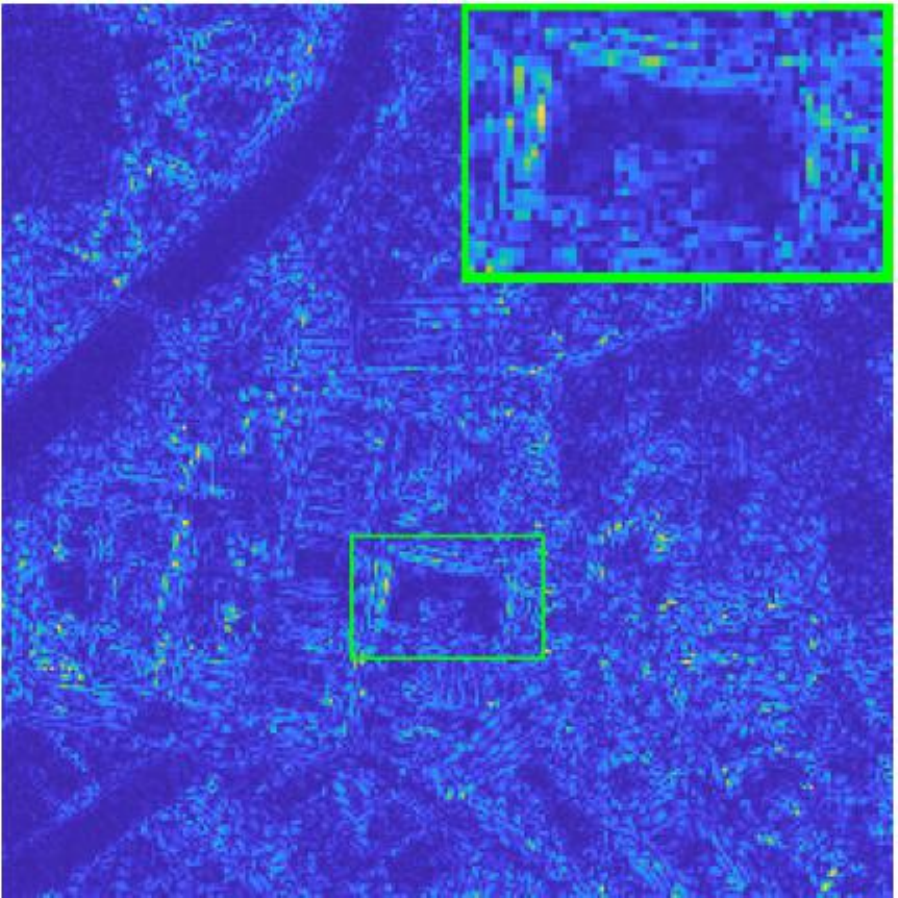}}
				\vspace{4pt}
				{FusionMamba}
				\centering
				
			\end{minipage}
			\begin{minipage}[t]{0.12\linewidth}
				{\includegraphics[width=1\linewidth]{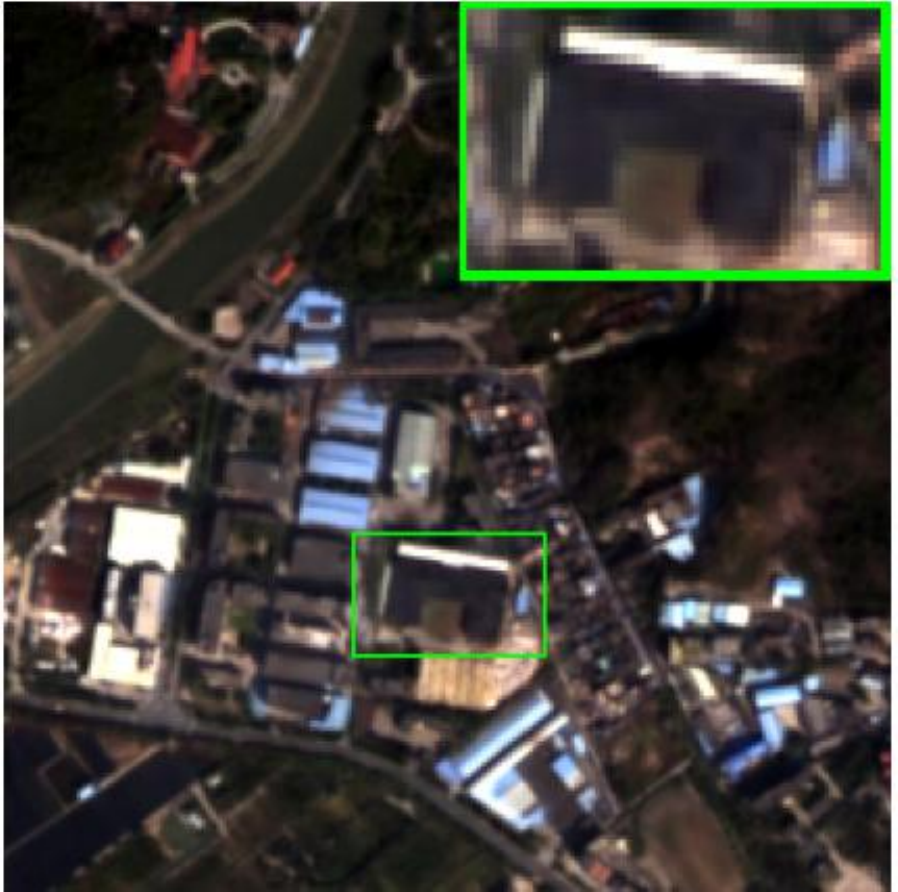}}
				{\includegraphics[width=1\linewidth]{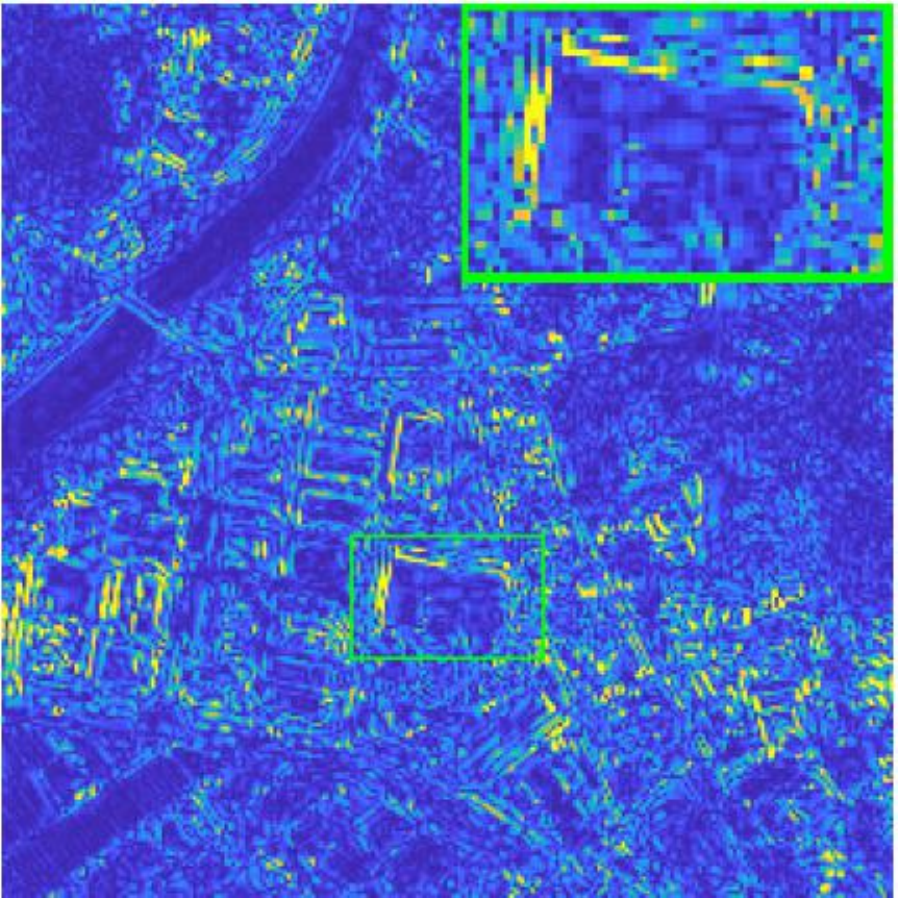}}
				\vspace{4pt}
				{FusionNet}
				{\includegraphics[width=1\linewidth]{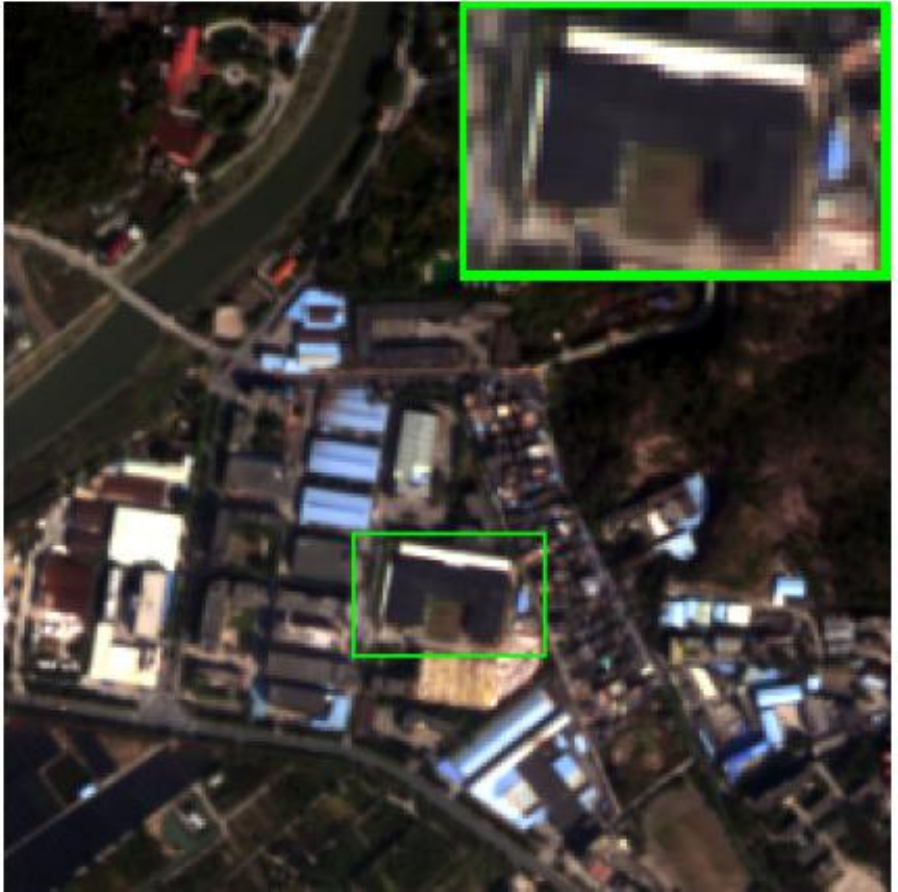}}
				{\includegraphics[width=1\linewidth]{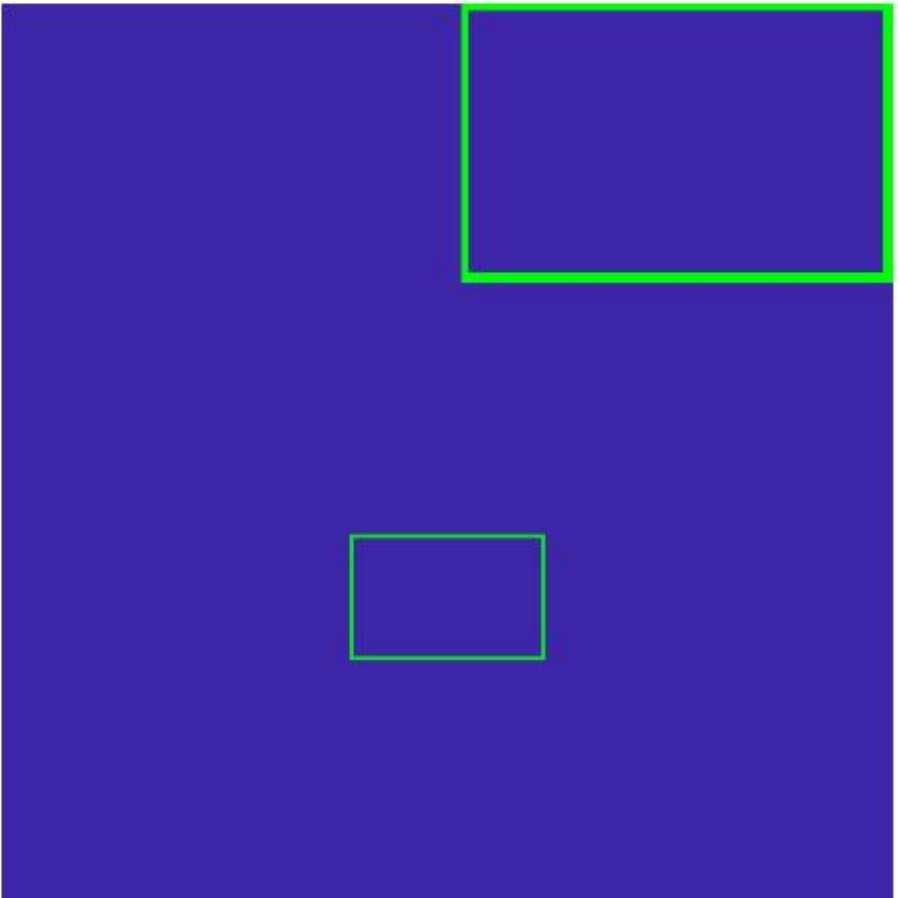}}
				\vspace{4pt}
				{GT}
				\centering
				
			\end{minipage}
		\end{minipage}
		\begin{minipage}[t]{1\linewidth}
			{\includegraphics[width=1\linewidth]{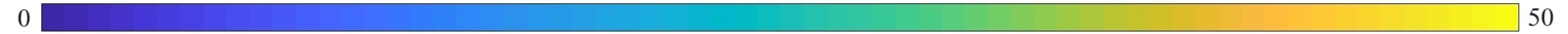}}
			\centering
		\end{minipage}
	\end{center}
    \vspace{-7pt}
	\caption{Results on a reduced-resolution example from the GF2 dataset. Rows 1 and 3: Natural color images. Rows 2 and 4: AEMs for spectral band 3. \label{gf2_v}}
\end{figure*}

\subsubsection{Datasets}
For the pansharpening task \cite{9245579}, we conduct experiments using the widely recognized WorldView-3 (WV3) and GaoFen-2 (GF2) datasets. The WV3 dataset consists of instances acquired by the sensor aboard the WV3 satellite. This sensor captures data across eight spectral bands, covering wavelengths from 0.4 to 1 ${\upmu}$m, with a spatial resolution of 1.2 meters. The images in the GF2 dataset are collected by the sensor onboard the GF2 satellite, which records data across four spectral bands within the wavelength range of 0.4 to 0.9 ${\upmu}$m. Additionally, this sensor provides a spatial resolution of 4 meters. Both datasets utilized in this study are sourced from the PanCollection\footnote{\url{https://github.com/liangjiandeng/PanCollection}}. The data generation process adheres strictly to Wald's protocol \cite{1997Fusion}, with comprehensive details provided in \cite{9844267}. 
Specifically, the WV3 dataset includes 10000 training samples, with 90\% allocated for training and 10\% for validation. Additionally, it contains 20 reduced-resolution and 20 full-resolution testing samples. Each training sample comprises an image triplet in the PAN/LRMS/GT format, with dimensions of $64\times64$, $16\times16\times8$, and $64\times64\times8$, respectively. The reduced-resolution testing samples include PAN/LRMS/GT image triplets sized $256\times256$, $64\times64\times8$, and $256\times256\times8$, respectively. Additionally, the full-resolution testing samples consist of image pairs in the PAN/LRMS format, with sizes of $512\times512$ and $128\times128\times8$, respectively.
In the GF2 dataset, there are 22010 training samples, divided into 90\% for training and 10\% for validation. Additionally, this dataset includes 20 reduced-resolution and 20 full-resolution testing samples. Each training sample contains a PAN/LRMS/GT image triplet of sizes $64\times64$, $16\times16\times4$, and $64\times64\times4$, respectively. The reduced-resolution testing samples comprise PAN/LRMS/GT image triplets sized $256\times256$, $64\times64\times4$, and $256\times256\times4$, respectively. Additionally, the full-resolution testing samples consist of PAN/LRMS image pairs sized $512\times512$ and $128\times128\times4$. The primary distinction between the WV3 and GF2 datasets lies in the number of spectral bands included in their multi-spectral images.

\subsubsection{Benchmarks}
We compare FusionMamba with representative pansharpening techniques, including four traditional approaches: TV \cite{palsson2013new}, GLP-HPM \cite{6616569}, GLP-FS \cite{vivone2018full}, and BDSD-PC \cite{2019Robust}; and ten DL-based methods: PanNet \cite{8237455}, MSDCNN \cite{8127731}, BDPN \cite{8667448}, FusionNet \cite{2020Detail}, MUCNN \cite{10.1145/3474085.3475600}, LAGNet \cite{jin2022aaai}, PMACNet \cite{9764690}, U2Net \cite{10.1145/3581783.3612084}, Pan-Mamba \cite{he2024pan}, and CANNet \cite{Duan_2024_CVPR}. For fairness, all DL-based methods are trained using the same Nvidia GPU 3090 and PyTorch environment.

\subsubsection{Quality Indices}
In accordance with the research standards of pansharpening, we utilize four quality indices, namely PSNR \cite{wang2004image}, Q2n \cite{2009Hypercomplex}, SAM \cite{yuhas1992discrimination}, and ERGAS \cite{1997Fusion}, to evaluate the results on reduced-resolution samples. The ideal values for these indices are +$\infty$, 1, 0, and 0, respectively. For full-resolution samples, we employ ${\rm{D}}_{\rm{\lambda}}$, ${\rm{D}}_{\rm{s}}$, and QNR \cite{6998089} as evaluation metrics, with ideal values of 0, 0, and 1, respectively. Notably, QNR, which combines ${\rm{D}}_{\rm{\lambda}}$ and ${\rm{D}}_{\rm{s}}$, provides a comprehensive measure of overall fusion quality.

\subsubsection{Settings} 
In the pansharpening task, we set ${C}$ to 32 and ${N}$ to 8. Additionally, we utilize the PixelShuffle technique \cite{Shi_2016_CVPR1} for up-sampling. During the training of our networks on the WV3 and GF2 datasets, the number of epochs is configured as 420 and 300, respectively. Besides, the batch size and initial learning rate are uniformly configured as 32 and $5\times 10^{-4}$, respectively. Furthermore, we employ the Adam optimizer, with the learning rate halved every 200 epochs. As for other DL-based methods, we adhere to the default settings specified in the related papers or source codes. 

\subsubsection{Results}
The quantitative evaluation results for the WV3 and GF2 datasets, respectively presented in Tables~\ref{rr} and \ref{gf2}, indicate that FusionMamba achieves the best overall performance on both the reduced-resolution and full-resolution testing samples. Given that the indicator values are approaching their limits, our method demonstrates significant improvements over other techniques. Additionally, the qualitative evaluation results on both datasets, as depicted in Figs.~\ref{rr_v} and \ref{gf2_v}, reveal that the FusionMamba's absolute error maps (AEMs) are the closest to the GT images. Consequently, our method exhibits superior effectiveness in the pansharpening task.

\subsection{Hyper-spectral Pansharpening}

\subsubsection{Datasets}
We conduct experiments on three widely used hyper-spectral pansharpening datasets \cite{7284770}: Pavia, Botswana, and Washington D.C. (WDC). The Pavia dataset includes images acquired by the Reflective Optics System Imaging Spectrometer (ROSIS) sensor, which records data across 115 spectral bands within the wavelength range of 0.4 to 0.9 $\upmu$m. Additionally, this sensor offers a spatial resolution of 1.3 meters. The images in the Botswana dataset are collected by the Hyperion sensor aboard the Earth Observing-1 (EO-1) satellite, operated by the National Aeronautics and Space Administration (NASA). This sensor captures data across 242 spectral bands, spanning wavelengths from 0.4 to 2.5 $\upmu$m, with a spatial resolution of 30 meters. The WDC dataset comprises images captured by the Hyper-spectral Digital Imagery Collection Experiment (HY-DICE) sensor, which records data across 210 spectral bands, covering a wavelength range from 0.4 to 2.4 $\upmu$m, with a spatial resolution of 1 meter. The hyper-spectral pansharpening datasets used in this study are sourced from the HyperPanCollection\footnote{\url{https://github.com/liangjiandeng/HyperPanCollection}}. The data generation process follows Wald's protocol \cite{1997Fusion}, with a detailed explanation provided in \cite{9870551}. 
Specifically, the Pavia dataset contains 1680 training samples, of which 90\% are allocated for training and 10\% for validation. Additionally, this dataset includes two testing samples. Each training sample consists of an image triplet in the PAN/LRHS/GT format, with sizes of $64\times64$, $16\times16\times102$, and $64\times64\times102$, respectively. The testing samples comprise PAN/LRHS/GT image triplets sized $400\times400$, $100\times100\times102$, and $400\times400\times102$, respectively.
The Botswana dataset contains 967 training samples, divided into 83\% for training and 17\% for validation, alongside four testing samples. Each training sample consists of a PAN/LRHS/GT image triplet with dimensions of $64\times64$, $16\times16\times145$, and $64\times64\times145$, respectively. The testing samples comprise PAN/LRHS/GT image triplets of sizes $128\times128$, $32\times32\times145$, and $128\times128\times145$. 
In the WDC dataset, there are 1024 training samples, divided into 90\% for training and 10\% for validation. Additionally, this dataset includes four testing samples. Each training sample contains a PAN/LRHS/GT image triplet of sizes $64\times64$, $16\times16\times191$, and $64\times64\times191$, respectively. The testing samples consist of PAN/LRHS/GT image triplets sized $128\times128$, $32\times32\times191$, and $128\times128\times191$.

\subsubsection{Benchmarks}
The proposed method is compared with several representative techniques, including four traditional approaches: GLP \cite{aiazzi2006mtf}, GSA \cite{4305344}, CNMF \cite{6049465}, and Hysure \cite{7000523}; as well as five DL-based methods: HyperPNN \cite{8731649}, HSpeNet series \cite{9200718}, FusionNet \cite{2020Detail}, Hyper-DSNet \cite{9870551}, and FPFNet \cite{10298274}. For a fair comparison, all DL-based methods are trained using the same Nvidia GPU 3090 and PyTorch environment.

\begin{table*}[t]	
	\centering\renewcommand\arraystretch{1.2}\setlength{\tabcolsep}{3.1pt}
	\belowrulesep=0pt\aboverulesep=0pt
	\caption{Results on testing samples of the WDC, Botswana, and Pavia datasets, which belong to the hyper-spectral pansharpening task.}
	\begin{tabular}{l|c|ccccc|ccccc|ccccc}
		\toprule
		
		\multirow{2}{*}{\textbf{Methods}} & 
		\multirow{2}{*}{\textbf{Params}} & 
		\multicolumn{5}{c|}{\textbf{Pavia}} & 
		\multicolumn{5}{c|}{\textbf{Botswana}} &
		\multicolumn{5}{c}{\textbf{WDC}}
		\\
		\cmidrule(lr){3-7}\cmidrule(lr){8-12}\cmidrule(lr){13-17}
		&\multicolumn{1}{c|}{} 
		&\multicolumn{1}{c}{PSNR} 
		&\multicolumn{1}{c}{CC} 
		&\multicolumn{1}{c}{SSIM} 
		&\multicolumn{1}{c}{SAM} 
		&\multicolumn{1}{c|}{ERGAS}
		&\multicolumn{1}{c}{PSNR} 
		&\multicolumn{1}{c}{CC} 
		&\multicolumn{1}{c}{SSIM} 
		&\multicolumn{1}{c}{SAM} 
		&\multicolumn{1}{c|}{ERGAS} 
		&\multicolumn{1}{c}{PSNR} 
		&\multicolumn{1}{c}{CC} 
		&\multicolumn{1}{c}{SSIM} 
		&\multicolumn{1}{c}{SAM} 
		&\multicolumn{1}{c}{ERGAS}  
		\\
		
		\midrule
		
		\textbf{GLP} \cite{aiazzi2006mtf} & $-$ 
		& 31.944 & 0.935 & 0.749 & 6.099 & 4.909
		& 32.559 & 0.951 & 0.837 & 1.383 & 1.207 
		& 27.946 & 0.934 & 0.761 & 6.546 & 5.110 
		\\ 
		\textbf{GSA} \cite{4305344} & $-$ 
		& 31.501 & 0.937 & 0.722 & 6.282 & 4.978 
		& 31.739 & 0.939 & 0.828 & 1.389 & 1.386
		& 24.462 & 0.906 & 0.671 & 7.846 & 6.079  
		\\ 
		\textbf{CNMF} \cite{6049465} & $-$ 
		& 31.184 & 0.894 & 0.659 & 6.953 & 6.263 
		& 30.220 & 0.917 & 0.788 & 1.934 & 1.718 
		& 24.604 & 0.890 & 0.678 & 8.441 & 6.682 
		\\ 
		\textbf{Hysure} \cite{7000523} & $-$ 
		& 32.208 & 0.921 & 0.730 & 6.240 & 5.474  
		& 30.610 & 0.928 & 0.796 & 1.747 & 1.595 
		& 25.598 & 0.913 & 0.718 & 7.254 & 5.834 
		\\ 
		\midrule
		\textbf{HyperPNN} \cite{8731649} & 0.13-0.14M
		& 33.394 & 0.963 & 0.827 & 4.566 & 3.750  
		& 33.114 & 0.961 & 0.873 & 1.366 & 1.195
		& 29.258 & 0.945 & 0.860 & 4.051 & 5.749 
		\\ 
		\textbf{HSpeNet1} \cite{9200718} & 0.18-0.19M 
		& 33.612 & 0.964 & 0.824 & 4.690 & 3.721  
		& 31.746 & 0.942 & 0.844 & 1.456 & 1.663 
		& 29.634 & 0.960 & 0.870 & 4.039 & 4.266 
		\\ 
		\textbf{HSpeNet2} \cite{9200718} & 0.11-0.13M 
		& 33.472 & 0.962 & 0.819 & 4.642 & 3.818  
		& 32.575 & 0.953 & 0.849 & 1.400 & 1.348 
		& 29.700 & 0.961 & 0.872 & 4.009 & 4.261 
		\\ 
		\textbf{FusionNet} \cite{2020Detail} & 0.21-0.26M 
		& \underline{34.739} & \underline{0.969} & 0.847 & 4.462 & 3.446 
		& 32.506 & 0.952 & 0.850 & 1.397 & 1.367 
		& 29.696 & 0.959 & 0.866 & \underline{3.917} & 4.339 
		\\ 
		\textbf{HyperDSNet} \cite{9870551} & 0.18-0.31M 
		& 34.376 & \underline{0.969} & \underline{0.849} 
		& \underline{4.295} & \underline{3.434}
		& \underline{33.538} & \underline{0.964} & \underline{0.876} 
		& \underline{1.305} & \underline{1.126} 
		& {30.232} & \underline{0.964} & \underline{0.875} 
		& {4.102} & \underline{3.943} 
		\\ 
		\textbf{FPFNet} \cite{10298274} & 3.00-3.06M 
		& 33.581 & 0.959 & 0.825 & 4.627 & 3.931 
		& 33.451 & 0.962 & 0.871 & 1.369 & 1.135 
		& \underline{30.291} & 0.957 & 0.855 & 4.440 & 4.250 
		\\    
		\textbf{FusionMamba} & 0.44-0.51M 
		& \textbf{35.628} & \textbf{0.973} & \textbf{0.872} & \textbf{3.963} & \textbf{3.171}  
		& \textbf{33.943} & \textbf{0.966} & \textbf{0.881} & \textbf{1.277} & \textbf{1.076}
		& \textbf{31.860} & \textbf{0.965} & \textbf{0.881} & \textbf{3.755} & \textbf{3.882}
		\\      
		\midrule
		\textbf{Ideal Values} 
		& $-$
		&\multicolumn{1}{c}{\textbf{+$\infty$}}
		&\multicolumn{1}{c}{\textbf{\textbf{1}}}
		&\multicolumn{1}{c}{\textbf{\textbf{1}}}
		&\multicolumn{1}{c}{\textbf{\textbf{0}}}
		&\multicolumn{1}{c|}{\textbf{\textbf{0}}}
		&\multicolumn{1}{c}{\textbf{+$\infty$}}
		&\multicolumn{1}{c}{\textbf{\textbf{1}}}
		&\multicolumn{1}{c}{\textbf{\textbf{1}}}
		&\multicolumn{1}{c}{\textbf{\textbf{0}}}
		&\multicolumn{1}{c|}{\textbf{\textbf{0}}}
		&\multicolumn{1}{c}{\textbf{+$\infty$}}
		&\multicolumn{1}{c}{\textbf{\textbf{1}}}
		&\multicolumn{1}{c}{\textbf{\textbf{1}}}
		&\multicolumn{1}{c}{\textbf{\textbf{0}}}
		&\multicolumn{1}{c}{\textbf{\textbf{0}}}
		\\ 
		\bottomrule
	\end{tabular}
	\label{hsp}	
\end{table*}

\begin{figure*}[t]
	\begin{center}
		\begin{minipage}[t]{1\linewidth}
			\begin{minipage}[t]{0.095\linewidth}
				{\includegraphics[width=1\linewidth]{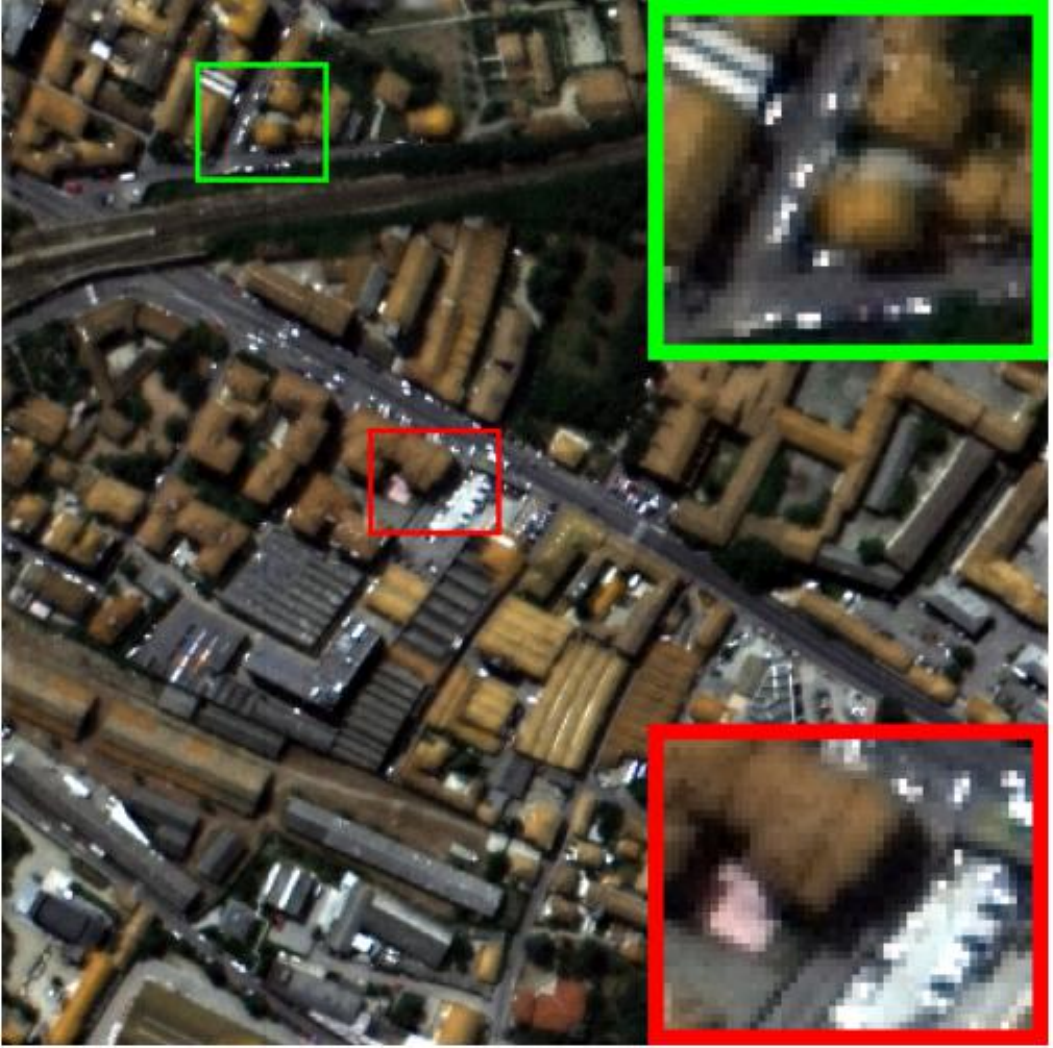}}
				\vspace{2pt}
				{\includegraphics[width=1\linewidth]{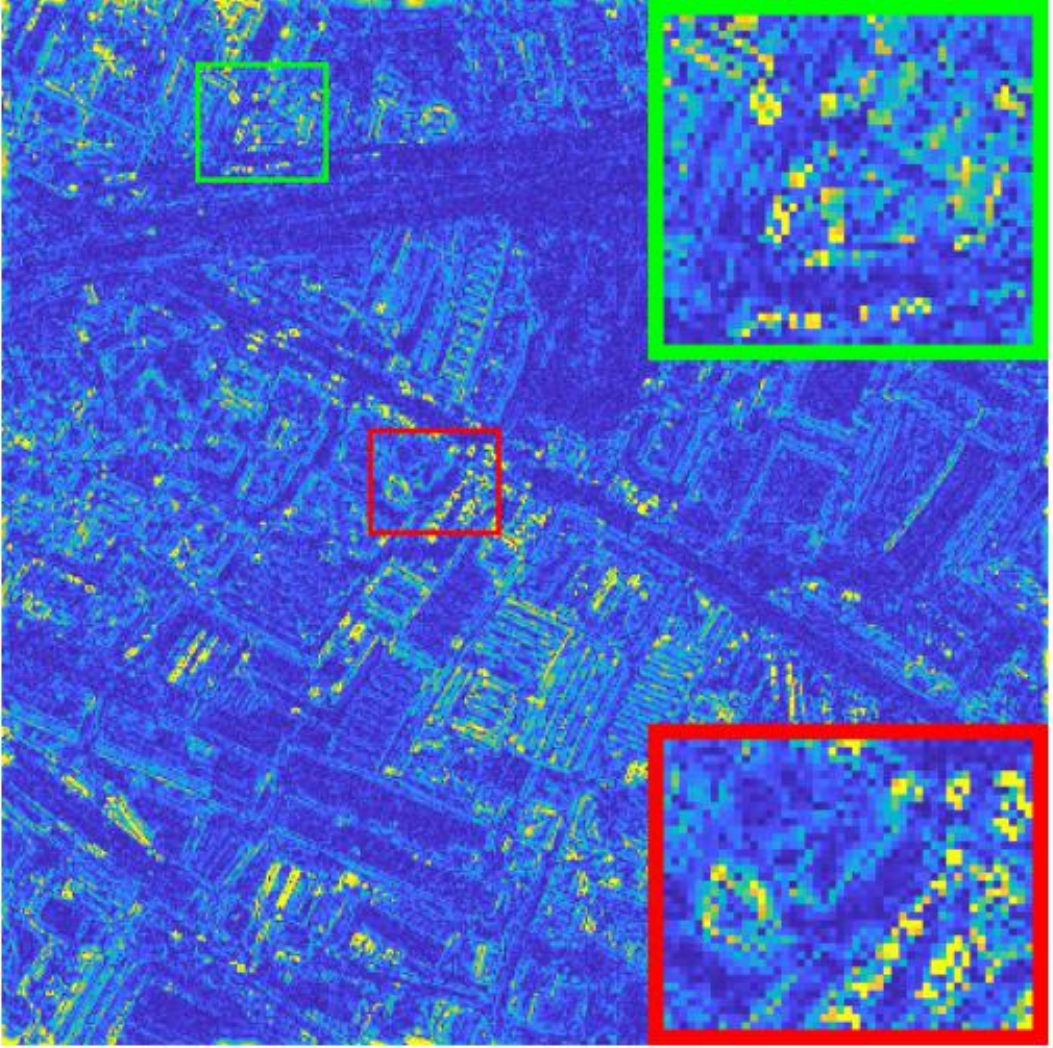}}
				{\includegraphics[width=1\linewidth]{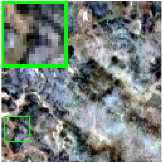}}
				\vspace{2pt}
				{\includegraphics[width=1\linewidth]{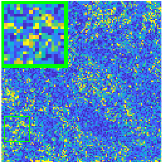}}
				{\includegraphics[width=1\linewidth]{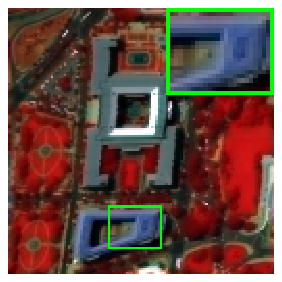}}
				{\includegraphics[width=1\linewidth]{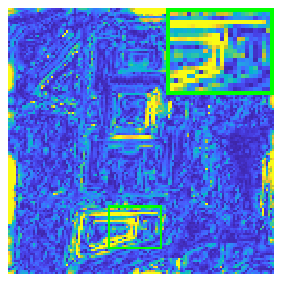}}
				\vspace{2pt}
				\scriptsize{GLP}
				\centering
				
			\end{minipage}
			\begin{minipage}[t]{0.095\linewidth}
				{\includegraphics[width=1\linewidth]{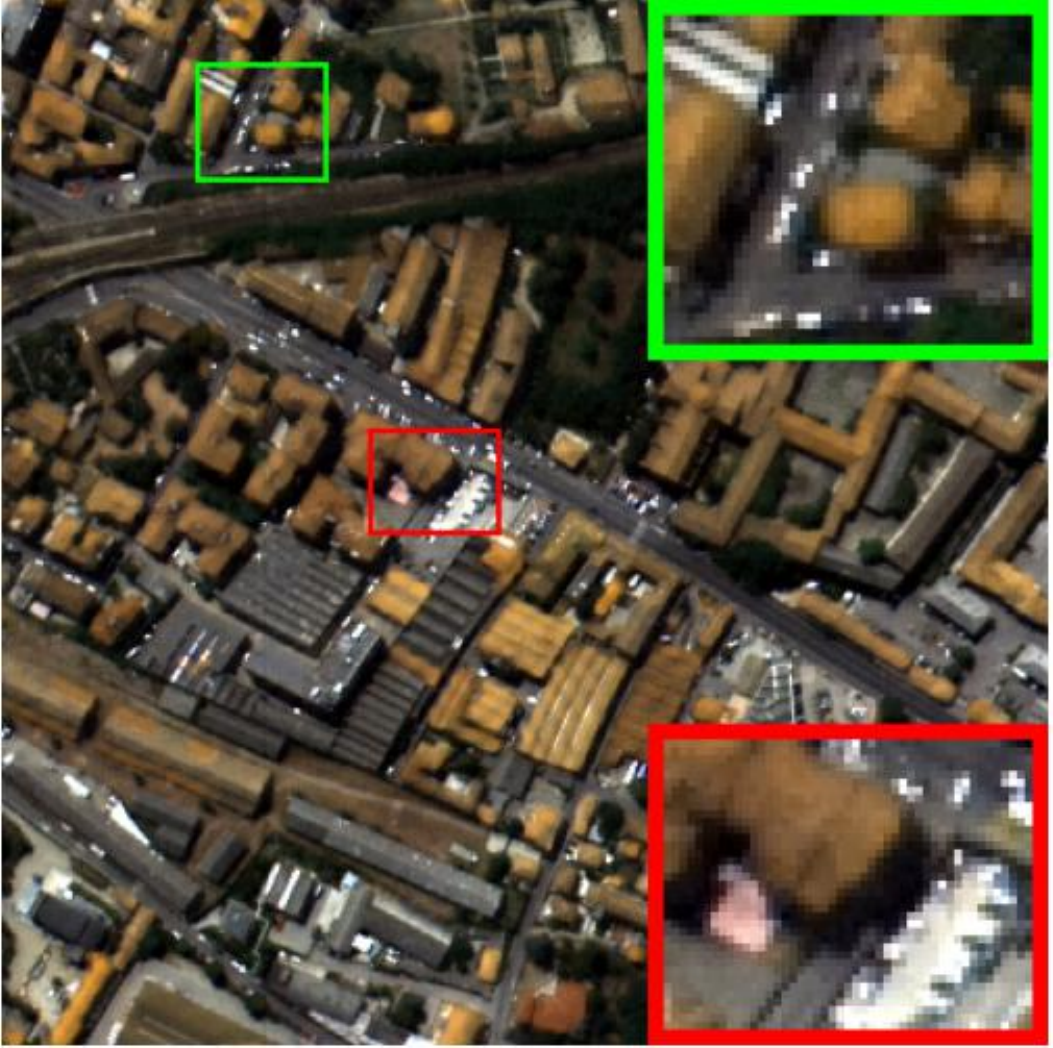}}
				\vspace{2pt}
				{\includegraphics[width=1\linewidth]{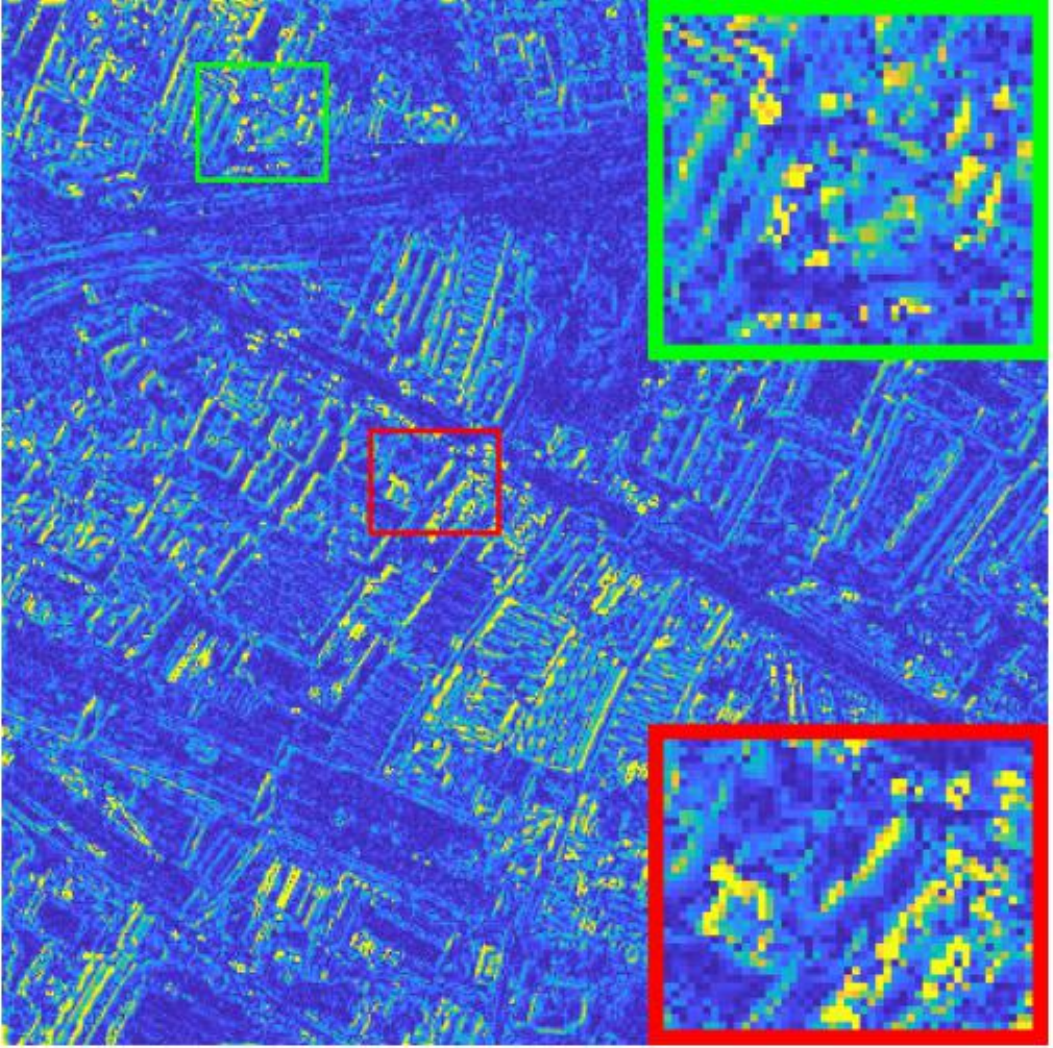}}
				{\includegraphics[width=1\linewidth]{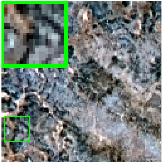}}
				\vspace{2pt}
				{\includegraphics[width=1\linewidth]{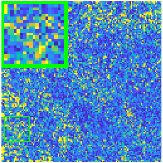}}
				{\includegraphics[width=1\linewidth]{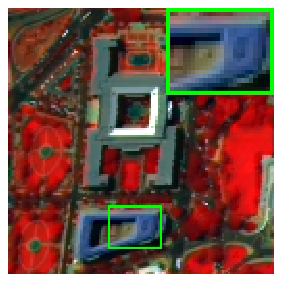}}
				{\includegraphics[width=1\linewidth]{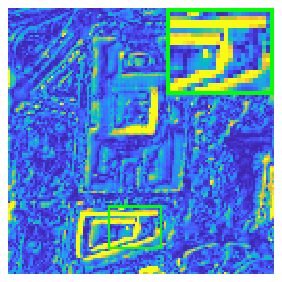}}
				\vspace{2pt}
				\scriptsize{HySure}
				\centering
				
			\end{minipage}
			\begin{minipage}[t]{0.095\linewidth}
				{\includegraphics[width=1\linewidth]{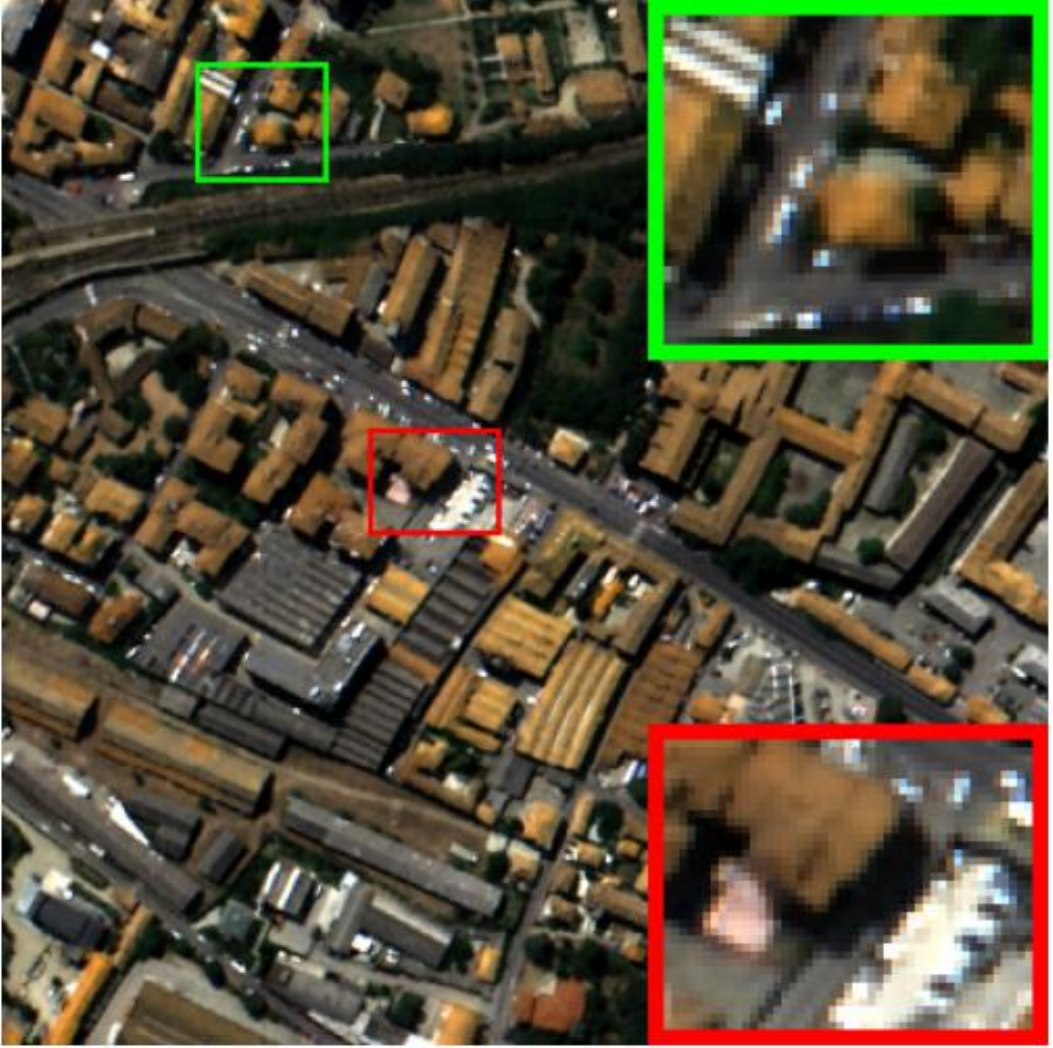}}
				\vspace{2pt}
				{\includegraphics[width=1\linewidth]{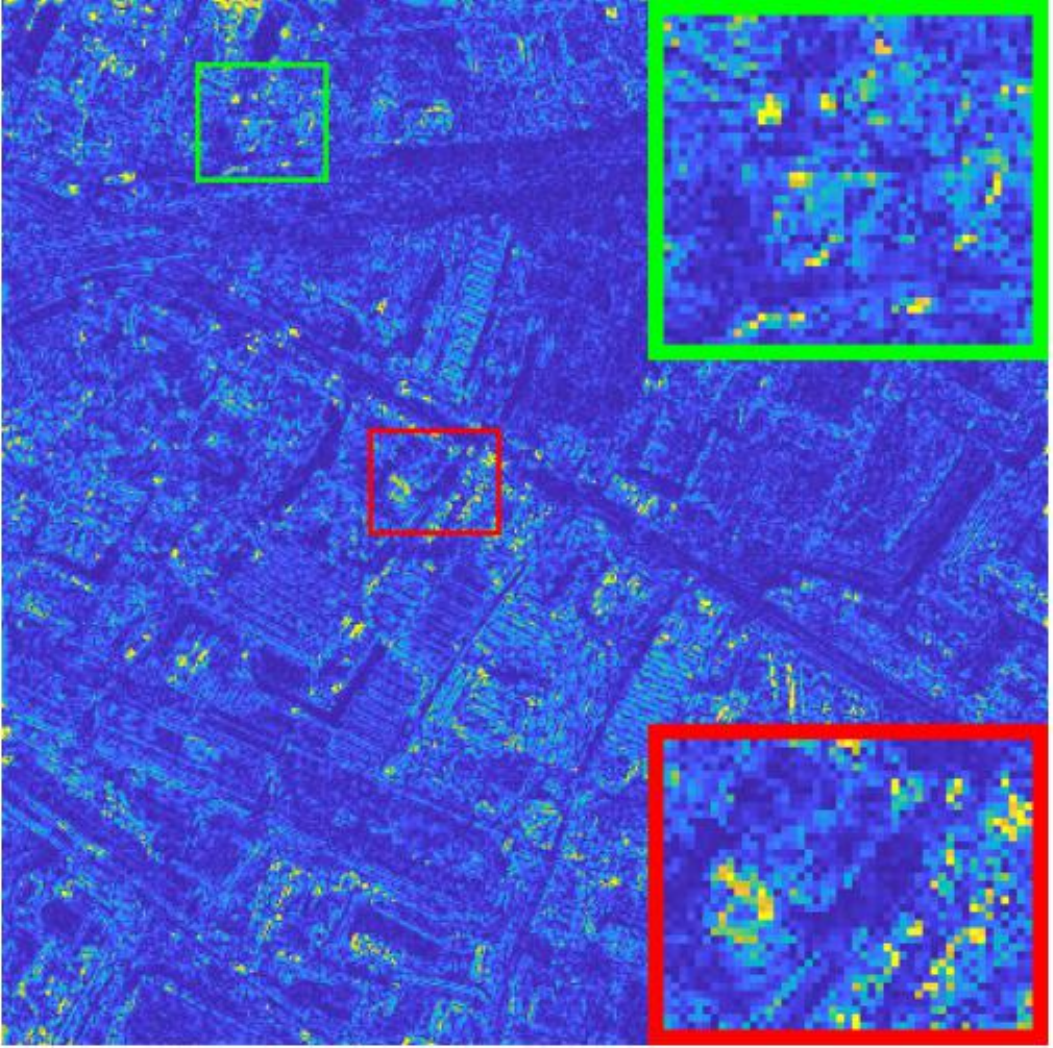}}
				{\includegraphics[width=1\linewidth]{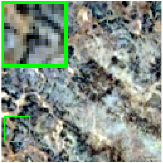}}
				\vspace{2pt}
				{\includegraphics[width=1\linewidth]{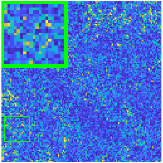}}
				{\includegraphics[width=1\linewidth]{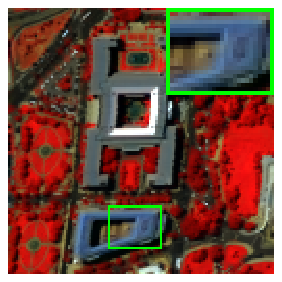}}
				{\includegraphics[width=1\linewidth]{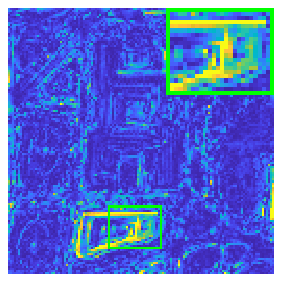}}
				\vspace{2pt}
				\scriptsize{HyperPNN}
				\centering
				
			\end{minipage}
			\begin{minipage}[t]{0.095\linewidth}
				{\includegraphics[width=1\linewidth]{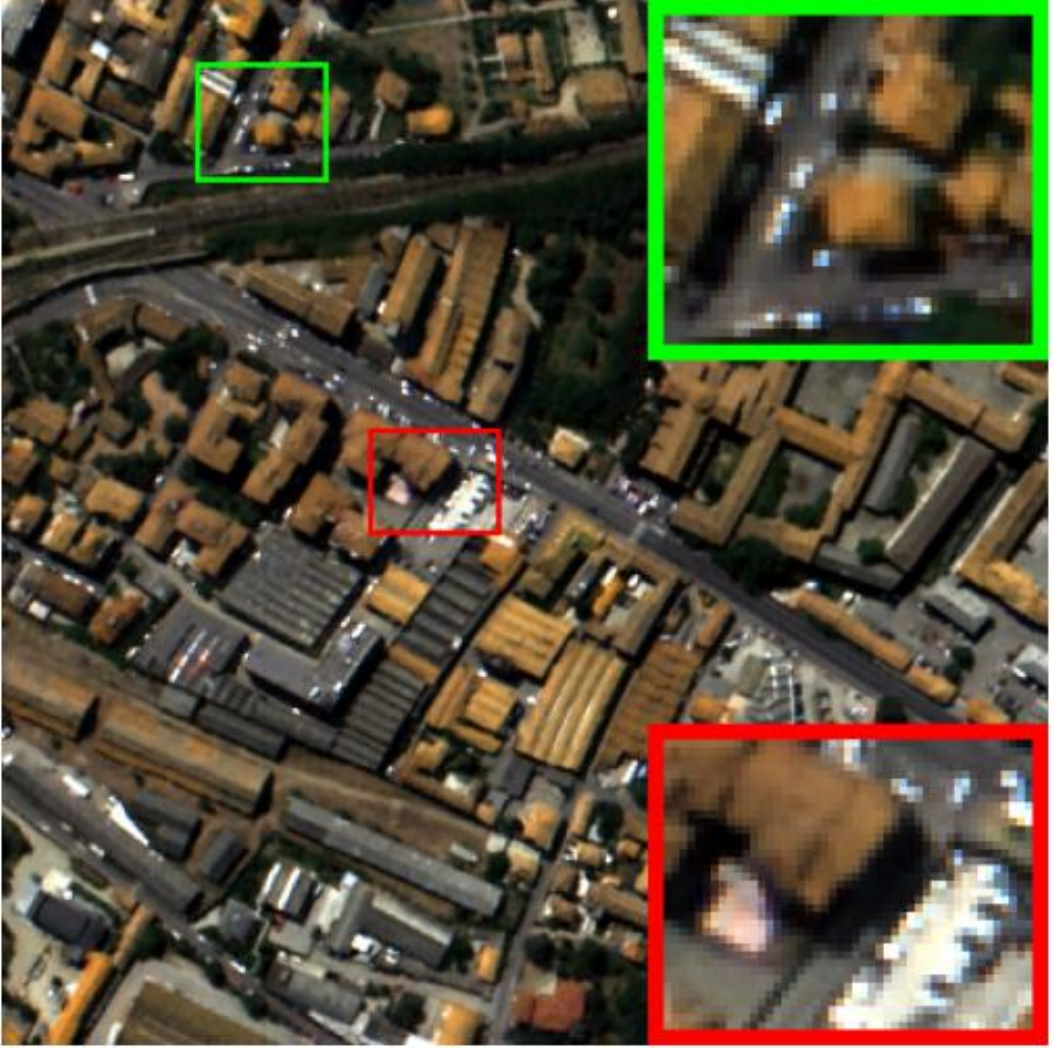}}
				\vspace{2pt}
				{\includegraphics[width=1\linewidth]{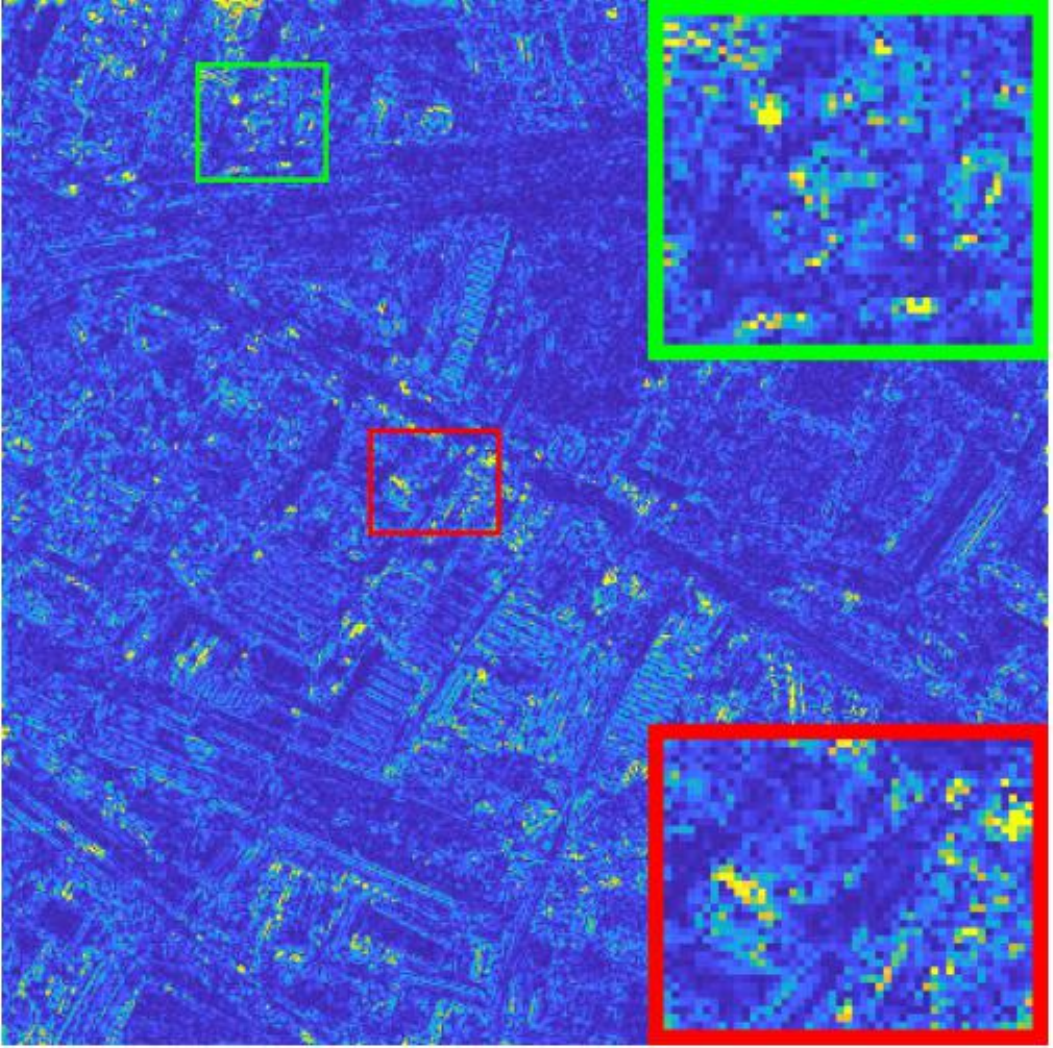}}
				{\includegraphics[width=1\linewidth]{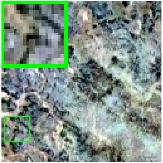}}
				\vspace{2pt}
				{\includegraphics[width=1\linewidth]{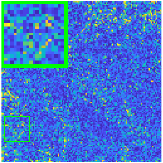}}
				{\includegraphics[width=1\linewidth]{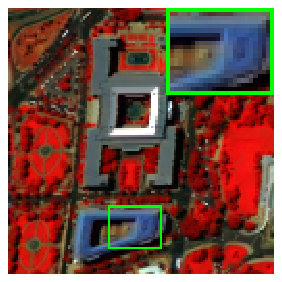}}
				{\includegraphics[width=1\linewidth]{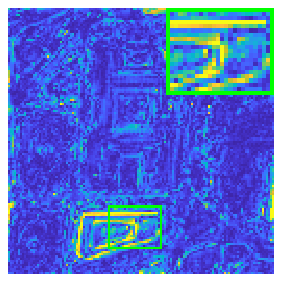}}
				\vspace{2pt}
				\scriptsize{HSpeNet1}
				\centering
				
			\end{minipage}
			\begin{minipage}[t]{0.095\linewidth}
				{\includegraphics[width=1\linewidth]{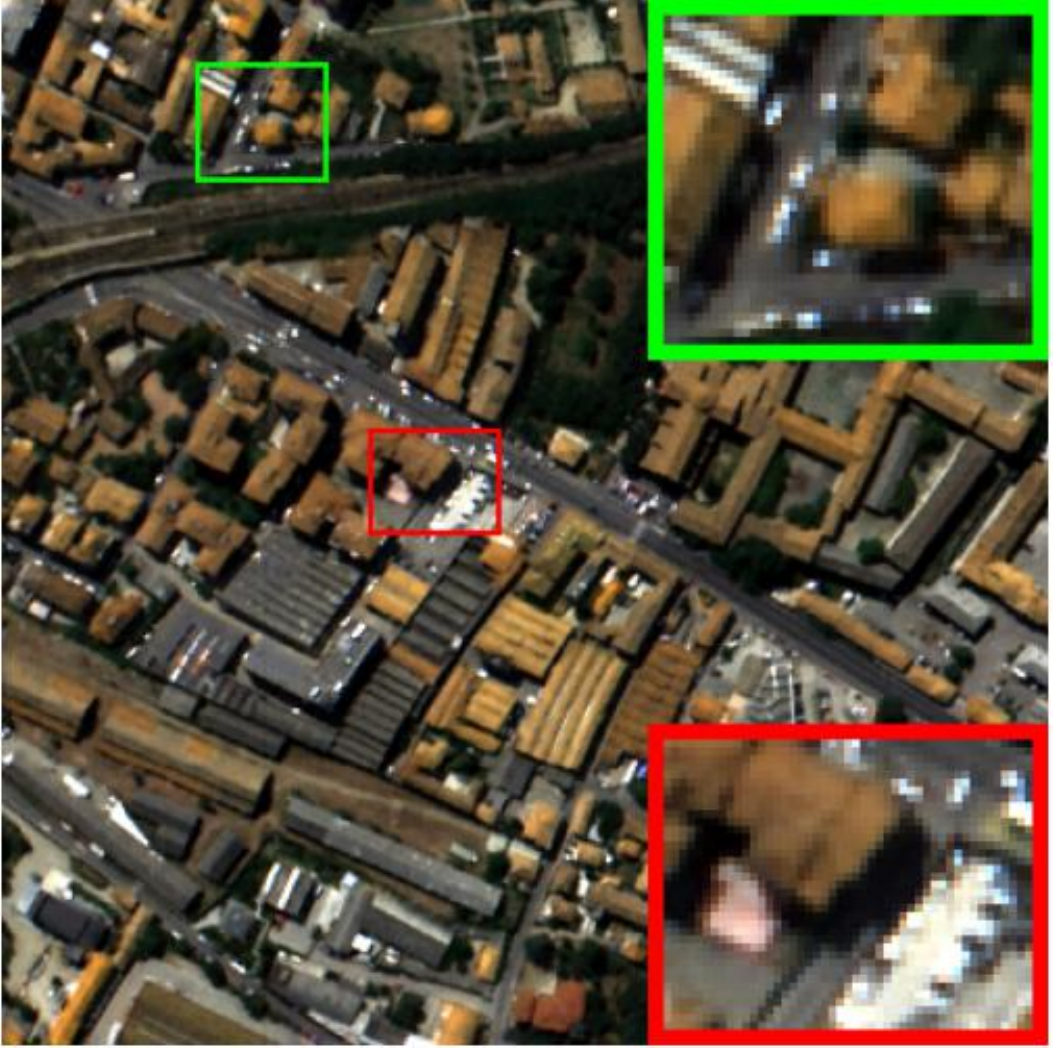}}
				\vspace{2pt}
				{\includegraphics[width=1\linewidth]{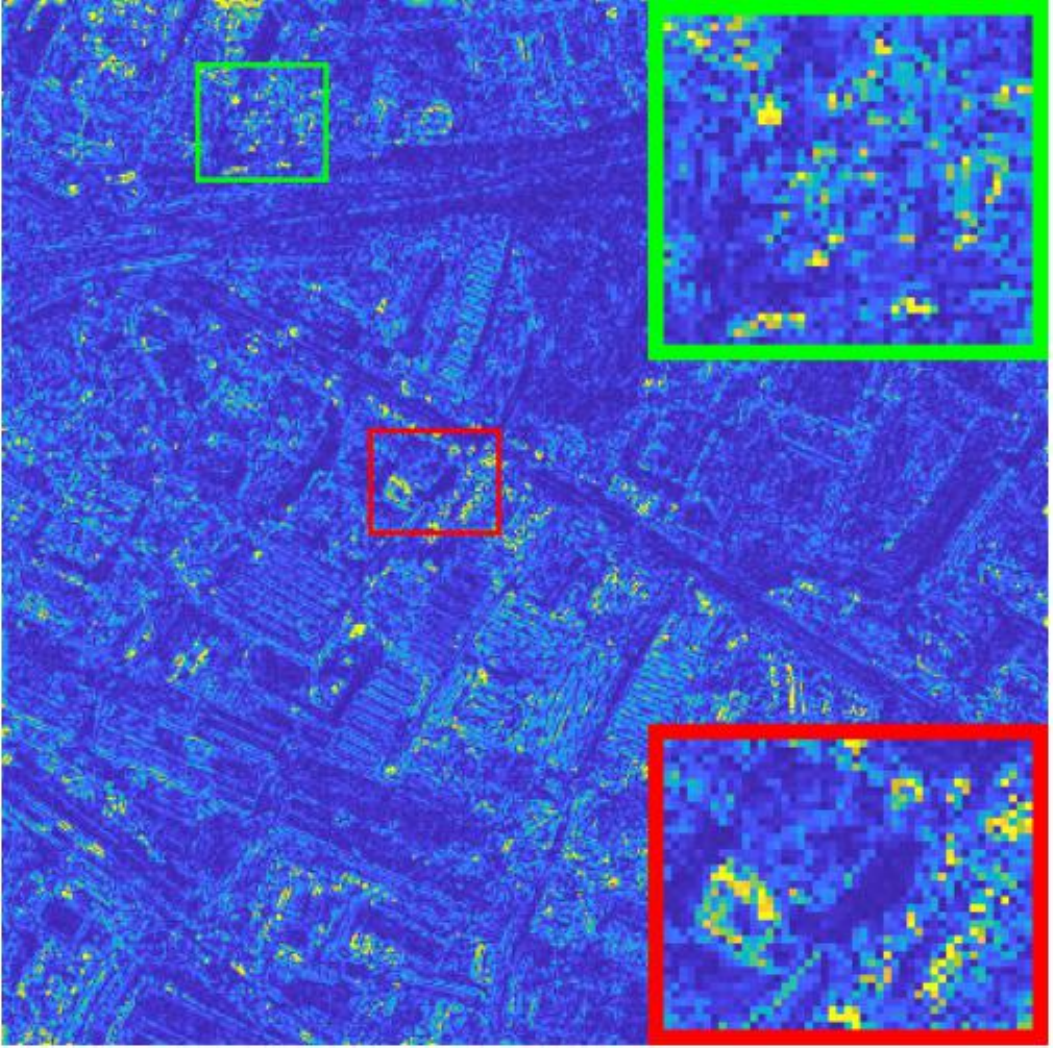}}
				{\includegraphics[width=1\linewidth]{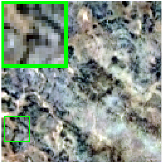}}
				\vspace{2pt}
				{\includegraphics[width=1\linewidth]{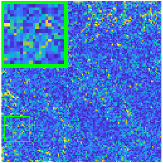}}
				{\includegraphics[width=1\linewidth]{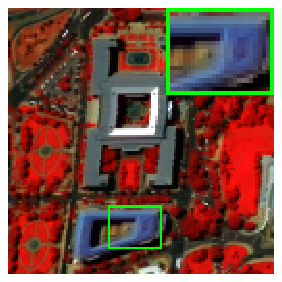}}
				{\includegraphics[width=1\linewidth]{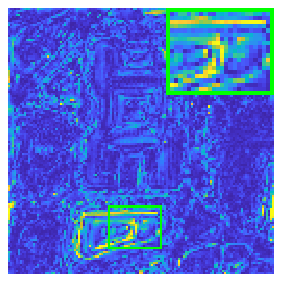}}
				\vspace{2pt}
				\scriptsize{HSpeNet2}
				\centering
				
			\end{minipage}
			\begin{minipage}[t]{0.095\linewidth}
				{\includegraphics[width=1\linewidth]{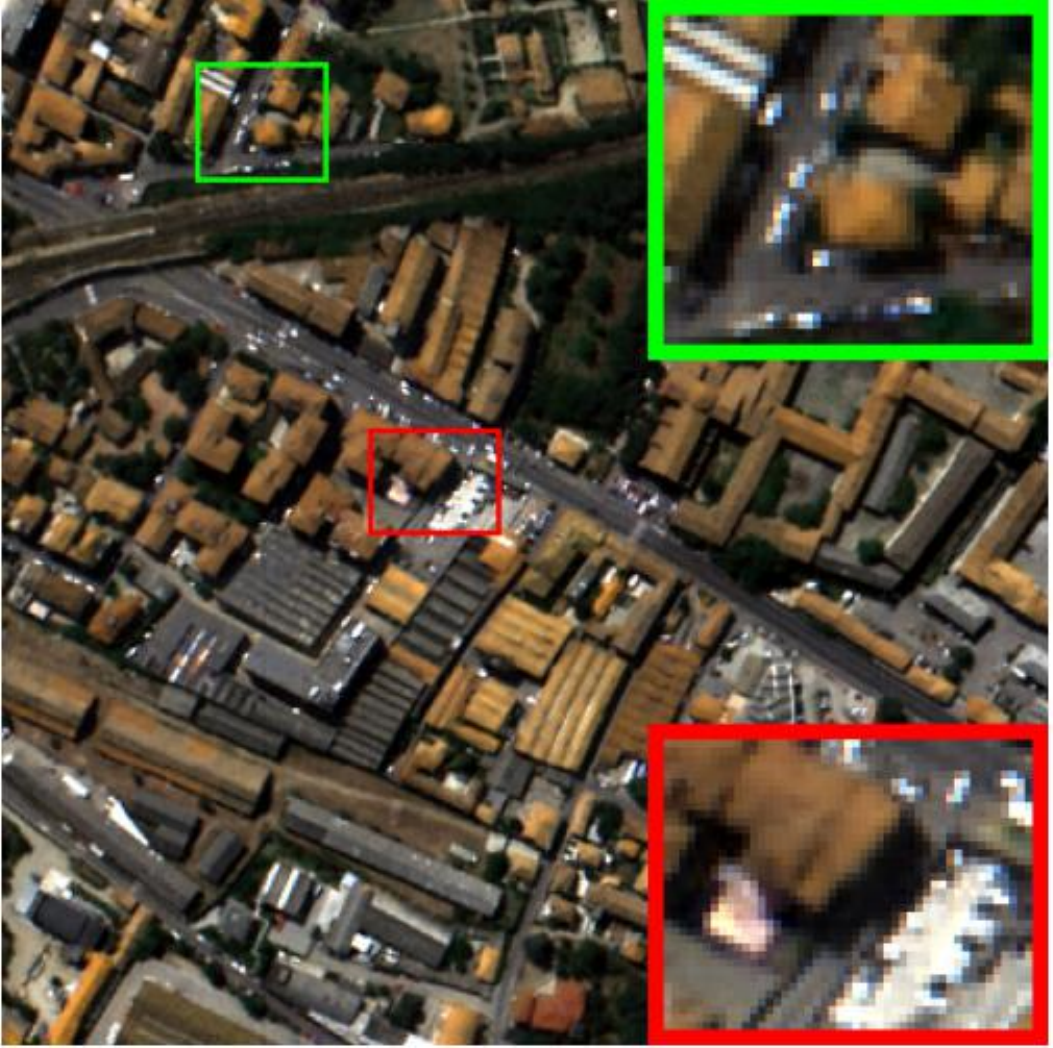}}
				\vspace{2pt}
				{\includegraphics[width=1\linewidth]{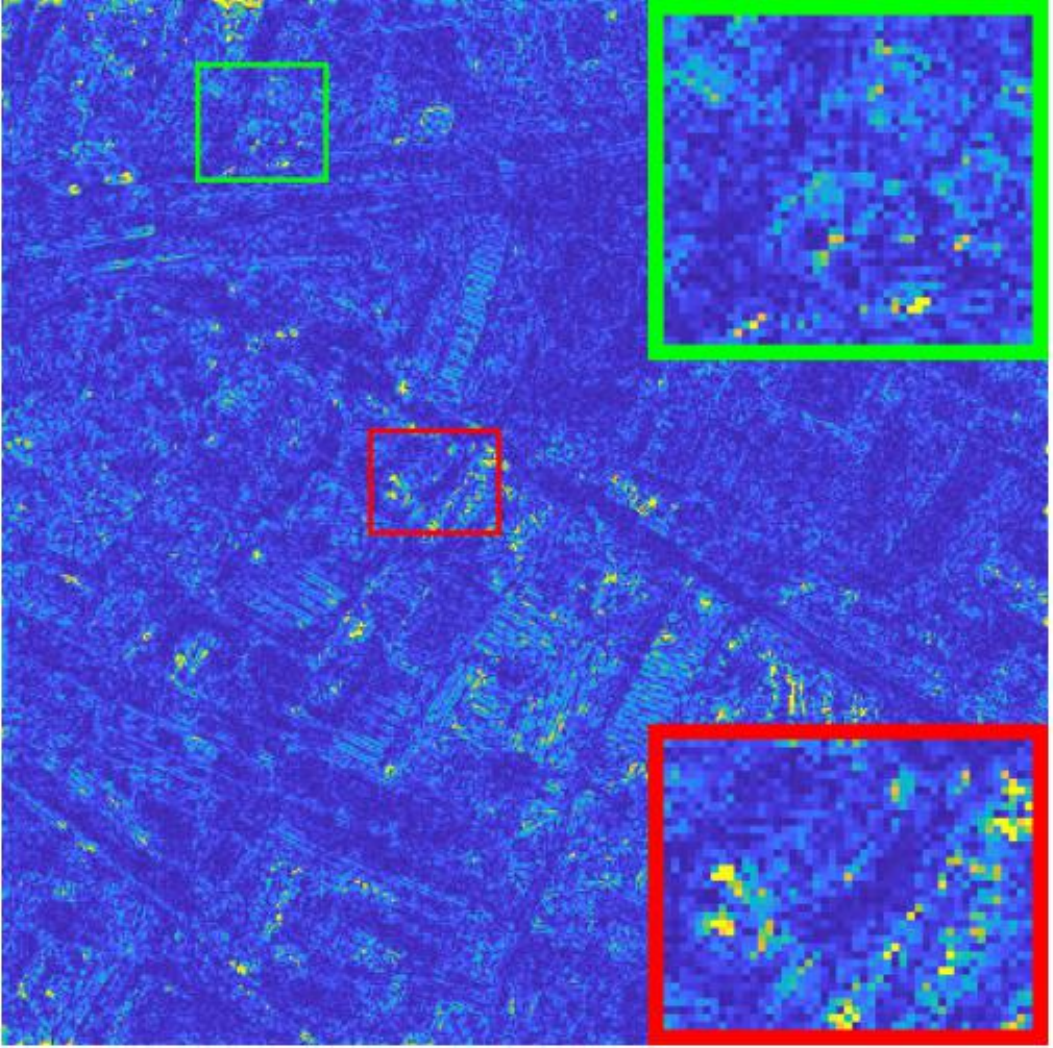}}
				{\includegraphics[width=1\linewidth]{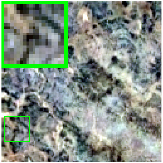}}
				\vspace{2pt}
				{\includegraphics[width=1\linewidth]{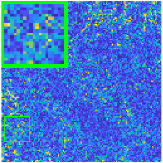}}
				{\includegraphics[width=1\linewidth]{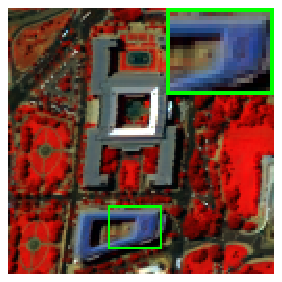}}
				{\includegraphics[width=1\linewidth]{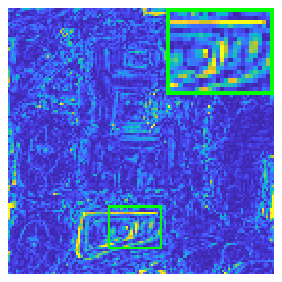}}
				\vspace{2pt}
				\scriptsize{FusionNet}
				\centering
				
			\end{minipage}
			\begin{minipage}[t]{0.095\linewidth}
				{\includegraphics[width=1\linewidth]{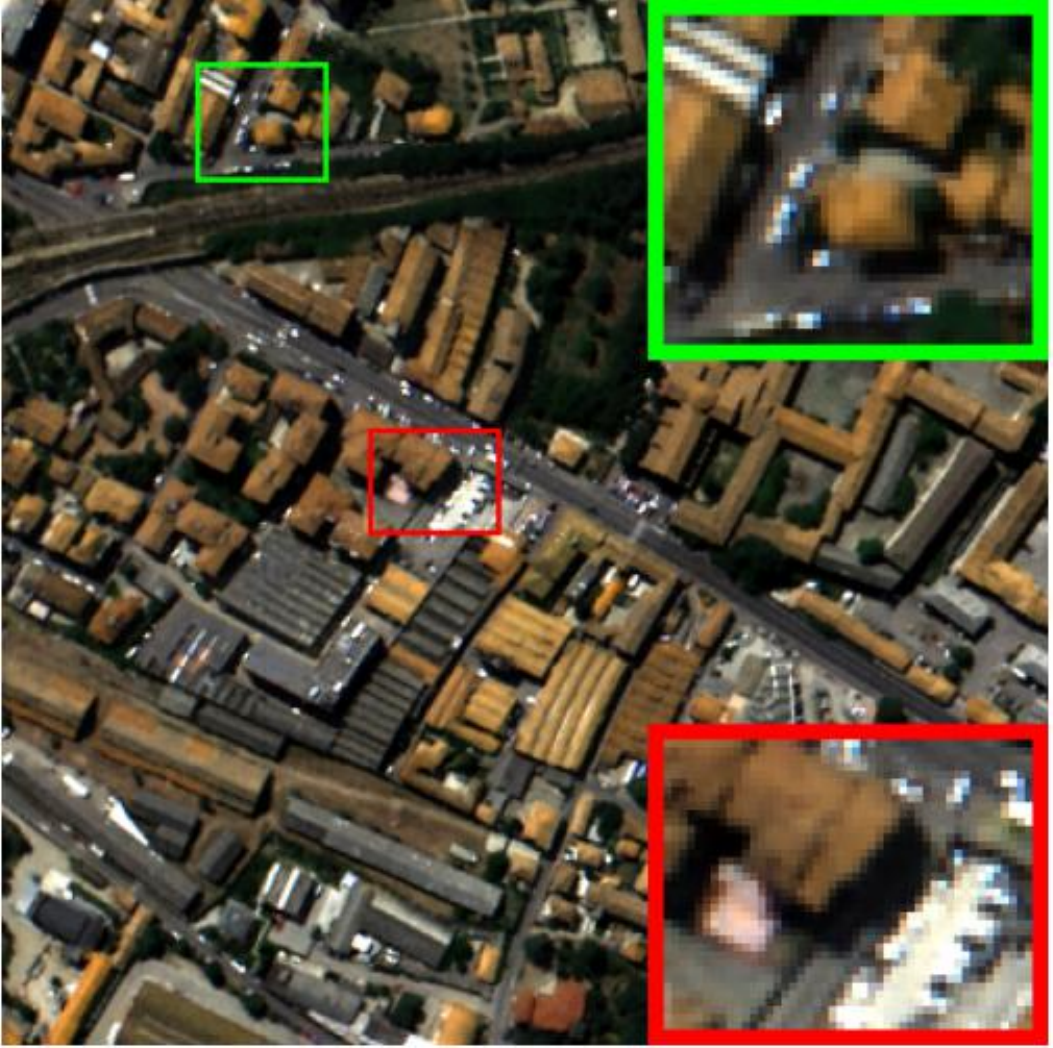}}
				\vspace{2pt}
				{\includegraphics[width=1\linewidth]{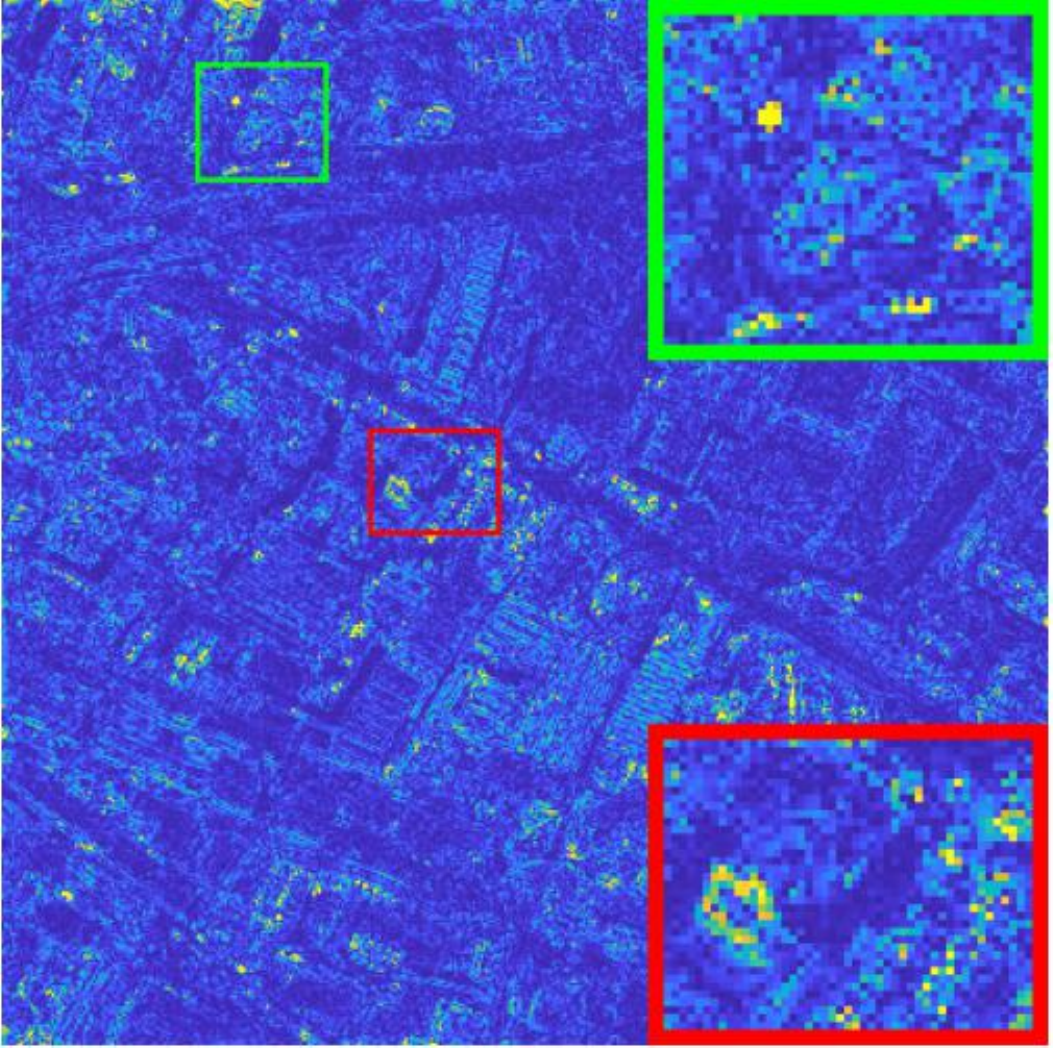}}
				{\includegraphics[width=1\linewidth]{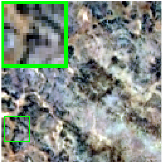}}
				\vspace{2pt}
				{\includegraphics[width=1\linewidth]{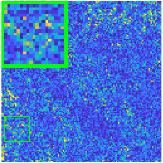}}
				{\includegraphics[width=1\linewidth]{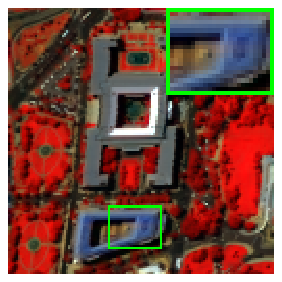}}
				{\includegraphics[width=1\linewidth]{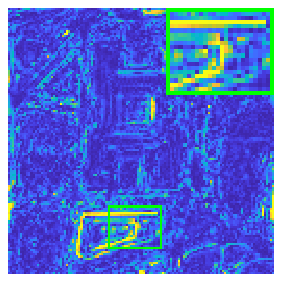}}
				\vspace{2pt}
				\scriptsize{Hyper-DSNet}
				\centering
				
			\end{minipage}
			\begin{minipage}[t]{0.095\linewidth}
				{\includegraphics[width=1\linewidth]{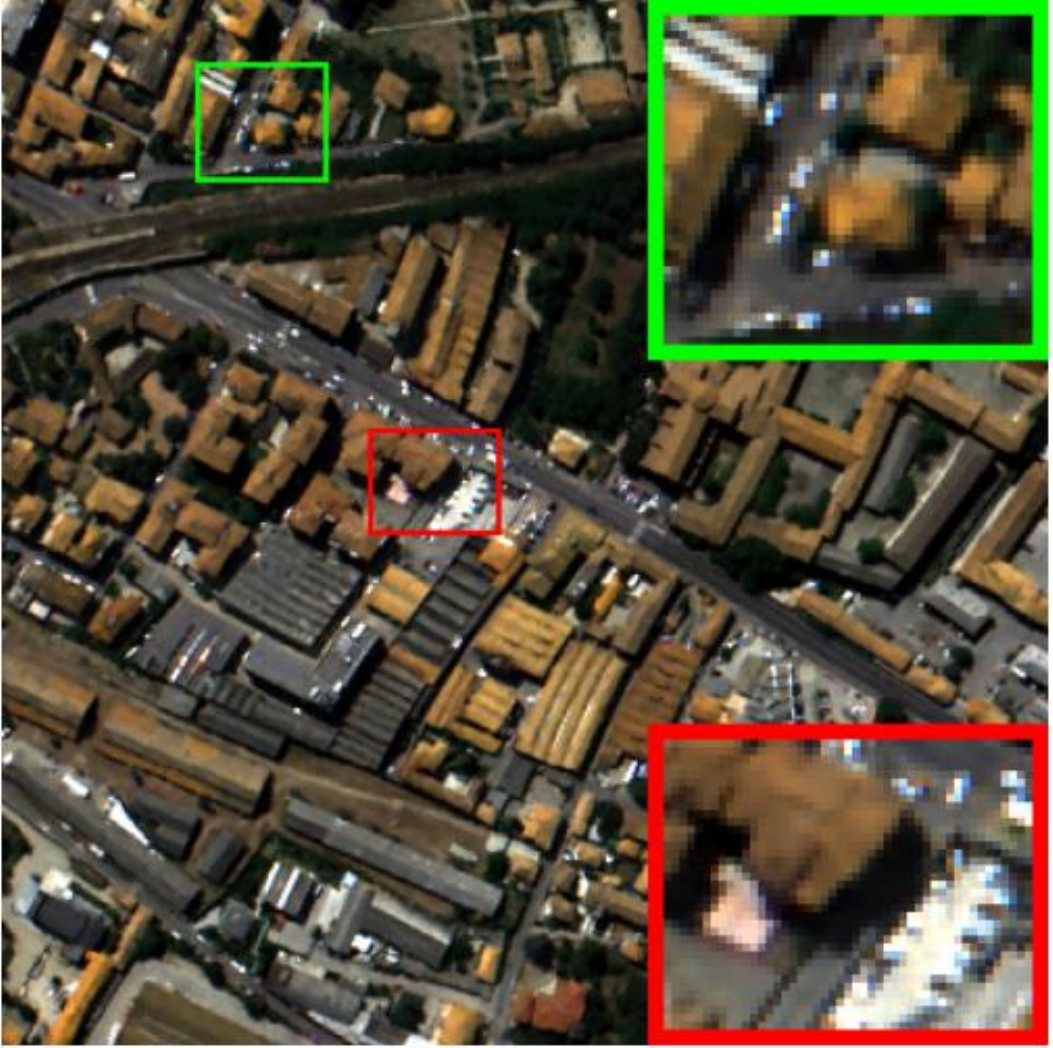}}
				\vspace{2pt}
				{\includegraphics[width=1\linewidth]{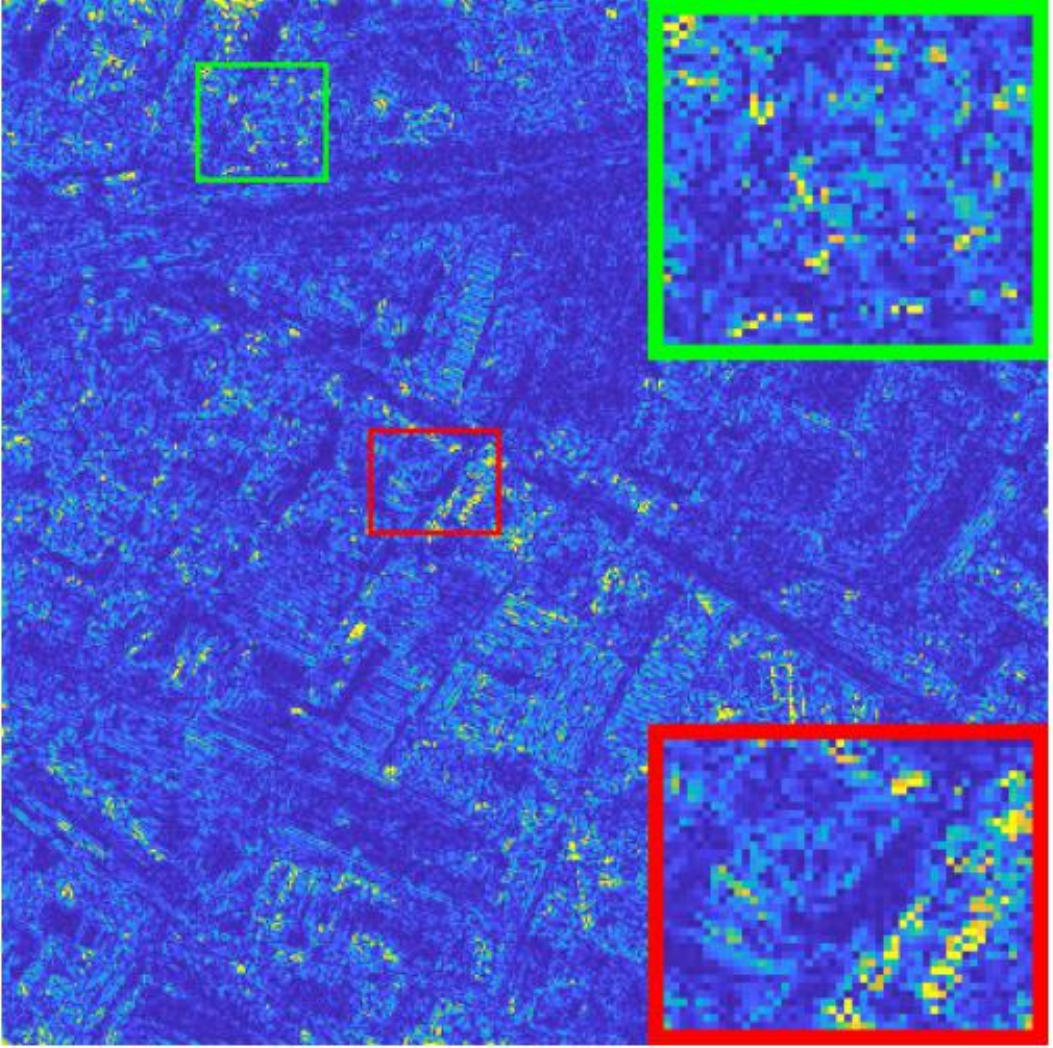}}
				{\includegraphics[width=1\linewidth]{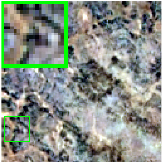}}
				\vspace{2pt}
				{\includegraphics[width=1\linewidth]{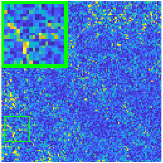}}
				{\includegraphics[width=1\linewidth]{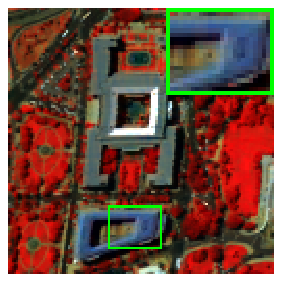}}
				{\includegraphics[width=1\linewidth]{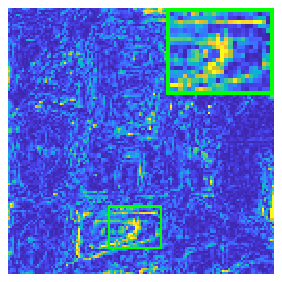}}
				\vspace{2pt}
				\scriptsize{FPFNet}
				\centering
				
			\end{minipage}
			\begin{minipage}[t]{0.095\linewidth}
				{\includegraphics[width=1\linewidth]{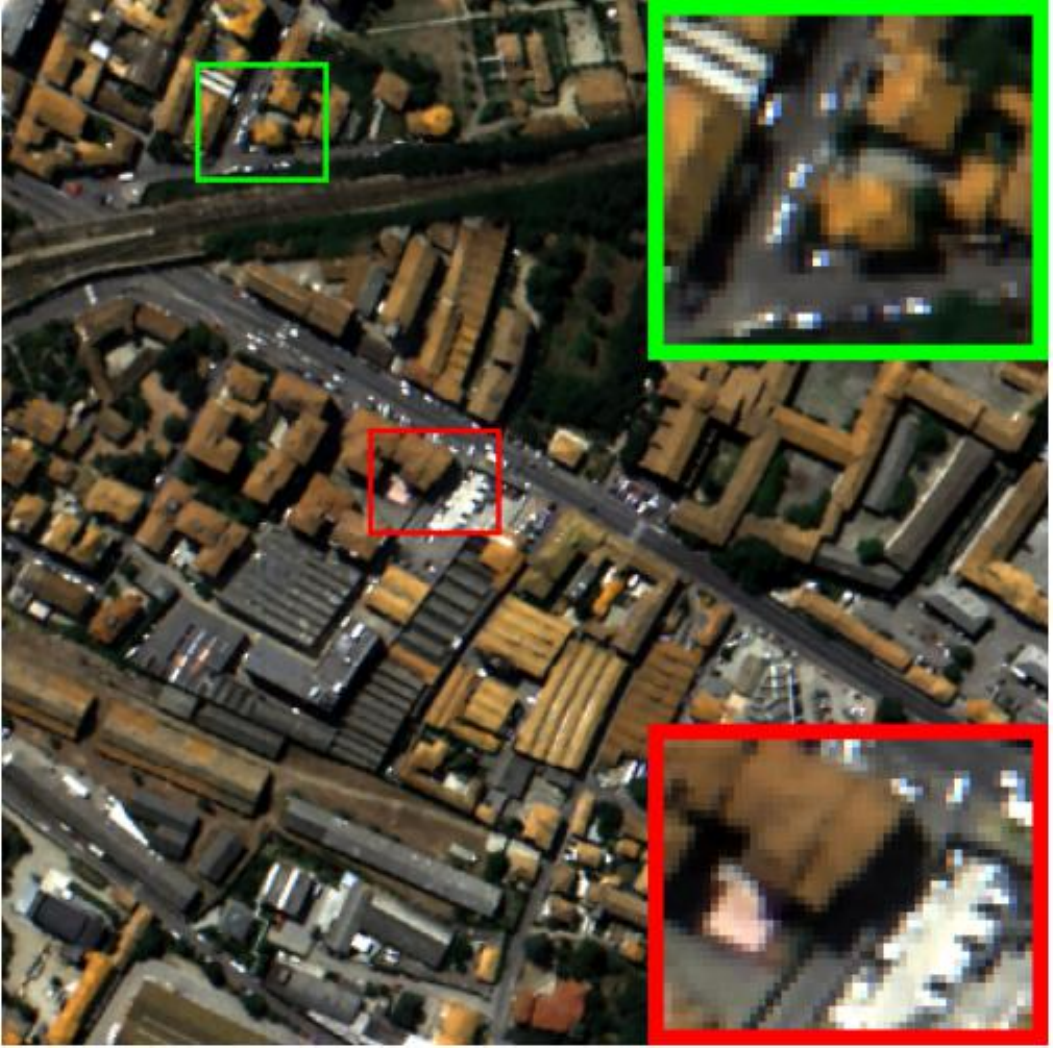}}
				\vspace{2pt}
				{\includegraphics[width=1\linewidth]{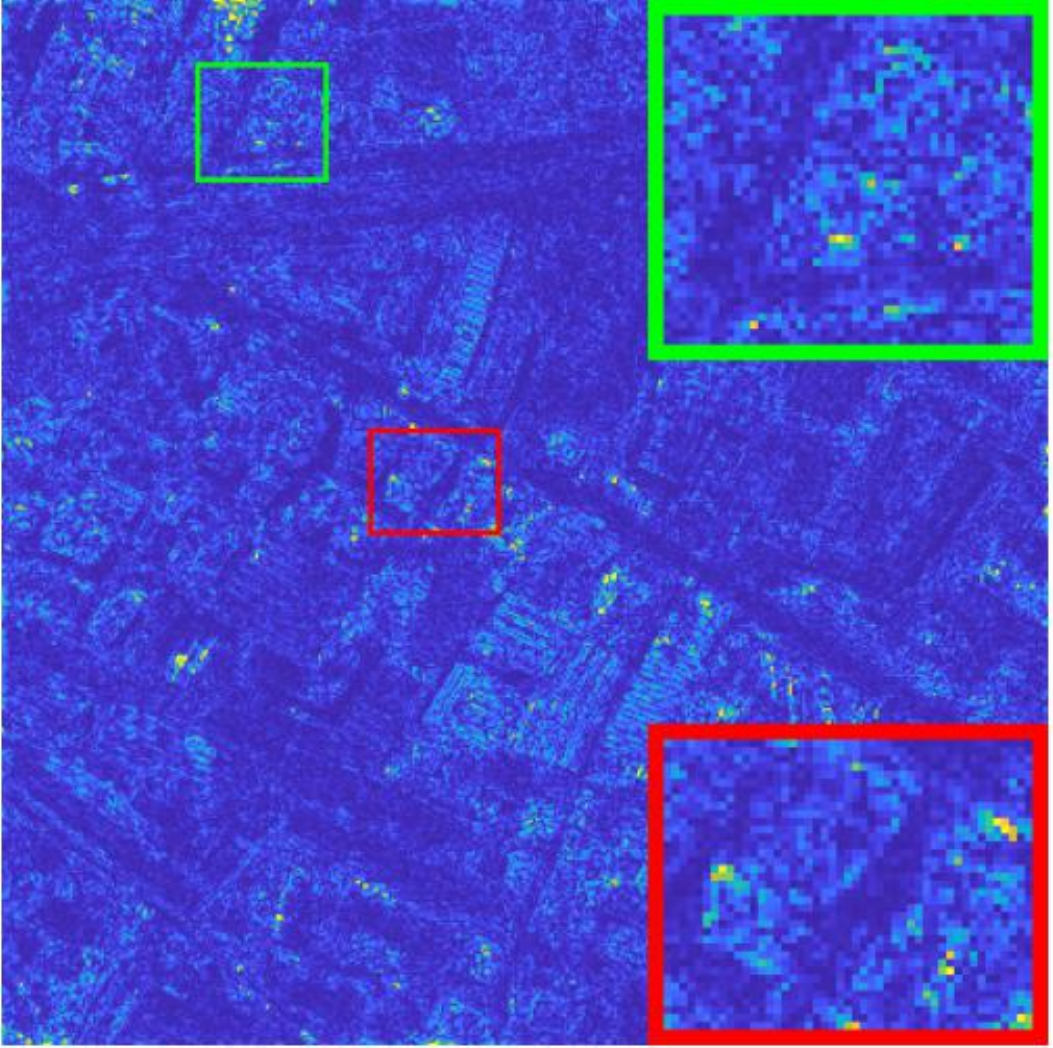}}
				{\includegraphics[width=1\linewidth]{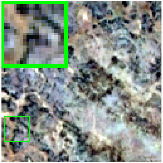}}
				\vspace{2pt}
				{\includegraphics[width=1\linewidth]{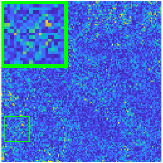}}
				{\includegraphics[width=1\linewidth]{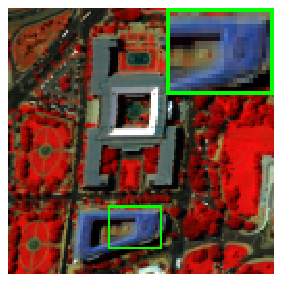}}
				{\includegraphics[width=1\linewidth]{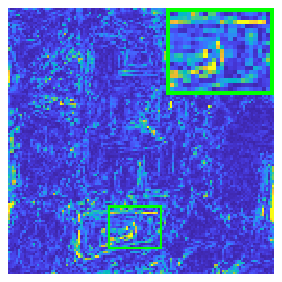}}
				\vspace{2pt}
				\scriptsize{FusionMamba}
				\centering
				
			\end{minipage}
			\begin{minipage}[t]{0.095\linewidth}
				{\includegraphics[width=1\linewidth]{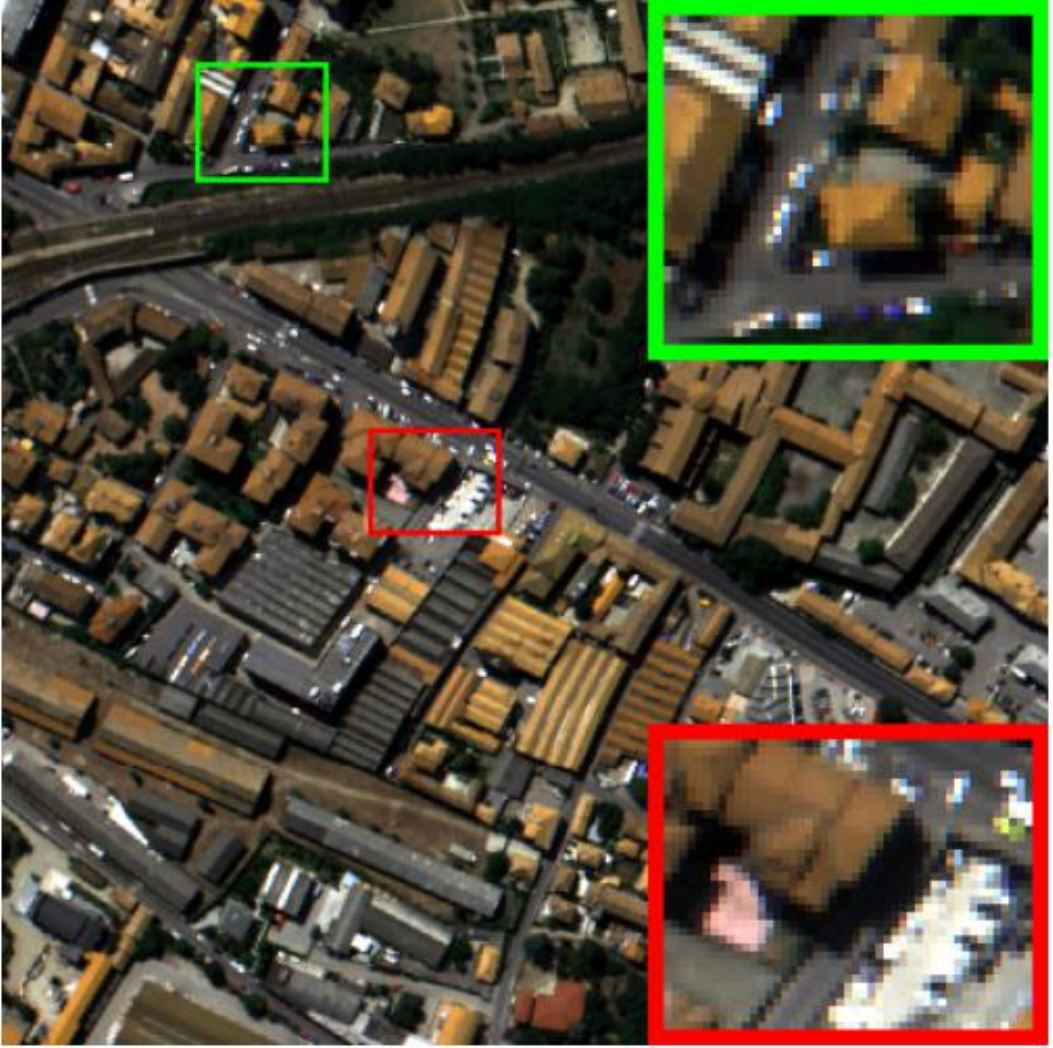}}
				\vspace{2pt}
				{\includegraphics[width=1\linewidth]{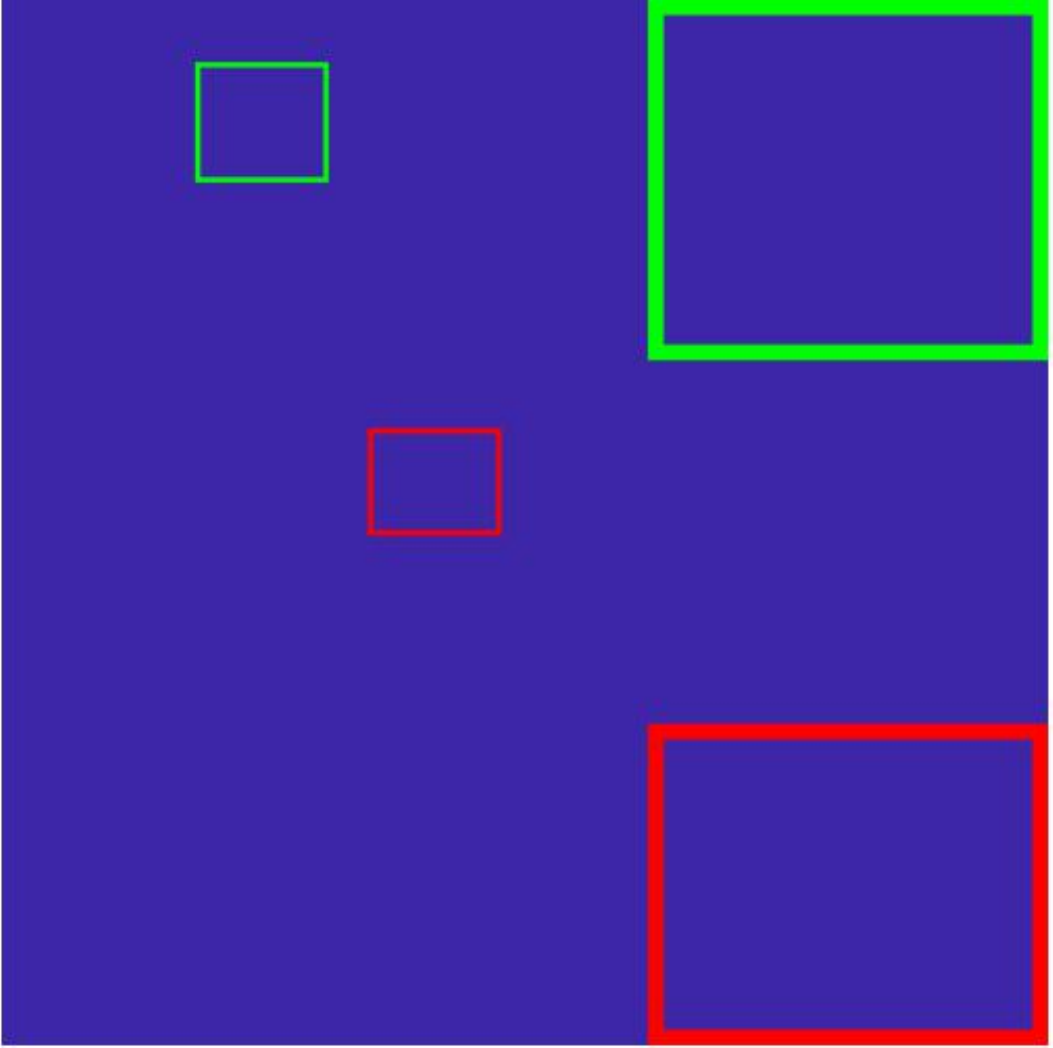}}
				{\includegraphics[width=1\linewidth]{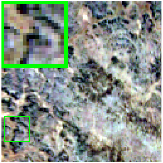}}
				\vspace{2pt}
				{\includegraphics[width=1\linewidth]{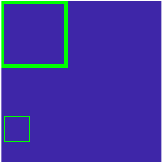}}
				{\includegraphics[width=1\linewidth]{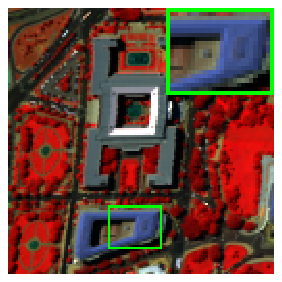}}
				{\includegraphics[width=1\linewidth]{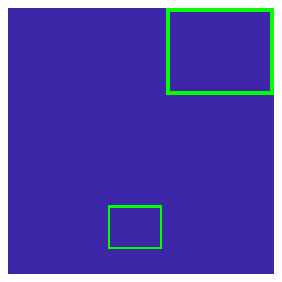}}
				\vspace{2pt}
				\scriptsize{GT}
				\centering
				
			\end{minipage}
		\end{minipage}
		\begin{minipage}[t]{1\linewidth}
			{\includegraphics[width=1\linewidth]{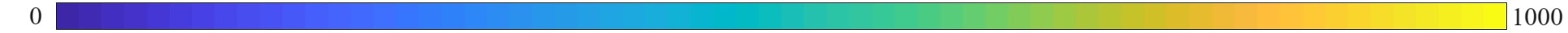}}
			\centering
		\end{minipage}
	\end{center}
    \vspace{-7pt}
	\caption{Qualitative evaluation results on the Pavia, Botswana, and WDC datasets. Row 1: Pseudo-color images for spectral bands 20, 40, and 60 from a testing sample in the Pavia dataset. Row 2: AEMs for spectral band 68 from the testing sample in row 1. Row 3: Pseudo-color images for spectral bands 30, 50, and 70 from a testing sample in the Botswana dataset. Row 4: AEMs for spectral band 34 from the testing sample in row 3. Row 5: Pseudo-color images for spectral bands 20, 50, and 80 from a testing sample in the WDC dataset. Row 6: AEMs for spectral band 25 from the testing sample in row 5. \label{hspp}}
\end{figure*}

\begin{figure*}[h]
	\begin{center}
		\begin{minipage}[t]{1\linewidth}
			\begin{minipage}[t]{0.33\linewidth}
				{\includegraphics[width=1\linewidth]{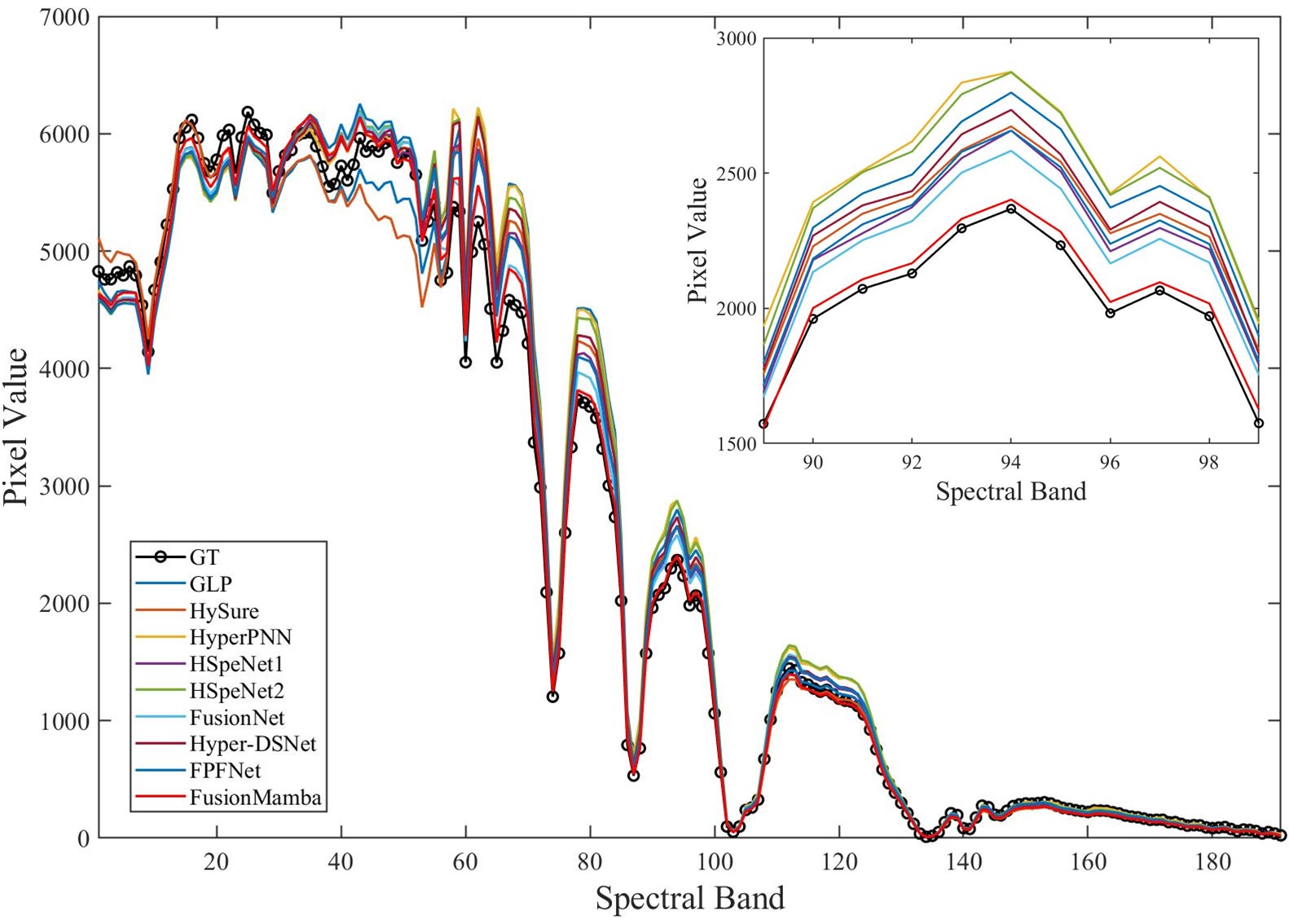}}
				{Spectral Vectors at $(44, 30)$}
				\centering
				
			\end{minipage}
			\begin{minipage}[t]{0.33\linewidth}
				{\includegraphics[width=1\linewidth]{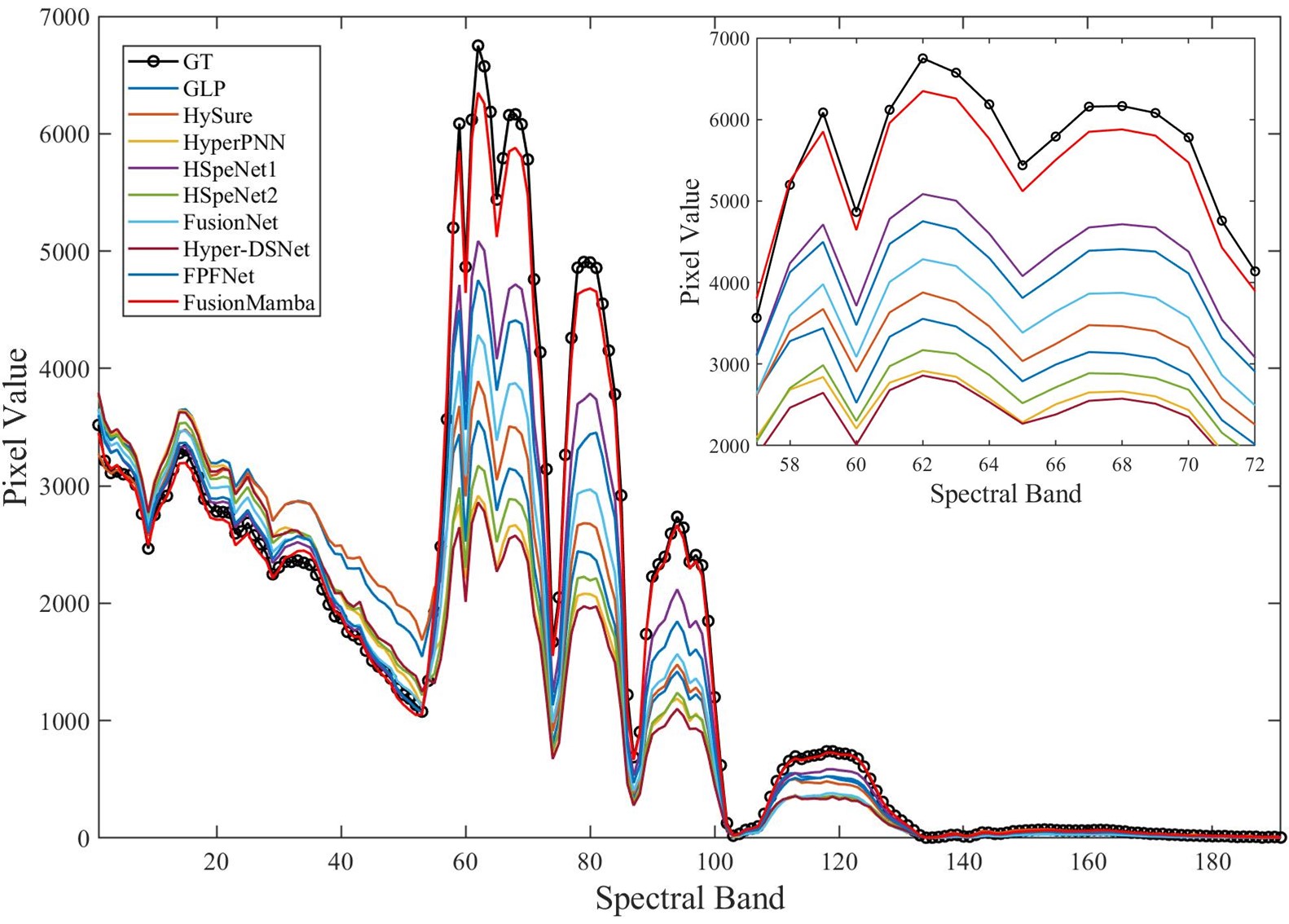}}
				{Spectral Vectors at $(80, 47)$}
				\centering
				
			\end{minipage}
			\begin{minipage}[t]{0.33\linewidth}
				{\includegraphics[width=1\linewidth]{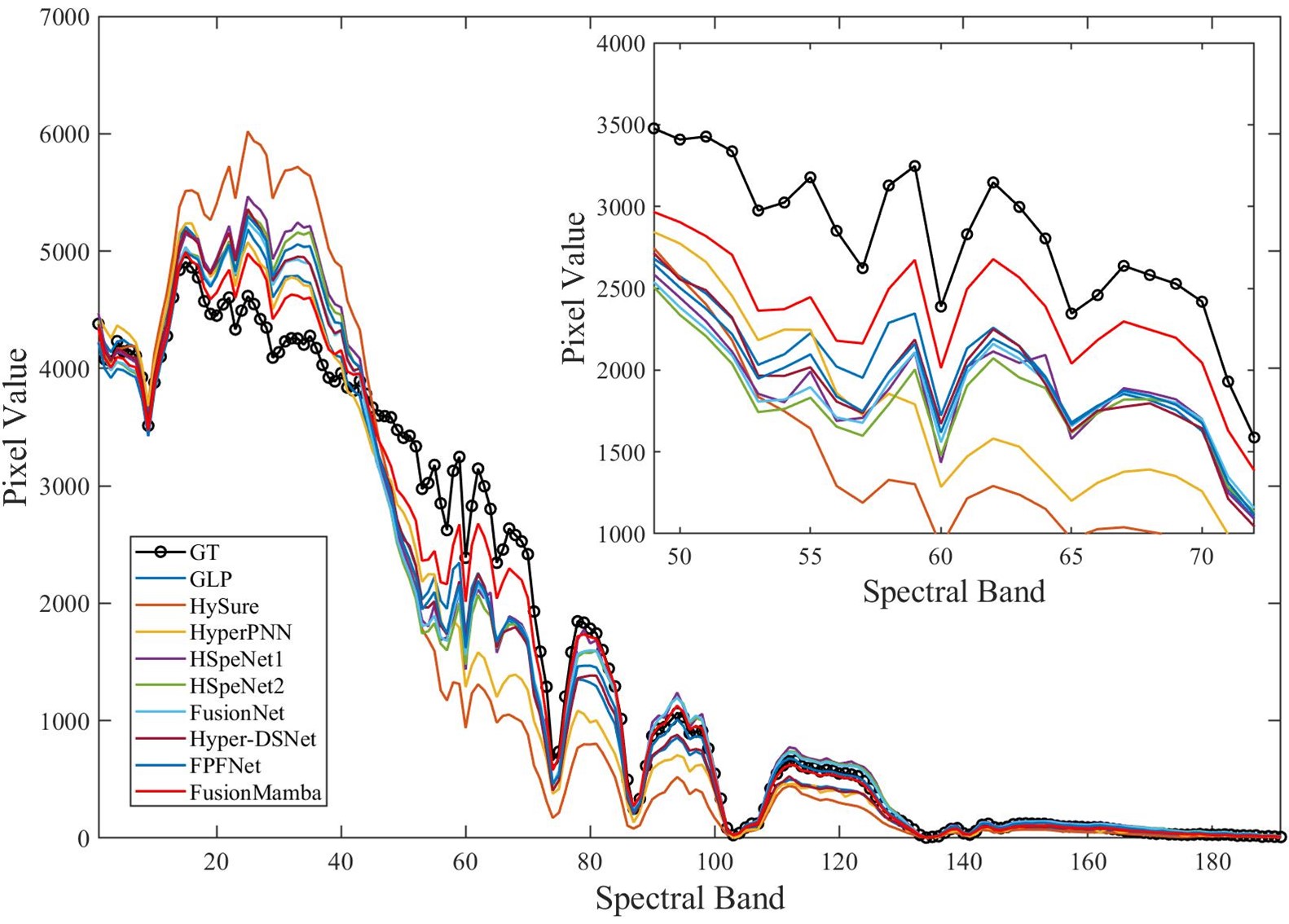}}
				{Spectral Vectors at $(115, 59)$}
				\centering
				
			\end{minipage}
		\end{minipage}
	\end{center}
    \vspace{-5pt}
	\caption{Comparison of spectral vectors at three randomly selected spatial locations from a testing sample in the WDC dataset.\label{wdccc}}
\end{figure*}

\subsubsection{Quality Indices}
In line with the research standards of the hyper-spectral pansharpening task, we select five widely used quality indices for evaluation, namely PSNR, cross-correlation (CC), SSIM \cite{wang2004image}, SAM, and ERGAS. The ideal values for these indices are +$\infty$, 1, 1, 0, and 0, respectively.

\subsubsection{Settings}
For hyper-spectral pansharpening, we set ${C}$ to 48 and ${N}$ to 4. Additionally, we employ the bicubic interpolation for up-sampling. Furthermore, in the U-shaped network branches, the number of channels in the feature maps remains constant across different stages, which slightly deviates from the depiction in Fig.~\ref{pipeline}. During the training of our networks on the Pavia, Botswana, and WDC datasets, the number of epochs is set to 1600, 3500, and 4000, respectively. Additionally, the batch size and initial learning rate are uniformly set to 32 and $2\times 10^{-4}$. Furthermore, we use the Adam optimizer, with the learning rate halving every 1000 epochs. As for other DL-based methods, we follow the default settings specified in the corresponding papers or source codes. 

\subsubsection{Results}
The quantitative evaluation results on three distinct datasets are presented in Table \ref{hsp}. Clearly, our method significantly outperforms other techniques across all quality indices. Additionally, the qualitative evaluation outcomes, shown in Fig.~\ref{hspp}, illustrate that FusionMamba produces fusion results that most closely resemble the GT images. Furthermore, Fig.~\ref{wdccc} displays spectral vectors from various spatial locations of a WDC testing sample, highlighting the minimal spectral distortion achieved by our method. These results indicate that FusionMamba excels in the hyper-spectral pansharpening task.

\subsection{Ablation Studies}

\begin{table}[t]
	\centering\renewcommand\arraystretch{1.2}\setlength{\tabcolsep}{3.3pt}
	\belowrulesep=0pt\aboverulesep=0pt
	\caption{Ablation study on our network architecture using 20 reduced-resolution samples from the WV3 dataset. All compared methods have an identical number of parameters.}\label{abl1}
	\begin{tabular}{l|c|cccc}
		\toprule
		\textbf{Methods} & GFLOPs & PSNR & Q2n & SAM & ERGAS \\  
		\midrule  
		\textbf{w/o U-shape} & 124 & 39.219 & \underline{0.921} & 2.855 & 2.134 \\
		\textbf{w/o Spatial Branch} & 31 & 39.255 & 0.920 & 2.869 & 2.121 \\
		\textbf{w/o Spectral Branch} & 31 & \underline{39.333} & \underline{0.921} & \underline{2.848} & \underline{2.097} \\
		{\textbf{w/o Combination Branch}} & {31} & {39.316} & {\underline{0.921}} & {2.853} & {2.101} \\
		\textbf{w/o MCA} & 31 & 39.324 & \underline{0.921} & 2.861 & 2.099 \\
		\textbf{w/ SENet} & 31 & 39.294 & 0.920 & 2.850 & 2.100 \\
		\textbf{FusionMamba} & 31 & \textbf{39.374} & \textbf{0.922} & \textbf{2.843} & \textbf{2.092} \\
		\midrule
		\textbf{Ideal Values} &  $-$ & \textbf{+$\infty$} & \textbf{1} & \textbf{0} & \textbf{0} \\ 
		\bottomrule
	\end{tabular}
\end{table}

\begin{table}[t]
	\centering\renewcommand\arraystretch{1.2}\setlength{\tabcolsep}{9.6pt}
	\belowrulesep=0pt\aboverulesep=0pt
	\caption{Ablation study on structures of Mamba and FusionMamba blocks using the reduced-resolution samples from the WV3 dataset. All methods have the same number of parameters.}\label{abl2}
	\begin{tabular}{l|cccc}
		\toprule
		\textbf{Methods} & PSNR & Q2n & SAM & ERGAS \\  
		\midrule  
		\textbf{OD Flattening} & 39.014 & 0.917 & 2.951 & 2.178 \\
		\textbf{BD Flattening} & 39.163 & 0.920 & 2.886 & 2.144 \\
		{\textbf{Shuffle Flattening}} & {39.117} & {0.919} & {2.898} & {2.152} \\
		\textbf{w/o $\mathbf{F}^{\rm{a}}_{\rm{out}}$} & \underline{39.309} & \underline{0.921} & \underline{2.855} & \underline{2.106} \\
		\textbf{w/o $\mathbf{F}^{\rm{b}}_{\rm{out}}$} & 39.293 & \underline{0.921} & 2.857 & 2.110 \\
		\textbf{FusionMamba} & \textbf{39.374} & \textbf{0.922} & \textbf{2.843} & \textbf{2.092} \\
		\midrule
		\textbf{Ideal Values} & \textbf{+$\infty$} & \textbf{1} & \textbf{0} & \textbf{0} \\ 
		\bottomrule
	\end{tabular}
\end{table}

\subsubsection{Network Architecture}
To validate the effectiveness of the proposed network architecture, we develop six variants of the FusionMamba and evaluate their performance using the reduced-resolution samples from the WV3 dataset, as detailed in Table~\ref{abl1}. For fairness, all compared methods are designed with an identical number of network parameters. Specifically, we maintain a consistent spatial resolution of feature maps across different stages (w/o U-shape) to assess the efficacy of hierarchical information learning. Additionally, we remove the four-directional Mamba blocks from either the spatial branch (w/o Spatial Branch) or the spectral branch (w/o Spectral Branch) to determine the effectiveness of separate feature extraction. {Furthermore, we assess the validity of the combination branch by incorporating the FusionMamba blocks into the spectral branch (w/o Combination Branch).} Finally, we remove the MCA module (w/o MCA) or replace it with the SENet \cite{Hu_2018_CVPR} (w/ SENet) to evaluate the contribution of the MCA. The results strongly support the validity of our designs.

\subsubsection{Structures of Mamba and FusionMamba Blocks}
To validate the effectiveness of our structural designs for the Mamba and FusionMamba blocks, we develop five variants and evaluate them using the reduced-resolution samples from the WV3 dataset, as presented in Table~\ref{abl2}. Specifically, we assess the performance of the four-directional flattening technique \cite{liu2024vmamba} employed in our Mamba and FusionMamba blocks by comparing it against the one-directional (OD) flattening method, the bidirectional (BD) flattening approach \cite{zhu2024vision}, {and the shuffle flattening technique \cite{10542538}.} Additionally, we remove either $\mathbf{F}_{\rm{out}}^{\rm{a}}$ (w/o $\mathbf{F}_{\rm{out}}^{\rm{a}}$) or $\mathbf{F}_{\rm{out}}^{\rm{b}}$ (w/o $\mathbf{F}_{\rm{out}}^{\rm{b}}$) from the FusionMamba block to verify the efficacy of our design. The quantitative results strongly affirm the validity of our structural designs for the Mamba and FusionMamba blocks.


\subsubsection{The Application of FusionMamba Block}
We investigate the potential of the FusionMamba block by incorporating it into several representative pansharpening frameworks, including PanNet \cite{8237455}, FusionNet \cite{2020Detail}, and U2Net \cite{10.1145/3581783.3612084}. In this process, we substitute the concatenation operation in PanNet and FusionNet, as well as the S2Block in U2Net, with the FusionMamba block. 
Table~\ref{abl3} showcases the evaluation results on the reduced-resolution samples from the WV3 dataset. The FusionMamba block demonstrates significant performance improvements, particularly when the baseline metrics are low. Moreover, even with high baseline performance, our method is still capable of exceeding the performance threshold. Consequently, the FusionMamba block proves to be an effective plug-and-play module for information integration.

\begin{table}[t]
	\centering\renewcommand\arraystretch{1.2}\setlength{\tabcolsep}{4.8pt}
	\belowrulesep=0pt\aboverulesep=0pt
	\caption{The application of the FusionMamba (FMamba) block in various pansharpening frameworks. All methods are assessed using the reduced-resolution samples from the WV3 dataset.}\label{abl3}
	\begin{tabular}{l|c|cccc}
		\toprule
		\textbf{Methods} & Params & PSNR & Q2n & SAM & ERGAS \\  
		\midrule  
		\textbf{PanNet \cite{8237455}} & 0.08M & 37.346 & 0.891 & 3.613 & 2.664 \\
		\textbf{PanNet + FMamba} & 0.09M & \textbf{38.178} & \textbf{0.904} & \textbf{3.236} & \textbf{2.418} \\
		\midrule
		\textbf{FusionNet \cite{2020Detail}} & 0.08M & 38.047 & 0.904 & 3.324 & 2.465 \\
		\textbf{FusionNet + FMamba}  & 0.09M & \textbf{38.604} & \textbf{0.914} & \textbf{3.092} & \textbf{2.294} \\
		\midrule
		\textbf{U2Net \cite{10.1145/3581783.3612084}} & 0.66M & 39.117 & \textbf{0.920} & 2.888 & 2.149 \\
		\textbf{U2Net + FMamba} & 0.83M & \textbf{39.181} & \textbf{0.920} & \textbf{2.885} & \textbf{2.132} \\
		\midrule
		\textbf{Ideal Values} & $-$ & \textbf{+$\infty$} & \textbf{1} & \textbf{0} & \textbf{0} \\ 
		\bottomrule
	\end{tabular}
\end{table}

\subsubsection{FSSM Block}
The main contribution of the FSSM block lies in its information interaction mechanism. Consequently, we investigate different interactive combinations of the projection parameters $\mathbf{B}$ and $\mathbf{C}$, along with the timescale parameter $\mathbf{\Delta}$. The quantitative evaluation results on 20 reduced-resolution samples from the WV3 dataset, as illustrated in Table~\ref{abl4}, demonstrate the correctness of our strategy: one input generates the projection and timescale parameters, while the other input serves as the 1D sequence to be processed.

\begin{table}[t]
	\centering\renewcommand\arraystretch{1.2}\setlength{\tabcolsep}{9.8pt}
	\belowrulesep=0pt\aboverulesep=0pt
	\caption{Ablation study on interactive combinations of the parameters $\mathbf{B}$, $\mathbf{C}$, and $\mathbf{\Delta}$ in the FSSM block. All methods are evaluated using 20 reduced-resolution samples from the WV3 dataset.}\label{abl4}
	\begin{tabular}{ccc|cccc}
		\toprule
		
		\multicolumn{3}{c|}{\textbf{Interactiveness}} &
		\multirow{2}{*}{PSNR} &
		\multirow{2}{*}{Q2n} &
		\multirow{2}{*}{SAM} &
		\multirow{2}{*}{ERGAS}
		\\
		\cmidrule(lr){1-3}
		\multicolumn{1}{c}{$\mathbf{B}$} &
		\multicolumn{1}{c}{$\mathbf{C}$} &
		\multicolumn{1}{c|}{$\mathbf{\Delta}$}
		\\
		\midrule  
		 $\times$ & $\times$ & $\times$ & 39.124 & 0.919 & 2.903 & 2.154 \\
		 \checked & $\times$ & $\times$ & {39.182} & 0.920 & {2.886} & {2.145} \\
		 $\times$ & \checked & $\times$& 39.157 & 0.919 & 2.893 & 2.150 \\
		 $\times$ & $\times$ & \checked & 39.116 & 0.919 & 2.907 & 2.159 \\
		 \checked & \checked & $\times$ & \underline{39.343} & \textbf{0.922} & \underline{2.846} & \underline{2.093} \\
		 \checked & $\times$ & \checked & {39.224} & \underline{0.921} & {2.864} & {2.122} \\
		 $\times$ & \checked & \checked & {39.265} & \underline{0.921} & {2.852} & {2.116} \\
		 \checked & \checked & \checked & \textbf{39.374} & \textbf{0.922} & \textbf{2.843} & \textbf{2.092} \\
		\midrule
		\multicolumn{3}{c|}{\textbf{Ideal Values}} & \textbf{+$\infty$} & \textbf{1} & \textbf{0} & \textbf{0} \\ 
		\bottomrule
	\end{tabular}
\end{table}

\subsubsection{Feature Extraction and Information Integration Combinations}
To emphasize the superiority of the Mamba and FusionMamba blocks, we systematically examine various combinations of feature extraction methods and information integration approaches. Specifically, the candidate feature extraction methods include the convolution (Conv) layer, self-attention (SA) module, and four-directional Mamba (Mamba) block. For information integration, we consider the concatenation (Concat) operation, cross-attention (CA) module, and FusionMamba (FMamba) block. Notably, a convolution layer is applied in Concat to adjust the number of channels. Table~\ref{abl5} presents quantitative evaluation results on 20 reduced-resolution samples from the WV3 dataset. To ensure a fair comparison, all combinations are designed to have the same number of network parameters. The combination of Mamba + FMamba achieves the best results across all quality indices while maintaining a relatively low FLOP consumption, underscoring both the efficacy and efficiency of our method. Additionally, combinations with Mamba outperform those without it, indicating the effectiveness of the Mamba block in feature extraction. Moreover, combinations incorporating FMamba far exceed those without it, demonstrating the superiority of the FusionMamba block in merging different types of information. 

\begin{table}[t]
	\centering\renewcommand\arraystretch{1.2}\setlength{\tabcolsep}{5.5pt}
	\belowrulesep=0pt\aboverulesep=0pt
	\caption{Different combinations of feature extraction methods and information integration approaches. All combinations have the same number of network parameters and are evaluated using 20 reduced-resolution samples from the WV3 dataset. Notably, combinations incorporating SA or CA require significantly higher FLOPs due to the quadratic complexity with respect to the number of input tokens.}\label{abl5}
	\begin{tabular}{l|c|cccc}
		\toprule
		\textbf{Methods} & GFLOPs & PSNR & Q2n & SAM & ERGAS \\  
		
		\midrule  
		\textbf{Conv + Concat} & 20 & 38.729 & 0.917 & 3.013 & 2.258 \\
		\textbf{Conv + CA} & 2511 & 38.885 & 0.917 & 2.995 & 2.211 \\
		\textbf{Conv + FMamba} & 26 & 39.111 & 0.918 & 2.919 & 2.152 \\
		\textbf{SA + Concat} & 2511 & 38.653 & 0.914 & 3.070 & 2.285 \\
		\textbf{SA + CA} & 5003 & 38.690 & 0.915 & 3.052 & 2.266 \\
		\textbf{SA + FMamba} & 2517 & \underline{39.206} & \underline{0.920} & \underline{2.867} & \underline{2.126} \\
		\textbf{Mamba + Concat} & 26 & 38.822 & 0.917 & 2.958 & 2.230 \\
		\textbf{Mamba + CA} & 2517 & 38.925 & 0.919 & 2.917 & 2.203 \\
		\textbf{Mamba + FMamba} & 31 & \textbf{39.374} & \textbf{0.922} & \textbf{2.843} & \textbf{2.092} \\
		\midrule
		\textbf{Ideal Values} &  $-$ & \textbf{+$\infty$} & \textbf{1} & \textbf{0} & \textbf{0} \\ 
		\bottomrule
	\end{tabular}
\end{table}

\section{Discussion}
\label{s5}

\begin{table}[t]	
	\centering\renewcommand\arraystretch{1.2}\setlength{\tabcolsep}{7pt}
	\belowrulesep=0pt\aboverulesep=0pt
	\caption{Quantitative evaluation results on the testing samples from the CAVE dataset, which belongs to the HISR task. \label{cave}}	
	\begin{tabular}{l|c|cccc}
			\toprule
			\textbf{Methods} & \textbf{Params} & PSNR & SSIM & SAM & ERGAS \\ 
			\midrule
			\textbf{LTMR} \cite{dian2019hyperspectral} &  $-$ & 36.543 & 0.963 & 6.711 & 5.387 \\ 
			\textbf{UTV} \cite{xujstars2020} &  $-$ & 38.615 & 0.941 & 8.649 & 4.519 \\
			\midrule
			\textbf{ResTFNet} \cite{2018Remote} &  2.39M & 45.584 & 0.994 & 2.764 & 2.313 \\ 
			\textbf{SSRNet} \cite{9186332} &  0.03M & 48.620 & 0.995& 2.542 & 1.636 \\ 
			\textbf{Fusformer} \cite{9841513} &  0.50M & 49.983 & 0.994 & 2.203 & 2.534 \\ 
			\textbf{3DT-Net} \cite{ma2023learning} &  3.16M & \underline{51.471}& \underline{0.997} & \underline{2.117} & \underline{1.119} \\ 
			{\textbf{PSRT}} \cite{deng2023psrt} &  0.25M & 50.595& \underline{0.997}& 2.146& 2.001 \\ 
			\textbf{U2Net} \cite{10.1145/3581783.3612084} &  2.65M & 50.433& \underline{0.997} & 2.187 & 1.277 \\
			\textbf{FusionMamba} &  2.58M & \textbf{51.658}& \textbf{0.998}& \textbf{2.021} & \textbf{1.081} \\ 
			\midrule
			\textbf{Ideal Values} &  $-$ & \textbf{+$\infty$} & \textbf{1} & \textbf{0} & \textbf{0} \\ 
			\bottomrule
		\end{tabular}
\end{table}

\begin{figure}[t]
	\begin{center}
			\begin{minipage}[t]{0.98\linewidth}
					\begin{minipage}[t]{0.155\linewidth}
							{\includegraphics[width=1\linewidth]{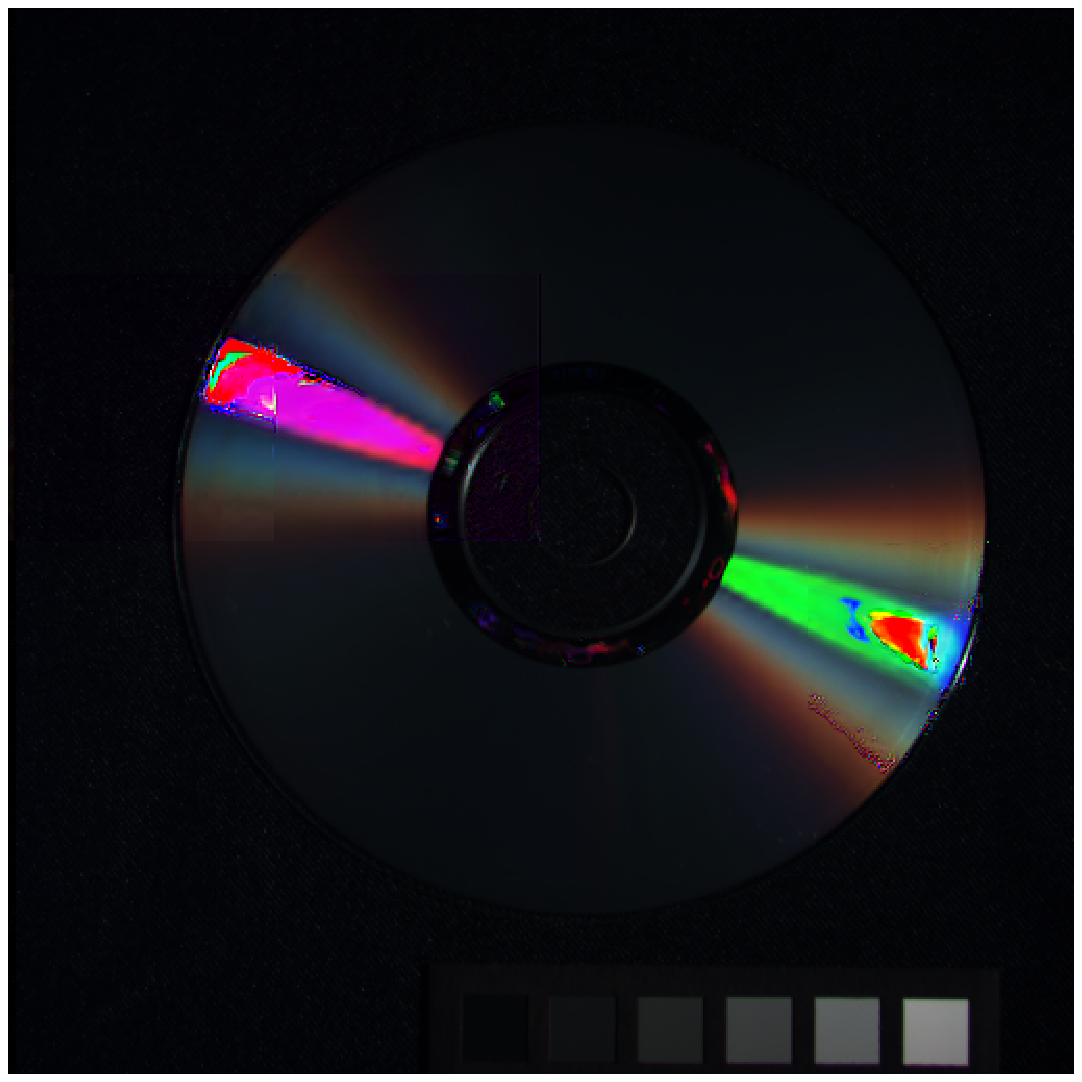}}
							\vspace{2pt}
							{\includegraphics[width=1\linewidth]{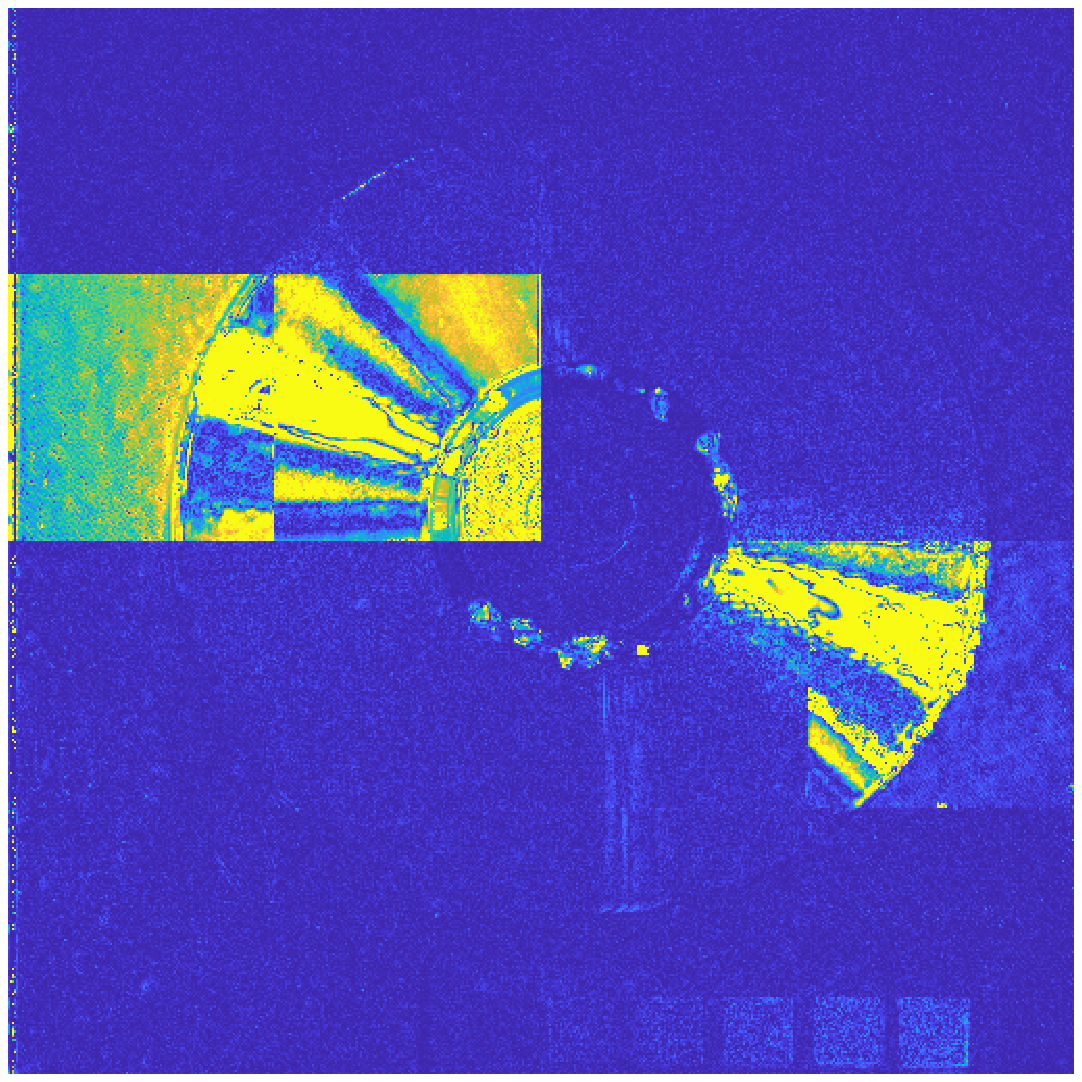}}
							{\includegraphics[width=1\linewidth]{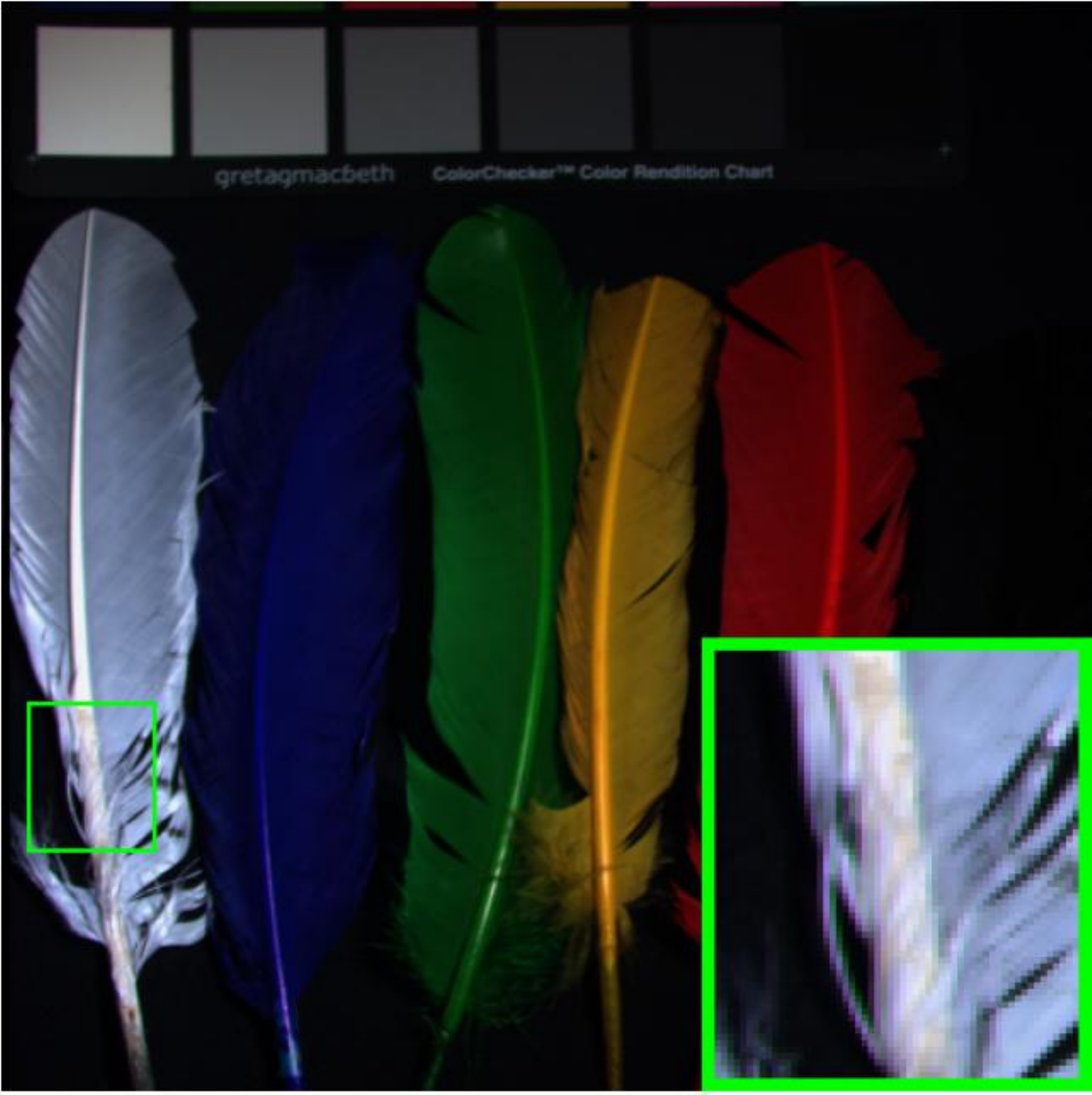}}
							\vspace{2pt}
							{\includegraphics[width=1\linewidth]{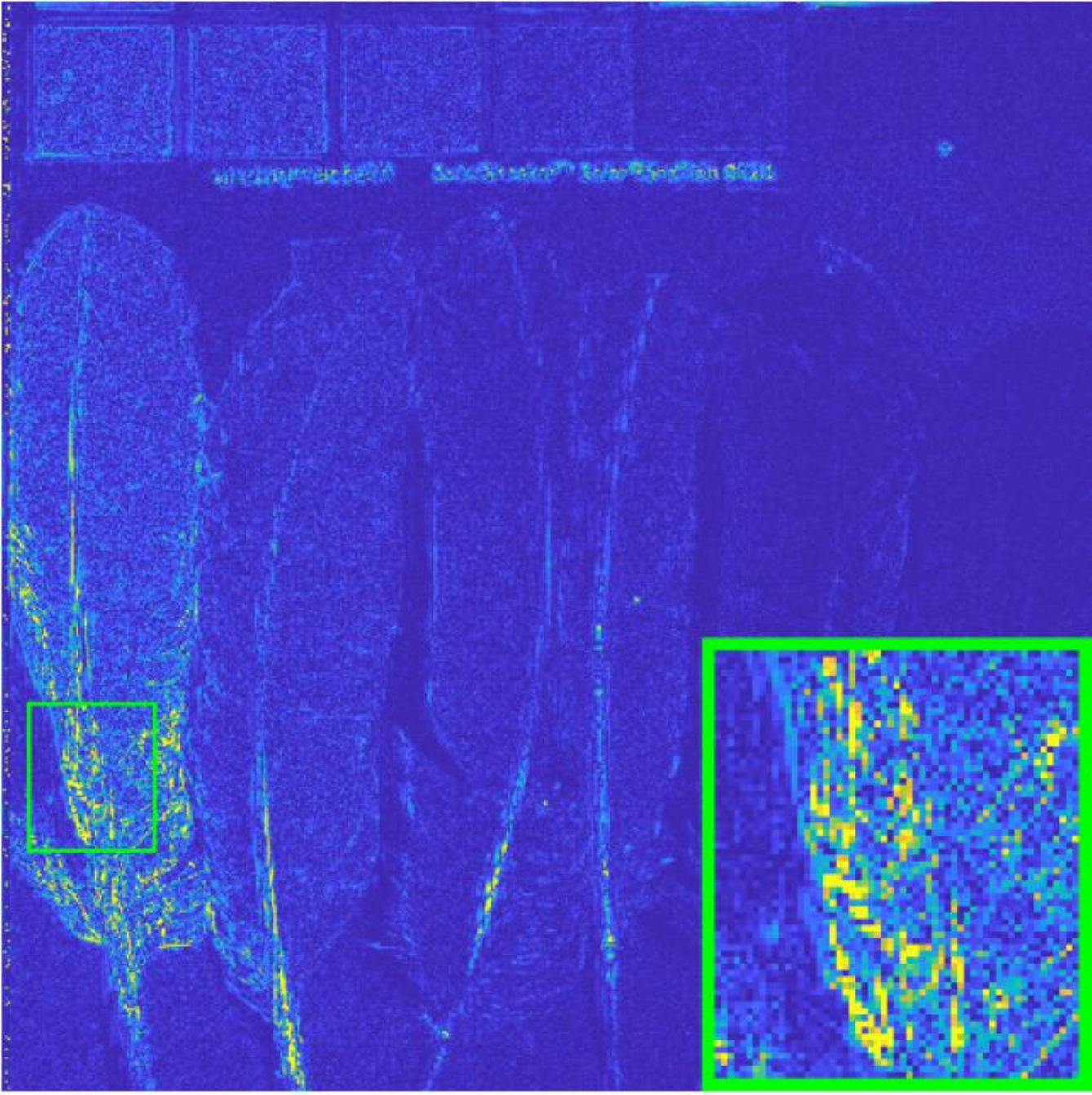}}
							{\includegraphics[width=1\linewidth]{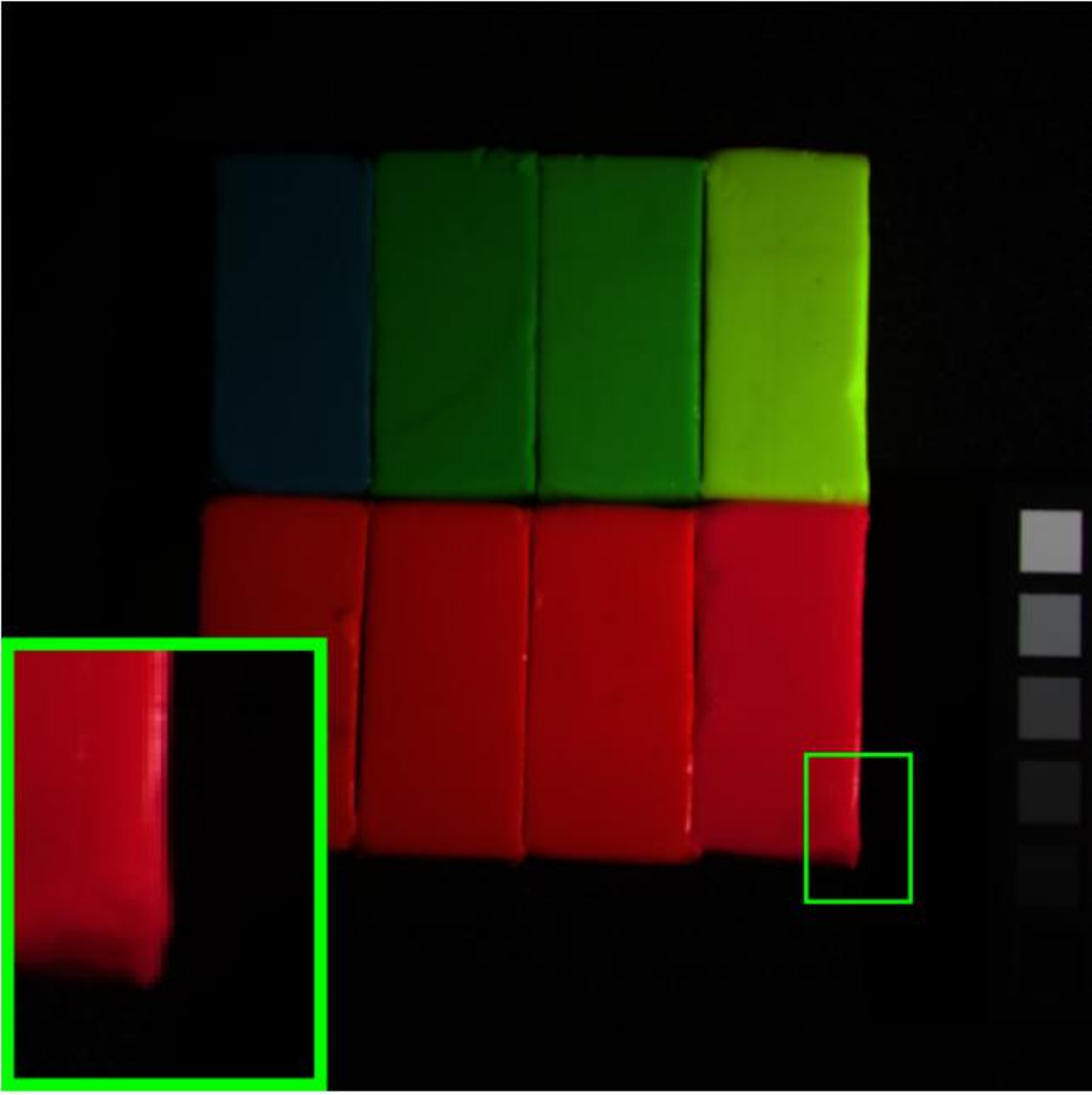}}
							{\includegraphics[width=1\linewidth]{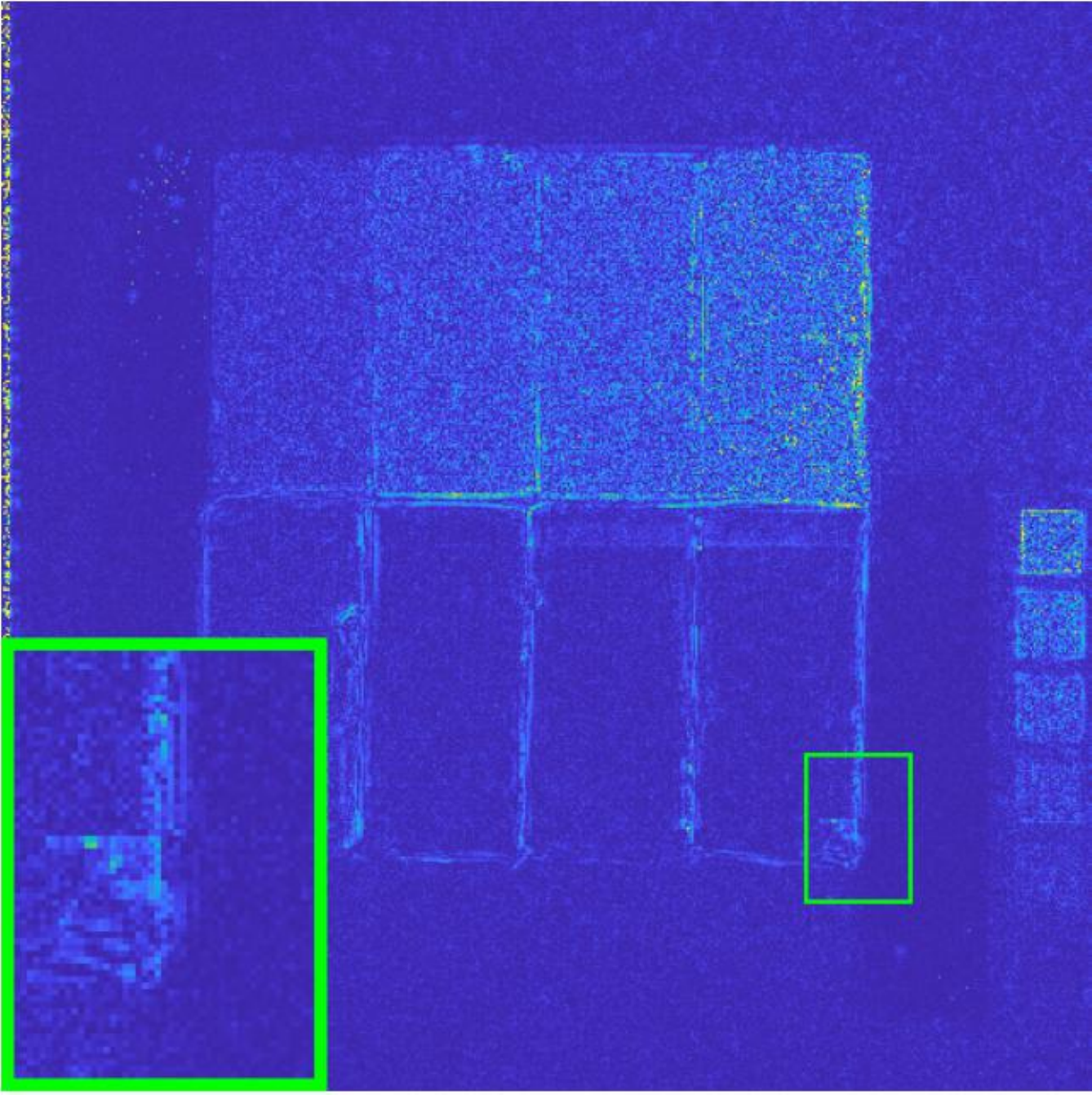}}
							\vspace{2pt}
							\scriptsize{Fusformer}
							\centering
							
						\end{minipage}
					\begin{minipage}[t]{0.155\linewidth}
							{\includegraphics[width=1\linewidth]{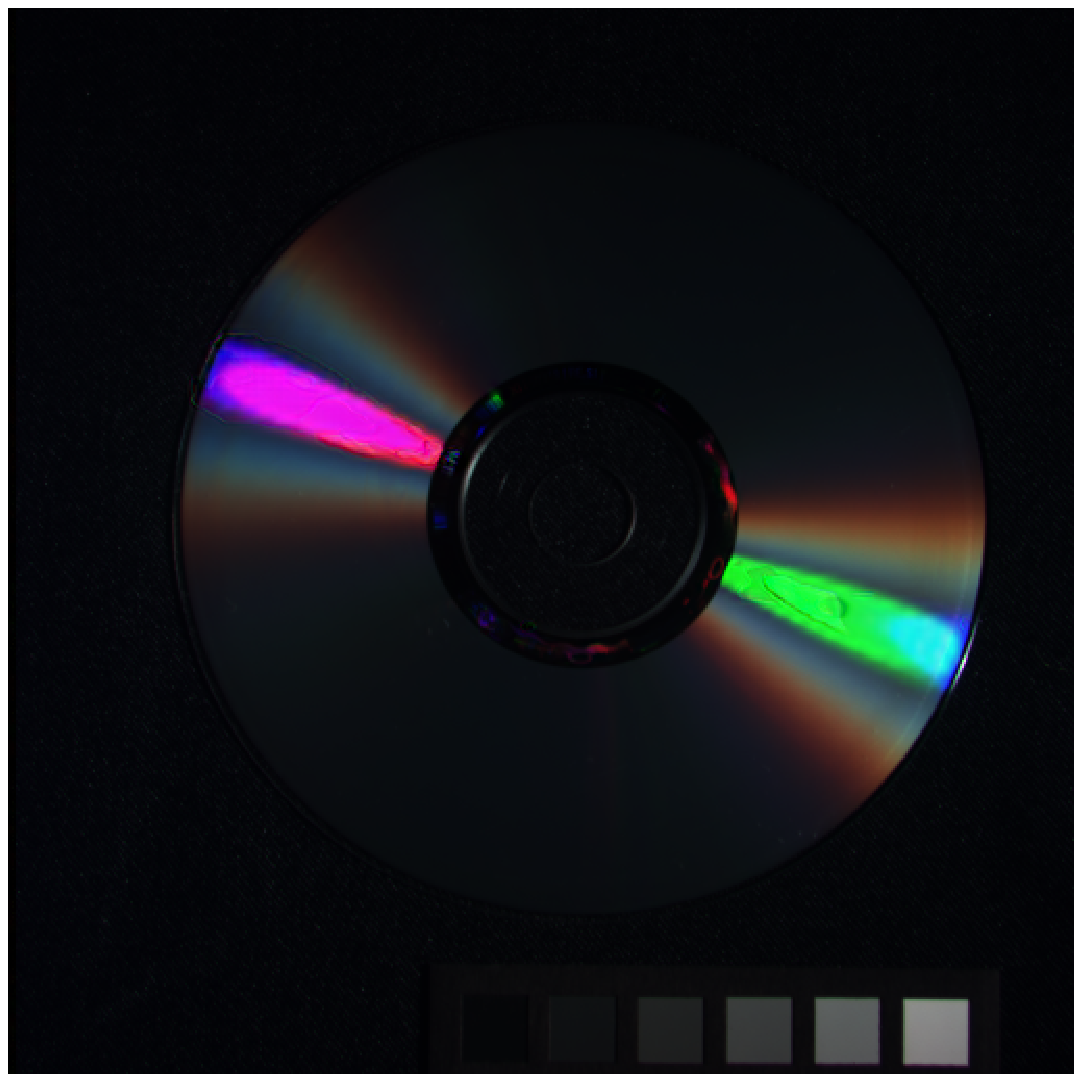}}
							\vspace{2pt}
							{\includegraphics[width=1\linewidth]{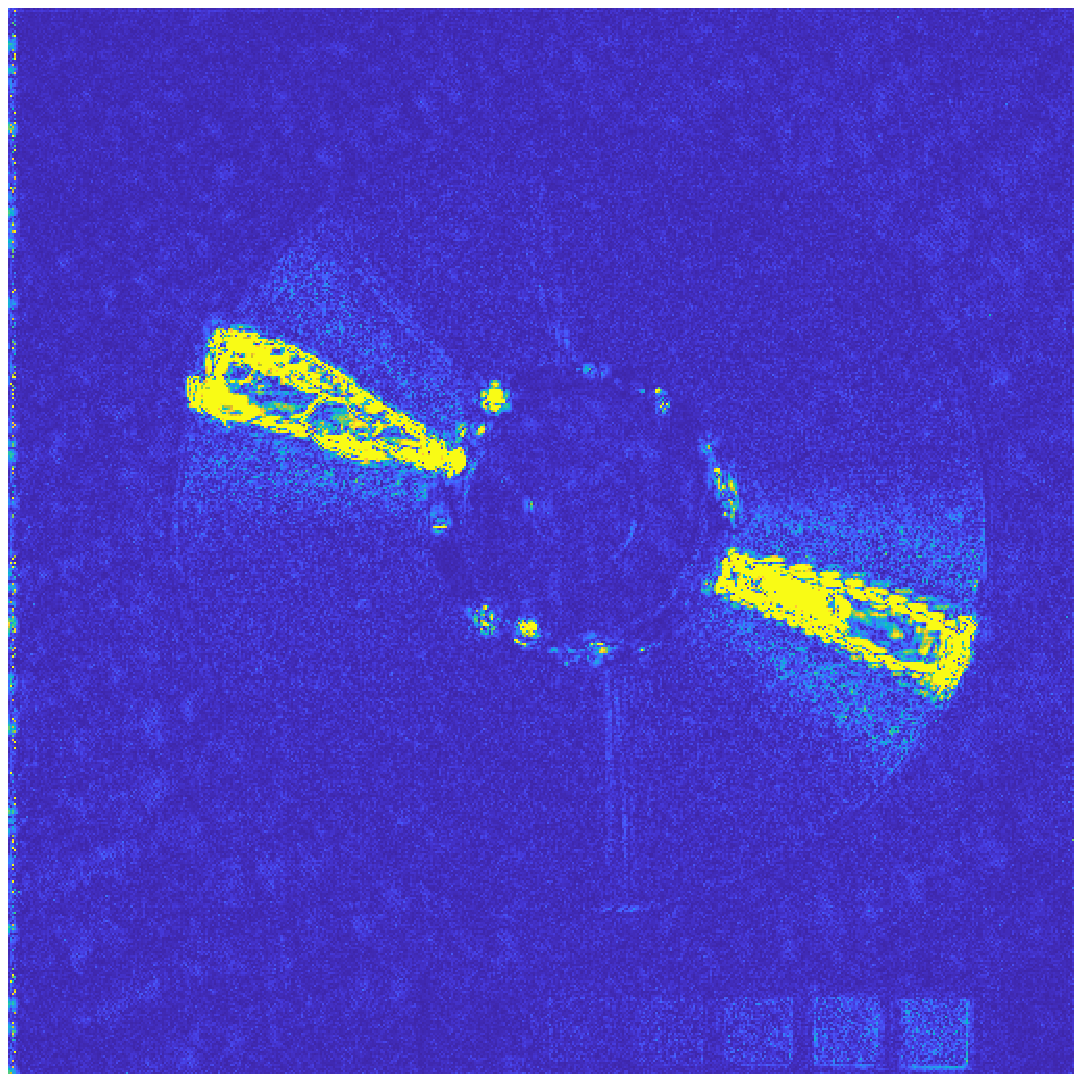}}
							{\includegraphics[width=1\linewidth]{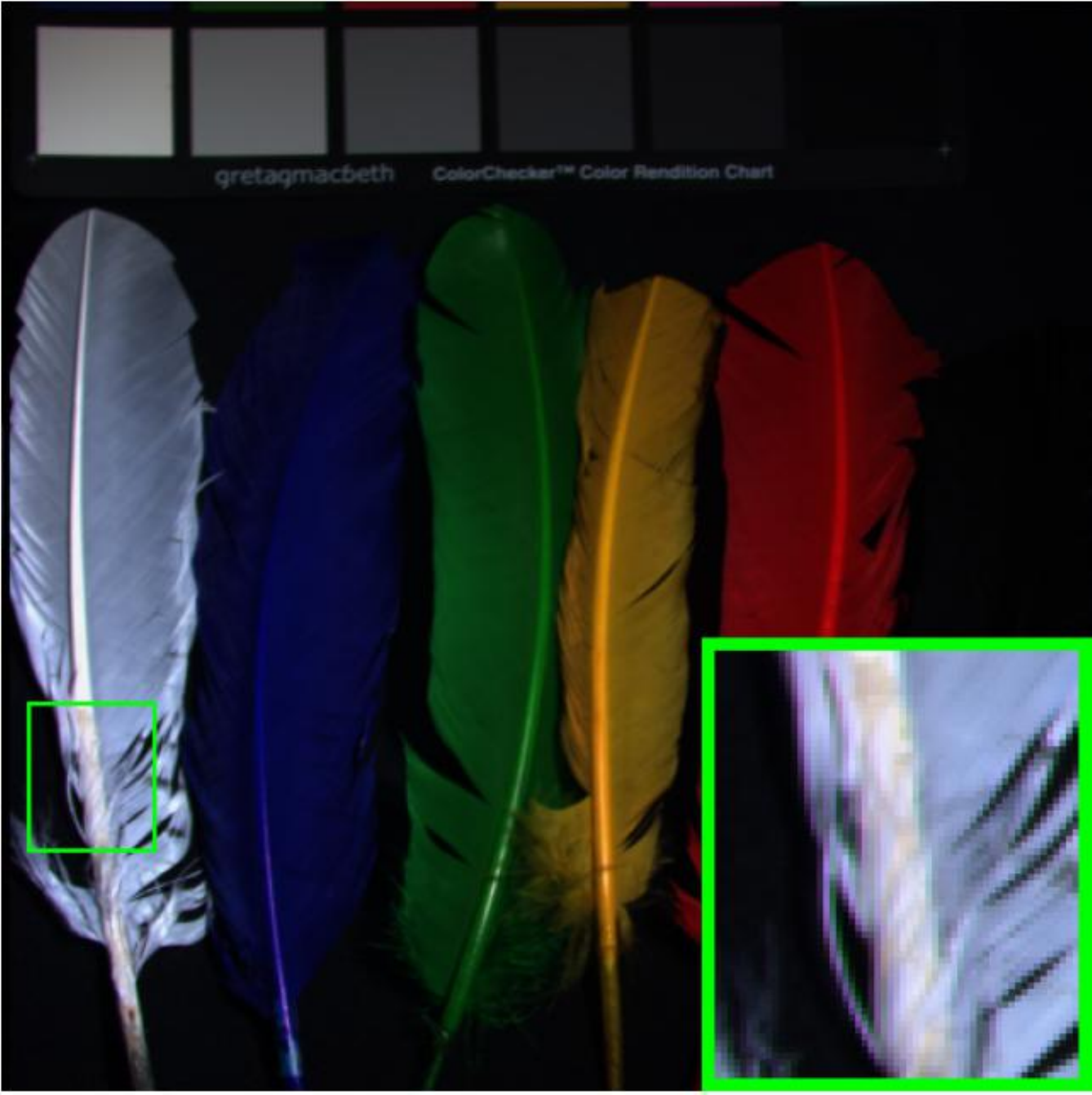}}
							\vspace{2pt}
							{\includegraphics[width=1\linewidth]{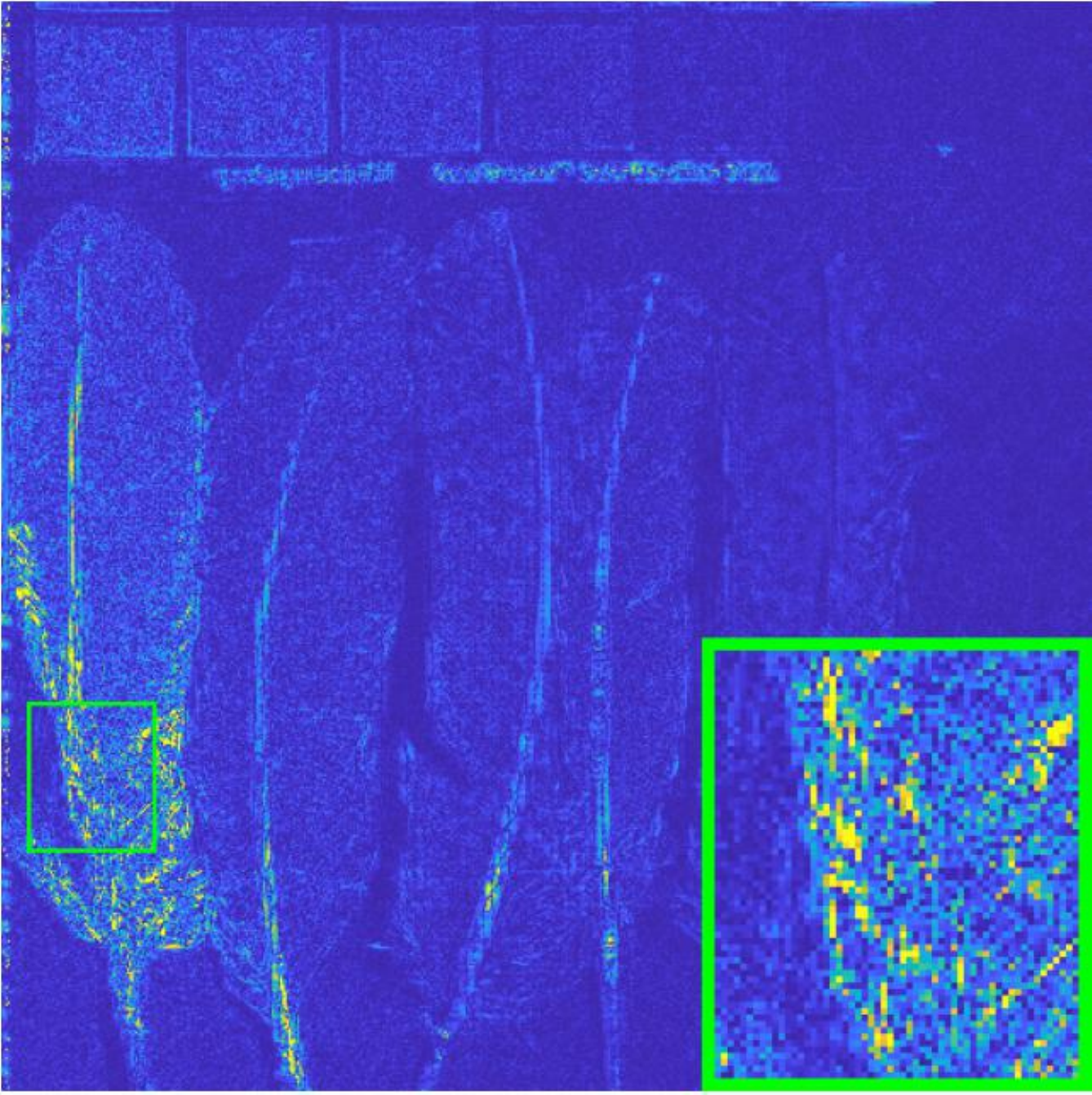}}
							{\includegraphics[width=1\linewidth]{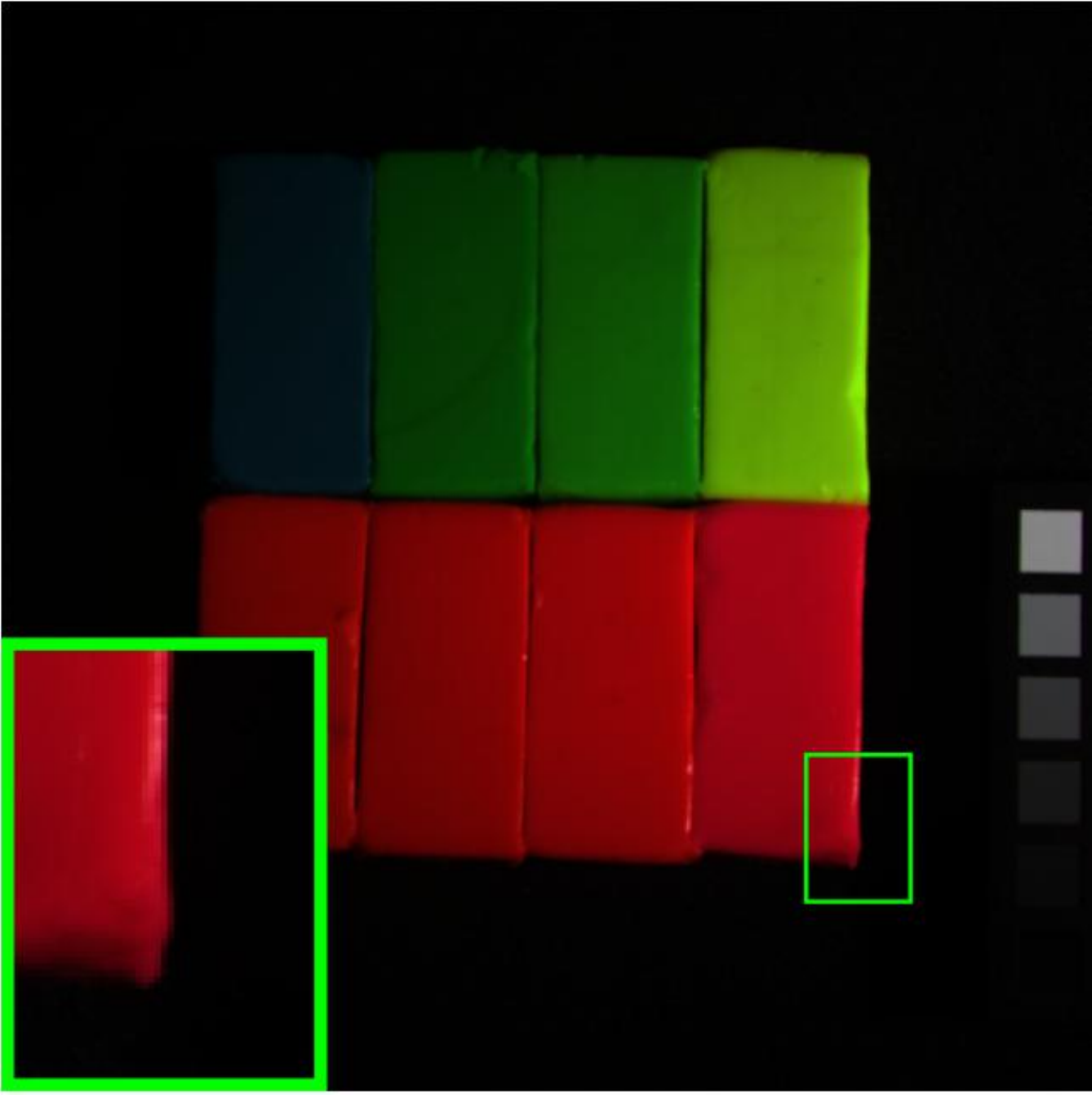}}
							{\includegraphics[width=1\linewidth]{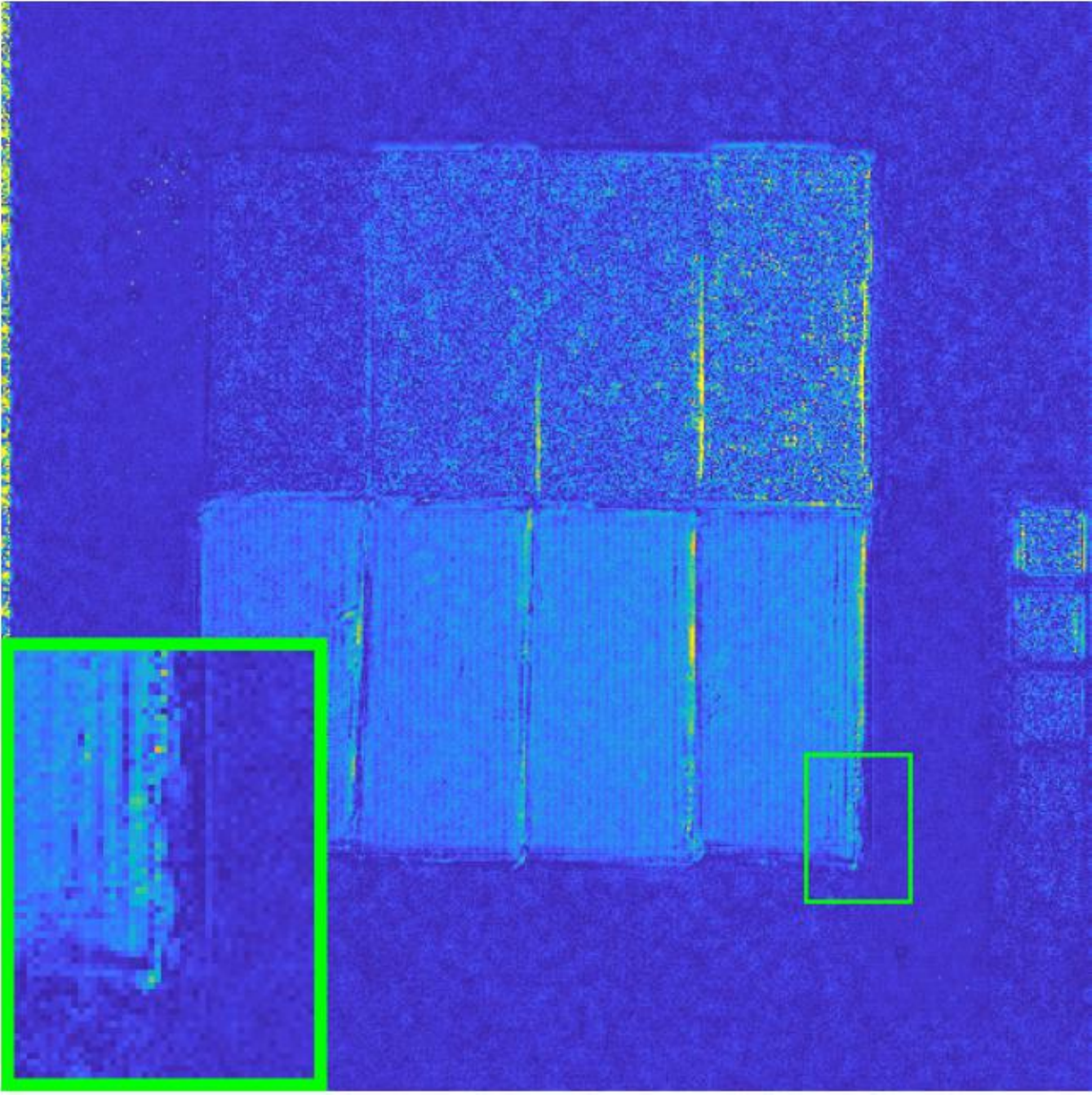}}
							\vspace{2pt}
							\scriptsize{3DT-Net}
							\centering
							
						\end{minipage}
					\begin{minipage}[t]{0.155\linewidth}
							{\includegraphics[width=1\linewidth]{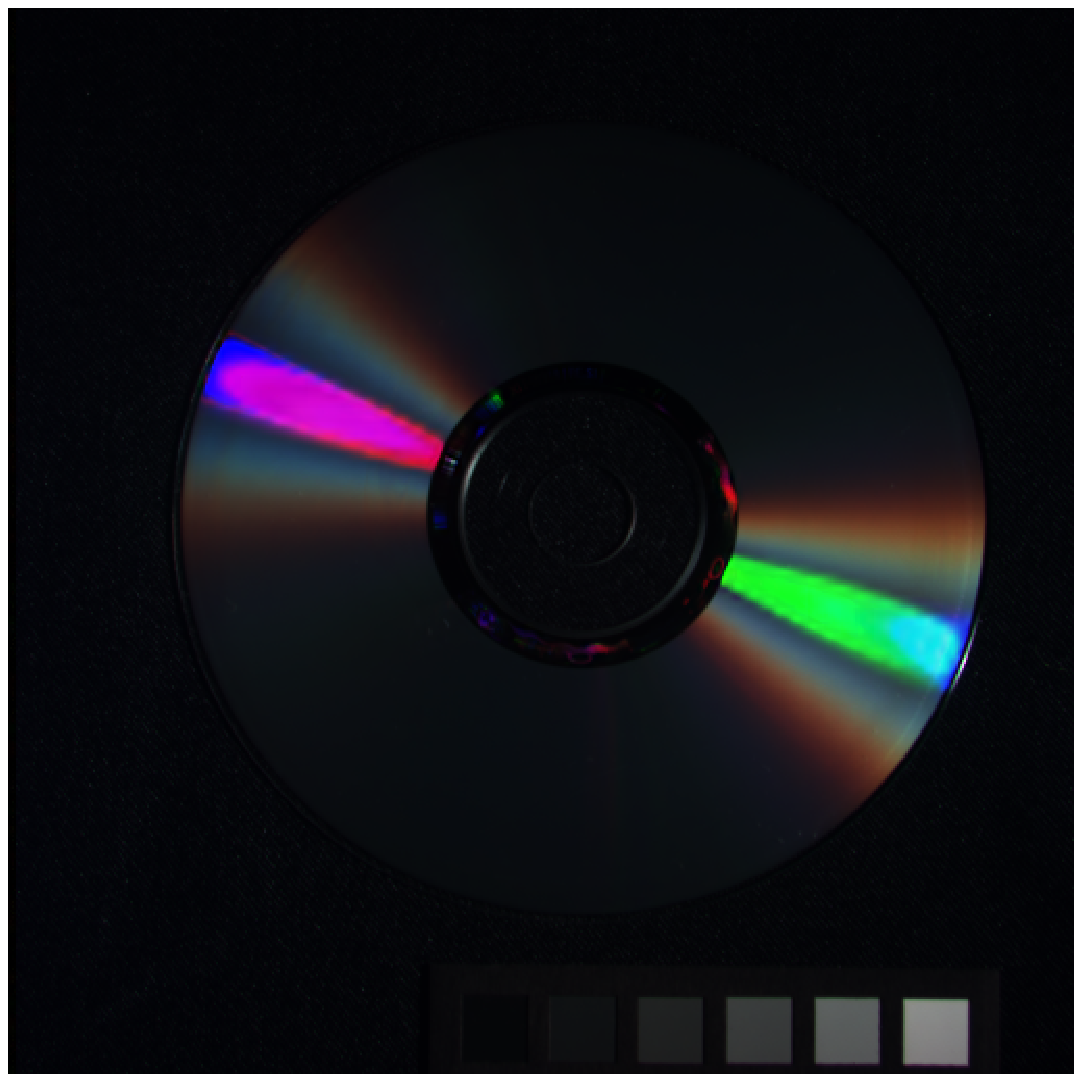}}
							\vspace{2pt}
							{\includegraphics[width=1\linewidth]{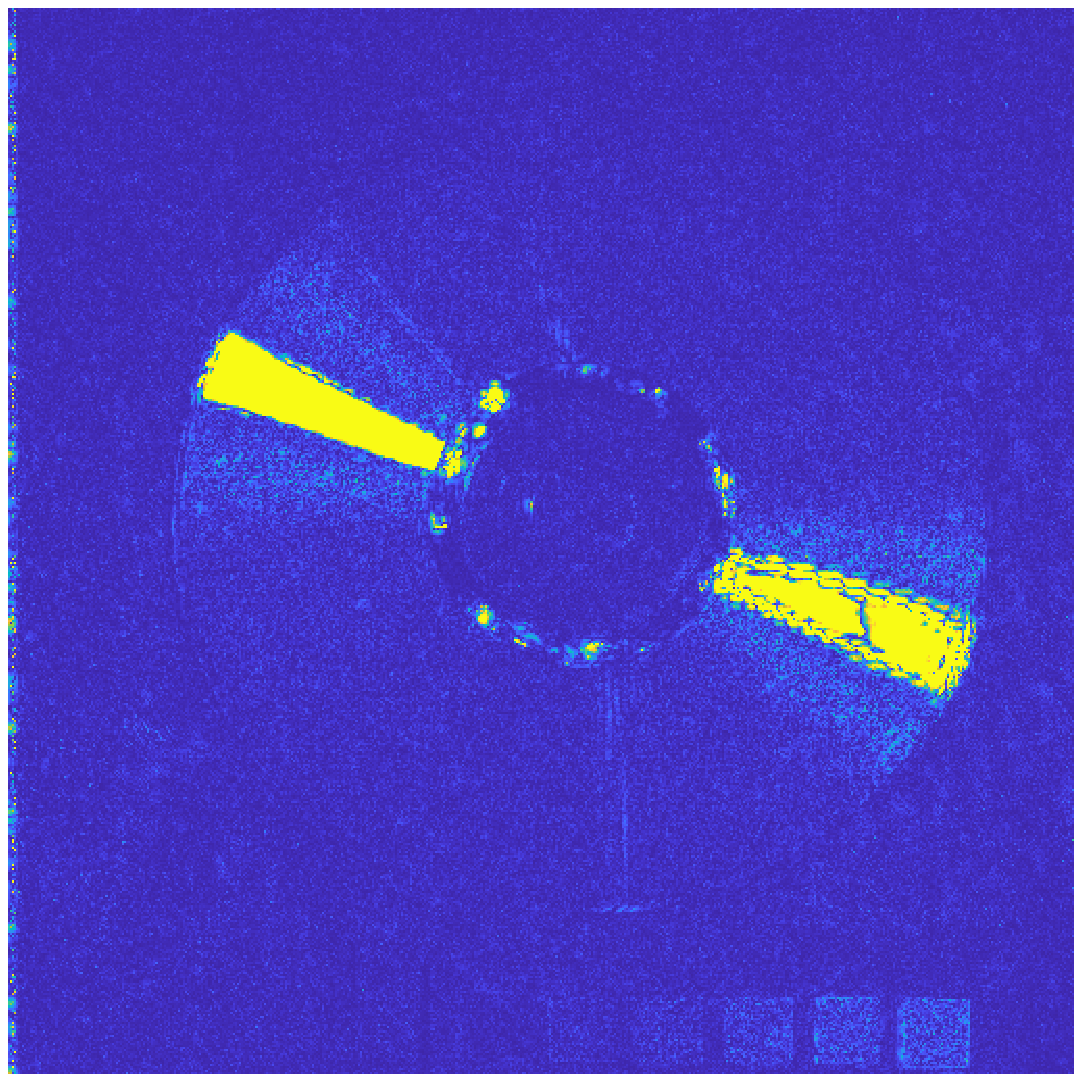}}
							{\includegraphics[width=1\linewidth]{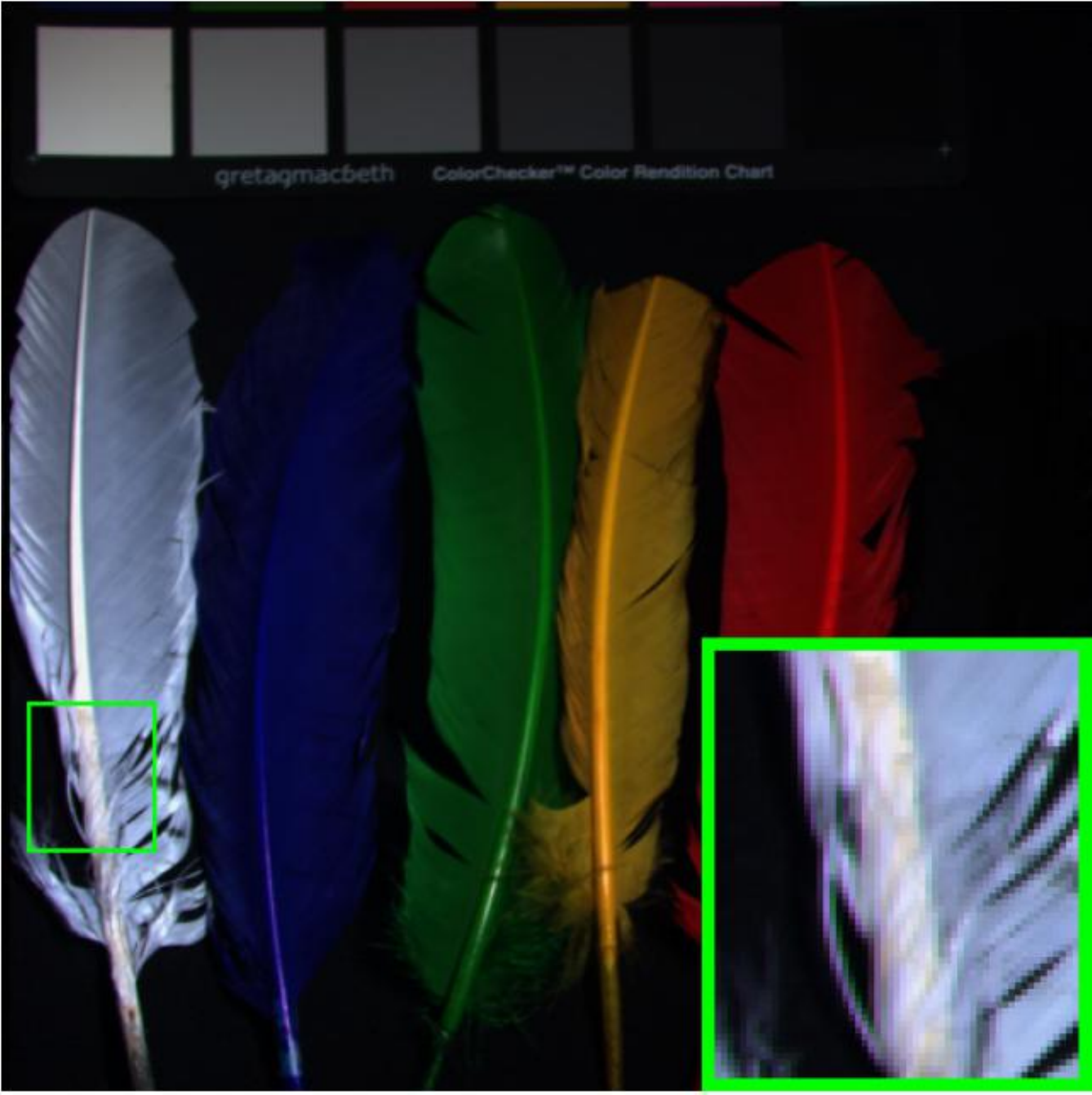}}
							\vspace{2pt}
							{\includegraphics[width=1\linewidth]{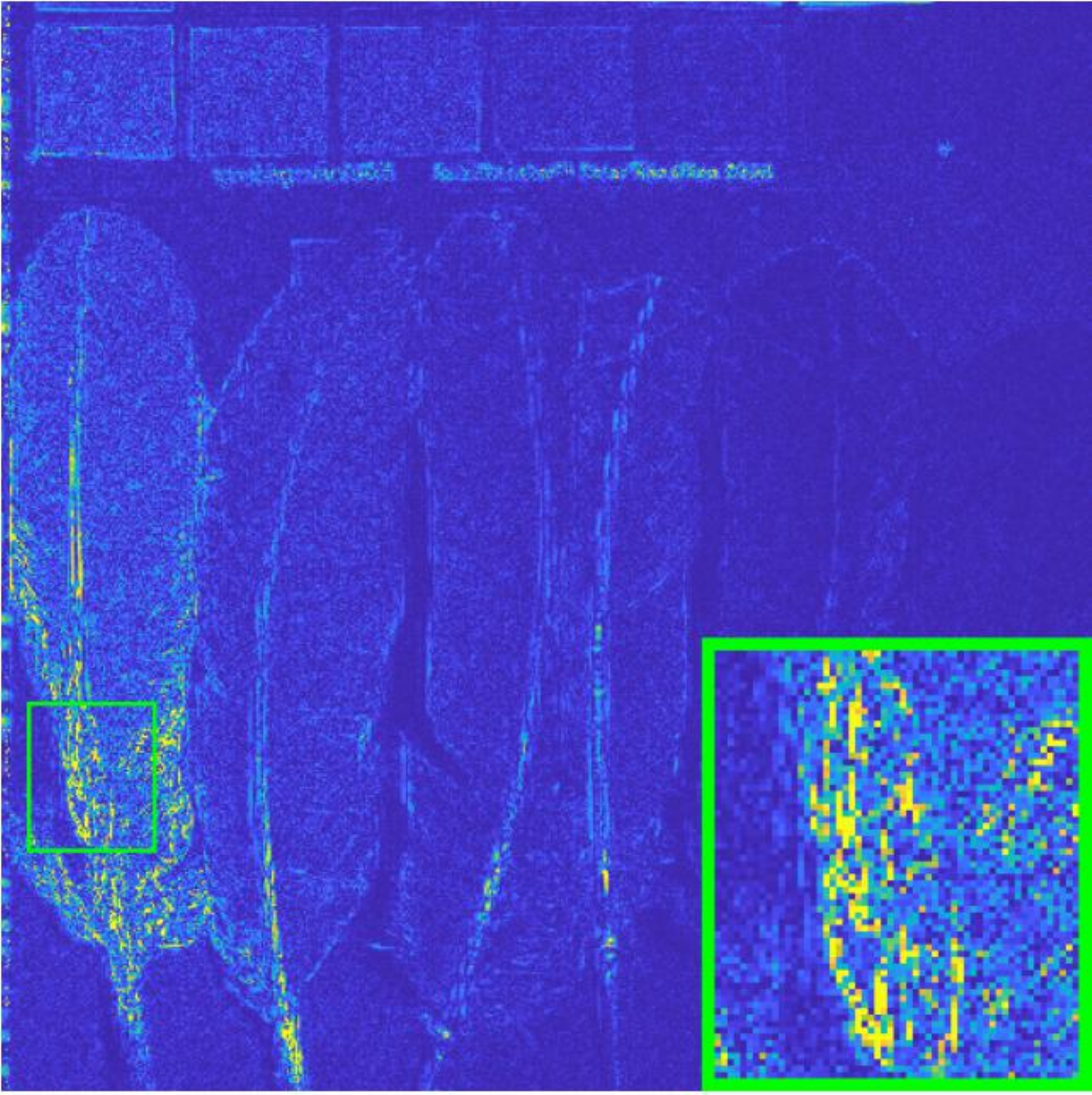}}
							{\includegraphics[width=1\linewidth]{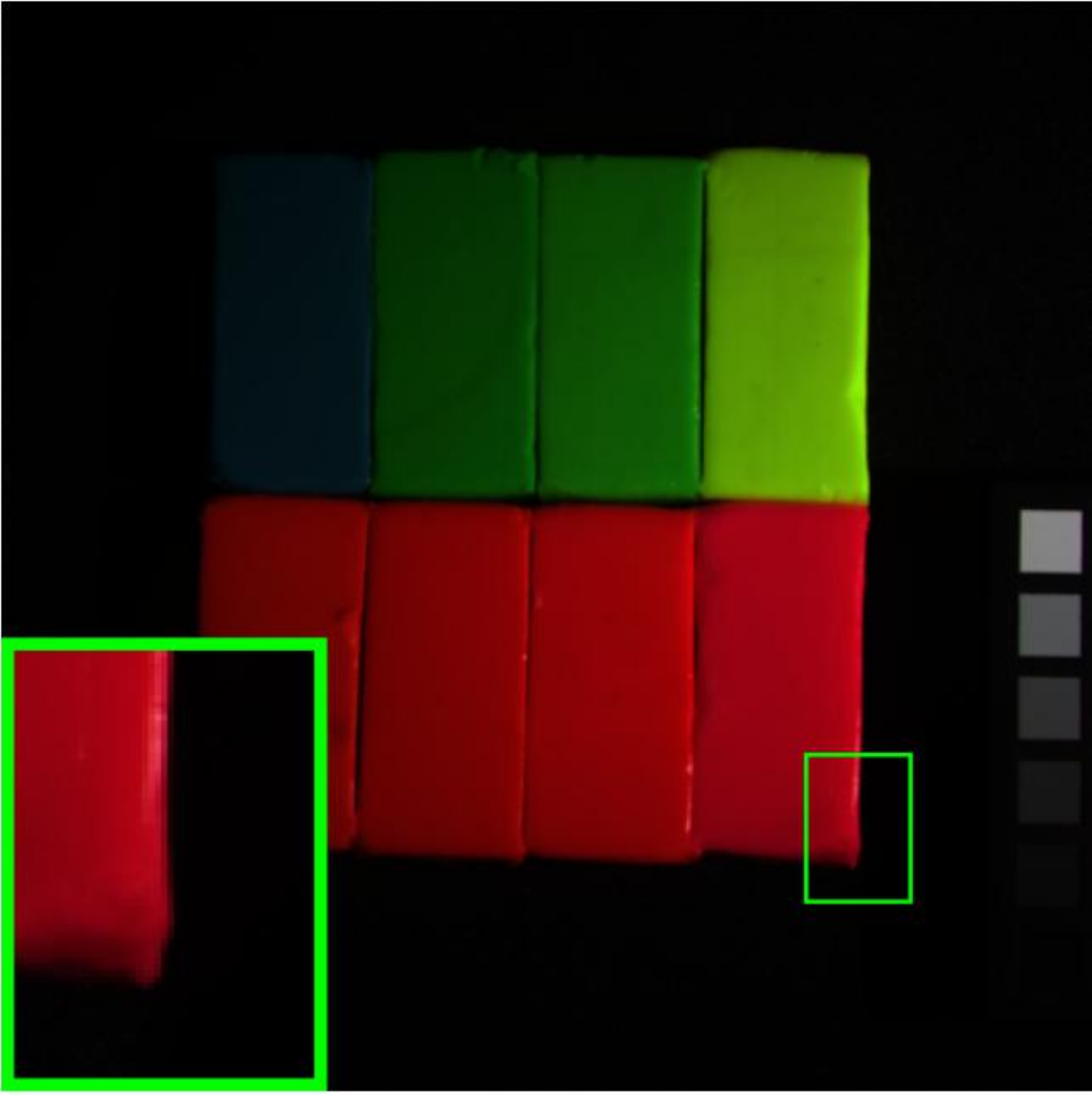}}
							{\includegraphics[width=1\linewidth]{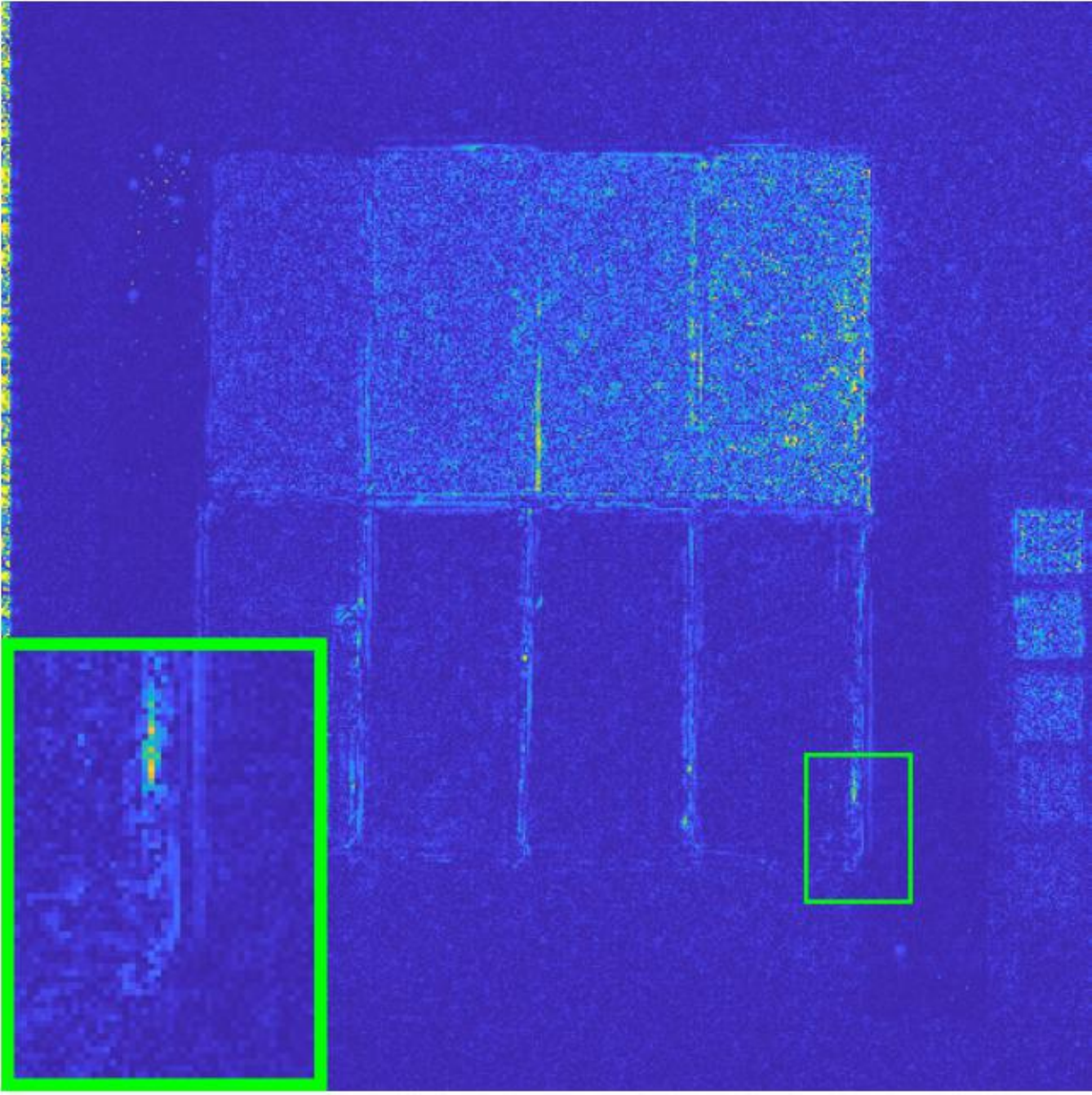}}
							\vspace{2pt}
							\scriptsize{PSRT}
							\centering
							
						\end{minipage}
					\begin{minipage}[t]{0.155\linewidth}
							{\includegraphics[width=1\linewidth]{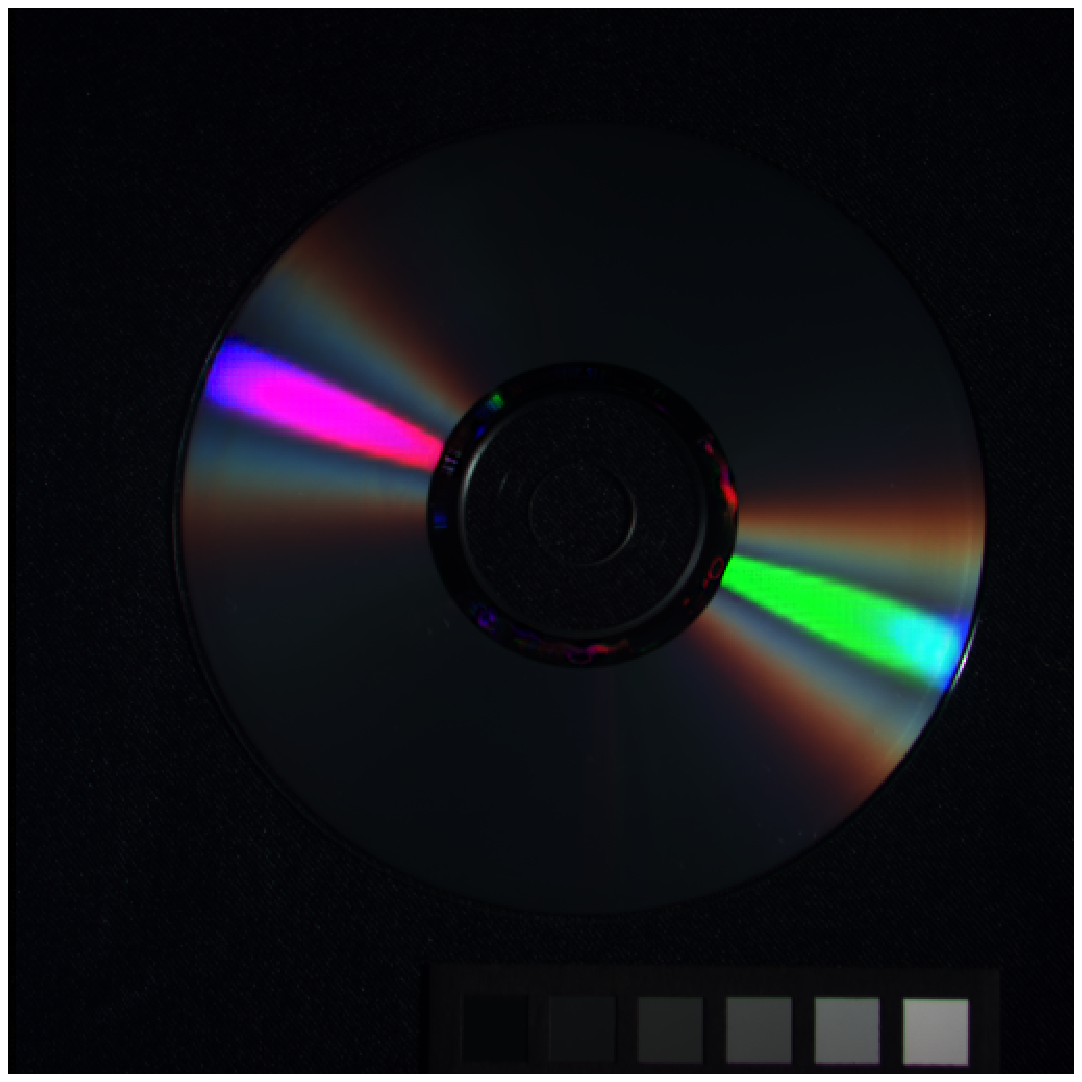}}
							\vspace{2pt}
							{\includegraphics[width=1\linewidth]{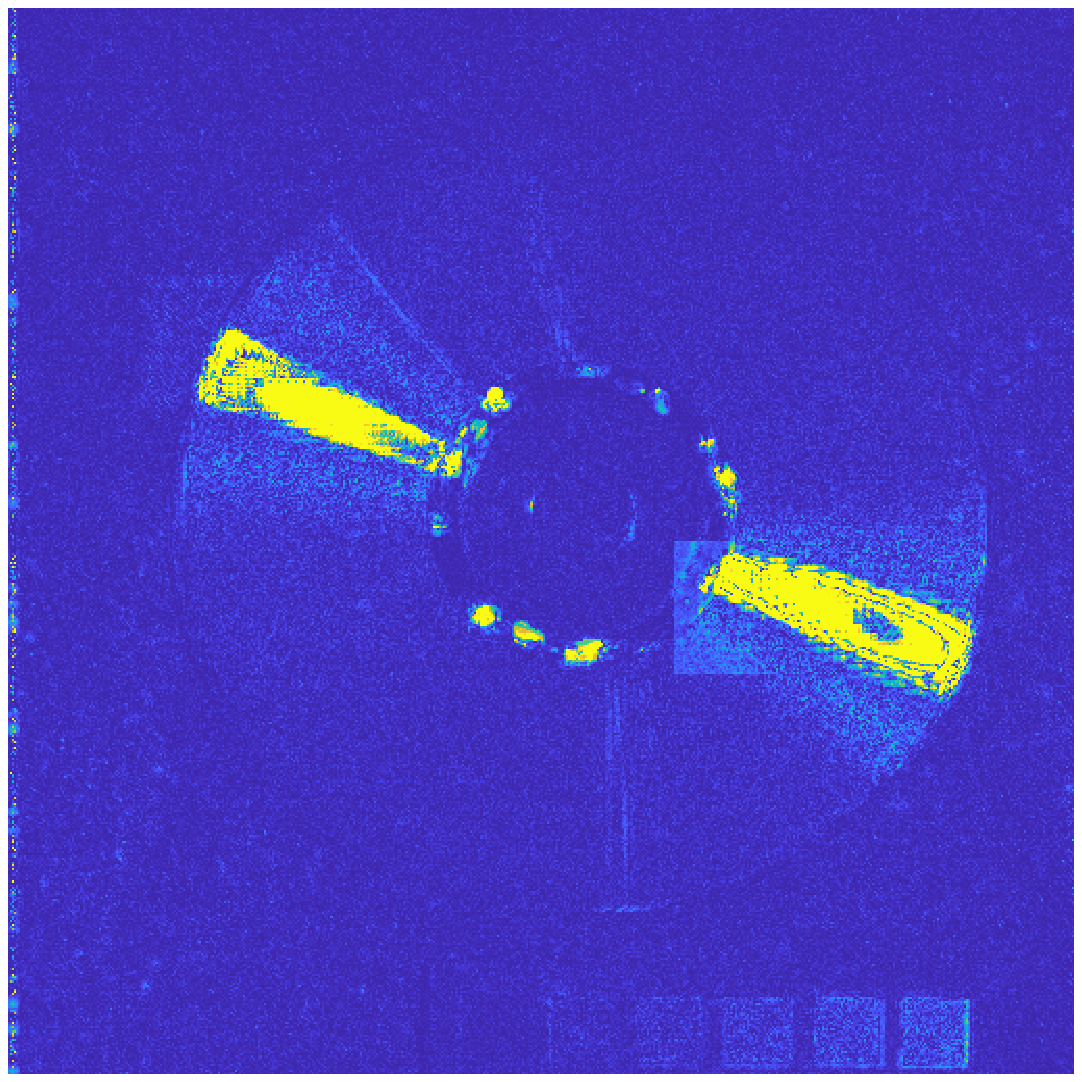}}
							{\includegraphics[width=1\linewidth]{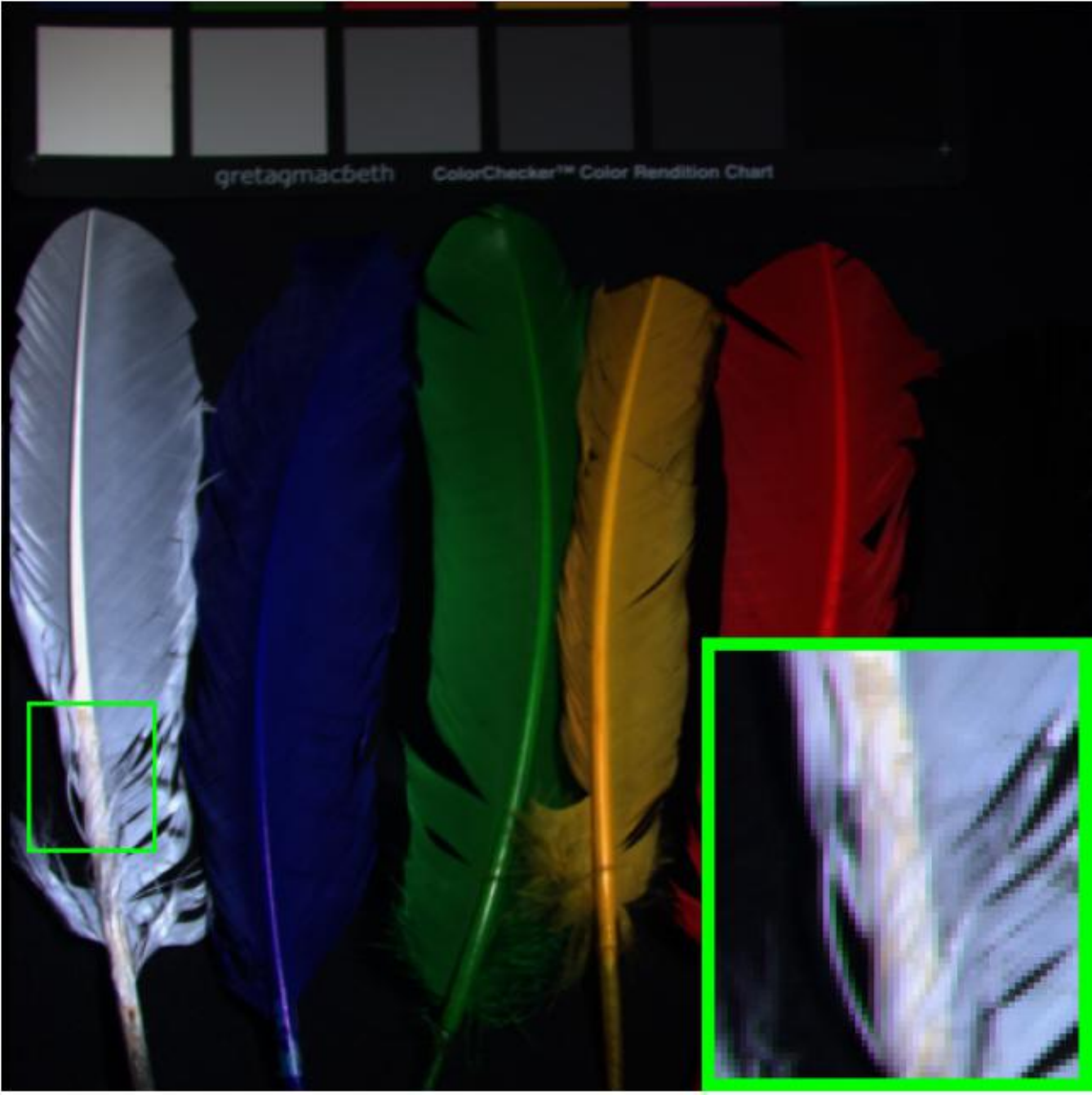}}
							\vspace{2pt}
							{\includegraphics[width=1\linewidth]{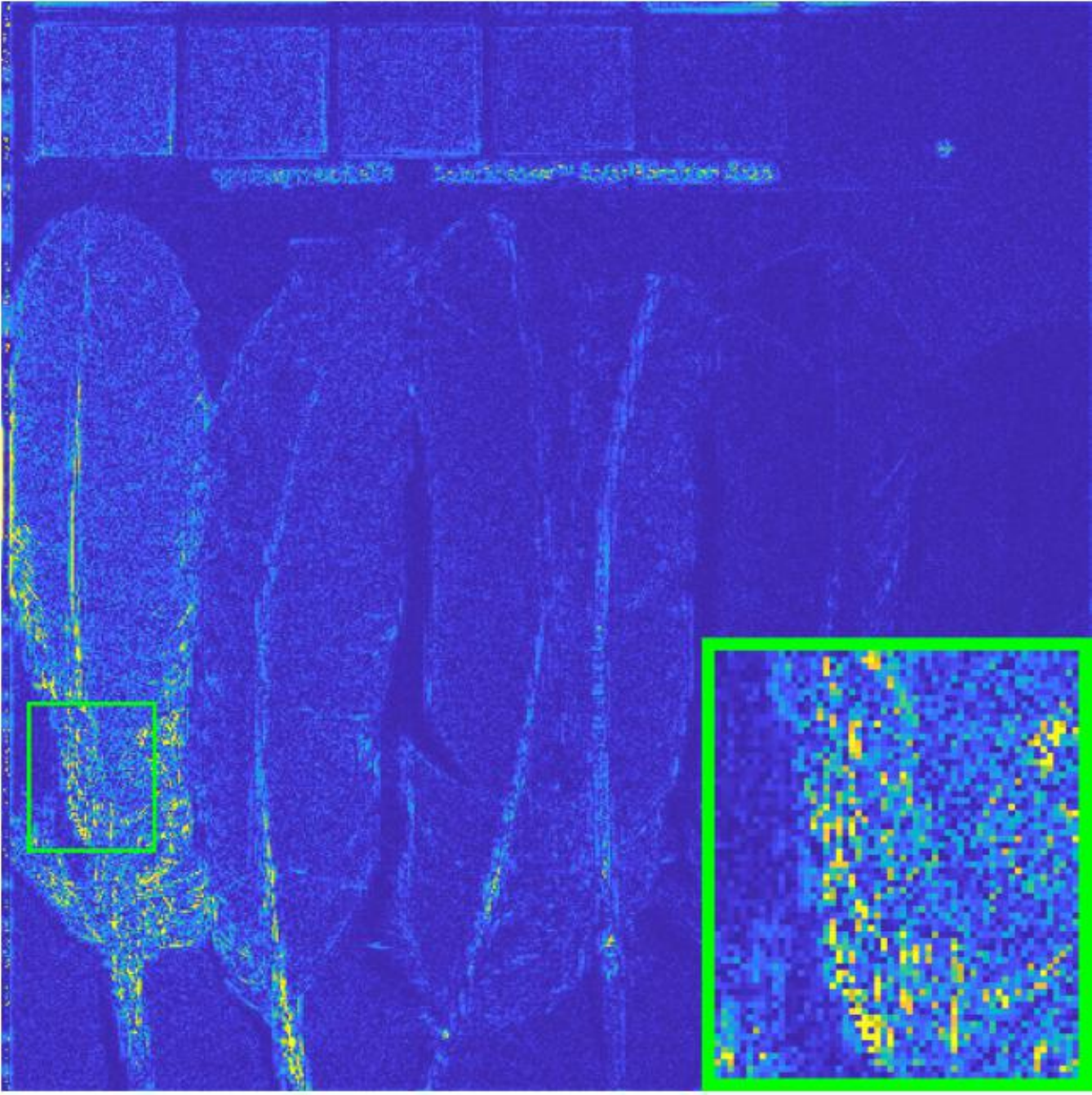}}
							{\includegraphics[width=1\linewidth]{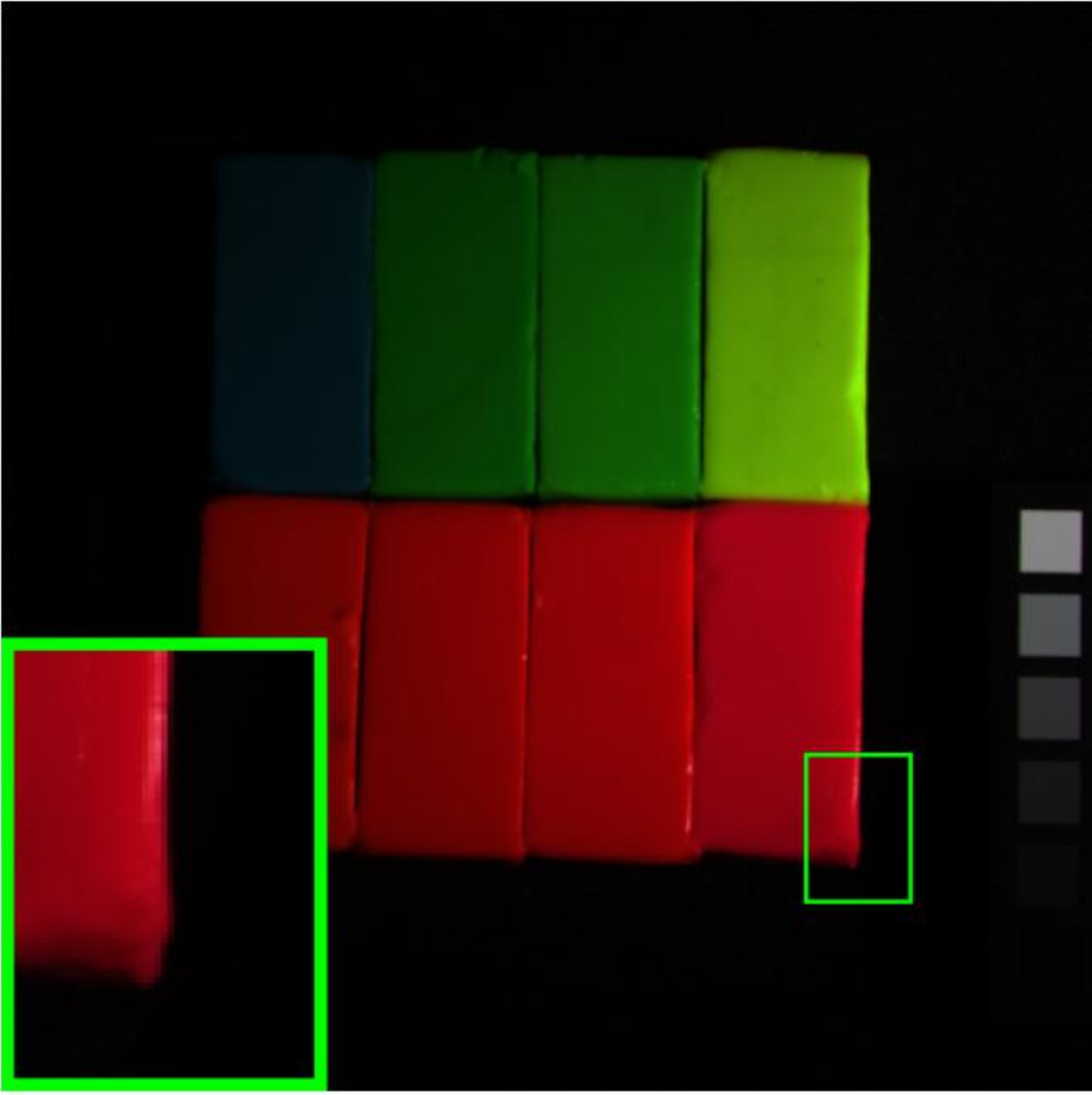}}
							{\includegraphics[width=1\linewidth]{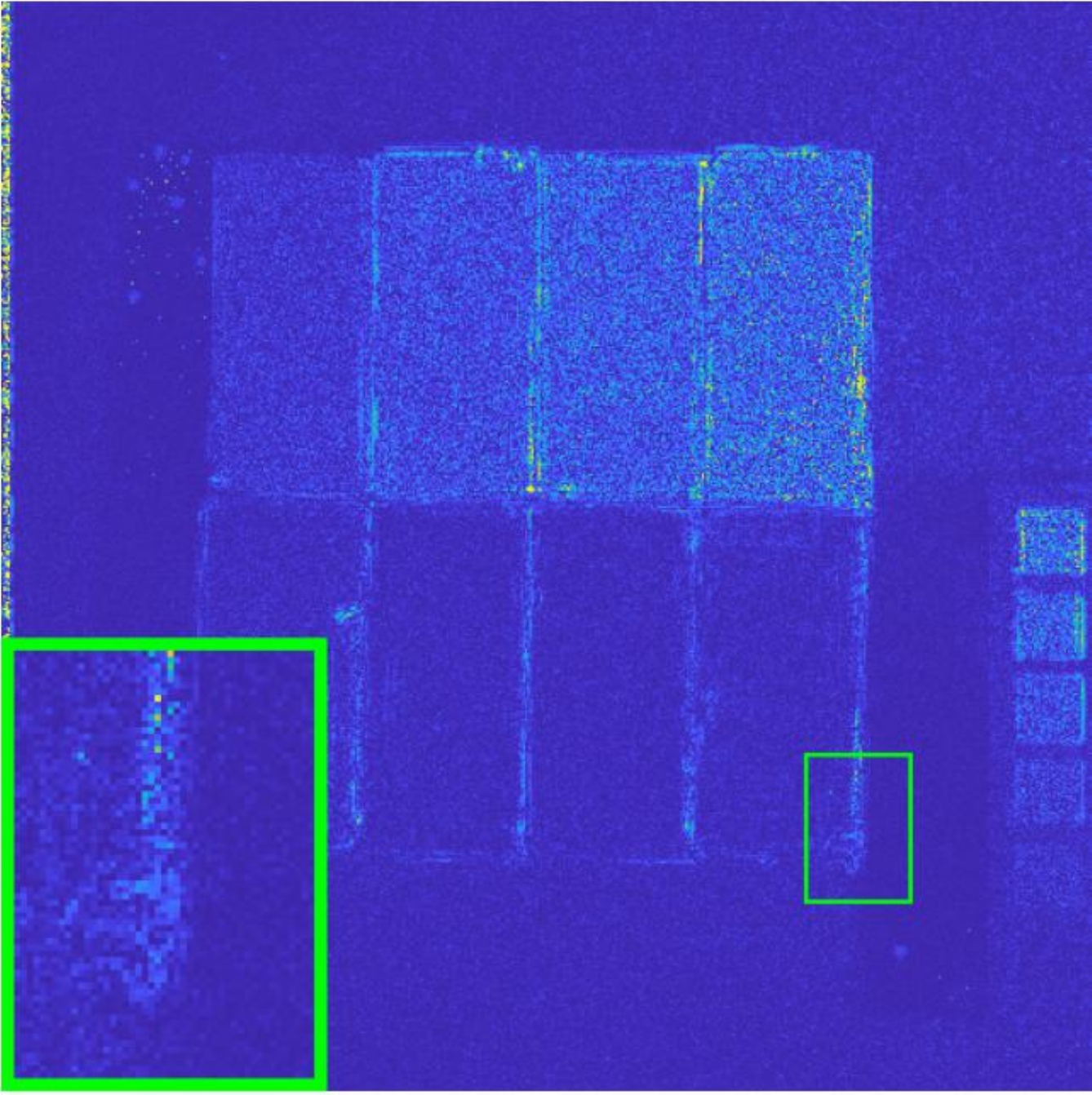}}
							\vspace{2pt}
							\scriptsize{U2Net}
							\centering
							
						\end{minipage}
					\begin{minipage}[t]{0.155\linewidth}
							{\includegraphics[width=1\linewidth]{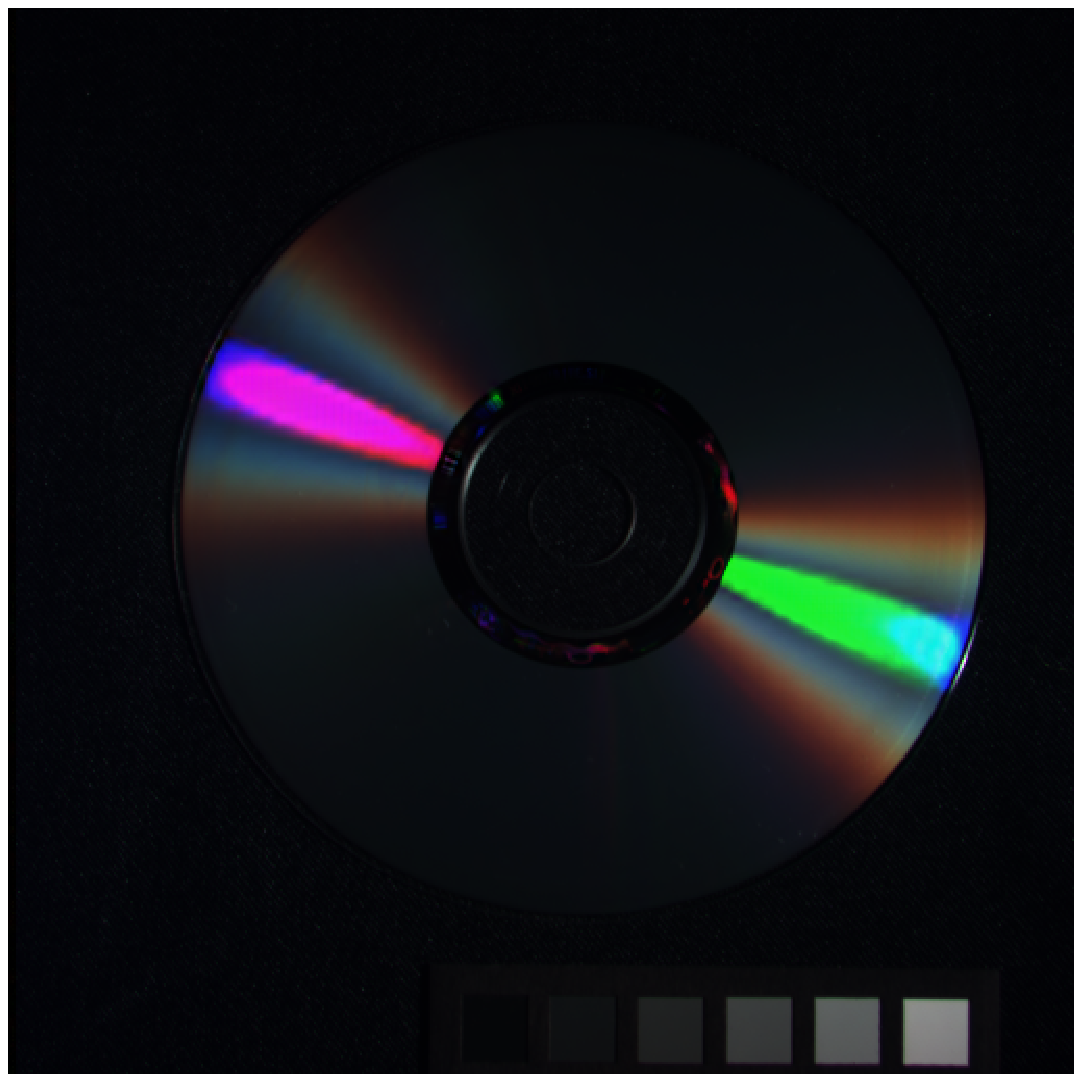}}
							\vspace{2pt}
							{\includegraphics[width=1\linewidth]{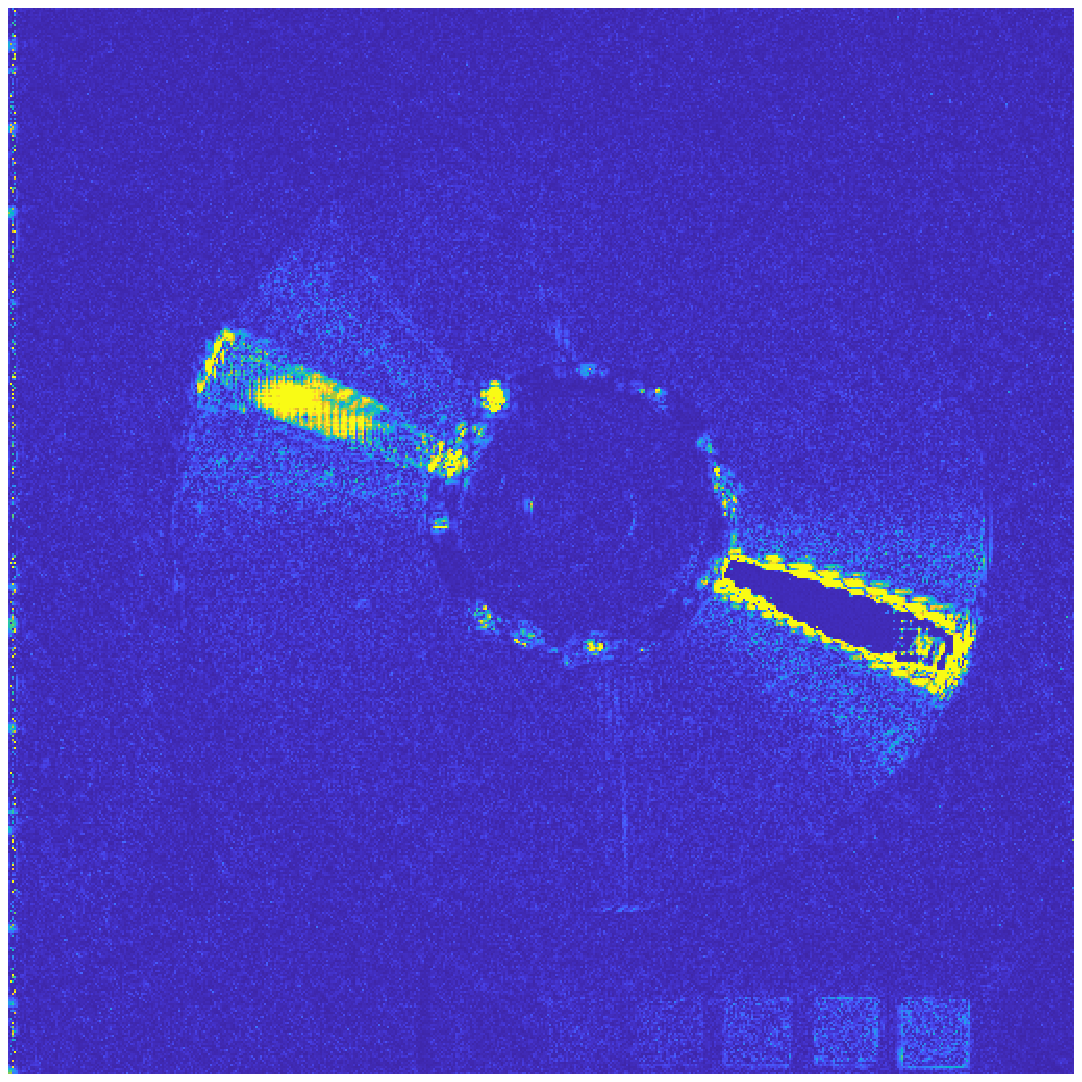}}
							{\includegraphics[width=1\linewidth]{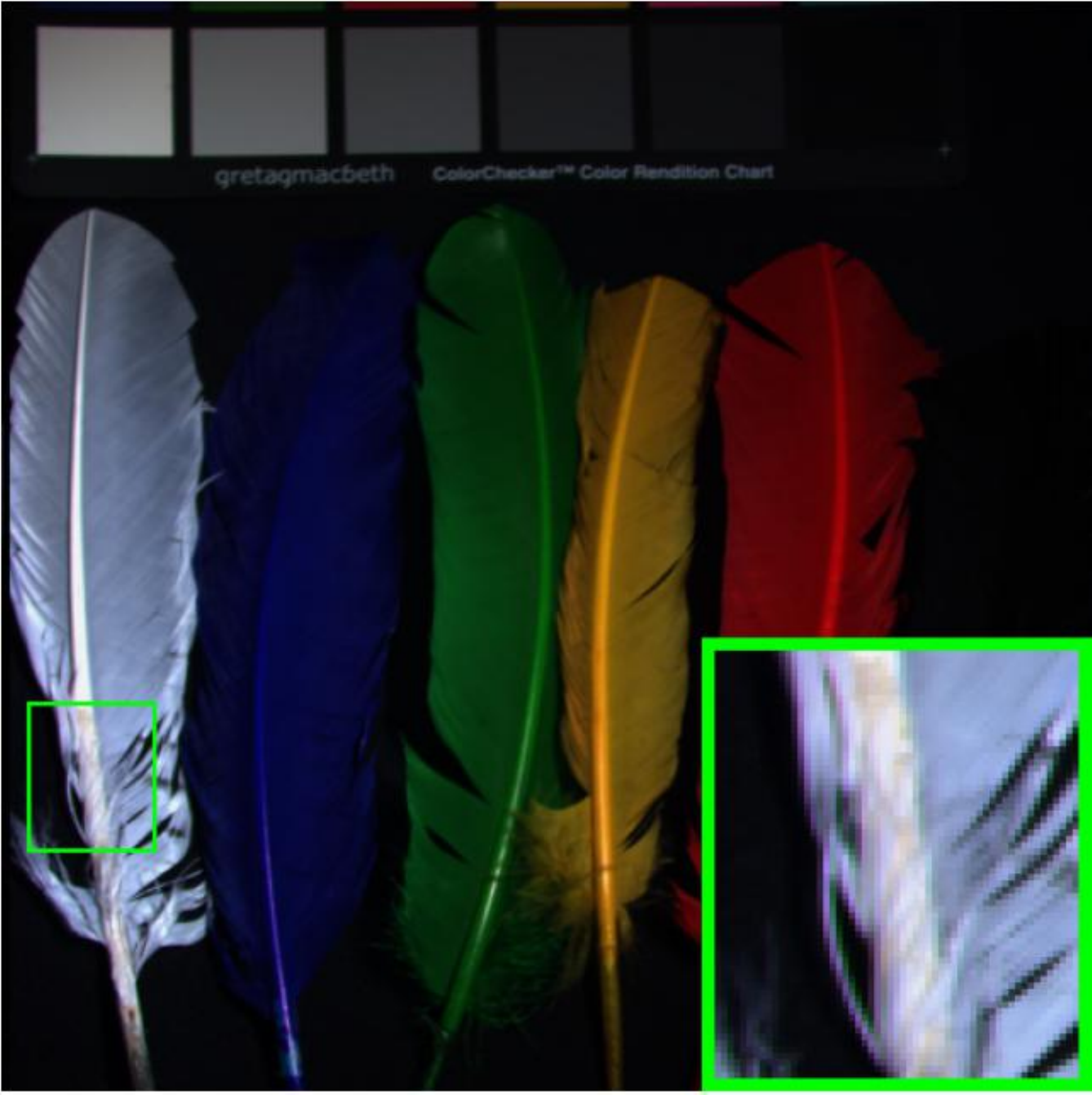}}
							\vspace{2pt}
							{\includegraphics[width=1\linewidth]{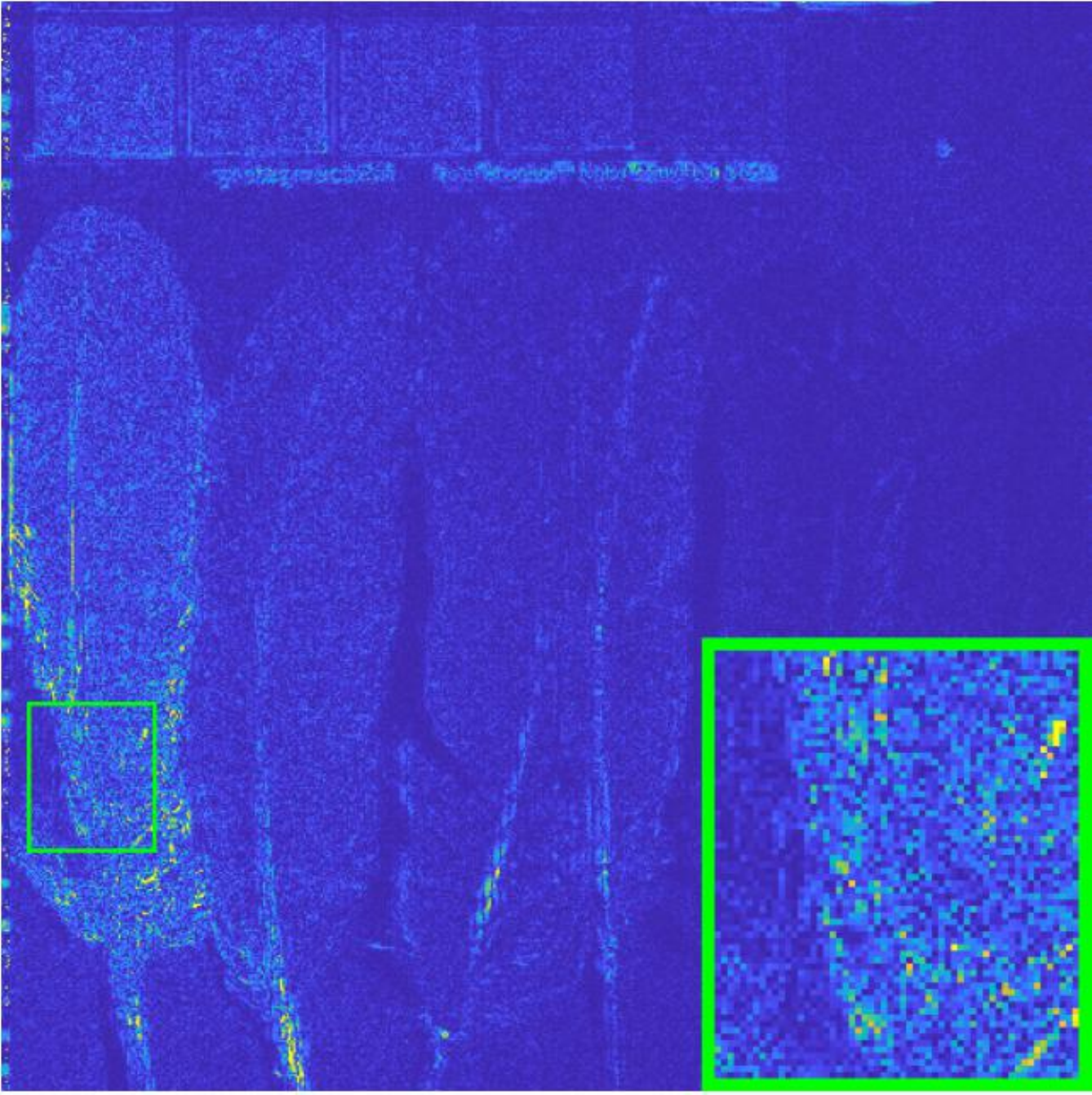}}
							{\includegraphics[width=1\linewidth]{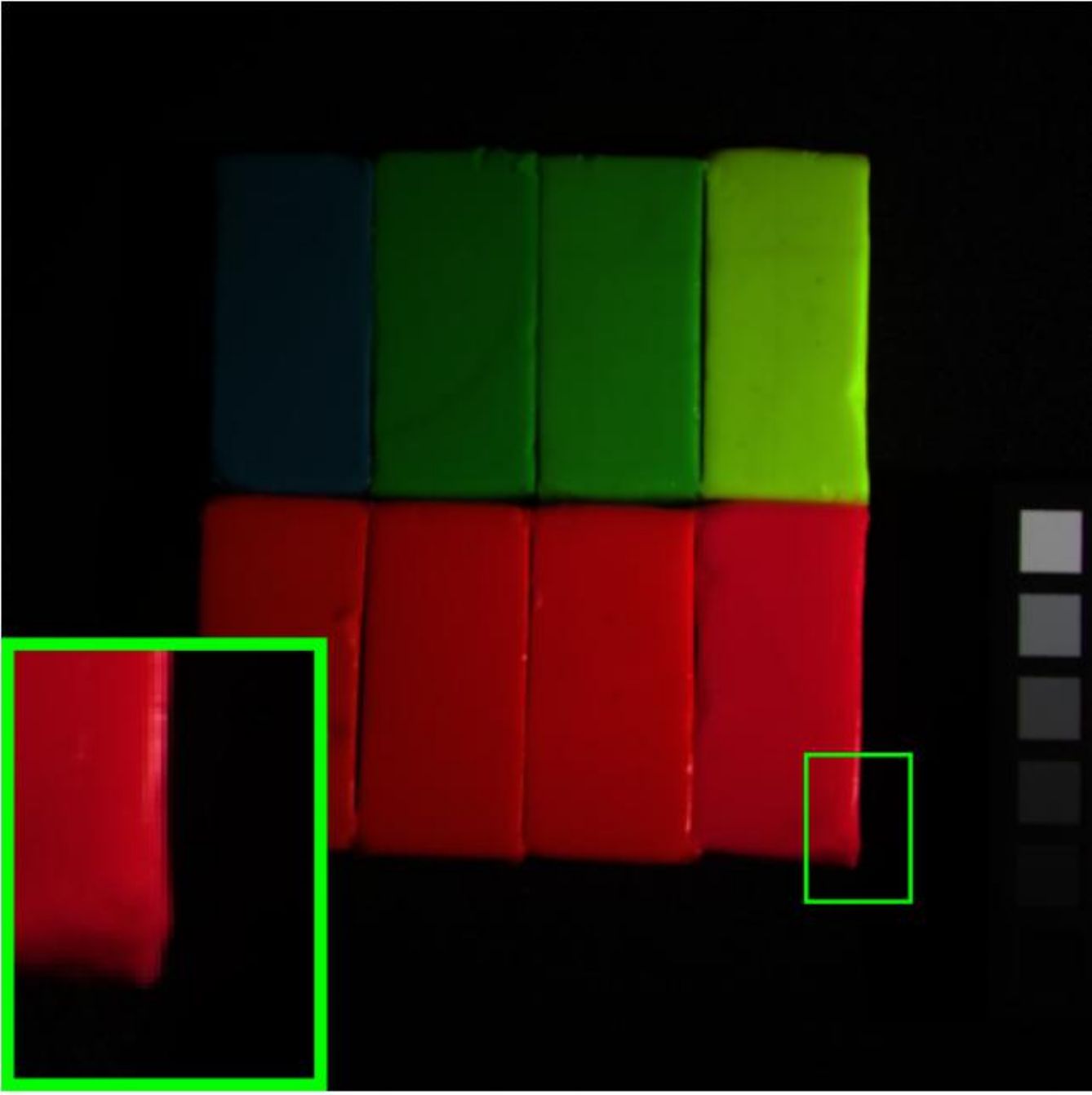}}
							{\includegraphics[width=1\linewidth]{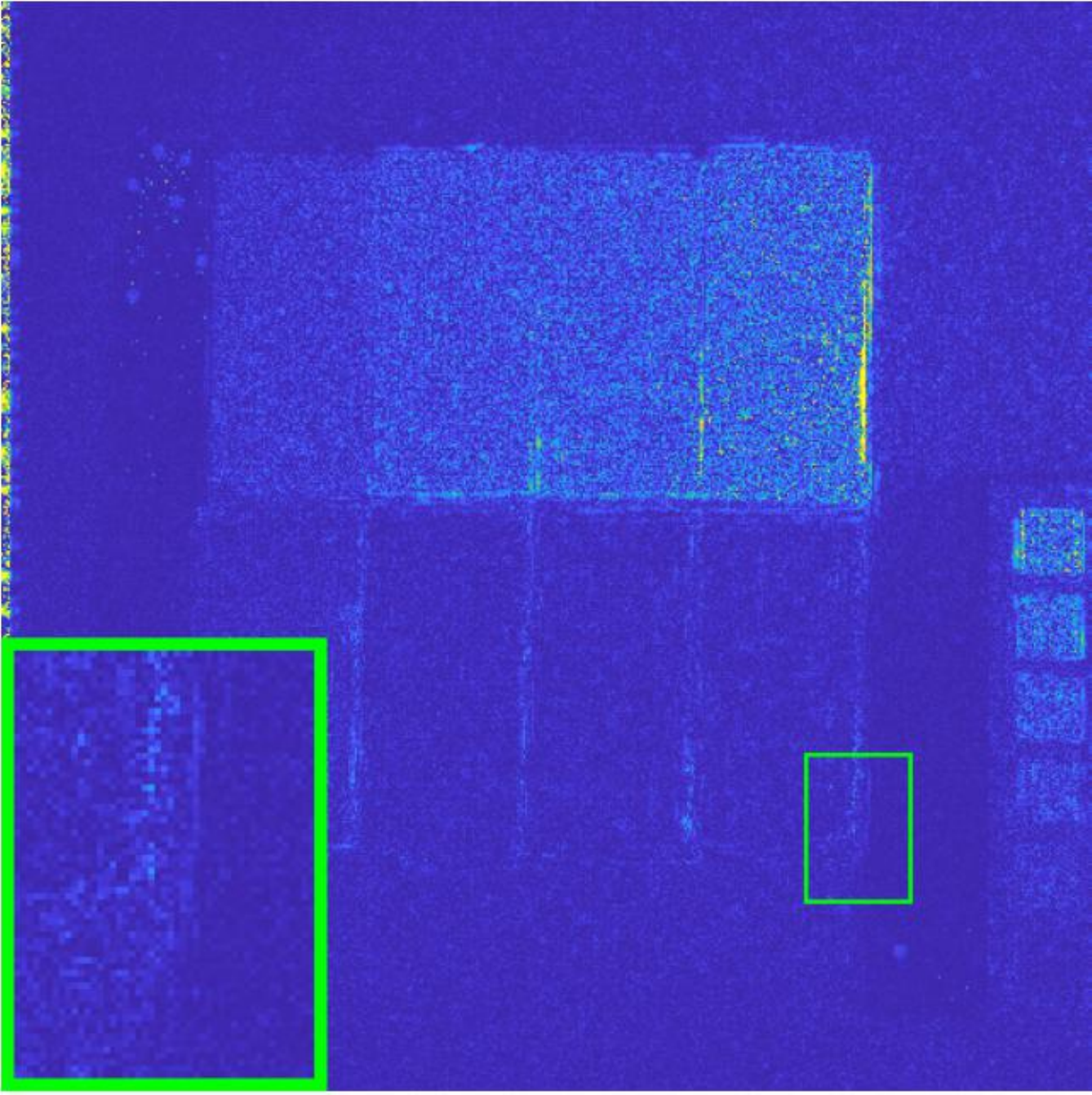}}
							\vspace{2pt}
							\scriptsize{Ours}
							\centering
							
						\end{minipage}
					\begin{minipage}[t]{0.155\linewidth}
							{\includegraphics[width=1\linewidth]{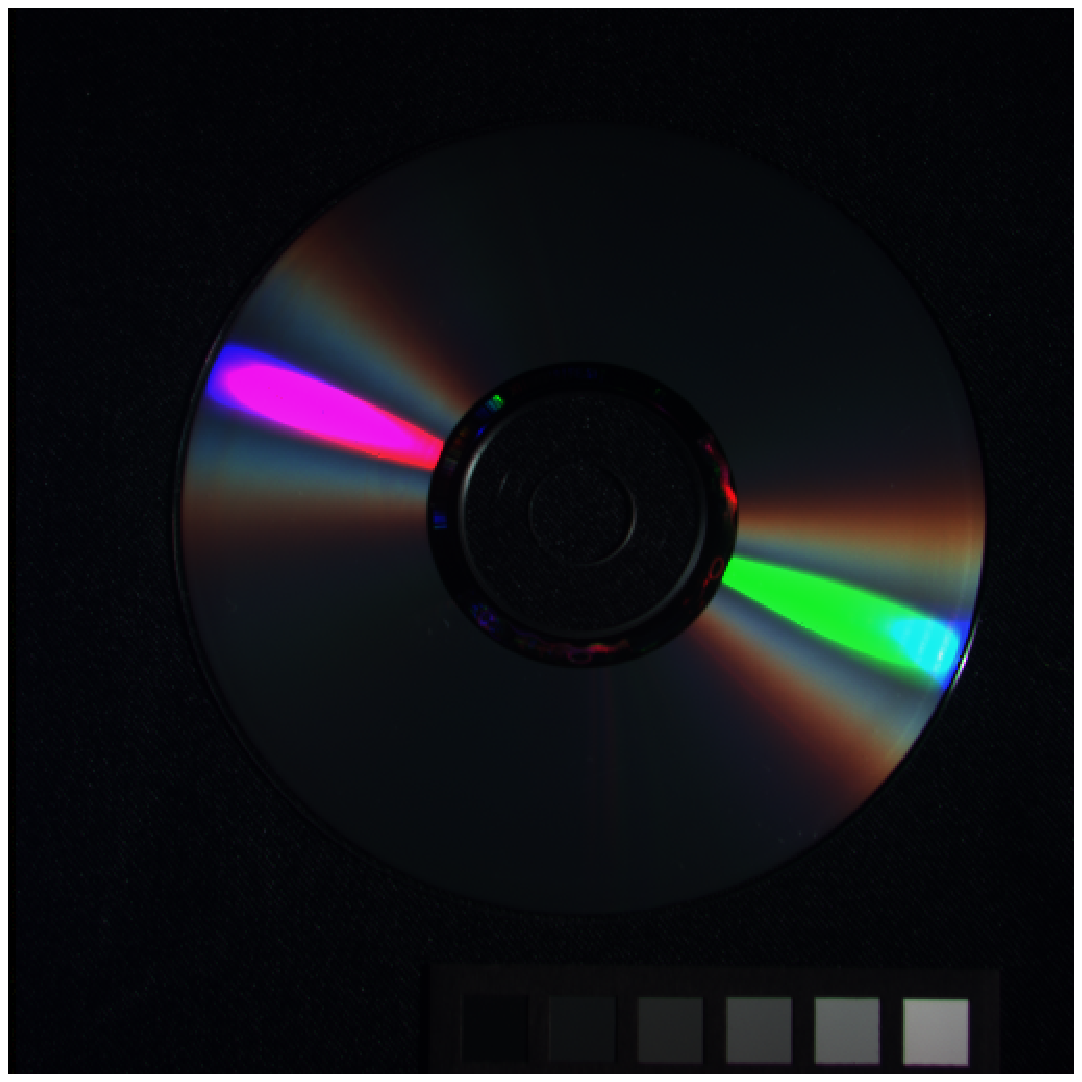}}
							\vspace{2pt}
							{\includegraphics[width=1\linewidth]{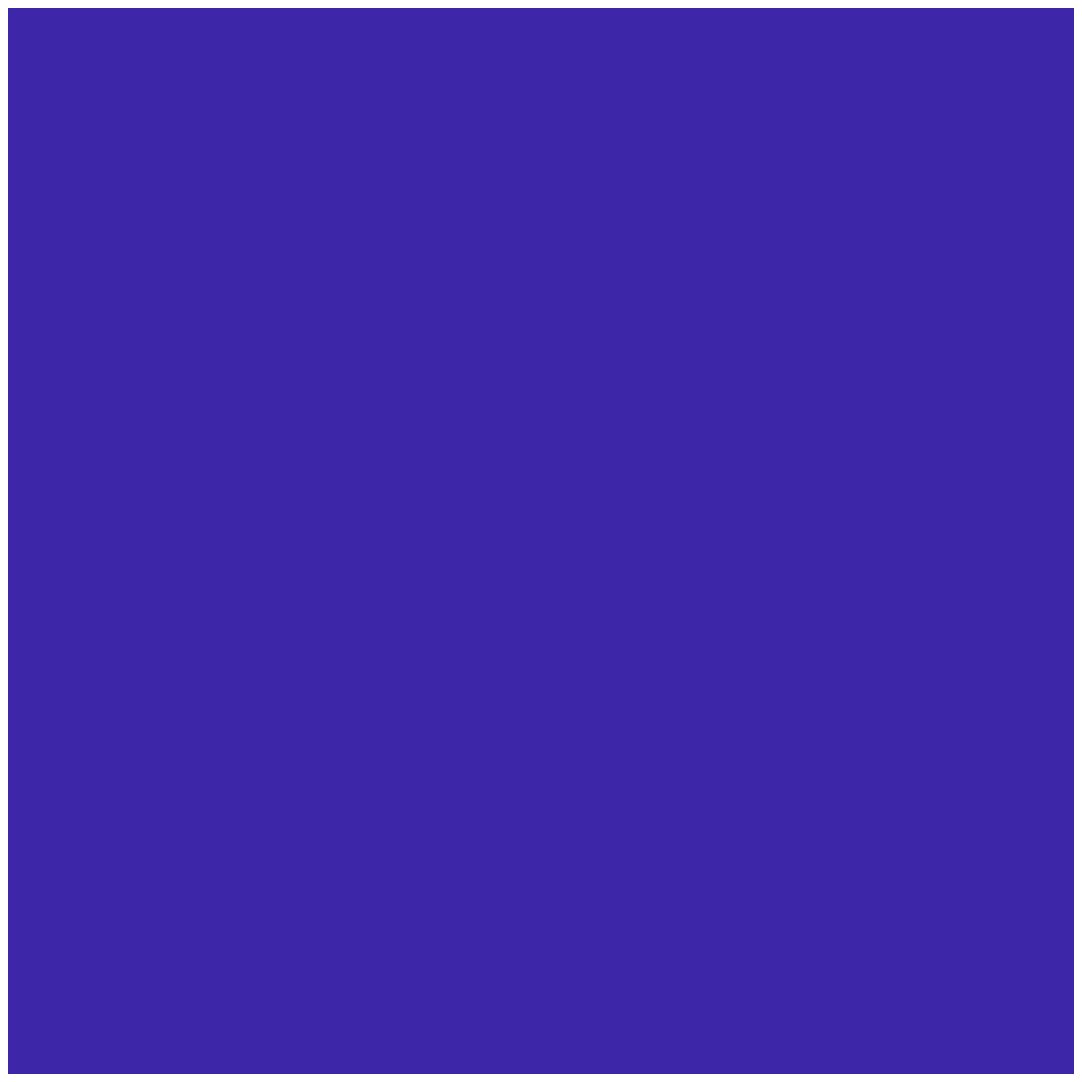}}
							{\includegraphics[width=1\linewidth]{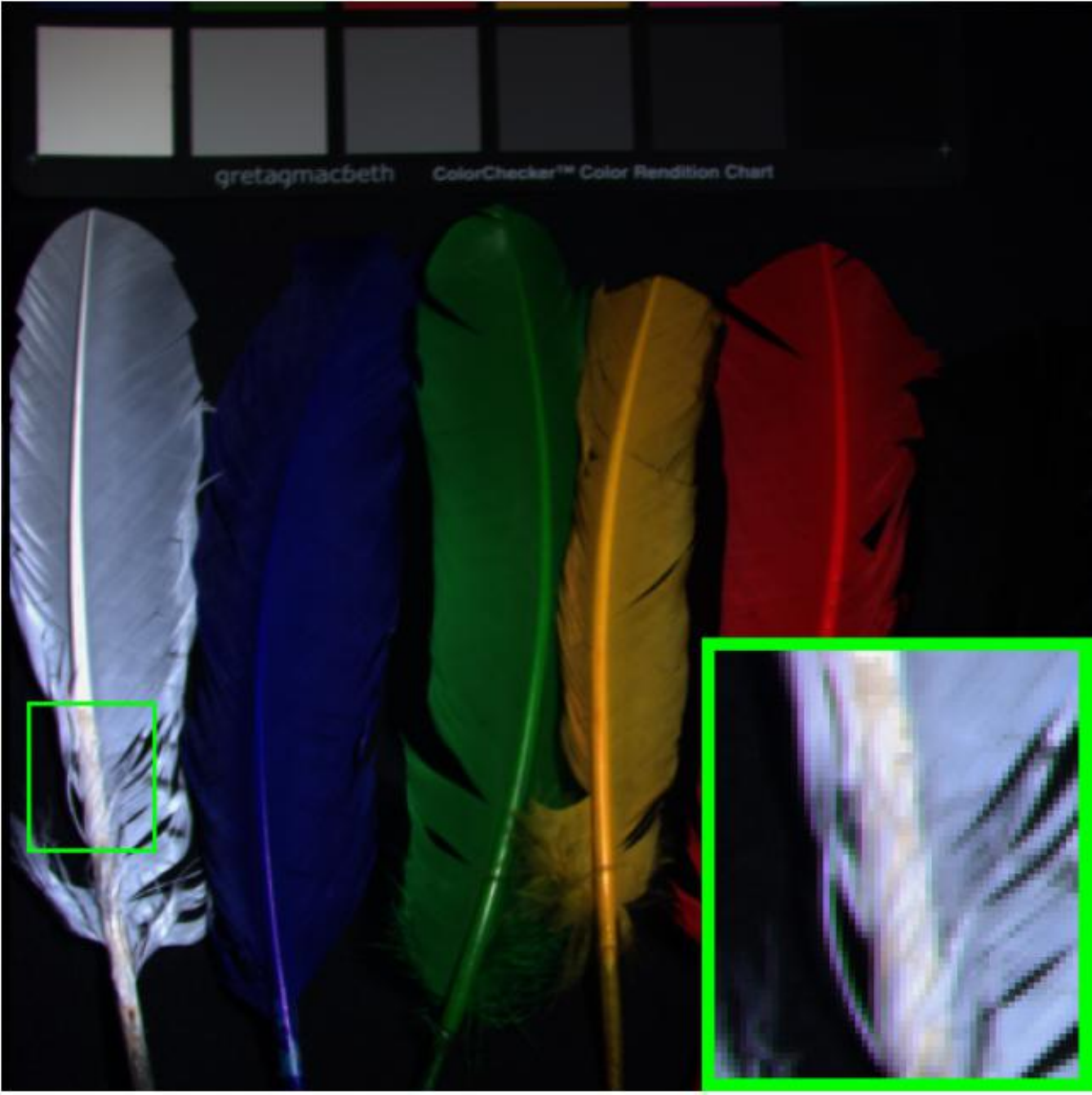}}
							\vspace{2pt}
							{\includegraphics[width=1\linewidth]{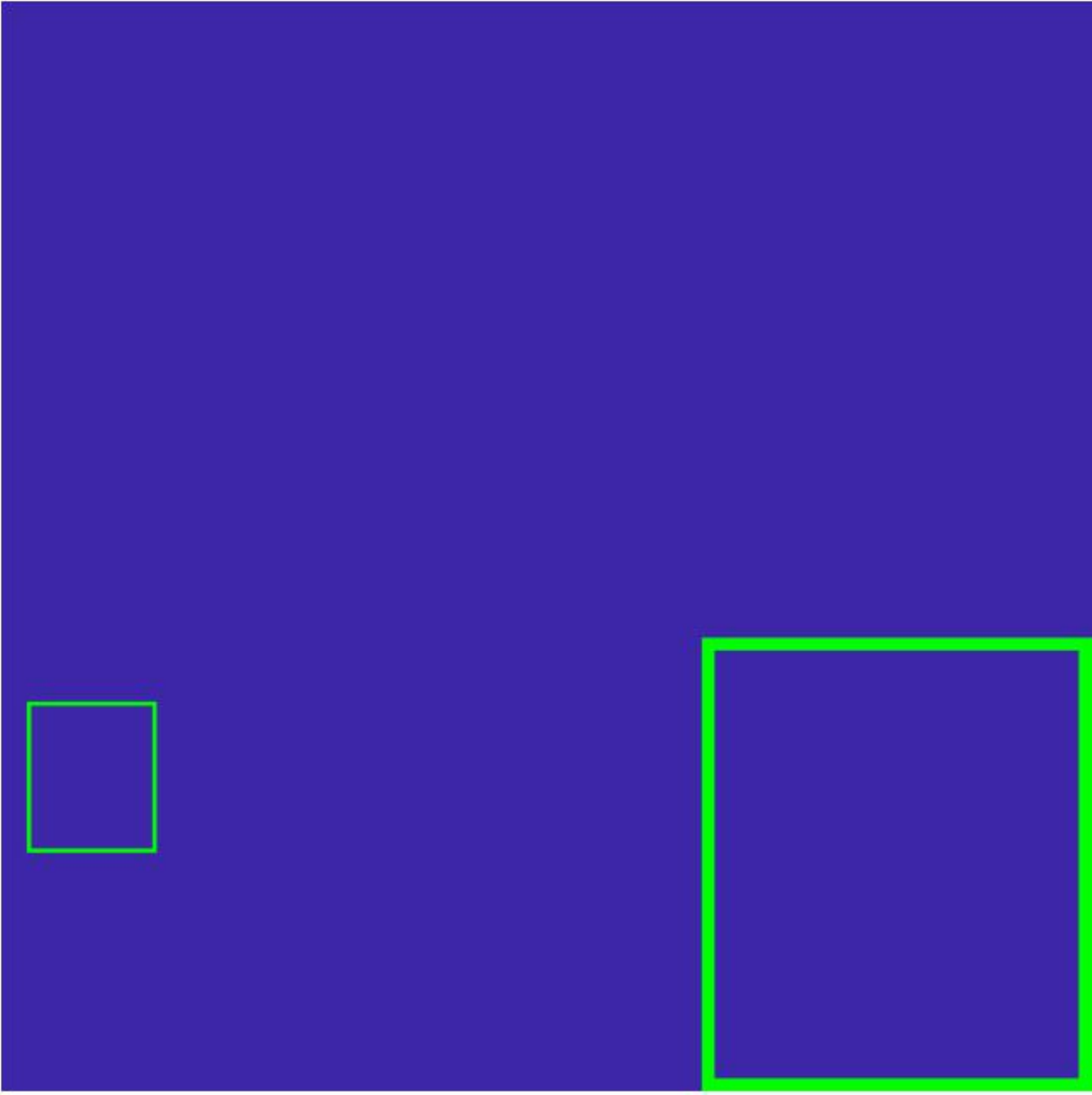}}
							{\includegraphics[width=1\linewidth]{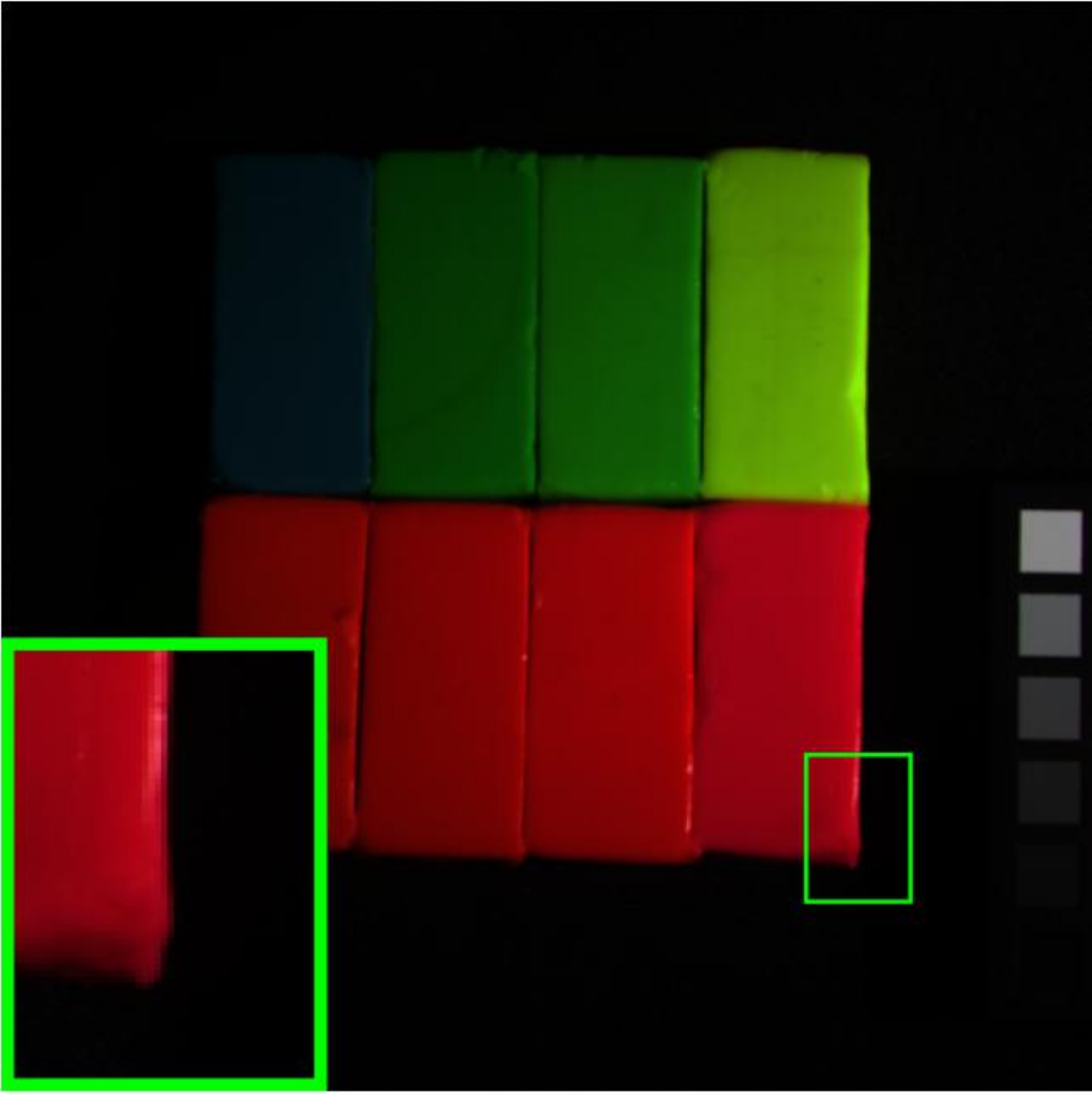}}
							{\includegraphics[width=1\linewidth]{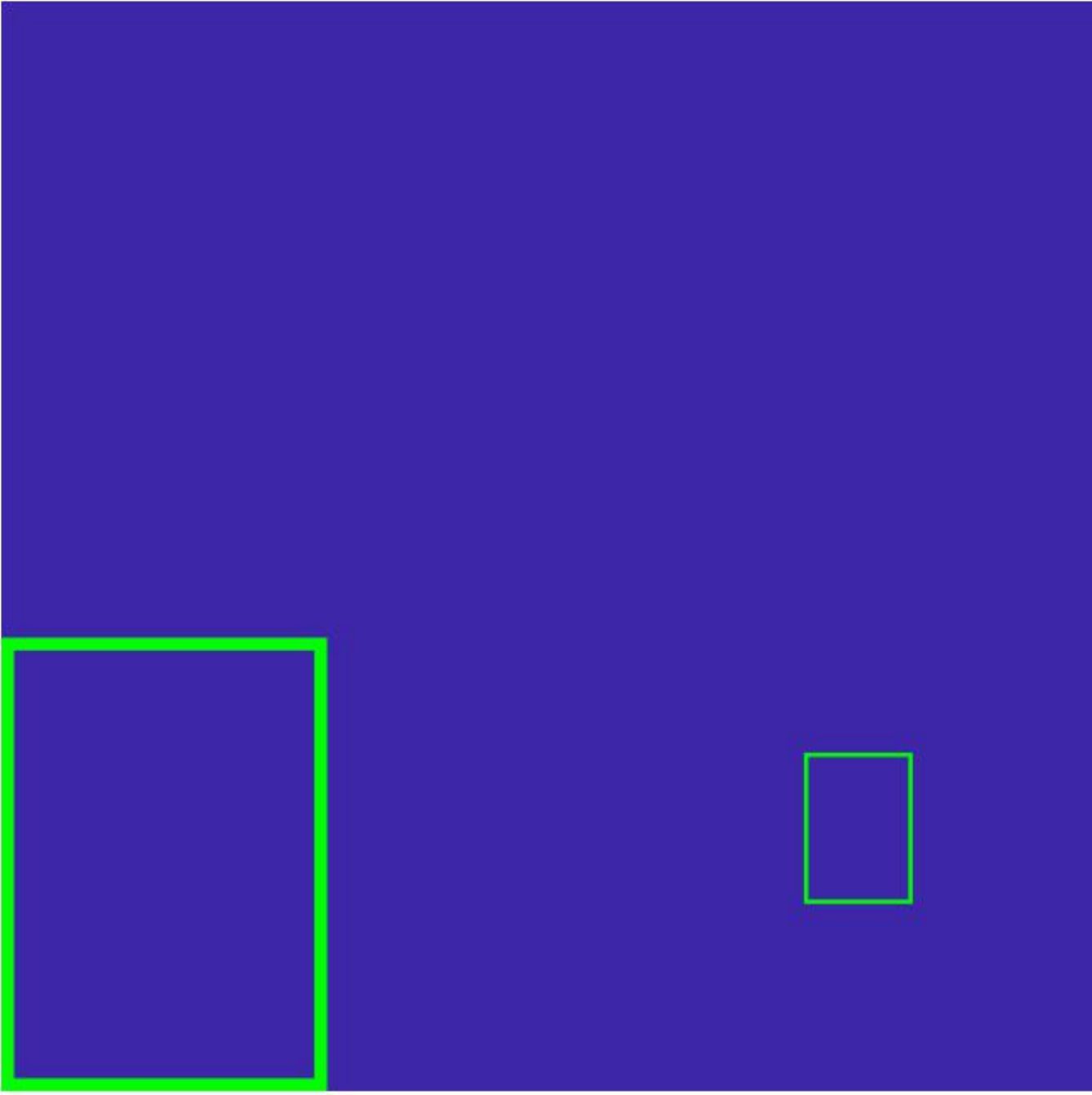}}
							\vspace{2pt}
							\scriptsize{GT}
							\centering
							
						\end{minipage}
				\end{minipage}
			\begin{minipage}[t]{0.98\linewidth}
					{\includegraphics[width=1\linewidth]{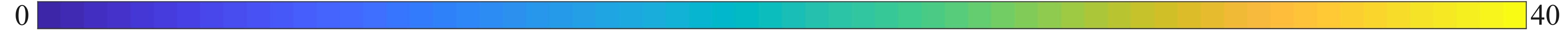}}
				\end{minipage}
		\end{center}
	\vspace{-6pt}
	\caption{Qualitative evaluation results of SOTA HISR methods on three testing samples from the CAVE dataset. Row 1: Pseudo-color images for spectral bands 6, 10, and 26 from the testing example \emph{cd}. Row 2: AEMs for spectral band 13 from \emph{cd}. Row 3: Pseudo-color images for spectral bands 6, 13, and 26 from the testing example \emph{feathers}. Row 4: AEMs for spectral band 14 from \emph{feathers}. Row 5: Pseudo-color images for spectral bands 6, 13, and 26 from the testing example \emph{clay}. Row 6: AEMs for spectral band 8 from \emph{clay}. \label{cave_v}}
\end{figure}

\subsection{Experiments for HISR}
\subsubsection{Dataset}

To validate the effectiveness of FusionMamba for image fusion tasks beyond remote sensing, we perform experiments using the CAVE dataset\footnote{\url{https://www.cs.columbia.edu/CAVE/databases/multispectral/}}, which belongs to the HISR task \cite{10194239}. Initially introduced in \cite{2010Generalized}, the CAVE dataset contains 32 RGB/HRHS image pairs with dimensions of $512 \times 512\times 3$ and $512 \times 512\times 31$, which are not directly suitable for training and testing purposes. 
{During the data generation phase, we select 20 samples for training and reserve the remaining samples for testing. From the training HRHS images, we extract 3920 overlapped patches of size $64 \times 64\times 31$ to serve as the GT images. Following this, a $3\times 3$ Gaussian blur kernel with a standard deviation of 0.5 is employed to down-sample the GT images, creating LRHS samples of size $16\times 16\times 31$. Additionally, we segment the training RGB images into 3920 overlapped patches, each sized $64\times 64\times3$, to match the spatial resolution of the GT images. This process generates 3920 training samples, each comprising an RGB/LRHS/GT image triplet of sizes $64\times 64\times 3$, $16\times 16\times 31$, and $64\times 64\times 31$, respectively. We process the testing data using a similar strategy, resulting in testing samples containing RGB/LRHS/GT image triplets sized $512\times 512\times 3$, $128\times 128\times 31$, and $512\times 512\times 31$.}

\subsubsection{Benchmarks, Metrics, and Settings}
We compare FusionMamba with several representative techniques for HISR, including two traditional approaches, namely LTMR \cite{dian2019hyperspectral} and UTV \cite{xujstars2020}, alongside six DL-based methods: ResTFNet \cite{2018Remote}, SSRNet \cite{9186332}, Fusformer \cite{9841513}, 3DT-Net \cite{ma2023learning}, PSRT \cite{deng2023psrt}, and U2Net \cite{10.1145/3581783.3612084}. 
In accordance with the research standards of HISR, we choose four quality indices for evaluation: PSNR, SSIM, SAM, and ERGAS. Their ideal values are $+\infty$, 1, 0, and 0, respectively. 
For the network configuration, we set $C$ and $N$ to 64 and 4, respectively. Additionally, we utilize the bicubic interpolation for up-sampling. During training, we set the batch size, number of epochs, and initial learning rate to 32, 1100, and $2\times 10^{-4}$, respectively. Besides, we employ the Adam optimizer and halve the learning rate every 500 epochs.

\subsubsection{Results} 
The quantitative evaluation results, as presented in Table~\ref{cave}, indicate that our method outperforms all others. Additionally, the qualitative evaluation outcomes, as depicted in Fig.~\ref{cave_v}, illustrate that the FusionMamba produces fusion outputs that most closely match the GT images. These findings strongly demonstrate the effectiveness of FusionMamba in image fusion tasks beyond remote sensing applications.

\subsection{Comparison between Transformers and Mamba}
{Both Transformers and Mamba demonstrate strong global perception capabilities, but Mamba consistently outperforms Transformers, as shown in Table~\ref{abl5}. This superiority can be attributed to two key factors. First, Mamba’s approach to global perception aligns more intuitively with image processing principles, where the influence of neighboring pixels on the output is generally more significant. Second, Mamba dynamically learns projection and timescale parameters from the input, making it content-aware. In contrast, while Transformers offer some adaptability through their self/cross-attention mechanisms, they lack input-derived parameters, which limits their content-awareness compared to Mamba. Furthermore, as discussed in Section~\ref{flopsexp}, Mamba exhibits significantly greater efficiency than Transformers. Consequently, Mamba emerges as a highly effective alternative to Transformers.}

\subsection{Comparison between Pan-Mamba and FusionMamba}
{As the pioneering application of the SSM in pansharpening, Pan-Mamba \cite{he2024pan} employs stacked one-directional Mamba blocks to extract spatial and spectral features. For information integration, it merely adds the outputs of two Mamba blocks, resulting in limited interaction between spatial and spectral data. In contrast, the FusionMamba incorporates four-directional Mamba blocks into two U-shape network branches, allowing for the effective and hierarchical learning of spatial and spectral characteristics. Furthermore, the proposed FusionMamba block enhances information integration through data interaction at the SSM level. Consequently, the FusionMamba is more powerful and interpretable than Pan-Mamba.}

\subsection{Strengths, Limitations, and Future Work}
\subsubsection{Strengths}
First, the proposed FusionMamba block enables the effective combination of spatial and spectral features with only linear computational complexity, representing a significant advancement in the application of the SSM for combining different types of information. With an identical number of parameters, our method outperforms existing fusion techniques like concatenation and cross-attention. Furthermore, experimental results in Table~\ref{abl3} demonstrate that the FusionMamba block can function as a plug-and-play module for information integration. Second, our interpretable network architecture allows for the separate and hierarchical learning of spatial and spectral features. This architecture also facilitates the progressive fusion of different information types and improves the representation of spectral characteristics. Therefore, the FusionMamba offers an optimal solution for image fusion.

\subsubsection{Limitations}
Our method exhibits two limitations. First, the FusionMamba block is designed to support only two inputs, restricting its applicability in fusion tasks that involve more than two inputs. Second, the FusionMamba block requires both feature maps to have the same number of channels, which is less flexible than the concatenation operation.

\subsubsection{Future Work}
In the future, we plan to expand the dual-input FusionMamba block to accommodate an arbitrary number of inputs, each with a variable number of feature channels. Additionally, we will delve deeply into the theories of the SSM to foster further groundbreaking innovations.

\section{Conclusion}
\label{s6}
In this paper, we propose FusionMamba, an innovative method for efficient remote sensing image fusion. To sufficiently merge spatial and spectral features, we expand the single-input Mamba block to accommodate dual inputs, creating the FusionMamba block. This novel module surpasses existing fusion techniques such as concatenation and cross-attention, representing a successful application of the SSM for information integration. Besides, the FusionMamba block can serve as a plug-and-play module, effectively merging different types of information. Additionally, our interpretable network architecture supports the separate and hierarchical learning of spatial and spectral characteristics, facilitates the progressive combination of different feature maps, and enhances the representation of spectral information. We evaluate the performance of FusionMamba across six datasets covering three image fusion tasks: pansharpening, hyper-spectral pansharpening, and HISR. Our method yields exceptional results, demonstrating the superiority of FusionMamba in image fusion.

\bibliographystyle{IEEEtran}
\bibliography{reference}


%
%
%
%

\vfill

\end{document}